# Feature Purification:
# How Adversarial Training Performs Robust Deep Learning

Zeyuan Allen-Zhu
zeyuan@csail.mit.edu
Microsoft Research Redmond

Yuanzhi Li
yuanzhil@andrew.cmu.edu
Carnegie Mellon University

March 16, 2020

(version 3)*

**Abstract**

Despite the empirical success of using *adversarial training* to defend deep learning models against adversarial perturbations, so far, it still remains rather unclear what the principles are behind the existence of adversarial perturbations, and what adversarial training does to the neural network to remove them.

In this paper, we present a principle that we call *feature purification*, where we show one of the causes of the existence of adversarial examples is the accumulation of certain small *dense mixtures* in the hidden weights during the training process of a neural network; and more importantly, one of the goals of adversarial training is to remove such mixtures to *purify* hidden weights. We present both experiments on the CIFAR-10 dataset to illustrate this principle, and a **theoretical result proving** that for certain natural classification tasks, training a two-layer neural network with ReLU activation using randomly initialized gradient descent *indeed* satisfies this principle.

Technically, we give, to the best of our knowledge, the first result *proving* that the following two can hold simultaneously for training a neural network with ReLU activation. (1) Training over the original data is indeed non-robust to small adversarial perturbations of some radius. (2) Adversarial training, even with an *empirical* perturbation algorithm such as FGM, can in fact be *provably* robust against *any* perturbations of the same radius. Finally, we also prove a complexity lower bound, showing that low complexity models such as linear classifiers, low-degree polynomials, or even the neural tangent kernel for this network, *cannot* defend against perturbations of this same radius, no matter what algorithms are used to train them.

---

*V1 of this paper was presented at IAS on this date, and can be viewed online: https://video.ias.edu/csdm/2020/0316-YuanzhiLi. We polished writing and experiments in V1.5, V2 and V3.

We would like to thank Sanjeev Arora and Hadi Salman for many useful feedbacks and discussions.

# 1 Introduction

Large scale neural networks have shown great power to learn from a training data set, and generalize to unseen data sampled from similar distributions for applications across different domains [42, 46, 53, 87]. However, recent study has discovered that these trained large models are extremely vulnerable to small "adversarial attacks" [19, 95]. It has been discovered that small perturbations to the input– often small enough to be invisible to humans– can create numerous errors in prediction. Such slightly perturbed inputs are often referred to as "adversarial examples".

Since the original discovery of "adversarial examples", a large body of works have been done emphasizing how to improve the robustness of the deep learning models against such perturbations [44, 63, 64, 83, 89]. One seminal approach is called *adversarial training* [65], where one iteratively computes adversarial examples from the training examples, and then retrain the model with these adversarial examples instead of the original examples (a.k.a. the *clean* examples). This approach was reported in [15] as the only approach that can defend against carefully designed adversarial attacks, and many follow-up works are built upon it [82, 111].

However, despite the great empirical success on improving the robustness of neural networks over various data sets, the theory of the adversarial examples is much less developed. In particular, we found that the following fundamental questions remain largely unaddressed:

> **Questions**
>
> *Why do adversarial examples exist when we train the neural networks using the original training data set? How can adversarial training further "robustify" the trained neural networks against these adversarial attacks?*

To answer these questions, one sequence of theoretical works try to explain the existence of adversarial examples using the high dimensional nature of the input space and the over-fitting behavior due to the sample size and sample noise [32, 33, 39, 40, 67, 86, 96], and treat adversarial training from the broader view of min-max optimization [24, 65, 92, 93, 103]. However, recent observations [48] indicate that these adversarial examples can also, and arguably often, arise from *features* (those that do generalize) rather *bugs* (those that do not generalize due to effect of poor statistical concentration). To the best of our knowledge, all existing works studying adversarial examples either (1) apply generally to the case of arbitrarily unstructured functions $f$ and only consider adversarial examples statistically, or (2) apply to a structured setting but only involving linear learners. These theoretical works, while shedding great lights to the study of adversarial examples, do not yet give concrete mathematical answers to the following questions regarding the specific hidden-layer structure of **neural networks**:

1. What are the *features* (i.e. the hidden weights) learned by the neural network via *clean training* (i.e., over the original data set)? Why are those features "non-robust"?

2. What are the *differences* between the features learned by clean training vs *adversarial training* (i.e., over a perturbed data set consisting of adversarial examples)?

3. Why do adversarial examples for a network transfer to other independently-trained networks?

Before going into the above questions regarding robustness, it is inevitable to first study *what the features are when learned by a neural network during clean training*. Theoretical studies are *also* limited in this direction. Most of existing works (1) only focus on the case when the training data is spherical Gaussian [16, 20, 22, 36, 52, 55, 58, 60, 62, 88, 90, 97, 100, 104, 112, 114], and some of them require heavy initialization using tensor decomposition, which might fail to capture the specific structure of the input and the property of a random initialization; or (2) only consider the neural

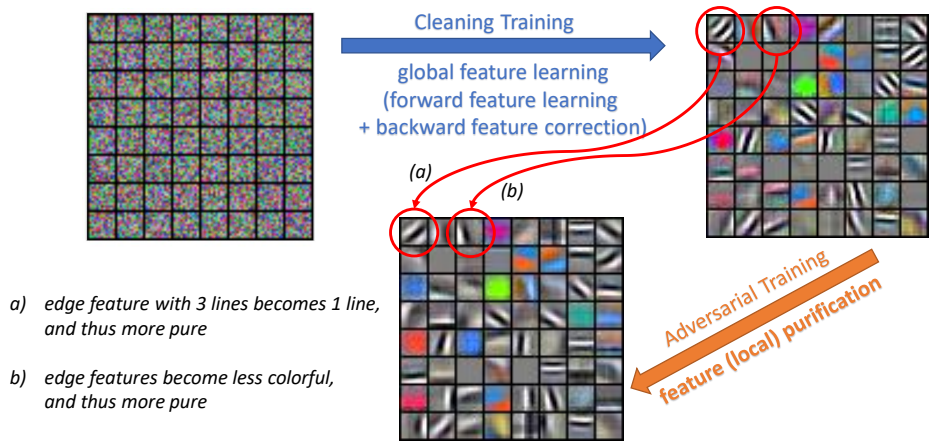

a) edge feature with 3 lines becomes 1 line, and thus more pure

b) edge features become less colorful, and thus more pure

Figure 1: Feature purification in adversarial training (for the first layer of AlexNet on CIFAR-10). Visualization of deeper layers of AlexNet as well as ResNet-34 can be found in Figure 6, Figure 10, and Figure 11.

tangent kernel regime, where the neural networks are linearized so the features are not learned (they stay at random initialization) [4, 6–8, 13, 14, 26, 26–29, 38, 45, 50, 57, 61, 105, 115, 116].

In this paper, we present a new routine that enables us to formally study the **learned features** (i.e. the *hidden weights*) of a neural network, when the inputs are more naturally structured than being Gaussians. Using this routine, we give, to the best of our knowledge, the *first theoretical result* towards answering the aforementioned fundamental questions of adversarial examples, for certain neural networks with ReLU activation functions.

**Our results.** We prove, for certain binary classification data set, when we train a *two-layer ReLU neural network* using *gradient descent*,[1] starting from random initialization,

1. Given polynomially manly training examples, in polynomially many iterations, the neural network will learn *well-generalizing features for the original data set*, and the learned network will have close-to-perfect prediction accuracy for the test data sampled from the same distribution.

2. However, even with a weight-decay regularizer to avoid over-fitting, even with *infinitely many training data*, and even when *super-polynomially many iterations* are used to train the neural network to convergence, the learned network still has near-zero robust accuracy against small-norm adversarial perturbations to the data. In other words, those *provably well-generalizing features* on the original data set are also *provably non-robust* to adversarial perturbations to the datas, so they cannot be due to having too few training samples [32, 33, 39, 40, 67, 86, 96].

3. Adversarial training, using perturbation algorithms such as Fast Gradient Method (FGM) [40], can *provably and efficiently* make the learned neural network achieve near-perfect robust accuracy, against even the worst-case norm-bounded adversarial perturbations, using a principle we refer to as "feature purification". We illustrate "feature purification" in Figure 1 by an experiment, and explain it in mathematical terms next.

**Feature purification: How adversarial training can perform robust deep learning.** In this work, we also give precise, mathematical characterizations on the *difference* between learned features by clean training versus adversarial training in the aforementioned setting, leading to (to our best knowledge) the first theory of *how, in certain learning tasks using ReLU neural networks, the provably non-robust features after clean training can be "robustified" via adversarial training.*

---

[1]Our theory extends to stochastic gradient descent (SGD) at the expense of complicating notations.

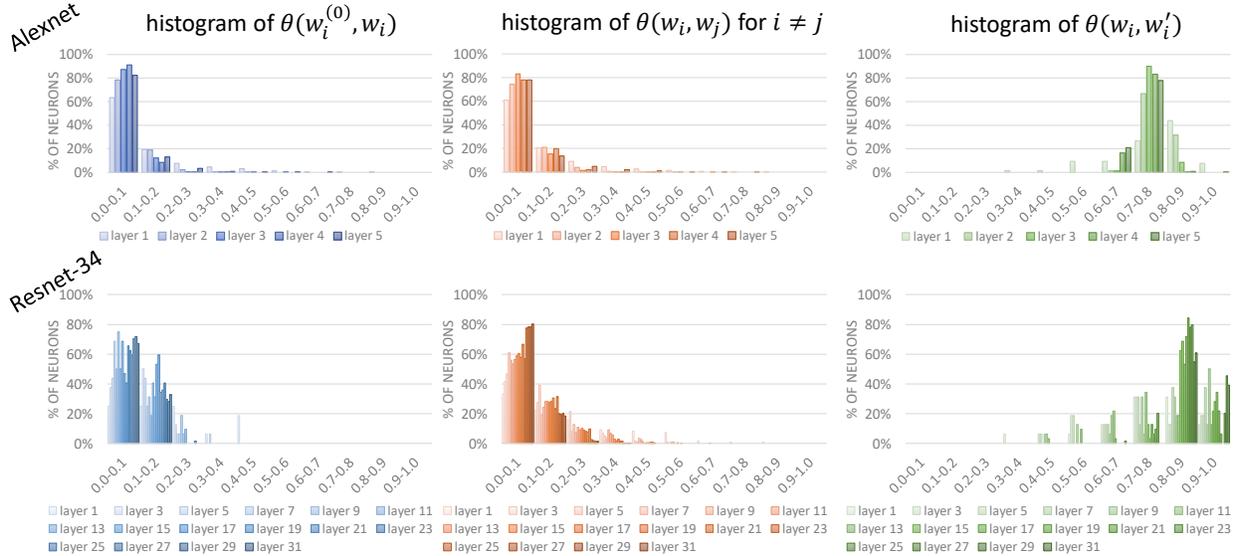

Figure 2: Measure of feature (local) purifications, for $\ell_2(1, 0.5)$ adversarial training on AlexNet and ResNet-34. For ResNet-34, the weights $\{w_i\}$ define networks with clean accuracy $> 80\%$ and robust accuracy $0\%$; while the weights $\{w_i'\}$ define networks with clean accuracy $> 65\%$ and robust accuracy $> 42\%$. Implementation details in Section 8.2.

We emphasize that prior theoretical works [35, 51, 84, 98, 113] mainly study adversarial examples in the context of *linear* models (such as linear regression, linear regression over prescribed feature mappings, or the neural tangent kernels). In those models, the features are *not* trained, so adversarial training only changes the weights associated with the linear combination of these features, but not the actual features themselves.

In contrast, this paper develops a theory showing that how, over certain learning tasks, adversarial training *can actually change the features* of certain neural networks to improve their robustness. We abstract this feature change in our setting into a general principle that we call *feature purification*, which although we only prove it for two-layer ReLU networks (see Theorem 5.1a+Theorem 5.3c), we *empirically observe that it occurs more generally to real-world, deep neural networks on real-world data sets*. We sketch its high-level idea as follows.

---
**The Principle of Feature Purification**

During adversarial training, the neural network will *neither* learn new, robust features *nor* remove existing, non-robust features learned over the original data set. Most of the works of adversarial training is done by *purifying a small part of each learned feature after clean training*.

---

Mathematically, as a *provisional* step to measure of change of features in a network, let us use (1) $w_i^{(0)}$ to denote the weight vector of the $i$-th neuron at initialization, (2) $w_i$ to denote its weight after clean training, and (3) $w_i'$ to denote its weight after adversarial training (using $w_i$ as initialization). The "feature purification" principle, in math, says if we use $\theta(z, z') := \frac{|\langle z, z'\rangle|}{\|z\|_2 \|z'\|_2}$ as a *provisional* measure of the correlation between "features", then (see Figure 2 for real-life experiments):

1. for most neurons: $\theta(w_i^{(0)}, w_i), \theta(w_i^{(0)}, w_i') \leq c$ for a small constant $c$ (such as 0.2);

2. for most neurons: $\theta(w_i, w_i') \geq C$ for a large constant $C$ (such as 0.8); and

3. for most pairs of different neurons: $\theta(w_i, w_j) \leq c$ for a small constant $c$ (such as 0.2).

In words, this says both clean training and adversarial training discover hidden weights $w_i, w_i'$ that are fundamentally different from initialization $w_i^{(0)}$. However, since $w_i$ and $w_i'$ are close, clean training must have already discovered a *big portion* of the robust features, and adversarial training merely needs to "purify" some small part of each original feature. In this paper:

- we prove this feature purification principle in the case of two-layer ReLU neural networks over certain data sets, with $c = o(1)$ and $C = 1 - o(1)$ (see Theorem 5.1a+Theorem 5.3c); and
- we provide empirical evidence that this feature purification principle holds also for deep neural networks used in real-life datasets (see Figure 2 as well as other experiments in the paper).

**Why clean training learns non-robust features? Which part of the features are "purified" during adversarial training?** In our setting, we also give mathematical characterizations of where the "non-robust" part of each feature comes from during clean training.

As we shall formally discuss in Section 6.2, training algorithms such as gradient descent will, at every step, add to the current parameters a direction that *maximally correlates with the labeling function on average*. For two-layer ReLU networks, we prove that such simple correlations will accumulate, in each neuron, a small part of its weight that correlates with the average of the training data, and we refer to it as the *dense mixture* (see Theorem 5.2). However, under natural assumptions of the data such as the *sparse coding model* — namely inputs come from sparse combinations of hidden dictionary words/vectors— such dense mixtures cannot have high correlation with any individual, clean example. Thus, even with these "dense mixtures" in the features, the network can still generalize well on the original data set. However, we show that these portions of the features are extremely vulnerable to small, adversarial perturbations along the "dense mixture" directions. As a result, one of the main goals of adversarial training, as we show, is to *purify* the neurons by removing such dense mixtures. This is the supporting theory behind our *feature purification* principle, as we also measure it empirically in the experiment section.

We believe our result extends the reach of traditional learning theory, where often statistical properties of the model (such as generalization, etc.) is separate from optimization (i.e., how the models are trained). *However, to understand adversarial examples in deep learning, one needs to admit that well-generalizing and adversarially robust neural networks do exist (and can even be found efficiently using adversarial training), thus it is also a global optimal solution of the clean training objective. It is rather a property of traditional clean training process using SGD which biases the network towards learning non-robust networks as another global optimal solution of the training objective.*

Moreover, in our setting, these dense mixtures in the hidden weights of the network come from the sparse coding structure of the data and the gradient descent algorithm. It is rather *independent* of the random initialization of the neural network. Thus, we prove that, at least in our scenario, **adversarial examples for one network do transfer to other independently trained ones**.

**Our contribution to computation complexity.** We also prove a lower bound that, for the same sparse coding data model, even when the original data is linearly-separable, any linear classifier, any low-degree polynomial, or even the corresponding neural tangent kernel (NTK) of our studied two-layer neural network, *cannot* achieve meaningful robust accuracy (although they can easily achieve high clean accuracy). Together with our upper bound, we have shown that using a *higher-complexity model* (such as a two-layer neural network with ReLU activation, comparing to NTK) can in fact achieve *better* robustness against adversarial perturbations. Thus, our theory *strongly supports* the experimental finding in [40, 65], where experts have noticed that robustness against adversarial examples requires a model with higher complexity. The main intuition is that low-complexity models, including the neural tangent kernel, lacks the power to *zero out* low magnitude signals to improve model robustness, as illustrated in Figure 3 and Section 3.

**Our experimental contributions.** We present quite a few experimental results supporting our theory. (1) Our sparse coding model can indeed capture real-world data to certain degree. (2) Our principle of feature purification also holds for architectures such as AlexNet and ResNet. (3) Adversarial training using adversarial examples indeed purify some "dense mixtures" in practice. (4) During clean training, *how* the features can emerge from random initialization by *wining the "lottery tickets"*, as predicted by our theory. We present our experiments following each of the theorem statements accordingly. We also include a whole Section 8 for more detailed experiments.

## 1.1 Related Works

**Adversarial examples: Empirical study.** Since the seminal paper [95] shows the existence of small adversarial perturbations to change the prediction of the neural networks, many empirical studies have been done to make the trained neural networks robust against perturbations [44, 63, 64, 83, 89] (and we refer to the citations therein). The recent study [15] shows that the seminal approach [65] of adversarial training is the most effective way to make the neural networks robust against adversarial perturbations.

**Adversarial examples: Theoretical study.** Existing theories mostly explain the existence of adversarial examples as the result of finite-sample data set over-fitting to high-dimensional learning problems [32, 33, 39, 40, 67, 86, 96]. Later, it is discovered by [48] that well-generalizing features can also be non-robust. Other theories focus on the Fourier perspective of the robustness [102, 108], showing that adversarial training might be preventing the network from learning the high frequency signals of the input image. Our theoretical work is fundamentally different from the aspect of poor statistical concentration over finite-sample data set, and our Theorem 5.1 and Theorem 5.3 **strongly supports** [48] that a well-trained, well-generalizing neural network can still be non-robust to adversarial attacks.

Other theories about adversarial examples focus on how adversarial training might require more training data comparing to clean training [84], and might decrease clean training accuracy [80, 98]. The works by [35, 113] focus on how adversarial training can be performed efficiently in the neural tangent kernel regime. The purpose of these results are also fundamentally different than ours.

**Sparse coding (data) model.** We use a data model called sparse coding, which is a popular model to model image, text and speech data [11, 12, 69, 77, 78, 101, 106, 107]. There are many existing theoretical works studying algorithm for sparse coding [9, 17, 43, 47, 54, 68, 85, 91, 94], however, these algorithms share little similarity to training a *neural network.*

The seminal work by [10] provides a neurally-plausible algorithm for learning sparse coding along with other works using alternative minimization [1, 2, 37, 56, 59]. However, all of these results require a (carefully picking) warm start, while our theory is for training a neural network starting from *random initialization.*

**Threshold degree and kernel lower bound.** We also provide, to the best of our knowledge, the first example when the original *classification* problem is learnable using a linear classifier but *no low-degree polynomial* can learn the problem *robustly* against small adversarial perturbations. Yet, the high-complexity neural networks can provably, efficiently and robustly learn the concept class. The lower bound for the classification accuracy using low-degree polynomials has been widely studied as the (approximate) *threshold degree* of a function or the sign-rank of a matrix [18, 21, 25, 41, 75, 81]. Our paper give the first example of a function with high (approximate) robust threshold degree, yet efficiently and robustly learnable by training a ReLU neural network using gradient descent.

Other related works prove lower bounds for kernel method in the regression case [3, 5]. Generally speaking, such lower bounds are about the actual (approximate) degree of the function, instead of the (approximate) threshold degree. It is well know that for general functions, the the actual degree can be arbitrary larger than the threshold degree.

## 2 Preliminaries

We use $\|x\|$ or $\|x\|_2$ to denote $\ell_2$ norm of a vector $x$, and $\|x\|_p$ to denote the $\ell_p$. For a matrix $\mathbf{M} \in \mathbb{R}^{d \times d}$, we use $\mathbf{M}_i$ to denote the $i$-th column of $\mathbf{M}$, and we use $\|\mathbf{M}\|_\infty$ to denote $\max_{i \in [d]} \sum_{j \in [d]} \mathbf{M}_{i,j}$ and $\|\mathbf{M}\|_1$ to denote $\max_{j \in [d]} \sum_{i \in [d]} \mathbf{M}_{i,j}$. We use $\mathsf{poly}(d)$ to denote $\Theta(d^C)$ when the degree $C$ is some not-specified constant. We use the term clean training to refer to the neural network found by training over the original data set, and the term robust training to refer to the neural network found by adversarial training. We let $\mathsf{sign}(x) = 1$ for $x \geq 0$ and $\mathsf{sign}(x) = -1$ for $x < 0$.

**Sparse coding model.** In this paper, we consider the training data $x \in \mathbb{R}^d$ generated from

$$x = \mathbf{M}z + \xi$$

for a dictionary $\mathbf{M} \in \mathbb{R}^{d \times D}$, where the *hidden vector* $z \in \mathbb{R}^D$ and $\xi \in \mathbb{R}^d$ is the noise. For simplicity, we focus on $D = d$ and $\mathbf{M}$ is a unitary matrix. Although our results extend trivially to the case of $D < d$ or when $\mathbf{M}$ is incoherent, we point out that the orthogonal setting is more principle: As we will argue in one of our main result, the clean training learns a non-robust "dense mixture" of features (See Eq 5.1). In our orthogonal setting, the accumulation of such mixture is clearly NOT due to the correlation between features, rather its an intrinsic property of gradient descent.

We assume the hidden vector $z$ is "sparse", in the following sense: for $k \leq d^{0.499}$, we have:

**Assumption 2.1** (distribution of hidden vector $z$). *The coordinates of $z$ are independent, symmetric random variables, such that $|z_i| \in \{0\} \cup [\frac{1}{\sqrt{k}}, 1]$. Moreover,*

$$\mathbb{E}[z_i^2] = \Theta\left(\frac{1}{d}\right), \quad \mathbf{Pr}[|z_i| = 1] = \Omega\left(\frac{1}{d}\right), \quad \mathbf{Pr}\left[|z_i| = \Theta\left(\frac{1}{\sqrt{k}}\right)\right] = \Omega\left(\frac{k}{d}\right)$$

The first condition is a regularity condition, which says that $\mathbb{E}[\|z\|_2^2] = \Theta(1)$. The second and third condition says that there is a non-trivial probability where $z_i$ attains the maximum value, and a (much) larger probability that $z$ is non-zero but has a small value (Remark: It could be the case that $z_i$ is neither maximum nor too small, for example, $|z_i|$ can also be $k^{-0.314}$ with probability $\frac{k^{0.628}}{d}$ as well, or $k^{-0.123}$ with probability $\frac{k^{0.0888}}{d}$). The main observation is that

**Fact 2.2.** *Under Assumption 2.1, w.h.p., $\|z\|_0 = \Theta(k)$ is a sparse vector.*

We study the simplest binary-classification problem, where the labeling function is linear over the hidden vector $z$:

$$y(x) = \mathsf{sign}\left(\langle w^\star, z \rangle\right)$$

For simplicity, we assume $\forall i \in [D], |w_i^\star| = \Theta(1)$, so all the coordinates of $z$ have relatively equal contributions. Our theorems extend to other $w^\star$ at the expense of complicating notations.

**Remark on sparse coding.** The sparse coding model is very natural and is widely used to model image, text and speech data [69, 77, 78, 101, 106, 107]. There certainly exist (provable) algorithms for dictionary learning based on sum of squares, linear programming [17, 91], but they *do not shed light* on the training process of neural networks. Even the neural algorithm for sparse coding [10] is still far away from training a neural network using SGD or its variants. **The main point of this paper is not to show neural networks can do sparse coding**. Instead, our main point is to distinguish the adversarial training and the clean training processes of neural networks using the sparse-coding model as a bridging tool.

**Noise model.** We have allowed the inputs $x = \mathbf{M}z + \xi$ to incorporate a noise vector $\xi$. Our lower bounds hold *even when* there is no noise ($\xi = 0$). Our upper bound theorems not only apply to $\xi = 0$, but *more generally* to "gaussian noise plus spike noise":

$$\xi = \xi' + \mathbf{M}\xi''$$

Here, the gaussian noise $\xi' \sim \mathcal{N}(0, \frac{\sigma_x^2}{d}\mathbf{I})$ where $\sigma_x \leq O(1)$ can be an arbitrary large constant. The spike noise $\xi''$ is any coordinate-wise independent, mean-zero random variable satisfying $\mathbb{E}[\xi_i''^2] \leq O(\frac{\sigma_x^2}{d})$ and $|\xi_i''| \leq \frac{1}{k^{0.501}}$ for every $i \in [d]$. Therefore, our upper bound theorems hold even under the following extreme circumstances:

- the noise $\xi$ can be of Euclidean norm $O(\sigma_x)$, larger than the signal $\|\mathbf{M}z\|_2 \approx 1$; and
- the spike noise $\xi_i''$ can be $\frac{1}{k^{0.501}}$ which is the maximum possible (because $z_i$ can be $\frac{1}{k^{0.5}}$).

We point out that there are *no dependencies* among the constants in those $O, \Theta$ and $\Omega$ notations of this section, except for the obvious ones (e.g. $\mathbf{Pr}[|z_i| = 1] \leq \mathbb{E}[|z_i|^2]$). In particular, $\sigma_x$ can be an arbitrarily large constant and $\mathbf{Pr}[|z_i| = 1]$ can be an arbitrary small constant times $1/d$.[2]

**Clean and robust error.** The goal of clean training is to learn a model $f$ so that $\mathsf{sign}(f(x))$ is as close to $y$ as possible. We define the classification error on the original data set as:

$$\text{clean error:} \quad \mathcal{E}^c(f) \stackrel{\text{def}}{=} \mathbf{Pr}_{x, y = y(x)}[\mathsf{sign}(f(x)) \neq y]$$

Next, we consider robust error against $\ell_p$ adversarial perturbations. For a value $\tau > 0$ and a norm $\|\cdot\|_p$, we define the robust error of the model $f$ (against $\ell_p$ perturbation of radius $\tau$) as:

$$\text{robust error:} \quad \mathcal{E}^r(f) = \mathbf{Pr}_{x, y = y(x)}[\exists \delta : \|\delta\|_p \leq \tau : \mathsf{sign}(f(x + \delta)) \neq y]$$

## 3 Warmup Intuitions

**Linear learners are not robust.** Given the setting of the data set, one direct approach is to use (the sign of) a linear classifier $f(x) = \langle w^\star, \mathbf{M}^\top x \rangle$ to predict the label of $x$. There are two issues of using such a classifier:

1. When $\sigma_x$ is as large as $\Theta(1)$, such classifier can not even classify $x$ in good *clean* accuracy. Recall $f(x) = \langle w^\star, \mathbf{M}^\top x \rangle = \langle w^\star, z \rangle + \langle \mathbf{M}w^\star, \xi \rangle$. By our assumption, typically $|\langle w^\star, z \rangle| = O(1)$ and $\langle \mathbf{M}w^\star, \xi' \rangle \sim \mathcal{N}(0, \Theta(\sigma_x^2))$. Thus, when $\sigma_x \geq \Theta(1)$, noise could be much larger than signal, and this linear classifier cannot be used to classify $x$ correctly. In this case, actually no linear classifier (or even constant-degree polynomials[3]) can give meaningful clean accuracy.

2. Even when $\sigma_x = 0$ so the original data is perfectly linearly-classifiable, linear classifier is also not robust to small perturbations. Since typically $|\langle w^\star, z \rangle| = O(1)$, one can design an adversarial perturbation $\delta = \frac{-Cy\mathbf{M}w^\star}{\|w^\star\|_2^2}$ for a large constant $C$, that can change the sign of the linear classifier $f(x) = \langle w^\star, \mathbf{M}^\top x \rangle$ for most inputs. Since $\|w^\star\|_2 = \Theta(\sqrt{d})$, this linear classifier

---

[2] Actually, our theorem extends trivially to the case even when $\mathbf{Pr}[|z_i| = 1] = \frac{1}{d^{1+o(1)}}$.

[3] One may think that using for example degree-3 polynomial $\sum_i w_i^\star \langle \mathbf{M}_i, x \rangle^3$ can reduce the level of noise, but due to the diversity in the value of $z_i$ when $z_i \neq 0$, one must use something close to linear when $|z_i|$ is large. Applying Markov brothers' inequality, one can show the low-degree polynomial must be close to a linear function.

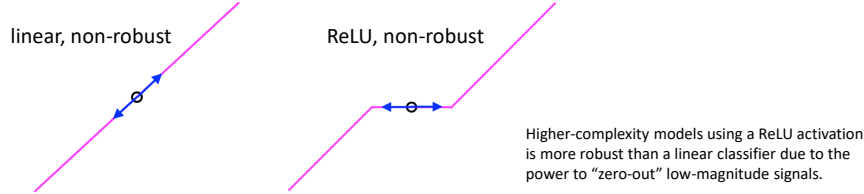

Figure 3: linear vs ReLU activation (our theorem is in symmetric ReLU for the sake of proof simplicity.)

*is not even robust* to adversarial perturbations of $\ell_2$ norm $\Theta\left(\frac{1}{\sqrt{d}}\right)$. In fact, no linear classifier can be robust to such small adversarial perturbations.

**High-complexity models are more robust.** Another choice to learn the labeling function is to use a higher-complexity model $f(x) = \sum_{i \in [d]} w_i^\star \langle \mathbf{M}_i, x \rangle \mathbb{1}_{|\langle \mathbf{M}_i, x \rangle| \geq \frac{1}{2\sqrt{k}}}$. Here, the "complexity" of $f$ is much higher because an indicator function is used.[4] Since $\langle \mathbf{M}_i, x \rangle = z_i + \langle \mathbf{M}_i, \xi \rangle$, by our noise model, as long as the signal $z_i \neq 0$ is non-zero, $|\langle \mathbf{M}_i, x \rangle| \geq \frac{1}{2\sqrt{k}}$ with high probability. Thus, this $f(x)$ is *equal to* the true labeling function $\langle w^\star, z \rangle$ w.h.p. over the original data set, so is (much) more *robust to noise* comparing to linear models.

Moreover, this $f$ is *also more robust to $\ell_2$ adversarial perturbations*. By Fact 2.2, w.h.p. the signal $z$ is $O(k)$-sparse, and thus there are at most $O(k)$ many coordinates $i \in [d]$ with $\mathbb{1}_{|\langle \mathbf{M}_i, x \rangle| \geq \frac{1}{2\sqrt{k}}} = 1$. Using this, one can derive that this high complexity model $f(x)$ has $1 - o(1)$ robust accuracy, against *any* adversarial perturbation of $\ell_2$ radius $o\left(\frac{1}{\sqrt{k}}\right)$. This is much larger than that of $O\left(\frac{1}{\sqrt{d}}\right)$ for a linear classifier, and it is actually information theoretically optimal.

To sum up, higher-complexity models (such as those using ReLU) have the power to *zero out* low-magnitude signals to improve adversarial robustness, as illustrated in Figure 3.

**Learning robust classifier using neural network.** Motivated by the above discussions between linear vs. high-complexity models, our goal is to show that a two-layer neural networks can (after adversarial training) learn a robust function $f(x)$ such as

$$f(x) \approx \sum_{i \in [d]} w_i^\star \left[ \mathsf{ReLU}(\langle \mathbf{M}_i, x \rangle - b) - \mathsf{ReLU}(-\langle \mathbf{M}_i, x \rangle - b) \right]$$

Here, $\mathsf{ReLU}(y) = \max\{y, 0\}$ is the ReLU function and $b \approx \frac{1}{2\sqrt{k}}$. In this paper, we present a theorem stating that adversarial training of a (wlog. symmetric) two-layer neural network can *indeed* recover a neural network of this form. In other words, after adversarial training, the features learned by the hidden layer of a neural network can indeed form a *basis* (namely, $\mathbf{M}_1, \ldots, \mathbf{M}_d$) of the input $x$ where the coefficients are *sparse*. We also present a theorem showing why, clean training **will not** learn this robust function. We also verify experimentally that the features learned by the first layer of AlexNet (after adversarial training) indeed form a sparse basis of the images, see Figure 4.

## 4 Learner Network and Adversarial Training

In this paper we consider a simple, two layer (symmetric)[5] neural network with ReLU activation.

$$f(x) = \sum_{i=1}^{m} a_i \left[ \mathsf{ReLU}(\langle w_i, x \rangle - b_i + \rho_i) - \mathsf{ReLU}(-\langle w_i, x \rangle - b_i + \rho_i) \right]$$

---

[4]One concrete measure of "higher complexity" is that $f$ cannot be well-approximated by low degree polynomial.
[5]We assume the neurons are symmetric (i.e., with $(w_i, -w_i)$ pairs) to simplify proofs.

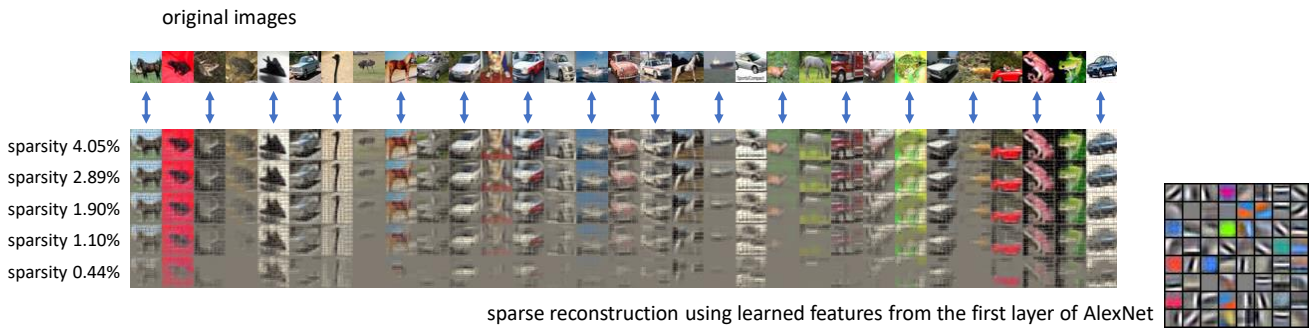

Figure 4: Reconstruct the original image using *sparse* linear combinations of the AlexNet's features (adversarially trained). The average sparsity is only 4.05% or less. More experiments in Section 8.3.

In this way we have $f(x) = -f(-x)$. We refer to $w_i \in \mathbb{R}^d$ as the hidden weight (or feature) of the $i$-th neuron, and $b_i > 0$ as the (negative) bias. Each $\rho_i \sim \mathcal{N}(0, \sigma_\rho^2)$ is a *smoothing* of the original ReLU, also known as the pre-activation noise. Equivalently, one can use the smoothed ReLU activation $\widetilde{\mathsf{ReLU}}(x) = \mathbb{E}_\rho \mathsf{ReLU}(x+\rho)$. In our result, $\sigma_\rho$ is always smaller than $b_i$ and much smaller than the typical value of $\langle w_i, x \rangle$. The main role of the pre-activation noise is simply to make the gradient of ReLU smooth: it simplifies our analysis for the sample complexity derivation (see Lemma A.2). In this paper, unless specially specified, we will use $\rho$ to denote $(\rho_i)_{i \in [m]}$.

To simplify analysis, we fix $a_i = 1$ throughout the training. We use $w_i^{(t)}$ to denote the hidden weights at time $t$, and use $f_t(w; x, \rho)$ to denote the network at iteration $t$

$$f_t(w; x, \rho) = \sum_{i=1}^{m} \left( \mathsf{ReLU}(\langle w_i^{(t)}, x \rangle + \rho_i - b_i^{(t)}) - \mathsf{ReLU}(-\langle w_i^{(t)}, x \rangle + \rho_i - b_i^{(t)}) \right)$$

Given a training set $\mathcal{Z} = \{x_j, y_j\}_{j \in [N]}$ together with one sample of pre-activation noise $\rho^{(j)}$ for each $(x_j, y_j)$, we define

$$\mathbf{Loss}_t(w; x, y, \rho) \stackrel{\text{def}}{=} \log(1 + e^{-y f_t(w; x, \rho)})$$

$$\mathbf{Loss}_t(w) \stackrel{\text{def}}{=} \mathbb{E}_{x, y=y(x), \rho}[\mathbf{Loss}_t(w; x, y, \rho)] \qquad \widetilde{\mathbf{Loss}}_t(w) \stackrel{\text{def}}{=} \frac{1}{N} \sum_{j \in [N]} [\mathbf{Loss}_t(w; x_j, y_j, \rho^{(j)})]$$

$$\mathbf{Obj}_t(w) \stackrel{\text{def}}{=} \mathbf{Loss}_t(w) + \lambda \sum_{i \in [m]} \mathbf{Reg}(w_i) \qquad \widetilde{\mathbf{Obj}}_t(w) \stackrel{\text{def}}{=} \widetilde{\mathbf{Loss}}_t(w) + \lambda \sum_{i \in [m]} \mathbf{Reg}(w_i)$$

Above, $\mathbf{Loss}_t(w; x, y, \rho)$ is the standard logistic loss, $\mathbf{Loss}_t(w)$ is the population risk and $\widetilde{\mathbf{Loss}}_t(w)$ is the empirical risk. We consider a strong, but *quite natural* regularizer to further avoid overfitting, given as $\mathbf{Reg}(w_i) \stackrel{\text{def}}{=} \left( \frac{\|w_i\|_2^2}{2} + \frac{\|w_i\|_2^3}{3} \right)$. Here, $\frac{\|w_i\|_2^2}{2}$ is known as *weight decay* in practice; the additional $\frac{\|w_i\|_2^3}{3}$ is an analog of weight decay combined with batch normalization [49].[6] We consider a fixed $\lambda = \frac{\log \log \log d}{d}$ for simplicity,[7] although our result trivially extends to other values of $\lambda$.

**Definition 4.1.** *In our case, the (clean, population) classification error at iteration $t$ is*

$$\mathcal{E}_t^c \stackrel{\text{def}}{=} \Pr_{x, y=y(x), \rho}[y \neq \mathsf{sign}(f_t(w^{(t)}; x, \rho))] \ . \tag{clean error}$$

---

[6]Indeed, for a function $f(\frac{w}{\|w\|_2}) + \frac{\lambda}{2}\|w\|_2$ over normalized $w$, its gradient with respect to $w$ is proportional to $(\mathbf{I} - \frac{ww^\top}{\|w\|_2^2})\nabla f(\frac{w}{\|w\|_2}) + \lambda w \cdot \|w\|_2$. Here, $\lambda w \cdot \|w\|_2$ can be viewed as the gradient of $\frac{\lambda}{3}\|w\|_2^3$.

[7]Throughout this paper, the purpose of any log log log $d$ factor  is to cancel out arbitrarily large constants so that we can present theorems and lemmas with simpler notations.

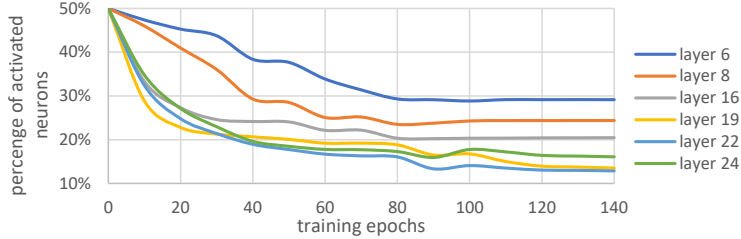

Figure 5: Neuron activation sparsity on clean training of WRN-28-10 over the CIFAR-10 dataset

We also introduce a notation

$$\ell'_t(w^{(t)}; x, y, \rho) \stackrel{\text{def}}{=} \frac{d}{ds}[\log(1+e^s)] \mid_{s=-yf_t(w^{(t)};x,\rho)}$$

and observe $\mathbb{E}_{x,y=y(x),\rho}[\ell'_t(w^{(t)}; x, y, \rho)] \geq \Omega(\mathcal{E}^c_t)$.

## 4.1 Clean Training

We consider *clean training* as the gradient descent algorithm with step length $\eta > 0$ on the hidden weights $w_1, \ldots, w_m$ over the original data set, see Algorithm 1. Our result extends to stochastic gradient descent at the expense of complicating notations. For simplicity, we assume the bias terms $b^{(t)}_1 = \cdots = b^{(t)}_m = b^{(t)}$ grow together.[8]

At initialization, we let $w^{(0)}_i \sim \mathcal{N}(0, \sigma_0^2 \mathbf{I})$ for $\sigma_0 = \frac{1}{\text{poly}(d)}$ and let $b^{(0)} = \Theta(\sigma_0 \sqrt{\log d})$. When near initialization, we manually increase the bias $b^{(t+1)} = b^{(t)} + \eta \mathfrak{B}$ where $\mathfrak{B} = \frac{c_b}{d}$ for some small constant $c_b > 0$— this corresponds to the "lottery ticket winning" phase to be discussed later in Section 6.1; and whenever $b^{(t)}$ reaches $\frac{1}{k^{0.5001}}$ we set $\mathfrak{B} = 0$— in this phase, the neurons that have won the "lottery ticket" will keep winning and grow significantly, to be discussed in Section 6.2.

We also choose pre-activation noise $\sigma^{(t)}_\rho = \frac{b^{(t)}}{\sqrt{\log d}} \cdot \Theta((\log \log \log d)^3)$ for $t \leq T_\mathsf{a} = \frac{1}{\text{poly}(d)\eta}$, and $\sigma^{(t)}_\rho = \frac{b^{(t)}}{\log d} \cdot \Theta((\log \log \log d)^3)$ for $t > T_\mathsf{a}$. The explicit choices of $\mathfrak{B}$ and $T_\mathsf{a}$ are given in the proofs.

## 4.2 Adversarial Training

We state the adversarial training algorithm in Algorithm 2. It takes as input a perturbation algorithm $A$, and repeatedly applies gradient descent over a perturbed data set (that comes from the original data set plus the perturbation given by $A$). Formally,

**Definition 4.2.** *An (adversarial) perturbation algorithm $A$ (a.k.a. attacker) maps the current network $f$ (which includes hidden weights $\{w_i\}$, output weights $\{a_i\}$, bias $\{b_i\}$ and smoothing parameter*

---

[8] We make several remarks about the bias growth.

- Our analysis does extend to the case of trainable bias $b_i$'s, when the spike noise is large (e.g. $\mathbb{E}[(\xi''_i)^2] \geq \Omega(1/d)$): in this case, by applying gradient descent on the bias, it will automatically grow until it is large enough to to de-noise. This shall significantly complicate the proofs, so we do not include it here.
- We grow bias to increase activation sparsity as training goes. This is very natural and simulates the actual training process in practice (see Figure 5).
- Alternatively, one can interpret our result as, in the practical sparse coding setting, even if the biases are well-tuned to simulate dictionary learning, clean learning over the weights $w_i$'s still accumulates still fails to learn the exact dictionary, rather it will accumulate dense mixtures in the neuron weights, leading to near-zero robust accuracy.

---

**Algorithm 1** clean training using gradient descent

1: begin with the randomly initialized $w^{(0)}$ and the starting bias $b^{(0)}$;
2: **for** $t \in \{0, 1, 2, \cdots, T_{\mathsf{f}} - 1\}$ **do**
3:     for each $(x_j, y_j) \in \mathcal{Z}$, sample pre-activation noise $\rho^{(j)}$ i.i.d. $\sim \mathcal{N}(0, (\sigma_\rho^{(t)})^2 \mathbf{I}_{m \times m})$.
4:     define empirical objective $\widetilde{\mathbf{Obj}}_t(w)$ at this iteration using $\{x_j, y_j, \rho^{(j)}\}_{j \in [N]}$.
5:     for each $i \in [m]$, update using gradient descent: $w_i^{(t+1)} \leftarrow w_i^{(t)} - \eta \nabla_{w_i} \widetilde{\mathbf{Obj}}_t(w^{(t)})$
6:     update $b^{(t+1)} \leftarrow b^{(t)} + \eta \mathfrak{B}$
7: **end for**

---

**Algorithm 2** adversarial training algorithm (against perturbation algorithm $A$)

1: begin with a network $f_{T_{\mathsf{f}}}$ learned through clean training in Algorithm 1.
2: **for** $t \in \{T_{\mathsf{f}}, T_{\mathsf{f}} + 1, \cdots, T_{\mathsf{f}} + T_{\mathsf{g}} - 1\}$ **do**
3:     for every $(x_j, y_j) \in \mathcal{Z}$, perturb $x_j^{(adv)} \leftarrow x_j + A(f_t, x_j, y_j, r_j)$.
4:     for each $(x_j, y_j) \in \mathcal{Z}$, sample pre-activation noise $\rho^{(j)}$ i.i.d. $\sim \mathcal{N}(0, (\sigma_\rho^{(t)})^2 \mathbf{I}_{m \times m})$.
5:     define empirical objective $\widetilde{\mathbf{Obj}}_t(w)$ at this iteration using $\{x_j^{(adv)}, y_j, \rho^{(j)}\}_{j \in [N]}$.
6:     for each $i \in [m]$, update using gradient descent: $w_i^{(t+1)} \leftarrow w_i^{(t)} - \eta \nabla_{w_i} \widetilde{\mathbf{Obj}}_t(w^{(t)})$
7: **end for**

---

$\sigma_\rho$), *an input $x$, a label $y$, and some internal random string $r$, to $\mathbb{R}^d$ satisfying*

$$\|A(f, x, y, r)\|_p \leq \tau \ .$$

*for some $\ell_p$ norm. We say $A$ is an $\ell_2$ perturbation algorithm of radius $\tau$ if $p = 2$, and $\ell_\infty$ perturbation algorithm of radius $\tau$ if $p = \infty$. For simplicity, we assume $A$ satisfies for fixed $f, y, r$, either $\|A(f, x, y, r)\|_p \leq \frac{1}{\mathsf{poly}(d)}$, or $A(f, x, y, r)$ is a $\mathsf{poly}(d)$-Lipschitz continuous function in $x$.*

One can verify that for our network, the fast gradient method (FGM) [40] satisfies the above properties. FGM is a widely used algorithm to find adversarial examples. In our language, FGM is simply given by:[9]

$$A(f, x, y) = \begin{cases} \arg\min_{\delta : \|\delta\|_p \leq \tau} \langle y \nabla_x f(x), \delta \rangle & \text{if } \|\nabla_x f(x)\|_q \geq \frac{1}{\mathsf{poly}(d)}; \\ 0 & \text{otherwise.} \end{cases}$$

**Definition 4.3.** *The robust error at iteration $t$, against arbitrary $\ell_p$ perturbation of radius $\tau$, is*

$$\mathcal{E}_t^r \stackrel{\text{def}}{=} \Pr_{x, y = y(x), \rho}[\exists \delta \in \mathbb{R}^d, \|\delta\|_p \leq \tau : \mathsf{sign}(f_t(x + \delta)) \neq y] \qquad \text{(robust error)}$$

*In contrast, the empirical robust classification error against algorithm $A$ is*

$$\widehat{\mathcal{E}_t^r} \stackrel{\text{def}}{=} \Pr_{x, y = y(x), \rho, r}[\mathsf{sign}(f_t(x + A(f_t, x, y, r))) \neq y] \qquad \text{(empirical robust error)}$$

Our upper bound theorems apply to all perturbation algorithms under Definition 4.2, and gives small empirical robust error $\widehat{\mathcal{E}^r}$. To obtain small (true) robust error $\mathcal{E}^r$, as we shall see, one can for instance let $A$ be the fast gradient method (FGM).

**Initialized from clean training.** In this paper, we assume adversarial training (i.e., Algorithm 2)

---

[9]Here, $\|\|_q$ is the dual norm of $\|\|_p$. In our case, due to the pre-activation noise, we define $\nabla_x f(x) = \nabla_x \mathbb{E}_\rho f(x; w, \rho)$. Also, we have zeroed out $A(f, x, y)$ when $\|\nabla_x f(x)\|_q$ is extremely small for the convenience of analysis, because otherwise $A(f, x, y)$ is not Lipscthiz continuous at those points.

is initialized from a network that is already clean-trained. In contrast, in practice, adversarial training usually begins directly with random initialization. We remark here that:

- First, in practice, adversarial training from a clean-trained initialization performs *no worse than* from a random initialization, see Table 1 on Page 21. In fact, it is sometimes even beneficial to begin with clean training and gradually switch to adversarial training (see e.g. [82]).

- Second, to prove our main conceptual message— feature purification— it is convenient to start from a clean-trained model, and then try to understand which part of the features are changed after robust training. Since neural nets have lots of equivalent transformations that are not very well-understood (even in two-layer case), if we adversarially train it from random initialization, then it is theoretically very hard to quantify how it is related to another clean-trained model learned from random initialization (since we need to understand all the invariants).[10]

- Yet, our theory still gives support to what happens in adversarial training from random initialization. As we shall prove, this "$o(1)$ feature change" comes from dense mixtures directions (see Theorem 5.2). Thus, adversarial training from random initialization should directly avoid learning such dense mixtures, as opposed to first learning them (by clean training) and then forgetting (by adversarial training). We illustrate this in Figure 9.

## 5 Statements and Explanations of Our Main Results

### 5.1 Clean Training, Adversarial Training and $\ell_2$ Robustness

For simplicity, in this subsection we sketch our main results for a special case $k = d^{0.36}$, although our theorems hold for a wider range of $k$ in the full appendix. Recall the learner network $f_t(x)$ and its random initialization are given in Section 4, the **clean training** using gradient descent is given in Algorithm 1, and $w_i^{(t)} \in \mathbb{R}^d$ is the hidden weight of the $i$-th hidden neuron at iteration $t$. We state the theorem for clean training as below:

**Theorem 5.1** (clean training). *There exists an absolute constants $C, c > 0$ such that for every constant $c_0 \in (0, c]$, every $d$ and $m$ with $m = d^{1+c_0}$, given $N \geq \Omega(d^C)$ many training data, for every random initialization weight $\sigma_0 = \frac{1}{\mathsf{poly}(d)}$, for every learning rate $\eta \in \left(0, \frac{1}{\Omega(d^C)}\right]$, if we define $T_{\mathsf{c}} := \Theta(\frac{d^{1.01}}{\eta})$, then for every $T_{\mathsf{f}} \in [T_{\mathsf{c}}, d^{\log d}/\eta]$, the following holds with high probability. The network $f_t$ with hidden weights $\{w_i^{(t)}\}_{i\in[m]}$ learned by **clean training** Algorithm 1 satisfies:*

(a) *Global feature learning: for every $t \in [T_{\mathsf{c}}, T_{\mathsf{f}}]$,*

$$\sum_{i\in[m]} \langle w_i^{(t)}, w_i^{(0)} \rangle^2 = o(1) \times \sum_{i\in[m]} \|w_i^{(t)}\|_2^2 \cdot \|w_i^{(0)}\|_2^2 \quad \text{and}$$

$$\sum_{i,j\in[m]} \langle w_i^{(t)}, w_j^{(t)} \rangle^2 = o(1) \times \left(\sum_{i\in[m]} \|w_i^{(t)}\|_2^2\right)^2 . \quad \text{(see Theorem C.2)}$$

(b) *Clean training has good clean accuracy: for every $t \in [T_{\mathsf{c}}, T_{\mathsf{f}}]$,*

$$\mathcal{E}_t^c = \Pr_{x, y=y(x), \rho}[y \neq \mathsf{sign}(f_t(w^{(t)}; x, \rho))] \leq o(1) . \quad \text{(see Theorem D.1)}$$

(c) *Clean training is not robust to small adversarial perturbations: for every $t \in [T_{\mathsf{c}}, T_{\mathsf{f}}]$, every*

---

[10]Even if one performs clean/adversarial training from the same random initialization, the additional randomness in SGD may quickly make the two models diverge from each other.

$\tau \geq \frac{1}{k^{0.5+10c}}$, using perturbation $\delta = -\tau \frac{y\mathbf{M}w^\star}{\|\mathbf{M}w^\star\|_2}$ (which does not depend on $f_t$),

$$\mathcal{E}_t^r \geq \Pr_{x,y,\rho}[f_t(w^{(t)}; x+\delta, \rho) \neq y] = 1 - o(1) \ . \qquad \text{(see Theorem E.1)}$$

Theorem 5.1 indicates that in our setting, clean training of the neural network has good clean accuracy but terrible robust accuracy. Such terrible robust accuracy is <u>not due to over-fitting</u>, as it holds even when a super-polynomially many iterations and infinitely many training examples are used to train the neural network. In the next theorem, we give a precise characterization of what the hidden weights $\{w_i\}$ are after clean training, and why they are not robust.

**Theorem 5.2** (clean training features). *For every neuron $i \in [m]$, there is a fixed subset $\mathcal{N}_i$ of size $|\mathcal{N}_i| = O(1)$ such that, for every $t \in [T_c, d^{\log d}/\eta)$,*

$$w_i^{(t)} = \sum_{j \in \mathcal{N}_i} \alpha_{i,j} w_j^\star \mathbf{M}_j + \sum_{j \notin \mathcal{N}_i} \beta_{i,j} w_j^\star \mathbf{M}_j \qquad \text{(see Theorem C.2)}$$

*where (1) $|\beta_{i,j}| < \frac{k}{d^{1-c}}$ for some small constant $c \in [0, 0.001]$, and (2) for at least $\Omega(d)$ many neurons $i \in [m]$, it satisfies $|\mathcal{N}_i| = 1$ and $\alpha_{i,j} > d^{-c}$. Moreover,*

$$\frac{1}{md} \sum_{i \in [m]} \sum_{j \in [d] \setminus \mathcal{N}_i} \beta_{i,j} \in \left[\frac{1}{d^c} \times \frac{k}{d},\ d^c \times \frac{k}{d}\right] \ . \qquad \text{(see Lemma E.2)}$$

Theorem 5.2 says that each neuron $w_i$ will learn constantly many (allegedly large) components in the directions $\{\mathbf{M}_j : j \in \mathcal{N}_j\}$, and its components in the remaining directions $\{\mathbf{M}_j : j \notin \mathcal{N}_j\}$ are all small. We emphasize that the sets $\mathcal{N}_i$ are independent of $t$ but are solely determined by random initialization. In other words, for each neuron $i$, which setset $\mathcal{N}_i$ it "wins" is completely determined by the "lottery ticket" (its random initialization). We discuss this in more details in Section 6.1.

Furthermore, Theorem 5.2 shows that instead of learning the pure, robust features $\{\mathbf{M}_j\}_{j \in [d]}$, intuitively, ignoring the small $d^c$ factors, focusing only on those neurons with $|\mathcal{N}_i| = 1$, and assuming for simplicity all the $\beta_{i,j}$'s are of similar (positive) magnitude, then, clean training will learn neurons:

$$w_i^{(t)} \approx \underbrace{\Theta(1)\mathbf{M}_j}_{\text{pure, robust feature}} + \underbrace{\sum_{j' \neq j} \left[\Theta\left(\frac{k}{d}\right) w_{j'}^\star \mathbf{M}_{j'}\right]}_{\text{dense mixture}} \qquad (5.1)$$

**Feature purification: mathematical reasoning.** Eq. (5.1) says that, after clean training, the neural network will be able to learn a big portion of the robust feature, $\Theta(1)\mathbf{M}_j$, plus some small *dense mixture* $v = \sum_{j' \neq j} \left[\Theta\left(\frac{k}{d}\right) w_{j'}^\star \mathbf{M}_{j'}\right]$. In our sparse coding model, each $x$ is of form $x = \mathbf{M}z + \xi$, where $z$ is a sparse vector and $\xi$ is the noise. One critical observation is that such dense mixture $v$ has *low correlation* with almost all inputs $x$ from the original distribution, so it has negligible effect for clean accuracy. However, such dense mixture is extremely vulnerable to *small but dense adversarial perturbations* of the input along this direction $v$, making the model non-robust.

As we point out, such "dense adversarial perturbation" directions do not exist in the original data.[11] Thus, one has to rely on adversarial training to remove dense mixtures to make the model robust. This is the main spirit of our *feature purification* principle, and we illustrate it in Figure 6.

**Where does dense mixture come from?** We shall explain in more details in Section 6.2, but at a high level, in each iteration, the gradient $\nabla \mathbf{Obj}$ will bias towards the direction that correlates with the labeling function $y = \text{sign}(\langle w^\star, z \rangle)$; and since $x = \mathbf{M}z + \xi$ in our model, such direction should be $\mathbf{M}w^\star = \sum_j w_j^\star \mathbf{M}_j$, so is a dense mixture direction and will be accumulated across time. The accumulation of dense mixture is consistent with the finding [23] (for linearly-separable data).

---

[11] One can try to add these dense mixtures directly to the training data set, which we conjecture to be similar to the approach in [48]

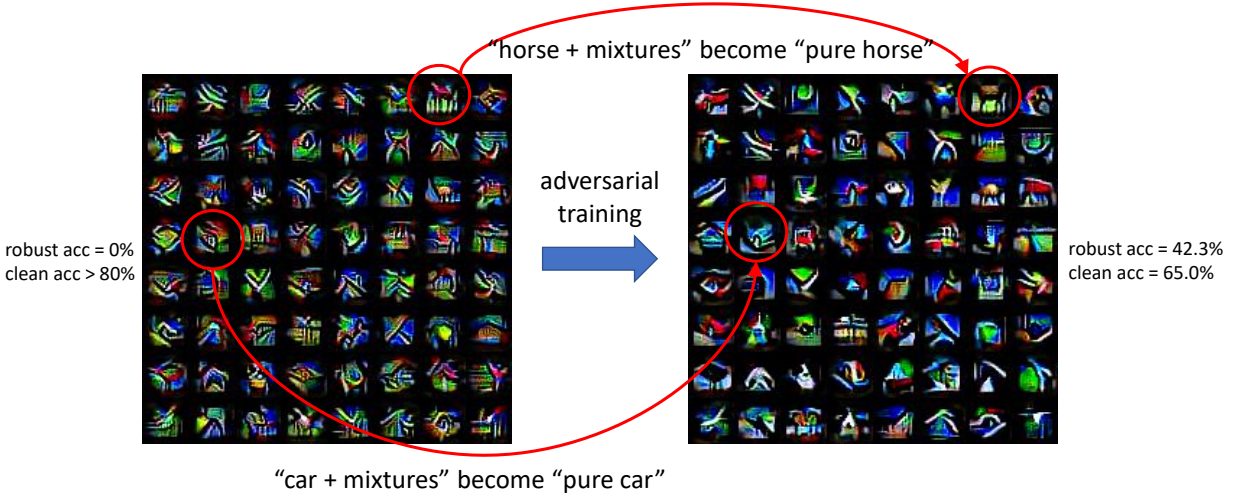

Figure 6: Experiments support our theory that adversarial training do purify dense mixtures, through visualizing some deep layer features of ResNet on CIFAR-10 data, trained against an $\ell_2(1, 0.25)$ attacker (see Section 8.2). More experiments for different layers of the network and different attackers, see Figure 11 and Figure 13.

However, as we have argued in Section 3, in our setting when the noise level $\sigma_x \geq \Omega(1)$ is large, such dense mixture direction cannot be used to given even good clean accuracy. Therefore, during clean training, the neural network has the incentive to discover features close to $\{\mathbf{M}_j\}_{j \in [d]}$ because they can "de-noise" $\xi$ better (see discussions in Section 3). Yet, our critical observation is that, even for well-trained neural network which aims to de-noise $\xi$, even when the neurons are close to being pure features $\{\mathbf{M}_j\}_{j \in [d]}$, the "dense direction" still *locally* correlates with the labeling function $y$, and thus can still be accumulated during the course of a **local training algorithm** such as gradient descent, leading to the small, non-robust part of each feature.

Next, we state the theorem for adversarial training (recall Algorithm 2). It shows that adversarial training indeed purifies the small dense mixtures, leading to *local changes* of the weights.

**Theorem 5.3** (adversarial training). *In the same setting as Theorem 5.1, suppose $A$ is an $\ell_2$ perturbation algorithm with radius $\tau \leq \frac{1}{k^{0.5+c}}$. Suppose we run **clean training** Algorithm 1 for $T_{\mathsf{f}} \geq T_{\mathsf{c}}$ iterations followed with **adversarial training** Algorithm 2 for $T_{\mathsf{g}} = \Theta(\frac{k^{2+c}}{\eta})$ iterations. The following holds with high probability.*

(a) *Empirical robust accuracy: for $t = T_{\mathsf{f}} + T_{\mathsf{g}}$*

$$\widehat{\mathcal{E}_t^r} = \Pr_{x, y = y(x), \rho, r}[\mathsf{sign}(f_t(x + A(f_t, x, y, r))) \neq y] \leq o(1) \qquad \text{(see Theorem F.1)}$$

(b) *Provable robust accuracy: when $A$ is the fast gradient method (FGM), for $t = T_{\mathsf{f}} + T_{\mathsf{g}}$*

$$\mathcal{E}_t^r \stackrel{\text{def}}{=} \Pr_{x, y = y(x), \rho}[\exists \delta \in \mathbb{R}^d, \|\delta\|_2 \leq \tau : \mathsf{sign}(f_t(x + \delta)) \neq y] \leq o(1) \qquad \text{(see Corollary F.2)}$$

(c) *Feature (local) purification: for every $t \in [T_{\mathsf{f}}, T_{\mathsf{f}} + T_{\mathsf{g}} - 1]$,*

$$\sum_{i \in [m]} \|w_i^{(T_{\mathsf{f}})} - w_i^{(t)}\|_2^2 = o(1) \times \sum_{i \in [m]} \|w_i^{(T_{\mathsf{f}})}\|_2^2 \qquad \text{(see (F.12))}$$

We emphasize that Theorem 5.3a holds for *any perturbation algorithm $A$* satisfying Definition 4.2, when we only concern the robustness of the network against $A$. Meaning that the *local* feature purification happens **regardless of which adversarial perturbation algorithm is used to find the adversarial examples**. More surprisingly, Theorem 5.3b says when a good perturbation algorithm such as FGM is used, then not only the robustness generalizes to unseen examples, it also generalizes to *any worst-case perturbation algorithm*.[12]

**Density of adversarial perturbation.** Our previous theorem suggests that one of the main goals of adversarial training is to remove dense mixtures to make the network more robust. Therefore, before adversarial training, the adversarial perturbations are dense in the basis of $\{\mathbf{M}_j\}_{j\in[d]}$; and after adversarial training, the adversarial perturbations ought to be more sparse and aligned with inputs from the original data set. Figure 7 has confirmed this theoretical finding using real-life data sets. Later in Section 8.3, we also present concrete measurements of the sparsity of these adversarial perturbations, and compare them in Figure 12.

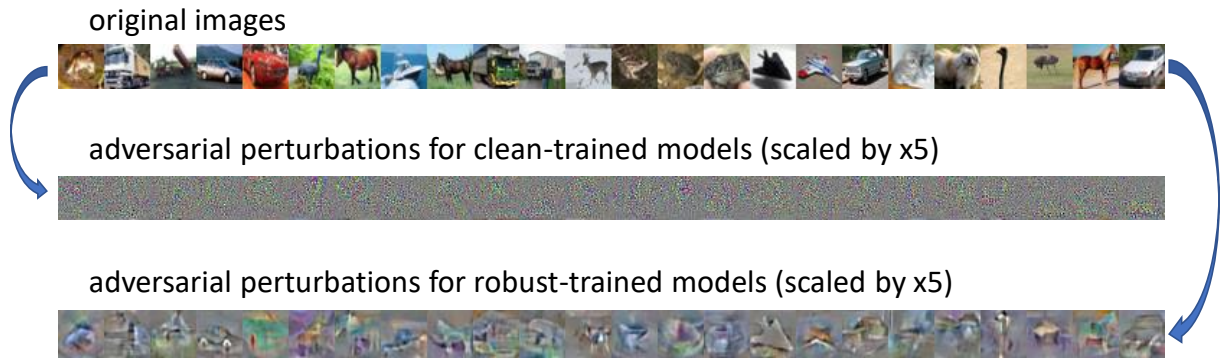

Figure 7: Adversarial perturbations before and after adversarial training; ResNet-34, CIFAR-10 data set. For clean-trained models, adversarial perturbations are "dense." After adversarial training, the "dense mixtures" are removed and the adversarial perturbations are more aligned with actual images. More experiments in Figure 12 and Section 8.3.

## 5.2 $\ell_\infty$ Robustness and Lower Bound for Low-Complexity Models

We also have the following theorem (stated in special case for simplicity) for $\ell_\infty$ robustness.

**Theorem 5.4** ($\ell_\infty$ adversarial training). *Suppose $\|\mathbf{M}\|_\infty, \|\mathbf{M}\|_1 = d^{o(1)}$. There exists constant $c_1 \in (0, c_0)$ such that, in the same setting as Theorem 5.3, except now for any $k \in [d^{c_1}, d^{0.399}]$, and A is an $\ell_\infty$-perturbation algorithm with radius $\tau = \frac{1}{k^{1.75+2c_0}}$. Then, the same Theorem 5.3 and Theorem 5.1 still hold and imply*

- *clean training is not robust again $\ell_\infty$ perturbation with radius $\frac{1}{k^{2-c_1}}$;*
- *adversarial training is robust against any $\ell_\infty$-perturbation of radius $\tau = \frac{1}{k^{1.75+2c_0}}$.*

*This gives a gap because $c_0, c_1$ can be made arbitrarily small. (see Theorem E.1 and Theorem F.4)*

We also show a lower bound that no low-degree polynomial, or even the corresponding neural tangent kernel (NTK), can robustly learn the concept class. Recall for our two-layer ReLU network,

---

[12] We point out that since our results hold for *any perturbation algorithm A*, it is impossible to characterize exactly what are the learned features after training (for example showing that they corresponds to the actual dictionary) since A might be a *bad adversarial perturbation finding algorithm* and the network does not even need to remove all the dense mixture in order to fool A. However, the true robustness given by Theorem 5.3b does imply that the dense mixture should be removed at least in terms of functionality, if one uses a good adversarial perturbation finding algorithm.

**Definition 5.5.** *The feature mapping of the* **neural tangent kernel** *for our two-layer network $f$ is*

$$\Phi(x) = \left( x \, \mathbb{E}_{\rho_i} \left( \mathbb{1}_{\langle w_i, x \rangle + \rho_i \geq b_i} - \mathbb{1}_{-\langle w_i, x \rangle + \rho_i \geq b_i} \right) \right)_{i=1}^{m}$$

*Therefore, given weights $\{v_i\}_{i \in [m]}$, the NTK function $p(x)$ is given as*

$$p(x) = \sum_{i \in [m]} \langle x, v_i \rangle \mathop{\mathbb{E}}_{\rho_i \sim \mathcal{N}(0, \sigma_\rho^2)} \left( \mathbb{1}_{\langle w_i, x \rangle + \rho_i \geq b_i} - \mathbb{1}_{-\langle w_i, x \rangle + \rho_i \geq b_i} \right)$$

*Without loss of generality, assume each $w_i \sim \mathcal{N}(0, \mathbf{I})$.*

In this paper, we consider a wide range of NTK parameters: $\rho_i \sim \mathcal{N}(0, \sigma_{\rho_i}^2)$ for arbitrary $\sigma_{\rho_i} \in [0, d^{o(1)}]$ and $|b_i| \leq d^{o(1)}$.

Our lower bound holds *even* for a most simple case $\mathbf{M} = \mathbf{I}$ and $\sigma_x = 0$, so the original concept class is linearly separable. We prove the following:

**Theorem 5.6** (lower bound). *For every constant $C > 1$, suppose $m \leq d^C$, then there is a constant $c > 0$ such that when $k = \frac{1}{d^c}$, considering $\ell_\infty$ perturbation with radius $\tau = \frac{1}{k^{100}}$, we have w.h.p. over the choice of $w_i$, for every $p(x)$ in the above definition, the robust error*

$$\mathcal{E}^r(p) \geq \frac{1 - o(1)}{2} \ . \qquad \text{(see Theorem G.1)}$$

(In contrast, Theorem 5.4 says adversarial training of neural network gives robust radius $\tau = \frac{1}{k^{1.76}}$.)

Since a poly-sized NTK kernel is known to be powerful enough to incorporate any low complexity functions (such as constant-degree polynomials) [6], we have the following corollary.

**Corollary 5.7** (lower bound). *In the same setting as Theorem 5.6, if $q(x)$ is a constant degree polynomial, then we also have the robust error $\mathcal{E}^r(q) \geq \frac{1-o(1)}{2}$.*

# 6 Overview of the Training Process

In this section, we present an overview of the proof for the training process, using gradient descent starting from random initialization. The complete proof is deferred to the Appendix.

## 6.1 Wining Lottery Tickets Near Random Initialization

Our proof begins by showing how the features in the neural network are emerged from random initialization. In this phase, the loss function is not sufficiently minimized yet, so the classification accuracy remains around 50%. However, we prove in this phase, gradient descent can already drive the neural network to learn a rich set of interesting features out of the random initialization. We call this process "lottery ticket winning" near random initialization, which is related to the study of [34].

*Remark.* This "lottery ticket winning" process is **fundamentally different from the neural tangent kernel analysis** (e.g. [4, 7, 8, 13, 14, 26–29, 38, 45, 50, 57, 61, 105, 116]). In this phase, although the loss is not sufficiently minimized, the activation patterns of the ReLU activations have changed *dramatically*, so that they have little correlations with the random initialization. Yet, we develop a new theoretical technique that allows us to control the change of the weights of the neurons, as we summarize below.

We derive the following property at random initialization. At iteration $t = 0$, the hidden weights are initialized as $w_i^{(0)} \sim \mathcal{N}(0, \sigma_0^2 \mathbf{I}_{d \times d})$. Using standard properties of Gaussians, we show the following critical property: as long as $m \geq d^{1.01}$, there exists small constants $c_3 > c_4 > 0$ such that

(i) For most of the neurons $i \in [m]$, $\max_{j \in [d]} \{\langle \mathbf{M}_j, w_i^{(0)} \rangle^2\} \leq 2\sigma_0^2 \log d$.

(ii) For at most $\frac{1}{d^{c_4}}$ fraction of of the neurons $i \in [m]$, there is a dimension $j \in [d]$ with $\langle \mathbf{M}_j, w_i^{(0)} \rangle^2 \geq 2.01\sigma_0^2 \log d$.

(iii) For at least $\frac{1}{d^{c_3}}$ fraction of of the neurons $i \in [m]$, there is one and only one $j \in [d]$ such that $\langle \mathbf{M}_j, w_i^{(0)} \rangle^2 \geq 2.02\sigma_0^2 \log d$, and all the other $j' \in [d]$ satisfies $\langle \mathbf{M}_{j'}, w_i^{(0)} \rangle^2 \leq 2.01\sigma_0^2 \log d$.

In other words, even with *very mild over-parameterization* $m \geq d^{1.01}$, by the property of random gaussian initialization, there will be some "potentially lucky neurons" in (ii), where the maximum correlation to one of the features $\mathbf{M}_j$ is **slightly higher than usual**. Moreover, there will be some "surely lucky neurons" in (iii), where such "slightly higher correlation" appears in one and only one of the target features $\mathbf{M}_j$.

In our proof, we denote the set of the neurons in (iii) whose correlation with $\mathbf{M}_j$ is slightly higher than usual as the set $\mathcal{S}_{j,sure}^{(0)}$, and denote those in (ii) as $\mathcal{S}_{j,pot}^{(0)}$. We will identify the following process during the training, as given in Theorem C.1:

---
**Lottery tickets winning process**

For every $j \in [d]$, at every iteration $t$, if $i \in \mathcal{S}_{j,sure}^{(0)}$, then $\langle \mathbf{M}_j, w_i^{(t)} \rangle^2$ will grow faster than $\langle \mathbf{M}_{j'}, w_i^{(t)} \rangle^2$ for every $t$, until $\langle \mathbf{M}_j, w_i^{(t)} \rangle^2$ becomes sufficiently larger than all the other $\langle \mathbf{M}_{j'}, w_i^{(t)} \rangle^2$.

---

In other words, if neuron $i$ *wins the lottery ticket* at random initialization, then eventually, it will deviate from random initialization and grow to a feature that is more close to (a scaling of) $\mathbf{M}_j$. Our other main observation is that if we slightly over-parameterize the network with $m \geq d^{1.001}$, then for each $j \in [d]$, $|\mathcal{S}_{j,sure}^{(0)}| \geq 1$ and $|\mathcal{S}_{j,pot}^{(0)}| \leq d^{0.01}$. Or in words, for each dimension $j \in [d]$, the number of lottery tickets across all neurons is at most $d^{0.01}$, but at least one neuron will win a lottery ticket (see Lemma B.2). We also illustrate the lottery ticket winning process experimentally in Figure 8.

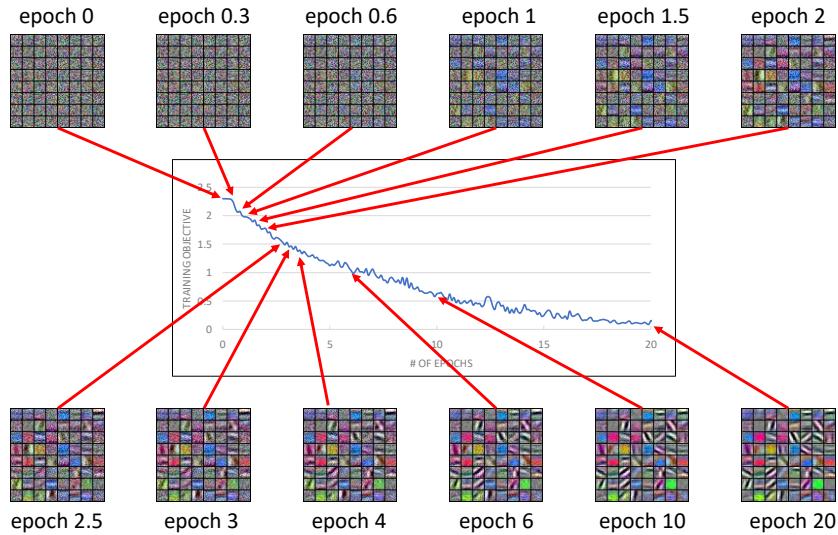

Figure 8: Lottery tickets winning process, AlexNet, CIFAR-10 data set.

## 6.2 The Formation of "Dense Mixtures" During Training

The next phase of our analysis begins when all the neurons already won their lottery tickets near random initialization. After that, the loss starts to decrease significantly, so the (clean) classification error starts to drop. We shall prove that in this phase, gradient descent will also accumulate, in each neuron, a small "dense mixture" that is extremely vulnerable to small but adversarial perturbations. To show this, we maintain the following critical property as given in Theorem C.2:

> If a neuron $i$ wins the lottery ticket for feature $\mathbf{M}_j$ near random initialization, then it will keep this "lottery ticket" throughout the training.

Or in math words, for each neuron $i$, after $\langle \mathbf{M}_j, w_i^{(t)} \rangle^2$ becomes sufficiently larger than all the other $\langle \mathbf{M}_{j'}, w_i^{(t)} \rangle^2$ at the first stage, it will stay much larger than other $\langle \mathbf{M}_{j'}, w_i^{(t)} \rangle^2$ for the remaining of the training process. To prove this, we introduce a careful coupling between the (directional) gradient of the neuron, and the (directional) Lipschitz continuity of the network $f_t$, this is given in Section C.4.2.

**The vulnerable dense mixtures.** The most critical observation in this phase is the formation of "dense mixtures", where we show that even for the "lucky neuron" that wins the lottery ticket, the hidden weight of this neuron will look like (see Theorem 5.2)

$$w_i \approx \alpha_t \left( \mathbf{M}_j + \Theta\left(\frac{k}{d}\right) \sum_{j' \neq j} w_{j'}^\star \mathbf{M}_{j'} \right) \tag{6.1}$$

In other words, up to scaling, these neurons will look like $w_i \approx \mathbf{M}_j + v_i$, where $v_i$ is a "dense mixture" $v_i = \Theta\left(\frac{k}{d}\right) \sum_{j' \neq j} w_{j'}^\star \mathbf{M}_{j'}$.

The key observation is that $v_i$ is *small and dense*, in the sense that it is a mixture of all the other features $\{\mathbf{M}_{j'}\}_{j' \in [d]}$, but each of the feature has a much smaller contribution comparing to the leading term $\mathbf{M}_j$. Recall in our sparse coding model, each input $x = \mathbf{M}z + \xi$; so *with high probability*:

$$|\langle v_i, x \rangle| \leq \widetilde{O}\left(\frac{k}{d} \|x\|_2\right) \tag{6.2}$$

This value is even smaller than $\frac{1}{k}$ when $k \leq \sqrt{d}$. Thus, this dense mixture will not be correlated with any particular natural input, and thus the existence of these mixtures will have negligible contribution to the output of $f_t$ on clean data.

However, if we perturb input $x$ along the dense direction $\delta \propto \sum_{j' \in [d]} \mathbf{M}_{j'}$, we can observe that:

$$|\langle v_i, \delta \rangle| = \Omega\left(\frac{k}{\sqrt{d}} \|\delta\|_2\right)$$

Comparing this with Eq (6.2), such "dense perturbation" can change the output of the neural network $f_t$ by *a lot*, using a small $\delta$ whose norm is much smaller than that of $x$. Thus, at this phase, even when the network has a good clean accuracy, it is still non-robust to these *small yet dense adversarial perturbations*. Moreover, this perturbation direction is "universal", in the sense that it does not depend on the randomness of the model at initialization, or the randomness we use during the training. This explains transfer attacks in practice: that is, the adversarial perturbation found in one model can also attack other models that are independently trained.

**Feature purification.** Since Eq. (6.2) suggests most original inputs have negligible correlations with each dense mixture, during clean training, gradient descent will have *no incentive to remove* *those mixtures*. Thus, we have to rely on adversarial training to *purify* those dense mixtures by introducing adversarial examples. Those examples have correlation with $v_i$'s that are higher than usual. As we prove in Theorem 5.1 and illustrate in Figure 1, such "purifications", albeit imposing only a small change to each neuron, will greatly improve the robustness of the neural network.

**The formation of the dense mixtures.** To further help the readers understand how those "dense mixtures" are formed, we sketch the proof of Theorem 5.2, which shows why clean training is provably non-robust. The main observation is that *when the dense mixtures are small*, the negative gradient of the (say, population) loss with respect to each neuron $w_i$ is approximately given by (recall $x = \mathbf{M}z + \xi$):

$$-\nabla_{w_i}\mathbf{Loss}(w^{(t)}) \approx \mathop{\mathbb{E}}_{x,y=y(x),\rho}\left[y\ell'_t(w^{(t)};x,y,\rho)\big(\mathbb{1}_{\langle w_i^{(t)},x\rangle+\rho_i\geq b^{(t)}} + \mathbb{1}_{-\langle w_i^{(t)},x\rangle+\rho_i\geq b^{(t)}}\big)\mathbf{M}z\right]$$

As a result,

$$-\langle\nabla_{w_i}\mathbf{Loss}(w^{(t)}),\mathbf{M}w^\star\rangle \approx \mathop{\mathbb{E}}_{x,y=y(x),\rho}\left[y\ell'_t(w^{(t)};x,y,\rho)\big(\mathbb{1}_{\langle w_i^{(t)},x\rangle+\rho_i\geq b^{(t)}} + \mathbb{1}_{-\langle w_i^{(t)},x\rangle+\rho_i\geq b^{(t)}}\big)\langle w^\star,z\rangle\right]$$

Since $y = \mathsf{sign}(\langle w^\star, z\rangle)$, we have $y\langle w^\star, z\rangle \geq 0$; together with the non-negativity $\ell' \geq 0$ and the indicators are always non-negative, we know $-\langle\nabla_{w_i}\mathbf{Loss}(w^{(t)}),\mathbf{M}w^\star\rangle$ is quite positive. Thus, during clean training, this dense direction $\mathbf{M}w^\star$ will naturally accumulate in each neuron.

We emphasize that this is indeed a *special property of gradient descent*. Consider again the case $\sigma_x = \Omega(1)$ discussed in Section 3, where $x = \mathbf{M}z + \xi$ with $\|\xi\|_2 = \Omega(1) = \Omega(\|\mathbf{M}z\|_2)$. With high probability, a linear classifier using direction $\mathbf{M}w^\star$ *cannot* be used to classify $x$ correctly. Yet, this direction $\mathbf{M}w^\star$ is still locally positively correlated with the labeling function $y$, *especially for well-trained, well-generalizing neural networks when the $\xi$ can be "de-noised"*. (Stochastic) gradient descent, as a local update algorithm, only exams the local correlation between the update direction and the labeling function, and it *does not exam* whether this direction can be used in the final result. Thus, this dense direction $\mathbf{M}w^\star$ will be accumulated step by step, leading to a non-robust part of each of the features during clean training. *In fact, even if we use $w_i = \mathbf{M}_i$ as initialization as opposed to random initialization, continuing clean training will still accumulate these small but dense mixtures.* We illustrate this in Figure 9.

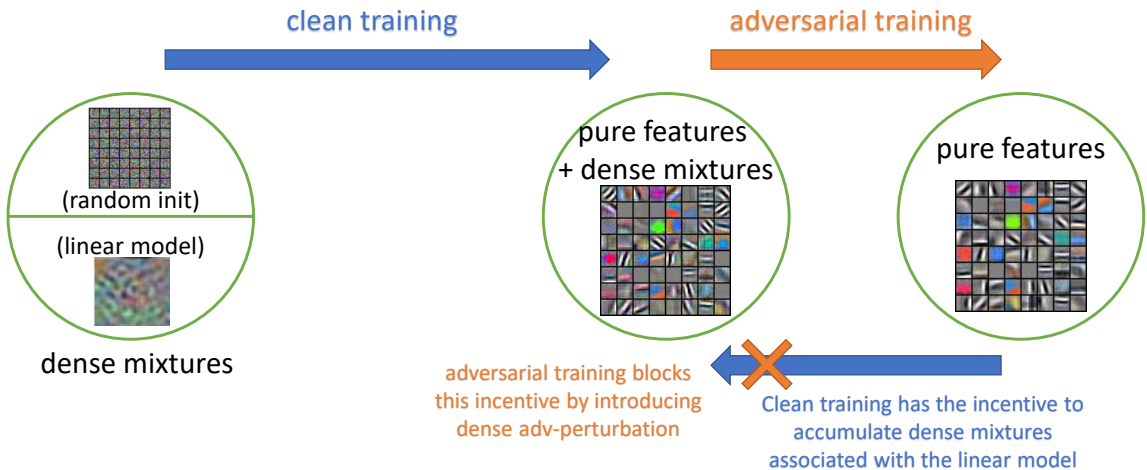

Figure 9: Overall summary of clean training, adversarial training, in the language of pure vs dense features. (Experiment based on AlexNet on CIFAR-10 dataset.)

# 7 Conclusion

In this paper, we made a first step towards understanding how, in principle, the features in a neural network are learned during the training process, and why after clean training, these *provably well-generalizing* features are still *provably non-robust*. Our main conclusion is that during the clean training process using (stochastic) gradient descent, neural network will accumulate, in all features, some "dense mixture directions" that have low correlations with any natural input, but are extremely vulnerable to (dense) adversarial perturbations. During adversarial training, such dense mixtures are purified to make the model more robust. Our results suggest that the non-robustness of clean training is mainly due to two reasons:

1. the inductive bias of (stochastic) gradient descent, and
2. the "sparse coding" structure of the data.

Both reasons are necessary in some sense. First, a robust model is also a global minimizer of the clean training objective (at least in our setting); but even with proper regularization and infinite training examples to avoid over-fitting, gradient descent still has *inductive bias* towards finding a non-robust model. Second, it is easy to come up with data sets— such as linear-classifier labels over well-conditioned mixture-of-Gaussians like inputs— where clean training using gradient descent directly achieves the best robust accuracy. Thus, to understand the non-robustness of neural networks, we more or less *have to* take into account the gradient descent algorithm and the structure of the inputs.

Indeed, our step is still very *provisional*. We immediately see a plethora of extensions from our work. First of all, natural images have much richer structures than sparsity; hence, those "non-robust mixtures" accumulated by clean training might also carry structural properties other than density. Moreover, we would like to extend our work to the clean and robust training of multi-layer neural networks, possibly with hierarchical feature purification processes. (Our experiments in Figure 10 have confirmed on such hierarchical feature purification phenomenon.) Indeed, understanding the whole picture of adversarial examples and adversarial training might require a complete understanding of deep learning.

# 8 Experiment Details

We perform experiments using three standard architectures, AlexNet, ResNet-16, and ResNet-34 with basic blocks, and tested on the CIFAR-10 dataset.[13]

We discover that learning rate 0.1 for good for ResNet and 0.02 is good for AlexNet; while weight decay 0.0001 is good for ResNet and 0.0005 is good for AlexNet (this was also recommended by the git repo authors). We use standard SGD with 0.9 momentum as the training algorithm. During adversarial training, we have implemented:

- The empirical $\ell_2$ perturbation algorithm (i.e., attacker) suggested by [82]. We choose two sets of parameters $\ell_2(1, 0.25)$ and $\ell_2(0.5, 0.12)$.[14]

---

[13]We used the implementations from https://github.com/bearpaw/pytorch-classification. We used their default random crop and random flip as data augmentation.

[14]We use their SMOOTHADV$_{\text{PGD}}$ attacker with following parameters. We use $\sigma = 0.25$ which is the random Gaussian perturbation added to the input; use MTRAIN $= 2$ which is the number of Gaussian noise samples used per training sample, use $T_{\text{PGD}} = 4$ which is the number of PGD attack steps, and use $\varepsilon = 1$ which is the $\ell_2$ radius for the PGD attacker. We also follow their instruction to perform 10 warmup epochs to gradually increase $\varepsilon$ from zero to $\varepsilon = 1$. We call this $\ell_2(1, 0.25)$. We have also implemented $\ell_2(0.5, 0.12)$.

- The empirical $\ell_\infty$ perturbation algorithm (i.e., attacker) [65], with $\ell_\infty$ radius 4/255 and 8/255, together with 7 steps of PGD attack. We call them $\ell_\infty(4/255)$ and $\ell_\infty(8/255)$ respectively.

We mostly focus on the $\ell_2(1, 0.25)$ attacker in this paper, but shall compare them in Section 8.4.

*Remark* 8.1. In Table 1, we present robust/clean accuracies against such attackers after vanilla clean training / vanilla adversarial training. We emphasize here that, in practice, one can also first perform clean training and then apply adversarial traing using the clean-trained weights as initialization (like we have theoretically studied in this paper). This does not affect the overall performance of both robust and clean accuracies.

| attacker | $\ell_\infty(4/255)$ | $\ell_\infty(8/255)$ | $\ell_2(0.5, 0.12)$ | $\ell_2(1, 0.25)$ |
|---|---|---|---|---|
| clean training | **0.0%** (93.9%) | **0.0%** (93.9%) | **0.0%** (93.9%) | **0.0%** (93.9%) |
| adversarial training (init. with random weights) | **61.4%** (82.3%) | **45.7%** (71.6%) | **59.8%** (78.3%) | **42.3%** (64.3%) |
| adversarial training (init. with clean-trained weights) | **61.9%** (83.9%) | **46.0%** (73.7%) | **60.0%** (79.2%) | **42.2%** (65.5%) |

Table 1: CIFAR-10 **robust accuracy %** (clean accuracy %) against different attackers using ResNet-34

## 8.1 Feature Visualization of Deeper Layers

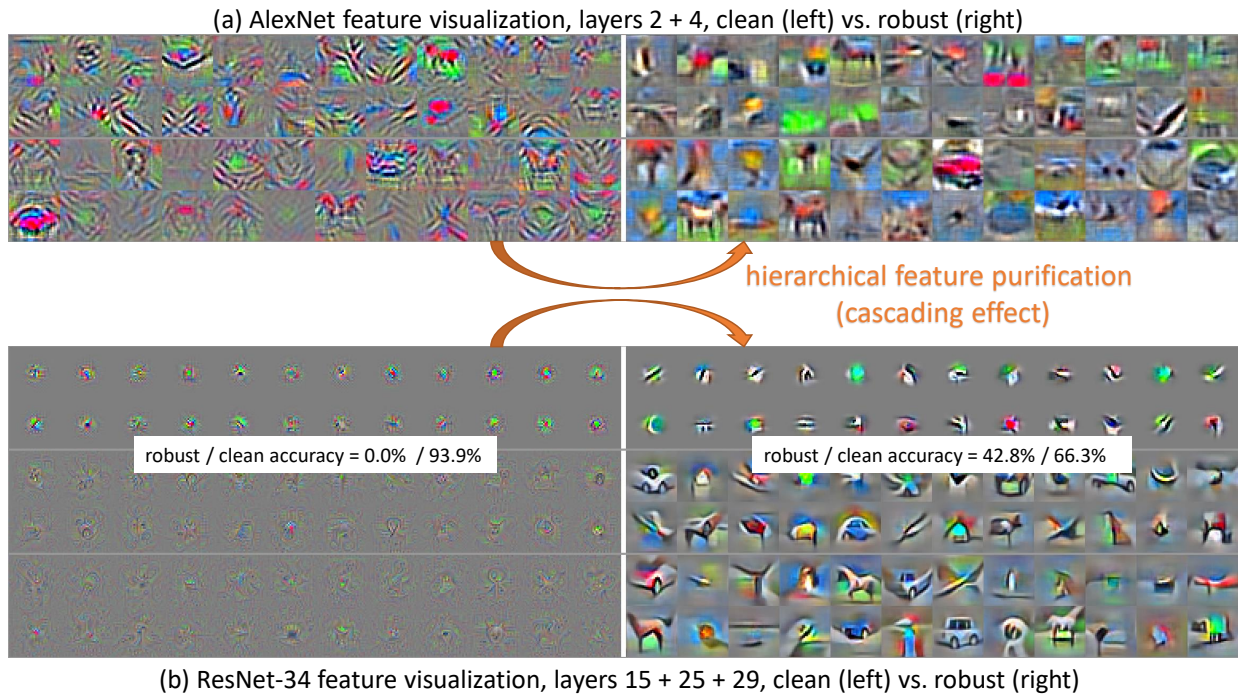

Figure 10: Visualization of deep features on cleanly-trained vs. robustly-trained models.
**Take-away message:** features from robustly-trained models are more "pure" and closer to the the real image space. We hope that our work can be extended to a "hierarchical feature purification" for multi-layer neural networks, using the recent advance in the theory of training deep neural networks efficiently and beyond NTKs [3, 5]

Visualizing the first layer of any trained architecture is trivial: for instance, for AlexNet, the weight tensor of the first layer is $3 \times 11 \times 11$ which gives the RGB color of $11 \times 11$ patches (and this was precisely what we presented in Figure 1). However, such visualization can be less meaningful for ResNet because the tensors are of dimension $3 \times 3 \times 3$.

Visualizing the features presented by *deeper* convolutional layers is an active research area, dating back at least to [31]. Perhaps the most naive approach is to start from a randomly initialize image (of size $3 \times 32 \times 32$), then take a specific neuron $n$ at some layer, and repeatedly take its gradient with respect to the image. If we keep adding this gradient to the input image, then ideally this gives us the image which "excites" $n$ the most. Unfortunately, it is a common knowledge in this area that this naive approach does not lead to "visually meaningful" images as we go (even slightly) deeper into a network (see e.g. the left column of Figure 10).

In existing literature, researchers have tried to various ways to resolve this issue (see e.g. an extensive survey by [76] and the references therein). At a high level, some penalizes the image to remove high-frequency noise [66, 70, 72, 79, 99]; some searches for images that can still excite the given neuron after jittering [70, 71, 79, 99]; and some searches only in the space of "real data" by building a model (e.g. using GAN) to capture the prior [70, 73, 74].

We observe that, if the model is robustly trained, then one can directly apply the naive approach to visualize features of the deep layers, and the resulting images can be "visually very meaningful." See Figure 10.[15] Our theory in fact explains this phenomenon: the dense mixtures accumulated during clean training are extremely harmful to the visualization effect, since they are "visually meaningless." After robust training, such dense mixtures are removed so the visualization starts to align better with human concepts.

Throughout this paper we stick to this naive approach for visualizing features of deep layers.[16]

## 8.2 Feature Purification at Deeper Layers

Since our theory shows the feature purification principle of a single layer, we perform the following experiment to verify it in practice, to **study the effect of *feature (local) purification* in each layer individually**. We consider the (pre-activation) ResNet-34 [109] architecture which has 31 convolutional layers. We select some convolutional layer $\ell$, and[17]

- (stage 1) perform $T$ epochs of adversarial training;
- freeze the weights of layers $1, 2, \ldots, \ell - 1$ and re-randomize weights of layers $\ell, \ell + 1, \ldots$;
- (stage 2) perform $T$ epochs of clean training (by training weights of layers $\ell, \ell + 1, \ldots$);
- (stage 3) perform $T$ epochs of adversarial training (by training weights of layers $\ell, \ell + 1, \ldots$).

Then, we visualize the features on layer $\ell$

- at the end of epoch $T$ (indicating layer $\ell$ is random),
- at the end of epoch $2T$ (indicating layer $\ell$ is clean trained), and
- at the end of epoch $3T$ (indicating layer $\ell$ is adversarially trained).

---

[15]This should not be surprising given that the "jittering" technique is known to work in practice on visualizing clean models.

[16]Specifically, starting from a random input image, we take 2000 gradient steps to update the image so that the given neuron at a specific layer is excited the most. We added a weight decay factor to incentivize the image to go to RGB (128,128,128) — except in Figure 6 we incentivize the image to go to RGB (0,0,0).

[17]We choose $T = 50$ with initial learning rate 0.1, and decay it to 0.01 at the end of the 40-th, 90-th and 140-th epoch. Note that since we have weight decay, if we run stage 3 indefinitely, then the individual neurons on the $\ell$-th layer may change too much (because two neurons can even swap their positions). Therefore, to illustrate our theoretical finding, we stop stage 3 *as long as* the robust (and clean) accuracy matches that of stage 1. This usually requires less than $T$ epochs.

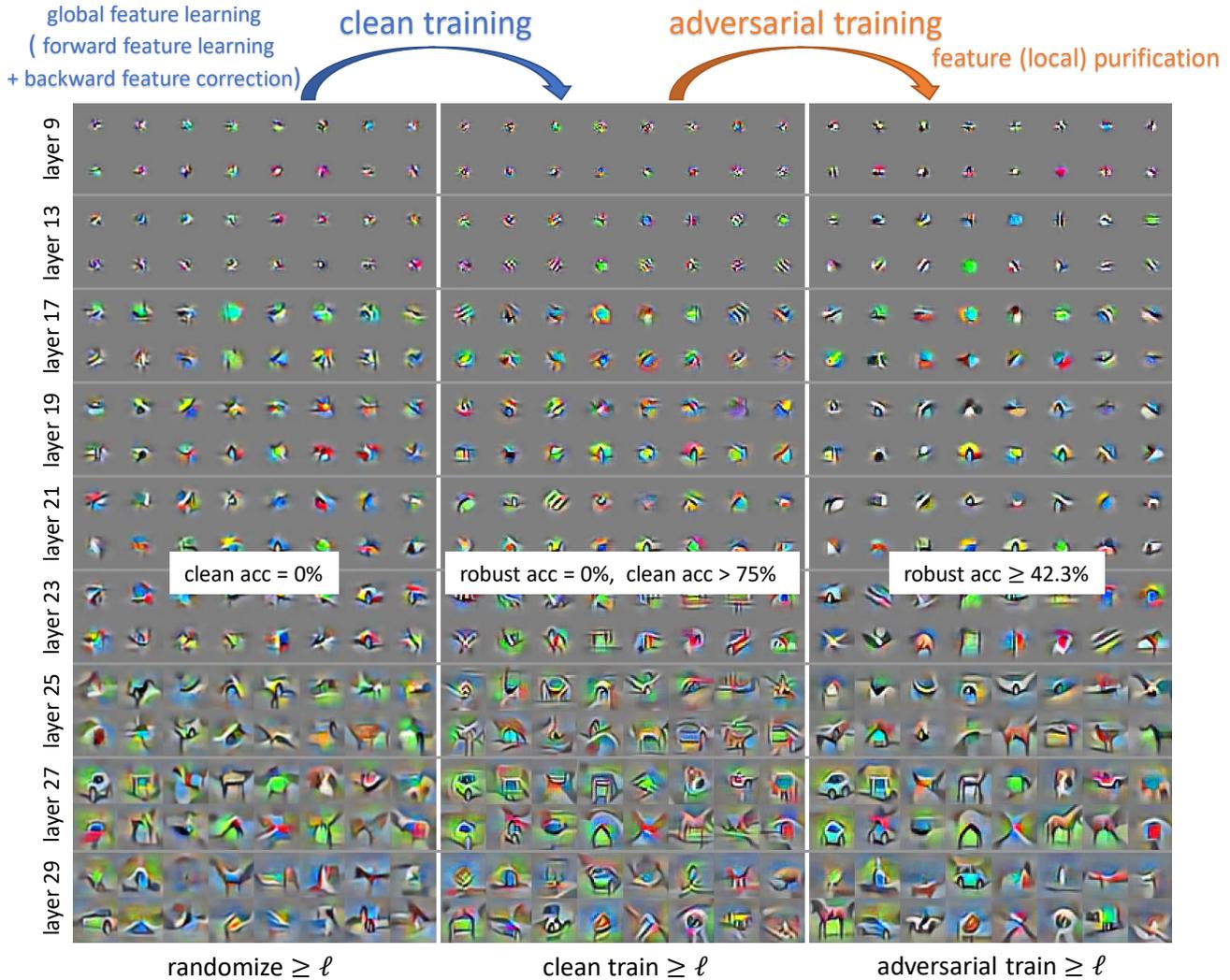

Figure 11: Visualization of the $\ell$-th layer features from ResNet-34 for $\ell \in \{9, 13, 17, 19, 21, 23, 25, 27, 29\}$. For each case of $\ell$, the layers $\leq \ell - 1$ are frozen at some pre-trained robust weights, and only layers $\geq \ell$ are trained. Left column refers to layers $\geq \ell$ are randomly initialized, "clean" refers to layers $\geq \ell$ are cleanly trained, and "robust" refers to layers $\geq \ell$ are adversarially trained.
**Take-away message:** feature purification happens even at deep layers of a neural network.

We present our findings in Figure 11, and report the correlation between neurons in Figure 2, for the $\ell_2(1, 0.25)$ adversarial attacker.

We also point out that with this training schedule, even when the first $(\ell - 1)$-layers are fixed to "robust features" and only the $\ell, \ell + 1, \cdots$ layers are trained, after clean training, the robust accuracy is still 0%.

### 8.3 Sparse Reconstruction of Input Data and of Adversarial Perturbation

Recall in Figure 4, we have shown that the input images can be sparsely reconstructed from the robust features. To better quantify this observation, we compare how sparse the input images can be reconstructed from (1) random features, (2) clean features, and (3) robust features. For each of the tasks, we use Lasso to reconstruct the 100 images, and sweep over all possible weights of the $\ell_1$ regularizer (which controls how sparse the reconstruction is).[18] The results are presented in the first row of Figure 12. As one can see, using clean features one can also sparsely reconstruct the input, but using robust features the reconstruction can be *even sparser*. This, to some extent, supports our theory that robust features are more "pure" than clean features.

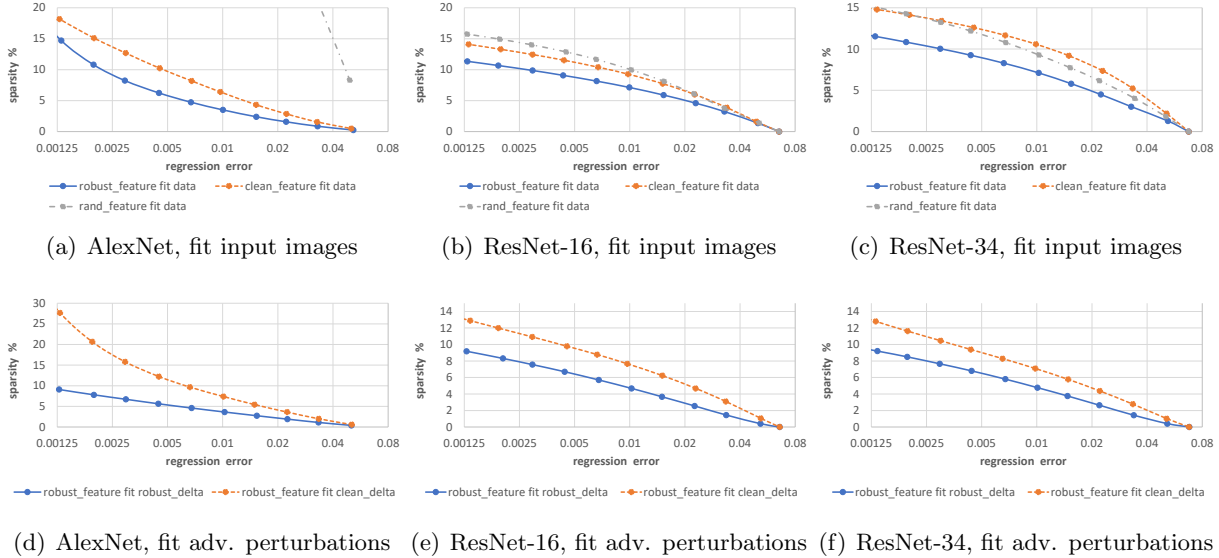

Figure 12: Sparse reconstruction of input mages and of adversarial perturbations.
> **Take-away message for the first row:** robust features can be used to reconstruct input images with better sparsity, suggesting that robust features are more pure.
> **Take-away message for the second row:** adversarial perturbations from a clean model are more "dense" comparing to those from a robust model (and in fact robust model's adversarial perturbations are (much) closer to real input images, see Figure 7).

Perhaps *more importantly*, our theory suggests that for clean-trained models, adversarial perturbations (we refer to as clean_delta) have "dense mixtures"; while for robust-trained models, adversarial perturbations (we refer to as robust_delta) are "more pure." This was visually illustrated in Figure 7. Now, to better quantify this observation, we compare how sparse clean_delta and robust_delta can be reconstructed from robust features. See the second row of Figure 12.[19] From this experiment, we confirm that in practice, adversarial perturbations on robust models are more "pure" and closer to real input images.

*Remark 8.2.* We point out when comparing how sparse clean_delta and robust_delta can be reconstructed from robust features, we *did not cheat*. For instance, in principle clean_delta may not lie in the span of robust features and if so, it cannot be (sparsely) reconstructed from them. In our experiments (namely, the second row of Figure 12), we noticed that clean_delta almost lies in the span of robust features (with regression error $< 0.00005$ for AlexNet and $< 10^{-9}$ for ResNet).

---

[18] Recall the Lasso objective is $\min_y \|Wy - x\|_2^2 + \lambda\|y\|_1$, where it uses $Wy$ to reconstruct given input $x$, and $\lambda$ is the weight of the regularizer to control how sparse $y$ is. The convolutional version of Lasso is analogous: the matrix $W$ becomes the "transpose" of the weight of the convolutional layer, which is for instance implemented as `nn.ConvTranspose2d` in PyTorch. In our implementation, we have shifted each input image so that it has zero mean in each of the three color channels. We have selected the first 100 images where the (trained) robust classifier gives correct labels; the plots are similar if one simply selects the first 100 training images.

[19] In fact, we have also re-scaled the perturbations so that they have similar mean and standard deviations comparing to real input images. This allows one to also compare the two rows of Figure 12.

## 8.4 Comparing Different Attackers

We also demonstrate in Figure 13 that feature purification occurs against several different attackers.

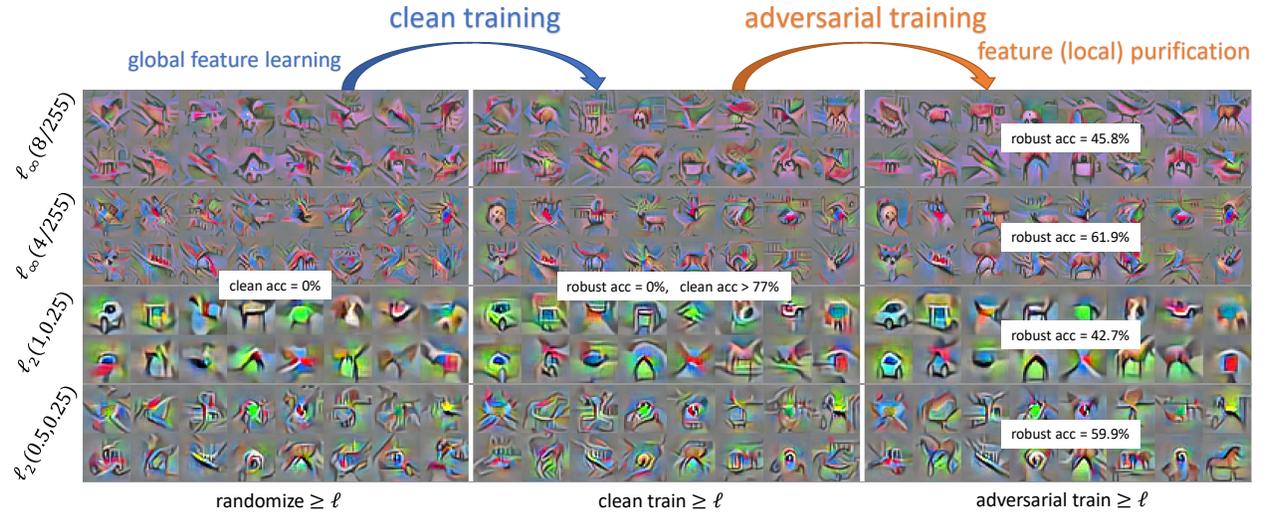

Figure 13: Visualization of the 27-th convolutional layer of ResNet-34 against different attackers.
**Take-away message:** feature purification happens against different attackers, and stronger attacker gives stronger effect of feature purification.

# Appendix: Complete Proofs

We give a quick overview of the structure of our appendix sections.

In Section A, we warm up the readers by calculating the gradient of the objective, and demonstrating that polynomially many samples are sufficient for the training.

In Section B, we formally introduce $\mathcal{S}_{j,pot}^{(t)}$, the set of "potentially lucky neurons" and $\mathcal{S}_{j,sure}^{(t)}$, the set of "surely lucky neurons" at iteration $t$. In particular, we shall emphasize on how those notions evolve as $t$ increases.

In Section C, we formally prove how "lucky neurons" continue to be lucky, and more importantly, for every neuron $i$ that is lucky in direction $j$, why it grows faster than other unlucky directions $j'$, and how much faster. Specifically, Theorem C.1 corresponds to the initial "lottery-winning" phase where the accuracy remains around 50%; and Theorem C.2 corresponds to the later phase where large signals become even larger and eventually most neurons become "pure + dense mix" of the form (6.1). This is the most difficult section of this paper.

In Section D, we prove that why clean training gives good clean (testing) accuracy. It is based on the structural theorem given by Theorem C.2, and requires some non-trivial manipulations of probability theory results (such as introducing a high-probability, Bernstein form of the McDiarmid's inequality).

In Section E, we prove that why the model obtained from clean training is non-robust. It formally shows how the "dense mixtures" become accumulated step by step during clean training.

In Section F, we prove our theorems for both $\ell_2$ and $\ell_\infty$ adversarial training. In particular, in this section we demonstrate why practical perturbation algorithms, such as the fast gradient method (FGM), can help the (adversarial) training process "kill" those "dense mixtures."

In Section G, we prove lower bounds for the neural tangent kernel model given by two-layer networks.

In Section H, we give missing details of some probability theory lemmas.

## A  Notations and Warmups

We find it perhaps a good exercise to do some simple calculations to warmup the readers with our notations, before going into the proofs.

**Global Assumptions.** Throughout the proof,

- We choose $m = d^{1+c_0}$ for a very small constant $c_0 \in (0, 1)$.

  (One should think of $c_0 = 0.0001$ for a simple reading. Our proof generalizers to larger $m = \mathsf{poly}(d)$ since having more neurons does not hurt performance, but we ignore the analysis so as to provide the simplest notations.)

- We assume $k < d^{(1-c_0)/2}$.

- We choose $\lambda = \frac{\log \log \log d}{d}$ for simplicity.

  (The purpose of $\log \log \log d$ factor is to simplify notations, and it can be tightened to constant.)

- Whenever we write "for random $x$", "for random $z$" or "for random $\xi$", we mean that the come from the distributions introduced in Section 2 with $x = \mathbf{M}z + \xi$.

**Network Gradient.** In every iteration $t$, the weights of the neurons are $w_1^{(t)}, \ldots, w_m^{(t)} \in \mathbb{R}^d$. Recall the output of the neural network on input $x \in \mathbb{R}^d$ is

$$f_t(w^{(t)}; x, \rho) = \sum_{i=1}^m \mathsf{ReLU}(\langle w_i^{(t)}, x \rangle + \rho_i - b^{(t)}) - \mathsf{ReLU}(-\langle w_i^{(t)}, x \rangle + \rho_i - b^{(t)}) \ .$$

**Fact A.1.** *If we denote by $\ell_t'(w^{(t)}; x, y, \rho) \stackrel{\text{def}}{=} \frac{d}{ds}[\log(1+e^s)]|_{s=-yf_t(w^{(t)};x,\rho)} = \frac{e^{-yf_t(w^{(t)};x,\rho)}}{1+e^{-yf_t(w^{(t)};x,\rho)}}$, then*

$$\nabla_{w_i} \mathbf{Loss}_t(w^{(t)}; x, y, \rho) = -y\ell_t'(w^{(t)}; x, y, \rho) \nabla_{w_i} f_t(w^{(t)}; x, \rho)$$
$$= -y\ell_t'(w^{(t)}; x, y, \rho) \left( \mathbb{1}_{\langle w_i^{(t)}, x \rangle + \rho_i \geq b^{(t)}} + \mathbb{1}_{-\langle w_i^{(t)}, x \rangle + \rho_i \geq b^{(t)}} \right) \cdot x$$

Let us also note that

$$\nabla \mathbf{Reg}(w_i) = (\|w_i\|_2 + 1) \cdot w_i$$

**Lemma A.2.** *Suppose $\mathcal{Z} = \{x^{(1)}, \ldots, x^{(N)}\}$ are i.i.d. samples from $\mathcal{D}$ and $y^{(i)} = y(x^{(i)})$, and suppose $N \geq \mathsf{poly}(d)$ for some sufficiently large polynomial. Let $f = f_t$ and suppose $b^{(t)} \leq \mathsf{poly}(d)$ and $\sigma_\rho \geq \frac{1}{\mathsf{poly}(d)}$. Then, for every $w_1, \ldots, w_N$ that may depend on the randomness of $\mathcal{Z}$ and satisfies $\|w_i\| \leq \mathsf{poly}(d)$, it satisfies*

$$\left| \frac{1}{N} \sum_{i \in [N]} \mathbb{E}_\rho \left[ \mathbf{Loss}(w; x^{(i)}, y^{(i)}, \rho) \right] - \mathbb{E}_{x \sim \mathcal{D}, y=y(x), \rho} \left[ \mathbf{Loss}(w; x, y, \rho) \right] \right| \leq \frac{1}{\mathsf{poly}(d)}$$

$$\left\| \frac{1}{N} \sum_{i \in [N]} \mathbb{E}_\rho \left[ \nabla_w \mathbf{Loss}(w; x^{(i)}, y^{(i)}, \rho) \right] - \mathbb{E}_{x \sim \mathcal{D}, y=y(x), \rho} \left[ \nabla_w \mathbf{Loss}(w; x, y, \rho) \right] \right\|_F \leq \frac{1}{\mathsf{poly}(d)}$$

*In addition, suppose for every $i \in [N]$, we have an i.i.d. random sample $\rho^{(i)} \sim \mathcal{N}(0, \sigma_\rho^2 \mathbf{I})$ that is independent of $\mathcal{Z}$ and $w$. Then, with probability at least $1 - e^{-\Omega(\log^2 d)}$ over $\rho$, we have*

$$\left| \frac{1}{N} \sum_{i \in [N]} \mathbb{E}_\rho \left[ \mathbf{Loss}(w; x^{(i)}, y^{(i)}, \rho) \right] - \frac{1}{N} \sum_{i \in [N]} \mathbf{Loss}(w; x^{(i)}, y^{(i)}, \rho^{(i)}) \right| \leq \frac{1}{\mathsf{poly}(d)}$$

$$\left\| \frac{1}{N} \sum_{i \in [N]} \mathbb{E}_\rho \left[ \nabla_w \mathbf{Loss}(w; x^{(i)}, y^{(i)}, \rho) \right] - \frac{1}{N} \sum_{i \in [N]} \nabla_w \mathbf{Loss}(w; x^{(i)}, y^{(i)}, \rho^{(i)}) \right\|_F \leq \frac{1}{\mathsf{poly}(d)}$$

*Proof.* The proof of the first part can be done by trivial VC dimension or Rademacher complexity arguments. For instance, the function $\mathbb{E}_\rho[\nabla \mathbf{Loss}(w; x, y, \rho)]$ is Lipschitz continuous in $w$ with Lipschitz parameter at most $\mathsf{poly}(d)$ (note that this relies on the fact that we take an expectation in $\rho$), and thus one can take an epsilon-net over all possible choices of $w$, and then apply a union bound over them.

The proof of the second part can be done by trivial Hoeffding bounds. □

## B  Neuron Structure and Initialization Properties

We consider $m = d^{1+c_0}$ for a very small constant $c_0 \in (0, 1)$, and consider constants $c_1 > c_2$ to be chosen shortly. Let us define a few notations to characterize each neuron's behavior.

**Definition B.1** (neuron characterization). *Recall $w_i^{(t)}$ is the weight for the $i$-th neuron at iteration $t$. We shall choose a parameter $\sigma_w^{(t)}$ at each iteration $t \geq 0$ and define the following notions. Consider any dimension $j \in [d]$.*

1. Let $\mathcal{S}_{j,sure}^{(t)} \subseteq [m]$ be those neurons $i \in [m]$ satisfying
    - $\langle w_i^{(t)}, \mathbf{M}_j \rangle^2 \geq (c_1 + c_2)(\sigma_w^{(t)})^2 \log d$,
    - $\langle w_i^{(t)}, \mathbf{M}_{j'} \rangle^2 < (c_1 - c_2)(\sigma_w^{(t)})^2 \log d$      for every $j' \neq j$,
    - $\mathsf{sign}(\langle w_i^{(t)}, \mathbf{M}_j \rangle) = \mathsf{sign}(w_j^\star)$.

2. Let $\mathcal{S}_{j,pot}^{(t)} \subseteq [m]$ be those neurons $i \in [m]$ satisfying
    - $\langle w_i^{(t)}, \mathbf{M}_j \rangle^2 \geq (c_1 - c_2)(\sigma_w^{(t)})^2 \log d$

3. Let $\mathcal{S}_{ept}^{(t)} \subseteq [m]$ be the set of neurons $i \in [m]$ satisfying
    - $\|w_i^{(t)}\|_2^2 \leq 2(\sigma_w^{(t)})^2 d$
    - $\langle w_i^{(t)}, \mathbf{M}_j \rangle^2 \geq (c_1 - c_2)(\sigma_w^{(t)})^2 \log d$      for at most $O(1)$ many $j \in [d]$.
    - $\langle w_i^{(t)}, \mathbf{M}_j \rangle^2 \geq 2(\sigma_w^{(t)})^2 \sqrt{\log d}$      for at most $2^{-\sqrt{\log d}} d$ many $j \in [d]$.
    - $|\langle w_i^{(t)}, \mathbf{M}_j \rangle| \leq \frac{\sigma_w^{(t)}}{\log d}$      for at least $\Omega(\frac{d}{\log d})$ many $j \in [d]$.

**Lemma B.2** (geometry at initialization). *Suppose each $w_i^{(0)} \sim \mathcal{N}(0, \sigma_0^2 \mathbf{I})$ and suppose $\sigma_w^{(0)} = \sigma_0$. For every constants $c_0 \in (0,1)$ and $\gamma \in (0, 0.1)$, by choosing $c_1 = 2 + 2(1-\gamma)c_0$ and $c_2 = \gamma c_0$, we have with probability $\geq 1 - o(1/d^3)$ over the random initialization, for all $j \in [d]$:*

$$|\mathcal{S}_{j,sure}^{(0)}| = \Omega\left(d^{\frac{\gamma}{4} c_0}\right) =: \Xi_1 \qquad |\mathcal{S}_{j,pot}^{(0)}| \leq O\left(d^{2\gamma c_0}\right) =: \Xi_2 \qquad \mathcal{S}_{ept}^{(0)} = [m]$$

> *Remark.* In the rest of the paper, we shall assume Lemma B.2 holds in all upper-bound related theorems/lemmas, and for simplicity, we assume $\gamma > 0$ is some small constant so that $(\Xi_2)^{100} \leq d^{c_0}$. The notations $\Xi_1$ and $\Xi_2$ shall be used throughout the paper.

**Definition B.3** (neuron characterization, continued). *Recall $b^{(t)}$ is the bias at iteration $t$, and let us introduce more notions.*

1. Let $\mathcal{S}_{ept+}^{(t)} \subseteq [m]$ be the set of neurons $i \in [m]$ satisfying
    - $\|w_i^{(t)}\|_2^2 \leq \frac{(\sigma_w^{(t)})^2 d}{\log^2 d}$,
    - $\left|\langle w_i^{(t)}, \mathbf{M}_j \rangle\right| \geq \frac{\sigma_w^{(t)}}{\log d}$      for at most $O(1)$ many $j \in [d]$.

2. Let $\mathcal{S}_{ept++}^{(t)} \subseteq [m]$ be the set of neurons $i \in [m]$ satisfying
    - $\|w_i^{(t)}\|_2^2 \leq \frac{(\sigma_w^{(t)})^2}{\beta^2}$      for $\beta \stackrel{\mathsf{def}}{=} \frac{1}{\sqrt{k}\Xi_2^{10}}$.

3. Let $\mathcal{S}_{j,pot+}^{(t)} \subseteq [m]$ be the set of neurons $i \in [m]$ satisfying
    - $|\langle w_i^{(t)}, \mathbf{M}_j \rangle| \geq \frac{k}{d\beta} b^{(t)}$.

4. Let $\mathcal{S}_{j,sure+}^{(t)} \subseteq [m]$ be the set of neurons $i \in [m]$ satisfying
    - $\langle w_i^{(t)}, \mathbf{M}_j \rangle^2 \geq 4k(b^{(t)})^2$,
    - $\mathsf{sign}(\langle w_i^{(t)}, \mathbf{M}_j \rangle) = \mathsf{sign}(w_j^\star)$.

Note that we do not have good properties on $\mathcal{S}_{ept+}^{(t)}$, $\mathcal{S}_{ept++}^{(t)}$, $\mathcal{S}_{j,pot+}^{(t)}$ or $\mathcal{S}_{j,sure+}^{(t)}$ at initialization $t = 0$; however, they will gradually begin to satisfy certain properties as the training process goes. See Section C for details.

## B.1 Proof of Lemma B.2

*Proof of Lemma B.2.* Recall if $g$ is standard Gaussian, then for every $t > 0$,

$$\frac{1}{\sqrt{2\pi}} \frac{t}{t^2+1} e^{-t^2/2} < \Pr_{g \sim \mathcal{N}(0,1)}[g > t] < \frac{1}{\sqrt{2\pi}} \frac{1}{t} e^{-t^2/2}$$

Therefore, for every $i \in [m]$ and $j \in [d]$,

- $p_1 = \mathbf{Pr}[\langle w_i^{(0)}, \mathbf{M}_j \rangle^2 \geq (c_1 + c_2)\sigma_0^2 \log d] = \Theta(\frac{1}{\log d}) \cdot \frac{1}{d^{(c_1+c_2)/2}} = \Theta(\frac{1}{\sqrt{\log d}}) \cdot \frac{1}{d \cdot d^{(1-\gamma/2)c_0}}$
- $p_2 = \mathbf{Pr}[\langle w_i^{(0)}, \mathbf{M}_j \rangle^2 \geq (c_1 - c_2)\sigma_0^2 \log d] = \Theta(\frac{1}{\log d}) \cdot \frac{1}{d^{(c_1-c_2)/2}} = \Theta(\frac{1}{\sqrt{\log d}}) \cdot \frac{1}{d \cdot d^{(1-3\gamma/2)c_0}}$

1. We first lower bound $|\mathcal{S}_{j,sure}^{(0)}|$. For every $i \in [m]$, with probability at least $p_1/2 \cdot (1-p_2)^{d-1} \geq \Omega(\frac{1}{\sqrt{\log d}}) \cdot \frac{d^{\frac{\gamma}{2}c_0}}{m}$ it satisfies

$$\langle w_i^{(0)}, \mathbf{M}_j \rangle^2 \geq (c_1+c_2)\sigma_0^2 \log d, \quad \mathsf{sign}(\langle w_i^{(0)}, \mathbf{M}_j \rangle)\mathsf{sign}(w_j^\star) \geq 0$$
$$\forall j' \neq j, \langle w_i^{(0)}, \mathbf{M}_{j'} \rangle^2 \leq (c_1-c_2)\sigma_0^2 \log d$$

By concentration with respect to all $m$ choices of $i \in [m]$, we know with probability at least $1 - o(\frac{1}{d^3})$ it satisfies $|\mathcal{S}_{j,sure}^{(0)}| = \Omega\left(d^{\frac{\gamma}{4}c_0}\right)$.

2. We next upper bound $|\mathcal{S}_{j,pot}^{(0)}|$. For every $i \in [m]$, with probability at most $p_2 < O(\frac{1}{\sqrt{\log d}}) \cdot \frac{d^{\frac{3\gamma}{2}c_0}}{m}$ it satisfies

$$\langle w_i^{(0)}, \mathbf{M}_j \rangle^2 \geq (c_1-c_2)\sigma_0^2 \log d$$

By concentration with respect to all $m$ choices of $i$, we know with probability at least $1 - o(\frac{1}{d^3})$ it satisfies $|\mathcal{S}_{j,pot}^{(0)}| = O(d^{2\gamma c_0})$.

3. As for $\mathcal{S}_{ept}^{(0)}$, we first note that for every $i \in [m]$, by chi-square distribution's tail bound, with probability at least $1 - o(1/d^3)$ it satisfies $\|w_i^{(0)}\|_2^2 \in \left[\frac{\sigma_0^2 d}{2}, 2\sigma_0^2 d\right]$.

For every $i \in [m]$, the probability of existing $q = 20/c_0$ different

$$j_1, \cdots, j_q \in [d] : s.t. \forall r \in [q] : \langle w_i^{(0)}, \mathbf{M}_{j_r} \rangle^2 \geq (c_1-c_2)\sigma_0^2 \log d$$

is at most $d^q \cdot (p_2)^q \leq d^{-q \cdot \frac{c_0}{2}} \leq \frac{1}{d^4 m}$. Union bounding over all possible $i \in [m]$ gives the proof that, with probability at least $1 - 1/d^4$, for all but at most $q = O(1)$ values of $j \in [d]$, it satisfies $\langle w_i^{(0)}, \mathbf{M}_j \rangle^2 < (c_1-c_2)(\sigma_w^{(t)})^2 \log d$.

For every $i \in [m]$ and $j \in [d]$, with probability at least $1 - e^{-\sqrt{\log d}}$ it satisfies $\langle w_i^{(0)}, \mathbf{M}_j \rangle^2 \leq 2\sigma_0^2 \sqrt{\log d}$. Therefore, with probability at least $1 - o(1/d^3)$, there are $\geq d(1 - 2^{-\sqrt{\log d}})$ indices $j \in [d]$ satisfying $\langle w_i^{(0)}, \mathbf{M}_j \rangle^2 \leq 2\sigma_0^2 \sqrt{\log d}$.

For every $i \in [m]$ and $j \in [d]$, with probability at least $\frac{1}{40000\sqrt{\log d}}$ it satisfies $|\langle w_i^{(0)}, \mathbf{M}_j \rangle| \leq \frac{\sigma_0}{10000\sqrt{\log d}}$. Therefore, with probability at least $1 - o(1/d^3)$, there are $\frac{1}{100000\sqrt{\log d}}$ indices $j \in [d]$ satisfying $|\langle w_i^{(0)}, \mathbf{M}_j \rangle| \leq \frac{\sigma_0}{10000\sqrt{\log d}}$. □

## C  Neuron Structure Change During Training

For analysis purpose, we consider two phases during training. In Phase I, the neurons have moved so little so that the accuracy remains 50% for binary classification; however, some neurons shall start to win lottery and form "singleton" structures. We summarize this as the following theorem.

**Theorem C.1** (phase I). *Suppose the high-probability initialization event in Lemma B.2 holds. Suppose $\eta, \sigma_0 \in (0, \frac{1}{\mathsf{poly}(d)})$ and $N \geq \mathsf{poly}(d)$. With probability at least $1 - e^{-\Omega(\log^2 d)}$, the following holds for all $t \leq T_{\mathsf{b}} \stackrel{\mathsf{def}}{=} \Theta\left(\frac{d^2 \sigma_0}{k\eta}\right)$*

1. $\mathcal{S}_{j,sure}^{(0)} \subseteq \mathcal{S}_{j,sure}^{(t)}$ for every $j \in [d]$.
2. $\mathcal{S}_{j,pot}^{(0)} \supseteq \mathcal{S}_{j,pot}^{(t)}$ for every $j \in [d]$.
3. $\mathcal{S}_{ept}^{(t)} = [m]$
4. $\mathcal{S}_{ept+}^{(t)} = [m]$ for every $t \geq T_{\mathsf{a}} \stackrel{\mathsf{def}}{=} \Theta\left(\frac{d\sigma_0 \log^{2.5} d}{\eta}\right)$.
5. $\mathcal{S}_{ept++}^{(t)} = [m]$ and $\mathcal{S}_{j,pot}^{(0)} \supseteq \mathcal{S}_{j,pot+}^{(t)}$ for every $j \in [d]$ at this iteration $t = T_{\mathsf{b}}$.

(Recall according to Definition B.1 we have $\mathcal{S}_{j,sure}^{(t)} \subseteq \mathcal{S}_{j,pot}^{(t)}$.)

In Phase II, the neurons start to move much more so that the network output becomes more meaningful; in phase II, the "singleton" neurons become even more singleton.

**Theorem C.2** (phase II). *In the same setting as Theorem C.1, with probability at least $1 - e^{-\Omega(\log^2 d)}$, the following holds for all $t \in \left[T_{\mathsf{b}}, d^{O(\log d)}/\eta\right]$.*

1. $\mathcal{S}_{ept++}^{(t)} = \mathcal{S}_{ept+}^{(t)} = [m]$.
2. $\mathcal{S}_{j,sure}^{(0)} \subseteq \mathcal{S}_{j,sure}^{(t)}$ for every $j \in [d]$.
3. $\mathcal{S}_{j,pot}^{(0)} \supseteq \mathcal{S}_{j,pot+}^{(t)} \supseteq \mathcal{S}_{j,pot}^{(t)}$ for every $j \in [d]$.
4. $\mathcal{S}_{j,sure}^{(0)} \subseteq \mathcal{S}_{j,sure+}^{(t)}$ for every $j \in [d]$, as long as $t \geq T_{\mathsf{e}} \stackrel{\mathsf{def}}{=} \Theta\left(\frac{d}{\eta \Xi_2 \log d}\right)$.

(Recall according to Definition B.3 we have $\mathcal{S}_{j,sure+}^{(t)} \subseteq \mathcal{S}_{j,pot}^{(t)}$.)

*Remark* C.3. Theorem C.2 immediately implies the first claim of Theorem 5.1 and the first claim of Theorem 5.2, after plugging in the definitions of those neuron structure sets introduced in Section B. For instance, we can write

$$w_i^{(t)} = \sum_{j \in [d]:\, i \in \mathcal{S}_{j,pot}^{(0)}} \langle w_i^{(t)}, \mathbf{M}_j \rangle \mathbf{M}_j + \sum_{j \in [d]:\, i \notin \mathcal{S}_{j,pot}^{(0)}} \langle w_i^{(t)}, \mathbf{M}_j \rangle \mathbf{M}_j$$

We make several observations, when $t \geq T_{\mathsf{e}}$:

- $\mathcal{S}_{ept}^{(t)} = [m]$ implies the cardinality of $\{j \in [d] \colon i \in \mathcal{S}_{j,pot}^{(0)}\}$ is $\leq O(1)$, so we can define it as $\mathcal{N}_i$.
- For every $i \notin \mathcal{S}_{j,pot}^{(0)}$, it also satisfies $i \notin \mathcal{S}_{j,pot+}^{(t)}$, so we have $|\langle w_i^{(t)}, \mathbf{M}_j \rangle| < \frac{k}{d\beta} b^{(t)} \leq \frac{k\Xi_2^2}{d}$ (using $b^{(t)} \leq \beta \Xi_2^2$ from Definition C.12).
- For every $i \in \mathcal{S}_{j,sure+}^{(t)} \subseteq \mathcal{S}_{j,pot}^{(0)}$, we have $\langle w_i^{(t)}, \mathbf{M}_j \rangle \cdot \mathsf{sign}(w_i^*) \geq 2\sqrt{k} b^{(t)} > \frac{1}{\Xi_2^8}$ (using $b^{(t)} = \beta \Xi_2^2$ from Definition C.12 and our choice of $\beta$).

They together imply the first claim of Theorem 5.2. One can similarly derive the first claim of Theorem 5.1.

## C.1 Auxiliary Lemma 1: Geometry of Crossing Boundary

We present a lemma to bound the size of the pre-activation signal.

**Lemma C.4** (pre-activation signal size). *For every $t \geq 0$, every $i \in [m]$, every $\lambda \geq 0$, every $j \in [d]$:*

(a) *If $i \in \mathcal{S}_{ept}^{(t)}$ then*

$$\mathbf{Pr}_{z,\xi}\left[\left\langle w_i^{(t)}, \sum_{j' \neq j} \mathbf{M}_{j'} z_{j'} + \xi \right\rangle^2 \geq \lambda^2 (\sigma_w^{(t)})^2 \right] \leq e^{-\Omega(\frac{\lambda}{\log^{1/4} d})} + e^{-\log^{1/4} d}$$

(b) *If $i \in \mathcal{S}_{ept+}^{(t)}$ then*

$$\mathbf{Pr}_{z,\xi}\left[\left\langle w_i^{(t)}, \sum_{j' \neq j} \mathbf{M}_{j'} z_{j'} + \xi \right\rangle^2 \geq \lambda^2 (\sigma_w^{(t)})^2 \right] \leq e^{-\Omega(\lambda \log d)} + e^{-\Omega(\lambda^2 \log d)} + O\left(\frac{k}{d}\right)$$

*Proof of Lemma C.4a.* Let $\mathcal{E}$ be the event where there exists $j' \in [d]$ with $|z_{j'}| \geq \frac{1}{\log^2 d}$ and $\langle w_i^{(t)}, \mathbf{M}_{j'} \rangle^2 \geq 2(\sigma_w^{(t)})^2 \sqrt{\log d}$. Since $\mathbb{E}[z_{j'}^2] = O\left(\frac{1}{d}\right)$, we know that $\mathbf{Pr}\left[|z_{j'}| \geq \frac{1}{\log^2 d}\right] \leq O\left(\frac{\log^4 d}{d}\right)$. By the definition of $\mathcal{S}_{ept}^{(t)}$ and union bound, we know

$$\mathbf{Pr}[\mathcal{E}] \leq O\left(\frac{\log^4 d}{d}\right) \times 2^{-\sqrt{\log d}} d \leq e^{-\log^{1/4} d}$$

Let $\mathcal{F}$ be the event where there exists $j' \in [d]$ with $z_{j'} \neq 0$ and $\langle w_i^{(t)}, \mathbf{M}_{j'} \rangle^2 \geq \Omega((\sigma_w^{(t)})^2 \log d)$. Again, by the definition of $\mathcal{S}_{ept}^{(t)}$, we know that

$$\mathbf{Pr}[\mathcal{F}] \leq O\left(\frac{k}{d}\right) \leq e^{-\log^{1/4} d}$$

Thus, when neither $\mathcal{E}$ or $\mathcal{F}$ happens, we have for every $j' \in [d]$:

$$\langle w_i^{(t)}, \mathbf{M}_{j'} z_{j'} \rangle^2 \leq \min\left\{2(\sigma_w^{(t)})^2 \sqrt{\log d} \cdot 1, O((\sigma_w^{(t)})^2 \log d) \cdot \frac{1}{\log^4 d}\right\} \leq 2(\sigma_w^{(t)})^2 \sqrt{\log d}$$

At the same time, we also have

$$\sum_{j' \in [d]} \mathbb{E}_{z_{j'}} \langle w_i^{(t)}, \mathbf{M}_{j'} z_{j'} \rangle^2 \leq \sum_{j' \in [d]} O(\frac{1}{d}) \langle w_i^{(t)}, \mathbf{M}_{j'} \rangle^2 \leq O((\sigma_w^{(t)})^2)$$

Apply Bernstein concentration bound we complete the proof that

$$\mathbf{Pr}_{z,\xi}\left[\left\langle w_i^{(t)}, \sum_{j' \neq j} \mathbf{M}_{j'} z_{j'} \right\rangle^2 \geq \frac{\lambda^2}{2} (\sigma_w^{(t)})^2 \right] \leq e^{-\Omega(\frac{\lambda}{\log^{1/4} d})} + e^{-\log^{1/4} d}$$

Finally, for the $\xi$ part, let us recall $\langle w_i^{(t)}, \xi \rangle$ variable with variance at most $O(\frac{\|w_i^{(t)}\|^2 \sigma_x^2}{d}) \leq O((\sigma_w^{(t)})^2)$ and each $|\langle w_i^{(t)}, \mathbf{M}_j \rangle \langle \mathbf{M}_j, \xi \rangle| \leq \frac{\sigma_w^{(t)}}{\log^2 d}$ w.h.p. Using Bernstein concentration of random variables, we finish the proof. $\square$

*Proof of Lemma C.4b.* Let $\mathcal{F}$ be the event where there exists $j' \in [d]$ with $z_{j'} \neq 0$ and $|\langle w_i^{(t)}, \mathbf{M}_{j'} \rangle| \geq$

$\Omega(\frac{\sigma_w^{(t)}}{\log d})$. By the definition of $\mathcal{S}_{ept+}^{(t)}$, we know that

$$\mathbf{Pr}[\mathcal{F}] \leq O\left(\frac{k}{d}\right)$$

When $\mathcal{F}$ does not happen, we have for every $j' \in [d]$: $|\langle w_i^{(t)}, \mathbf{M}_{j'} z_{j'}\rangle| \leq O(\frac{\sigma_w^{(t)}}{\log d})$ and at the same time

$$\sum_{j' \in [d]} \mathbb{E}_{z_{j'}} \langle w_i^{(t)}, \mathbf{M}_{j'} z_{j'}\rangle^2 \leq \sum_{j' \in [d]} O(\frac{1}{d}) \langle w_i^{(t)}, \mathbf{M}_{j'}\rangle^2 \leq O(\frac{(\sigma_w^{(t)})^2}{\log d})$$

Apply Bernstein concentration bound we complete the proof that

$$\mathbf{Pr}_{z,\xi}\left[\left\langle w_i^{(t)}, \sum_{j' \neq j} \mathbf{M}_{j'} z_{j'} \right\rangle^2 \geq \frac{\lambda^2}{2}(\sigma_w^{(t)})^2\right] \leq e^{-\Omega(\lambda \log d)} + e^{-\Omega(\lambda^2 \log d)} + O\left(\frac{k}{d}\right)$$

Finally, for the $\xi$ part, let us recall $\langle w_i^{(t)}, \xi\rangle$ is a random variable with variance at most $O(\frac{\|w_i^{(t)}\|^2 \sigma_x^2}{d}) \leq O(\frac{(\sigma_w^{(t)})^2}{\log^2 d})$. Using the Bernstein concentration bound, we finish the proof. $\square$

## C.2 Auxiliary Lemma 2: A Critical Lemma for Gradient Bound

In this section we present a critical lemma that shall be used multiple times to bound the gradient in many of the following sections. Recall $\frac{c_2}{c_1} \in (0, 0.1)$ is a constant from Lemma B.2.

**Lemma C.5** (critical). *Let $Y(z, S_1): \mathbb{R} \times \mathbb{R}^p \to [-1, 1]$ be a center symmetric function, meaning $Y(z, S_1) = Y(-z, -S_1)$. Let $S_1 \in \mathbb{R}^p$ and $S_2 \in \mathbb{R}$ be random variables, where $(S_1, S_2)$ is symmetrically distributed, meaning $(S_1, S_2)$ distributes the same as $(-S_1, -S_2)$.*

*For every $\alpha > 0$, suppose $\rho \sim \mathcal{N}(0, \sigma_\rho^2)$, define quantity*

$$\Delta := \mathbb{E}_{S_1, S_2, \rho}\left[Y(1, S_1)\mathbb{1}_{\alpha+S_2+\rho \geq b} - Y(-1, S_1)\mathbb{1}_{-\alpha+S_2+\rho \geq b}\right]$$

*and define parameters $V := \mathbb{E}[(S_2)^2]$ and $L := \mathbb{E}_{S_1}[|Y(1, S_1) - Y(0, S_1)|]$. Then,*

*(a) it always satisfies $|\Delta| \leq O\left(\frac{\sqrt{V}}{\sigma_\rho} + L\right)$*

*(b) if $Y(z, S_1)$ is a monotonically non-decreasing in $z \in \mathbb{R}$ for every $S_1 \in \mathbb{R}^p$, then $\Delta \geq -\Omega(\frac{\sqrt{V}}{\sigma_\rho})$*

*Furthermore, suppose we can write $S_1 = (S_1', S_1'')$ and $S_2 = S_2' + S_2''$ for $(S_1', S_2')$ and $(S_1'', S_2'')$ being independent (although $S_1', S_2'$ may be dependent, and $S_1'', S_2''$ may be dependent). Then, we have*

*(c) if $\alpha \leq b(1 - \frac{c_2}{2c_1})$, then $|\Delta| \leq \left(e^{-\Omega(b^2/\sigma_\rho^2)} + \Gamma\right)\left(\min\{1, O(\frac{\alpha}{\sigma_\rho})\} + L_y\right) + \Gamma_y$*

*where the parameters*

- $\Gamma := \mathbf{Pr}\left[|S_2| \geq \frac{c_2}{10c_1} \cdot b\right]$ and $\Gamma_y := \mathbf{Pr}\left[|S_2''| \geq \frac{c_2}{10c_1} \cdot b\right]$
- $L_y := \max_{S_1'}\left\{\mathbb{E}_{S_1''}[|Y(1, S_1', S_1'') - Y(0, S_1', S_1'')|]\right\} \leq 1$

*Proof of Lemma C.5.* We first focus on $Y(1, S_1)\mathbb{1}_{\alpha+S_2+\rho \geq b}$, and write

$$Y(1, S_1)\mathbb{1}_{\alpha+S_2+\rho \geq b} = Y(0, S_1)\mathbb{1}_{\alpha+S_2+\rho \geq b} + \big(Y(1, S_1) - Y(0, S_1)\big)\mathbb{1}_{\alpha+S_2+\rho \geq b} \tag{C.1}$$

$$= Y(0, S_1)\mathbb{1}_{\alpha+S_2+\rho \geq b} \pm |Y(1, S_1) - Y(0, S_1)|\mathbb{1}_{\alpha+S_2+\rho \geq b} \tag{C.2}$$

Focusing on the term $Y(0, S_1)\mathbb{1}_{\alpha+S_2+\rho \geq b}$, by the symmetric properties of $Y$ and $(S_1, S_2)$, we have

$$|\mathbb{E}[Y(0, S_1)\mathbb{1}_{\alpha+S_2+\rho \geq b}]| = \frac{1}{2} \left|\mathbb{E}\left[Y(0, S_1)\mathbb{1}_{\alpha+S_2+\rho \geq b} + Y(0, -S_1)\mathbb{1}_{\alpha-S_2+\rho \geq b}\right]\right|$$

$$= \frac{1}{2} \left|\mathbb{E}\left[Y(0, S_1)\left(\mathbb{1}_{\alpha+S_2+\rho \geq b} - \mathbb{1}_{\alpha-S_2+\rho \geq b}\right)\right]\right|$$

$$\leq \mathbf{Pr}[\mathbb{1}_{\alpha+S_2+\rho \geq b} \neq \mathbb{1}_{\alpha-S_2+\rho \geq b}]$$

$$\leq \mathbf{Pr}[\rho \in [b - \alpha - S_2, b - \alpha + S_2]]$$

$$= O\left(\frac{\mathbb{E}[|S_2|]}{\sigma_\rho}\right) = O\left(\frac{\sqrt{\mathbb{E}[S_2^2]}}{\sigma_\rho}\right) = O\left(\frac{\sqrt{V}}{\sigma_\rho}\right) \quad \text{(C.3)}$$

Putting (C.3) into (C.2), applying the bound $L = \mathbb{E}_{S_1}[|Y(1, S_1) - Y(0, S_1)|]$, and repeating the same analysis for $Y(-1, S_1)\mathbb{1}_{-\alpha+S_2+\rho \geq b}$ gives

$$|\Delta| \leq O\left(\frac{\sqrt{V}}{\sigma_\rho} + L\right).$$

This proves Lemma C.5a. When $Y(z, S_1)$ is a monotone non-decreasing in $z \in \mathbb{R}$, we have

$$Y(1, S_1)\mathbb{1}_{\alpha+S_2+\rho \geq b} \geq Y(0, S_1)\mathbb{1}_{\alpha+S_2+\rho \geq b}$$
$$Y(-1, S_1)\mathbb{1}_{-\alpha+S_2+\rho \geq b} \leq Y(0, S_1)\mathbb{1}_{-\alpha+S_2+\rho \geq b}$$

so we can go back to (C.1) (and repeating for $Y(-1, S_1)$) to derive that

$$\Delta \geq -\Omega\left(\frac{\sqrt{V}}{\sigma_\rho}\right)$$

This proves Lemma C.5b. Finally, when $\alpha \leq b(1 - \frac{c_2}{2c_1})$, we can bound $\Delta$ differently

$$|\Delta| \leq 2\mathbf{Pr}[\mathbb{1}_{\alpha+S_2+\rho \geq b} \neq \mathbb{1}_{-\alpha+S_2+\rho \geq b}] + \mathbb{E}\left[|Y(1, S_1) - Y(-1, S_1)| \cdot \mathbb{1}_{\alpha+S_2+\rho \geq b}\right]$$

$$= 2\mathbf{Pr}[\rho \in [b - S_2 - \alpha, b - S_2 + \alpha]] + \mathbb{E}\left[|Y(1, S_1) - Y(-1, S_1)| \cdot \mathbb{1}_{\alpha+S_2+\rho \geq b}\right] \quad \text{(C.4)}$$

To bound the first term in (C.4) we consider two cases.:

- when $|S_2| \leq \frac{b}{4}$, we have $\mathbf{Pr}_\rho[\rho \in [b - S_2 - \alpha, b - S_2 + \alpha]] \leq \min\{1, \frac{\alpha}{\sigma_\rho}\}e^{-\Omega(b^2/\sigma_\rho^2)}$;

- when $|S_2| \geq \frac{b}{4}$ (happening w.p. $\leq \Gamma$), we have $\mathbf{Pr}_\rho[\rho \in [b - S_2 - \alpha, b - S_2 + \alpha]] \leq \min\{1, O(\frac{\alpha}{\sigma_\rho})\}$.

Putting together, we know that

$$\mathbf{Pr}[\rho \in [b - S_2 - \alpha, b - S_2 + \alpha]] \leq \min\{1, \frac{\alpha}{\sigma_\rho}\} \cdot \left(e^{-\Omega(b^2/\sigma_\rho^2)} + \Gamma\right) \quad \text{(C.5)}$$

To bound the second term in (C.4), first recall $S_1 = (S_1', S_1'')$ and $S_2 = S_2' + S_2''$, so we can write

$$\mathbb{E}\left[|Y(1, S_1', S_1'') - Y(-1, S_1', S_1'')| \cdot \mathbb{1}_{\alpha+S_2'+S_2''+\rho \geq b}\right]$$

$$\leq \mathbb{E}\left[|Y(1, S_1', S_1'') - Y(-1, S_1', S_1'')| \cdot \left(\mathbb{1}_{|\alpha+S_2'+\rho| \geq (1-\frac{c_2}{10c_1}) \cdot b} + \mathbb{1}_{|S_2''| \geq \frac{c_2}{10c_1} \cdot b}\right)\right]$$

$$\leq \mathbb{E}\left[|Y(1, S_1', S_1'') - Y(-1, S_1', S_1'')| \cdot \mathbb{1}_{|\alpha+S_2'+\rho| \geq (1-\frac{c_2}{10c_1}) \cdot b}\right] + \Gamma_y \quad \text{(C.6)}$$

To bound the first term in (C.6), we can take expectation over $S_1''$ and use the bound $L_y$ to derive

$$\mathbb{E}\left[|Y(1, S_1', S_1'') - Y(-1, S_1', S_1'')| \cdot \mathbb{1}_{|\alpha+S_2'+\rho| \geq (1-\frac{c_2}{10c_1}) \cdot b}\right] \leq L_y \mathbf{Pr}\left[|\alpha + S_2' + \rho| \geq (1 - \frac{c_2}{10c_1}) \cdot b\right]$$

but since $\alpha \leq (1 - \frac{c_2}{2c_1}) \cdot b$ and $S_2' = S_2 - S_2''$, we can further bound

$$\mathbf{Pr}\left[|\alpha + S_2' + \rho| \geq (1 - \frac{c_2}{10c_1}) \cdot b\right] \leq \mathbf{Pr}\left[|S_2''| \geq \frac{c_2}{10c_1} \cdot b\right] + \mathbf{Pr}\left[|S_2| \geq \frac{c_2}{10c_1} \cdot b\right] + \mathbf{Pr}\left[\rho \geq \frac{c_2}{10c_1} \cdot b\right]$$

$$\leq \Gamma_y + \Gamma + e^{-\Omega(b^2/\sigma_\rho^2)}$$

Putting these back to (C.6), we have

$$\mathbb{E}\left[|Y(1, S_1', S_1'') - Y(-1, S_1', S_1'')|\mathbb{1}_{\alpha + S_2' + S_2'' + \rho \geq b}\right] \leq \left(e^{-\Omega(b^2/\sigma_\rho^2)} + \Gamma + \Gamma_y\right) L_y + \Gamma_y \quad (C.7)$$

Putting (C.5) and (C.7) back to (C.4), we conclude the when $\alpha \leq b(1 - \frac{c_2}{2c_1})$, we have

$$|\Delta| \leq \left(e^{-\Omega(b^2/\sigma_\rho^2)} + \Gamma\right)\left(O(\frac{\alpha}{\sigma_\rho}) + L_y\right) + \Gamma_y \qquad \square$$

## C.3 Phase I: Winning lottery tickets near initialization

**Definition C.6.** *In Phase I, we have two sub-phases:*

- *In Phase I.1, we pick $b^{(t)} = \sqrt{c_1}\sigma_w^{(t)}\sqrt{\log d}$ and $\sigma_\rho^{(t)} = \sigma_w^{(t)}(\log\log\log d)^3$*
  
  *We grow $b^{(t+1)} = b^{(t)} + \frac{C\eta}{d}$ for $T_{\mathsf{a}} = \Theta\left(\frac{d\sigma_0 \log^{2.5} d}{\eta}\right)$ iterations.*

- *In Phase I.2, we pick $b^{(t)} = \sqrt{c_1}\sigma_w^{(t)}\sqrt{\log d}$ and $\sigma_\rho^{(t)} = \sigma_w^{(t)} \cdot \frac{(\log\log\log d)^3}{\sqrt{\log d}}$*
  
  *We grow $b^{(t+1)} = b^{(t)} + \frac{C\eta}{d}$ for $T_{\mathsf{b}} = \Theta\left(\frac{d^2 \sigma_0}{k\eta}\right)$ iterations.*

### C.3.1 Activation Probability

Recall $x = \sum_j \mathbf{M}_j z_j + \xi$. Recall also $\frac{c_2}{c_1} \in (0, 0.1)$ is a constant from Lemma B.2.

**Lemma C.7** (activation probability). *We define $\Gamma_t$ to be any value such that*

- $\mathbf{Pr}_x\left[|\langle w_i^{(t)}, \sum_{j' \neq j} \mathbf{M}_{j'} z_{j'} + \xi\rangle| \geq \frac{c_2}{10c_1} b^{(t)}\right] \leq \Gamma_t$ *for every $i \in [m]$ and $j \in [d]$;*

- $\mathbf{Pr}_x\left[|\langle w_i^{(t)}, x\rangle| \geq \frac{c_2}{10c_1} b^{(t)}\right] \leq \Gamma_t$ *for every $i \in [m]$*

- $\mathbf{Pr}_x\left[|\rho_i| \geq \frac{c_2}{10c_1} b^{(t)}\right] \leq \Gamma_t$ *for every $i \in [m]$*

*We define $\Gamma_{t,y}$ to be any value such that*

- *for every $i \in [m]$ and $j \in [d]$, there exists $\Lambda \subseteq [d] \setminus \{j\}$ with $|\Lambda| \geq \Omega(\frac{d}{\sqrt{\log d}})$ satisfying*

$$\mathbf{Pr}_x\left[\left|\langle w_i^{(t)}, \sum_{j' \in \Lambda} \mathbf{M}_{j'} z_{j'}\rangle\right| \geq \frac{c_2}{10c_1} b^{(t)}\right] \leq \Gamma_{t,y}$$

*Then,*

- *If we are in Phase I.1 and $\mathcal{S}_{ept}^{(t)} = [m]$, then we can choose $\Gamma_t = e^{-\Omega(\log^{1/4} d)}$ and $\Gamma_{t,y} = \frac{1}{d^{10}}$.*

- *If we are in Phase I.2 and $\mathcal{S}_{ept+}^{(t)} = [m]$, then we can choose $\Gamma_t = O(\frac{k}{d})$ and $\Gamma_{t,y} = \frac{1}{d^{10}}$.*

*Proof.* Recall $b^{(t)} = \Theta(\sigma_w^{(t)}\sqrt{\log d})$. Applying Lemma C.4a and Lemma C.4b we immediately have

- If $\mathcal{S}_{ept}^{(t)} = [m]$, then $\mathbf{Pr}_x\left[\left|\langle w_i^{(t)}, \sum_{j' \neq j} \mathbf{M}_{j'} z_{j'} + \xi\rangle\right| \geq \frac{c_2}{10c_1} b^{(t)}\right] \leq e^{-\Omega(\log^{1/4} d)}$;

- If $\mathcal{S}_{ept+}^{(t)} = [m]$, then $\mathbf{Pr}_x\left[\left|\langle w_i^{(t)}, \sum_{j' \neq j} \mathbf{M}_{j'} z_{j'} + \xi\rangle\right| \geq \frac{c_2}{10c_1} b^{(t)}\right] \leq O\left(\frac{k}{d}\right)$.

Now, recall $x = \sum_{j' \in [d]} \mathbf{M}_{j'} z_{j'} + \xi$ so it differs from $\sum_{j' \neq j} \mathbf{M}_{j'} z_{j'} + \xi$ only by one term. Therefore, we have the same bound on $\mathbf{Pr}_x \left[ \left| \langle w_i^{(t)}, x \rangle \right| \geq \frac{c_2}{10c_1} b^{(t)} \right]$ by modifying the statements of Lemma C.4a and Lemma C.4b (without changing the proofs) to include this missing term.

- If $\mathcal{S}_{ept}^{(t)} = [m]$, $\mathbf{Pr}_{x,\rho} \left[ \left| \langle w_i^{(t)}, x \rangle \right| \geq \frac{c_2}{10c_1} b^{(t)} \right] \leq e^{-\Omega(\log^{1/4} d)}$
- If $\mathcal{S}_{ept+}^{(t)} = [m]$, $\mathbf{Pr}_{x,\rho} \left[ \left| \langle w_i^{(t)}, x \rangle \right| \geq \frac{c_2}{10c_1} b^{(t)} \right] \leq O\left(\frac{k}{d}\right)$

At the same time, using $\rho_i \sim \mathcal{N}(0, (\sigma_\rho^{(t)})^2)$, we also have

- In Phase I.1, because $\sigma_\rho^{(t)} = \Theta(\frac{\sigma_w^{(t)} (\log \log \log d)^3}{\sqrt{\log d}}) b^{(t)}$, we have $\mathbf{Pr}_\rho \left[ |\rho_i| \geq \frac{c_2}{10c_1} b^{(t)} \right] \ll e^{-\Omega(\log^{1/4} d)}$
- In Phase I.2, because $\sigma_\rho^{(t)} = \Theta(\frac{\sigma_w^{(t)} (\log \log \log d)^3}{\log d}) b^{(t)}$, we have $\mathbf{Pr}_\rho \left[ |\rho_i| \geq \frac{c_2}{10c_1} b^{(t)} \right] \ll O\left(\frac{k}{d}\right)$

As for the bound on $\Gamma_{t,y}$, for every $i \in [m]$, $j \in [d]$, let $\Lambda \subseteq [d] \setminus \{j\}$ be the subset containing all $j' \in [d] \setminus \{j\}$ with $|\langle w_i^{(t)}, \mathbf{M}_{j'} \rangle| \leq q \stackrel{\text{def}}{=} \frac{\sigma_w^{(t)}}{\log d}$. By the assumption $\mathcal{S}_{ept}^{(t)} = [m]$, we know $|\Lambda| \geq \Omega(d/\log d)$.

Since $|\langle w_i^{(t)}, \mathbf{M}_{j'} \rangle| \leq q$, $\mathbb{E}[z_j^2] = \Theta(1/d)$ and $|z_j| \leq 1$, by Bernstein's inequality, we have

$$\mathbf{Pr}_x \left[ \left| \left\langle w_i^{(t)}, \sum_{j' \in \Lambda} \mathbf{M}_{j'} z_{j'} \right\rangle \right| \geq \frac{c_2}{10c_1} b^{(t)} \right] \leq e^{-\Omega\left(\frac{(b^{(t)})^2}{q^2 |\Lambda|/d + q \cdot b^{(t)}}\right)} \leq e^{-\Omega(\log^{1.5} d)} \quad . \qquad \square$$

### C.3.2 Growth Lemmas

Our first lemma here shall be used to (lower) bound how $\langle w_i^{(t)}, \mathbf{M}_j \rangle$ (i.e., the weight with respect to neuron $i$ in direction $\mathbf{M}_j$) grows for those $i \in \mathcal{S}_{j,sure}$.

**Lemma C.8** (signal growth). *Suppose we (1) either are in Phase I.1 with $\mathcal{S}_{ept}^{(t)} = [m]$, (2) or are in Phase I.2 with $\mathcal{S}_{ept+}^{(t)} = [m]$. Then, for every $j \in [d]$, every $i \in \mathcal{S}_{j,sure}^{(t)}$, as long as $|\langle w_i^{(t)}, \mathbf{M}_j \rangle| = O(b^{(t)} \log \log \log d)$, the following holds:*

$$\mathbb{E}_{x,y,\rho} \left[ y \mathbb{1}_{\langle w_i^{(t)}, x \rangle + \rho_i \geq b^{(t)}} z_j \right] = \Theta\left(\frac{1}{d}\right)$$

*Proof of Lemma C.8.* Recall that $i \in \mathcal{S}_{j,sure}^{(t)}$ means $\text{sign}(\langle w_i^{(t)}, \mathbf{M}_j \rangle) = \text{sign}(w_j^\star)$. Without loss of generality, let us assume $\text{sign}(\langle w_i^{(t)}, \mathbf{M}_j \rangle) = \text{sign}(w_j^\star) = 1$.

First consider the case when $|z_j| = 1$. Since $j \in \mathcal{S}_{j,sure}^{(t)}$, we have $\langle w_i^{(t)}, \mathbf{M}_j \rangle \geq b^{(t)} \sqrt{1 + \frac{c_2}{c_1}}$ so applying Lemma C.7,

$$\mathbf{Pr}[\langle w_i^{(t)}, x \rangle + \rho_i \geq b^{(t)} \mid z_j = 1] \geq 1 - 2\Gamma_t = 1 - o(1)$$
$$\mathbf{Pr}[\langle w_i^{(t)}, x \rangle + \rho_i \geq b^{(t)} \mid z_j = -1] \leq 2\Gamma_t = o(1)$$

Moreover, a simple calculation using $|w_j^\star| = \Theta(1)$ gives us $\mathbb{E}_{x,y}[y \mid z_j = 1] = \Theta(1)$ (can be proven by Lemma H.1) and therefore

$$\mathbb{E}_{x,y,\rho} \left[ y \mathbb{1}_{\langle w_i^{(t)}, x \rangle + \rho_i \geq b^{(t)}} z_j \mid |z_j| = 1 \right] = \Theta(1)$$

For all other non-zero value $|z_j| = s > 0$, we have $s \geq \frac{1}{\sqrt{k}}$ and wish to apply Lemma C.5 to bound
$$\Delta_s := \mathbb{E}_{x,y,\rho}\left[y\mathbb{1}_{\langle w_i^{(t)}, x\rangle + \rho_i \geq b^{(t)}} \mathsf{sign}(z_j) \mid |z_j| = s\right]$$

In Phase I.1, to apply Lemma C.5, we choose parameters as follows:

- $Y = y$, $S_1 = \sum_{j' \neq j} w_{j'}^\star z_{j'}$, $S_2 = \left\langle w_i^{(t)}, \sum_{j' \neq j} \mathbf{M}_{j'} z_{j'} + \xi \right\rangle$, $\alpha = \langle w_i^{(t)}, \mathbf{M}_j \rangle \cdot s > 0$, $\rho = \rho_i$,

- $V = \mathbb{E}[S_2^2] = O((\sigma_w^{(t)})^2)$, $L = \Theta(s)$ (using Lemma H.1), $\Gamma = \Gamma_t$ (using Lemma C.7),

- let $\Lambda$ be the subset defined in Lemma C.7, then we can let $S_1'' = (z_j)_{j \in \Lambda}$ and $S_2'' = \left\langle w_i^{(t)}, \sum_{j' \in \Lambda} \mathbf{M}_{j'} z_{j'} \right\rangle$

- we have $\Gamma_y = \Gamma_{t,y} = \frac{1}{d^{10}}$ (from Lemma C.7) and

$$L_y = \max_{z_{j'} \text{ for } j' \in [d] \setminus \{j\} \setminus \Lambda} \left\{ \mathbb{E}_{z_{j'} \text{ for } j' \in \Lambda} [|\mathsf{sign}(w_j^\star z_j + S_1) - \mathsf{sign}(S_1)|] \right\}$$
$$\leq \max_{z_{j'} \text{ for } j' \in [d] \setminus \{j\} \setminus \Lambda} \left\{ \mathbf{Pr}_{z_{j'} \text{ for } j' \in \Lambda} [S_1'' \in [-S_1' - |w_j^\star z_j|, -S_1' + |w_j^\star z_j|]] \right\}$$
$$\overset{\text{①}}{\leq} O\left(\frac{s}{\sqrt{|\Lambda|/d}} + \frac{1}{\sqrt{|\Lambda|k/d}}\right) \leq O\left((s + \frac{1}{\sqrt{k}})\sqrt{\log d}\right) \leq O(s \cdot \sqrt{\log d})$$

where inequality ① uses Lemma H.1a and $|\Lambda| \geq \Omega(\frac{d}{\log d})$ from Lemma C.7.

Hence, invoking Lemma C.5, we have

- $\Delta_s \geq -\frac{\sigma_w^{(t)}}{\sigma_\rho^{(t)}} \geq -\frac{O(1)}{(\log\log\log d)^3}$ and $|\Delta_s| \leq \frac{\sigma_w^{(t)}}{\sigma_\rho^{(t)}} + s \leq \frac{O(1)}{(\log\log\log d)^3} + s$ when $s = \Omega\left(\frac{1}{\log\log\log d}\right)$

- $|\Delta_s| \leq \left(e^{-\Omega(b^2/\sigma_\rho^2)} + \Gamma_t\right)\left(O(\frac{\alpha}{\sigma_\rho}) + L_y\right) + \Gamma_{t,y} \leq e^{-\Omega(\log^{1/4} d)} \cdot s$ when $s = O\left(\frac{1}{\log\log\log d}\right)$ (which implies $\alpha < \frac{b^{(t)}}{4}$)

Notice that $\mathbb{E}[z_j^2] = O(1/d)$, which implies that
$$\mathbf{Pr}\left[|z_j| = \Omega\left(\frac{1}{\log\log\log d}\right)\right] = O\left(\frac{(\log\log\log d)^2}{d}\right)$$

This together gives us the bound that
$$-O\left(\frac{1}{d \cdot \log\log\log d}\right) \leq \mathbb{E}_{x,y,\rho}\left[y\mathbb{1}_{\langle w_i^{(t)}, x\rangle + \rho_i \geq b^{(t)}} z_j \mid |z_j| < 1\right] \leq O\left(\frac{1}{d}\right)$$

In Phase I.2, the analysis is similar with different parameters: in particular,

- $V = \mathbb{E}[S_2^2] = O(\frac{(\sigma_w^{(t)})^2}{\log d})$ which is tighter,

Therefore, we have

- $\Delta_s \geq -\frac{\sigma_w^{(t)}}{\sigma_\rho^{(t)}\sqrt{\log d}} \geq -\frac{O(1)}{(\log\log\log d)^3}$ and $|\Delta_s| \leq \frac{\sigma_w^{(t)}}{\sigma_\rho^{(t)}\sqrt{\log d}} + s \leq \frac{O(1)}{(\log\log\log d)^3} + s$ when $s = \Omega\left(\frac{1}{\log\log\log d}\right)$

- $|\Delta_s| \leq \left(e^{-\Omega(b^2/\sigma_\rho^2)} + \Gamma_t\right)\left(O(\frac{\alpha}{\sigma_\rho}) + L_y\right) + \Gamma_{t,y} \leq O(\frac{ks\log d}{d})$ when $s = O\left(\frac{1}{\log\log\log d}\right)$ (which implies $\alpha < \frac{b^{(t)}}{4}$).
 (This uses $\Gamma_t = O(k/d)$ and $\frac{\alpha}{\sigma_\rho} \leq o(s\log d)$.)

Taking expectation over $z_j$ as before, and using $k < d^{(1-c_0)/2}$ finishes the proof. $\square$

Our next lemma shall be used to upper bound how $\langle w_i^{(t)}, \mathbf{M}_j \rangle$ can grown for every $i \in [m]$.

**Lemma C.9** (maximum growth). *Suppose we (1) either are in Phase I.1 with $\mathcal{S}_{ept}^{(t)} = [m]$, (2) or are in Phase I.2 with $\mathcal{S}_{ept+}^{(t)} = [m]$. Then, for every $j \in [d]$, every $i \in [m]$, the following holds:*

$$\left| \mathop{\mathbb{E}}_{x,y,\rho} \left[ y \mathbb{1}_{\langle w_i^{(t)}, x \rangle + \rho_i \geq b^{(t)}} z_j \right] \right| = O\left(\frac{1}{d}\right) .$$

*Proof.* Proof is analogous to that of Lemma C.8, and the reason we no longer need the requirement $|\langle w_i^{(t)}, \mathbf{M}_j \rangle| = O(b^{(t)} \log \log \log d)$ is because, when invoking Lemma C.5, it suffices for us to apply Lemma C.5a for every non-zero values of $z$ (as opposed to only those $z = \Omega(\frac{1}{\log \log \log d})$) which no longer requires $\alpha \leq b$. □

Our next lemma shall be used to upper bound how $\langle w_i^{(t)}, \mathbf{M}_j \rangle$ can grown for every $i \in [m] \setminus \mathcal{S}_{j,pot}^{(t)}$.

**Lemma C.10** (non-signal growth). *Suppose we (1) either are in Phase I.1 with $\mathcal{S}_{ept}^{(t)} = [m]$, (2) or are in Phase I.2 with $\mathcal{S}_{ept+}^{(t)} = [m]$. Then, for every $j \in [d]$, every $i \in [m] \setminus \mathcal{S}_{j,pot}^{(t)}$, the following holds:*

$$\left| \mathop{\mathbb{E}}_{x,y,\rho} \left[ y \mathbb{1}_{\langle w_i^{(t)}, x \rangle + \rho_i \geq b^{(t)}} z_j \right] \right| = O(\frac{\Gamma_t \cdot \log d}{d})$$

*where $\Gamma_t$ is given from Lemma C.7.*

*Proof of Lemma C.10.* Suppose $|z_j| = s$ and without loss of generality $\langle w_i^{(t)}, \mathbf{M}_j \rangle \geq 0$. We choose $\alpha = \langle w_i^{(t)}, \mathbf{M}_j \rangle \cdot s$ as before. Then, we have $\alpha \leq \sqrt{c_1 - c_2} \sigma_w^{(t)}) \sqrt{\log d} = b^{(t)} \sqrt{1 - \frac{c_2}{c_1}} \leq b^{(t)} (1 - \frac{c_2}{2c_1})$ always holds.

Therefore, using the same notation as the proof of Lemma C.8, we always have the bound

$$|\Delta_s| \leq \left( e^{-\Omega(b^2/\sigma_\rho^2)} + \Gamma_t \right) \left( O(\frac{\alpha}{\sigma_\rho}) + L_y \right) + \Gamma_{t,y}$$

Plugging in the parameters we finish the proof. □

Our final lemma shall be used to upper bound how $\langle w_i^{(t)}, \mathbf{M}_j \rangle$ can grown with respect to the noise $\xi$ in the input.

**Lemma C.11** (noise growth). *For every $i \in [m]$, every $j \in [d]$, the following holds:*

$$\left| \mathop{\mathbb{E}}_{x,y,\rho} \left[ y \mathbb{1}_{\langle w_i^{(t)}, x \rangle + \rho_i \geq b^{(t)}} \langle \xi, \mathbf{M}_j \rangle \right] \right| = O\left( \Gamma_t \frac{\sigma_x^2}{d \sigma_\rho^{(t)}} |\langle w_i^{(t)}, \mathbf{M}_j \rangle| \right)$$

*Proof of Lemma C.11.* We can define $\alpha = |\langle \mathbf{M}_j, \xi \rangle|$ and study

$$\Delta_s := \mathop{\mathbb{E}}_{x,y,\rho} \left[ y \mathbb{1}_{\langle w_i^{(t)}, x \rangle + \rho_i \geq b^{(t)}} \mathsf{sign}(\langle \xi, \mathbf{M}_j \rangle) \,\Big|\, |\langle \xi, \mathbf{M}_j \rangle| = \alpha \right] \tag{C.8}$$

This time we have $L = L_y = 0$, $\Gamma_y = 0$, $V = \mathbb{E}[S_2^2] = O((\sigma_w^{(t)})^2)$, so applying Lemma C.5 we have,

- when $\alpha \leq b^{(t)}/4$, $|\Delta_s| \leq \frac{|\langle w_i^{(t)}, \mathbf{M}_j \rangle| |\langle \mathbf{M}_j, \xi \rangle|}{\|\mathbf{M}_j\|_2 \sigma_\rho^{(t)}} \cdot \left( e^{-\Omega(b^2/\sigma_\rho^2)} + \Gamma_t \right)$;

- when $\alpha > b^{(t)}/4$ (which happens with exponentially small prob.), $|\Delta_s| \leq O(\frac{\sigma_w^{(t)}}{\sigma_\rho})$.

Together, using the fact that $\mathbb{E}[|\Delta_s|] \leq \sqrt{\mathbb{E}[\Delta_s^2]}$ we have:

$$\left| \mathop{\mathbb{E}}_{x,y,\rho} \left[ y \mathbb{1}_{\langle w_i^{(t)}, x \rangle + \rho_i \geq b^{(t)}} \langle \xi, \mathbf{M}_j \rangle \right] \right| = O\left( \Gamma_t \frac{\sigma_x^2}{d \sigma_\rho^{(t)}} |\langle w_i^{(t)}, \mathbf{M}_j \rangle| \right)$$

$\square$

### C.3.3 Proof of Theorem C.1

Suppose in Lemma C.8 the hidden constant is $20C$ for the lower bound, that is,

$$\mathop{\mathbb{E}}_{x,y,\rho} \left[ y \mathbb{1}_{\langle w_i^{(t)}, x \rangle + \rho_i \geq b^{(t)}} z_j \right] \geq \frac{20C}{d}$$

*Proof of Theorem C.1.* Let us prove by induction with respect to $t$. Suppose the properties all hold at $t = 0$. Recall from Fact A.1, for iteration $t$, for every neuron $i \in [m]$,

$$\nabla_{w_i} \mathbf{Loss}_t(w^{(t)}; x, y, \rho) = -y \ell_t'(w^{(t)}; x, y, \rho) \left( \mathbb{1}_{\langle w_i^{(t)}, x \rangle + \rho_i \geq b^{(t)}} + \mathbb{1}_{-\langle w_i^{(t)}, x \rangle + \rho_i \geq b^{(t)}} \right) \cdot x \ .$$

Using the bound on $\|w_i^{(t)}\|_2$ (from $\mathcal{S}_{ept}^{(t)} = [m]$) and the fact $\sigma_0 \leq \frac{1}{\mathsf{poly}(d)}$, we know $\ell_t'(w^{(t)}; x, y, \rho) = \frac{1}{2} \pm \frac{1}{\mathsf{poly}(d)}$. Also, recall also from Lemma A.2 that

$$\mathop{\mathbb{E}}_{x \sim \mathcal{D}, y=y(x), \rho} \left[ \nabla_{w_i} \mathbf{Loss}_t(w^{(t)}; x, y, \rho) \right] = \nabla_{w_i} \widetilde{\mathbf{Loss}}_t(w^{(t)}) \pm \frac{1}{\mathsf{poly}(d)} \ .$$

Together, we have a clean formulation for our gradient update rule:

$$w_i^{(t+1)} = w_i^{(t)}(1 - \eta \lambda - \eta \lambda \|w_i^{(t)}\|_2) + \mathop{\mathbb{E}}_{x,y=y(x),\rho} \left[ y \left( \mathbb{1}_{\langle w_i^{(t)}, x \rangle + \rho_i \geq b^{(t)}} + \mathbb{1}_{-\langle w_i^{(t)}, x \rangle + \rho_i \geq b^{(t)}} \right) \cdot x \right] \pm \frac{\eta}{\mathsf{poly}(d)} \ .$$

and as a result for every $j \in [d]$

$$\langle w_i^{(t+1)}, \mathbf{M}_j \rangle = \langle w_i^{(t)}, \mathbf{M}_j \rangle (1 - \eta \lambda - \eta \lambda \|w_i^{(t)}\|_2) \pm \frac{\eta}{\mathsf{poly}(d)}$$
$$+ \mathop{\mathbb{E}}_{x,y=y(x),\rho} \left[ y \left( \mathbb{1}_{\langle w_i^{(t)}, x \rangle + \rho_i \geq b^{(t)}} + \mathbb{1}_{-\langle w_i^{(t)}, x \rangle + \rho_i \geq b^{(t)}} \right) \cdot (z_j + \langle \xi, \mathbf{M}_j \rangle) \right] \ . \quad (C.9)$$

We now prove each statement separately (and note our proofs apply both to Phase I.1 and I.2).

1. For every $i \notin \mathcal{S}_{j,pot}^{(t)}$, by substituting Lemma C.10 and Lemma C.11 into (C.9), we have

$$\langle w_i^{(t+1)} - w_i^{(t)}, \mathbf{M}_j \rangle \leq \frac{\eta \cdot e^{-\Omega(\log^{1/4} d)}}{d} \ll \sqrt{c_1 - c_2}(\sigma_w^{(t+1)} - \sigma_w^{(t)}) \sqrt{\log d} \quad (C.10)$$

so we also have $\langle w_i^{(t+1)}, \mathbf{M}_j \rangle < \sqrt{c_1 - c_2} \sigma_w^{(t+1)} \sqrt{\log d}$ and thus $i \notin \mathcal{S}_{j,pot}^{(t+1)}$.

2. For every $i \in \mathcal{S}_{j,sure}^{(t)}$, suppose wlog $\langle w_i^{(t)}, \mathbf{M}_j \rangle$ is positive. Then, either $\langle w_i^{(t)}, \mathbf{M}_j \rangle > \Omega(b^{(t)} \log \log \log d)$ in such a case we still have $\langle w_i^{(t+1)}, \mathbf{M}_j \rangle \geq \sqrt{c_1 + c_2} \sigma_w^{(t+1)} \sqrt{\log d}$. Otherwise, if $\langle w_i^{(t)}, \mathbf{M}_j \rangle \leq \Omega(b^{(t)} \log \log \log d)$ then by substituting Lemma C.8 and Lemma C.11 into (C.9), we have (using $\sigma_0 \leq \frac{1}{\mathsf{poly}(d)}$ and $\lambda \leq \frac{\log d}{d}$)

$$\langle w_i^{(t+1)}, \mathbf{M}_j \rangle \geq (1 - \eta \lambda) \langle w_i^{(t)}, \mathbf{M}_j \rangle + \frac{20 C \eta}{d} \geq \langle w_i^{(t)}, \mathbf{M}_j \rangle + \sqrt{c_1 + c_2}(\sigma_w^{(t+1)} - \sigma_w^{(t)}) \sqrt{\log d}$$

so by induction we also have $\langle w_i^{(t+1)}, \mathbf{M}_j \rangle \geq \sqrt{c_1 + c_2} \sigma_w^{(t+1)} \sqrt{\log d}$. Combining this with $\mathcal{S}_{j,pot}^{(0)} \supseteq \mathcal{S}_{j,pot}^{(t+1)}$, we conclude that $i \in \mathcal{S}_{j,sure}^{(t+1)}$.

3. To check $\mathcal{S}_{ept}^{(t+1)} = [m]$, we need to verify four things:

   - $\langle w_i^{(t)}, \mathbf{M}_j \rangle^2 \geq (c_1 - c_2)(\sigma_w^{(t)})^2 \log d$      for at most $O(1)$ many $j \in [d]$.
     This is so because $\mathcal{S}_{ept}^{(0)} = [m]$ and $\mathcal{S}_{j,pot}^{(0)} \supseteq \mathcal{S}_{j,pot}^{(t+1)}$.
   - $\langle w_i^{(t+1)}, \mathbf{M}_j \rangle^2 \geq 2(\sigma_w^{(t+1)})^2 \sqrt{\log d}$      for at most $2^{-\sqrt{\log d}} d$ many $j \in [d]$.
     This can be derived from (C.10) in the same way.
   - $|\langle w_i^{(t+1)}, \mathbf{M}_j \rangle| \leq \frac{\sigma_w^{(t+1)}}{\log d}$      for at least $\Omega(\frac{d}{\log d})$ many $j \in [d]$.
     This can be derived from (C.10) in the same way.
   - $\|w_i^{(t)}\|_2^2 \leq 2(\sigma_w^{(t)})^2 d$

     For every $i \in [m]$, suppose wlog $\langle w_i^{(t)}, \mathbf{M}_j \rangle$ is positive. Then, by substituting Lemma C.9 and Lemma C.11 into (C.9), we have
     $$\langle w_i^{(t+1)} - w_i^{(t)}, \mathbf{M}_j \rangle \leq O(\frac{\eta}{d})$$
     Applying this formula for $t+1$ times, we derive that
     $$|\langle w_i^{(t+1)}, \mathbf{M}_j \rangle| \leq O\big(\frac{\eta(t+1)}{d}\big) + |\langle w_i^{(0)}, \mathbf{M}_j \rangle| \tag{C.11}$$
     and therefore applying this together with (C.10),
     $$\|w_i^{(t+1)}\|_2^2 = \sum_{j:\, i \in \mathcal{S}_{j,pot}^{(0)}} |\langle w_i^{(t+1)}, \mathbf{M}_j \rangle|^2 + \sum_{j:\, i \notin \mathcal{S}_{j,pot}^{(0)}} |\langle w_i^{(t+1)}, \mathbf{M}_j \rangle|^2$$
     $$\overset{①}{\leq} 1.5 \|w_i^{(0)}\|_2^2 + O(1) \cdot \big(\frac{\eta(t+1)}{d}\big)^2 + d \cdot \big(\frac{\eta(t+1)}{d}\big)^2 \cdot e^{-\Omega(\log^{1/4} d)} \leq 2(\sigma_w^{(t+1)})^2 d$$
     (Above, inequality ① uses that there are at most $O(1)$ indices $j \in [d]$ such that $i \in \mathcal{S}_{j,pot}^{(0)}$.)

4. Finally, to check $\mathcal{S}_{ept+}^{(t+1)} = [m]$ for $t \geq T_{\mathsf{a}} = \Theta(\frac{d\sigma_0 \log^{2.5} d}{\eta})$, we first derive that
   $$\sigma_w^{(t+1)} = \sigma_0 + \Theta\big(\frac{\eta}{d\sqrt{\log d}}\big) \cdot (t+1) \geq \sigma_0 \cdot \Omega(\log^2 d) \,.$$

   - For every $i \notin \mathcal{S}_{j,pot}^{(t+1)}$, (C.10) gives
     $$|\langle w_i^{(t+1)}, \mathbf{M}_j \rangle| \leq |\langle w_i^{(0)}, \mathbf{M}_j \rangle| + \frac{e^{-\Omega(\log^{1/4} d)}}{d} \cdot \eta(t+1)$$
     $$\leq O(\sigma_0 \sqrt{\log d}) + \frac{e^{-\Omega(\log^{1/4} d)}}{d} \cdot \eta(t+1) \leq O\big(\frac{\sigma_w^{(t+1)}}{\log^{1.5} d}\big)$$
     In particular, this together with $\mathcal{S}_{ept}^{(0)} = [m]$ ensures that for every $i \in [m]$, $\left|\langle w_i^{(t+1)}, \mathbf{M}_j \rangle\right| \geq \frac{\sigma_w^{(t+1)}}{\log d}$ for at most $O(1)$ many $j \in [d]$.
   - For any $i \in \mathcal{S}_{j,pot}^{(0)}$, using (C.11) we have
     $$|\langle w_i^{(t+1)}, \mathbf{M}_j \rangle| \leq |\langle w_i^{(0)}, \mathbf{M}_j \rangle| + O\big(\frac{\eta(t+1)}{d}\big) \leq O(\sigma_0 \sqrt{\log d}) + O\big(\frac{\eta(t+1)}{d}\big) = \Theta(\sigma_w^{(t+1)} \sqrt{\log d}) \tag{C.12}$$
     Using this together with the previous item, as well as $|\mathcal{S}_{j,pot}^{(0)}| \leq O(1)$, we have $\|w_i^{(t+1)}\|^2 \leq O\big(\frac{(\sigma_w^{(t+1)})^2 d}{\log^3 d}\big)$.

Putting them together we have $\mathcal{S}_{ept+}^{(t+1)} = [m]$ for every $t \geq T_{\mathsf{a}}$.

5. After $t = T_{\mathsf{b}}$ iterations, we have $\sigma_w^{(t)} = \Theta(\sigma_0 + \frac{\eta}{d\sqrt{\log d}} \cdot T_{\mathsf{b}})$, for every $i \notin \mathcal{S}_{j,pot}^{(0)}$, by Lemma C.10 and Lemma C.11

$$|\langle w_i^{(t)}, \mathbf{M}_j \rangle| \leq |\langle w_i^{(T_{\mathsf{a}})}, \mathbf{M}_j \rangle| + \frac{O(k/d) \cdot \sqrt{\log d}}{d} \cdot \eta(T_{\mathsf{b}} - T_{\mathsf{a}}) \leq O(\sigma_w^{(t)} \cdot \frac{k \log d}{d})$$

Combining this with (C.11), we immediately have

$$\|w_i^{(t)}\|^2 = \sum_{j:\, i \in \mathcal{S}_{j,pot}^{(0)}} |\langle w_i^{(t+1)}, \mathbf{M}_j \rangle|^2 + \sum_{j:\, i \notin \mathcal{S}_{j,pot}^{(0)}} |\langle w_i^{(t+1)}, \mathbf{M}_j \rangle|^2$$

$$\leq (\sigma_w^{(t)})^2 \cdot O(\frac{k^2 \log^2 d}{d} + \log d)$$

This implies $\mathcal{S}_{ept++}^{(t)} = [m]$ and $\mathcal{S}_{j,pot}^{(0)} \supseteq \mathcal{S}_{j,pot+}^{(t)}$ at this iteration $t$.

$\square$

## C.4 Phase II: Signal Growth After Winning Lottery

**Definition C.12.** *In phase II we make the following parameter choices.*

- *In Phase II, we pick $b^{(t)} = \sqrt{c_1}\sigma_w^{(t)}\sqrt{\log d}$ and $\sigma_\rho^{(t)} = \sigma_w^{(t)} \cdot \frac{(\log \log \log d)^3}{\sqrt{\log d}}$.*

  *We grow $b^{(t+1)} = b^{(t)} + \frac{C\eta}{d}$ as before (the same as phase I.2 in Definition C.6) for each iteration, but stop growing $b^{(t)}$ when it reaches a threshold $b^{(t)} = \beta\Xi_2^2$.*

We first introduce a notation on a (high-probability) version of the coordinate Lipscthiz continuity.

**Definition C.13** (coordinate Lipschitzness)**.** *At every iteration $t$, for every $j \in [d]$, we define $L_{t,j} > e^{-\Omega(\log^2 d)}$ to be the smallest value such that w.p. at least $1 - e^{-\Omega(\log^2 d)}$ over the choice of $\{z_{j'}\}_{j' \neq j}$ and $\xi$, for every $z \in [-1,1]$ and $z = (z_1, \cdots, z_{j-1}, z, z_{j+1}, \cdots, z_d)$, $z' = (z_1, \cdots, z_{j-1}, 0, z_{j+1}, \cdots, z_d)$, $x = \mathbf{M}z + \xi$ and $x' = \mathbf{M}z' + \xi$:*

$$|f_t(x) - f_t(x')| \leq L_{t,j}|z|\ .$$

### C.4.1 Growth Lemmas

In this subsection, we provide new growth lemmas Lemma C.14, Lemma C.15, Lemma C.16, Lemma C.17 that are specific to Phase II, to replace the user of the old growth lemmas Lemma C.8, Lemma C.9, Lemma C.10, Lemma C.11 from Phase I.

**Lemma C.14** (signal growth II)**.** *Suppose we $\mathcal{S}_{ept+}^{(t)} = \mathcal{S}_{ept++}^{(t)} = [m]$. Then, for every $j \in [d]$, every $i \in \mathcal{S}_{j,sure}^{(t)}$, the following holds:*

$$\mathbb{E}_{x,y,\rho}\left[y\ell_t'(w^{(t)}; x, y, \rho)\mathbb{1}_{\langle w_i^{(t)}, x \rangle + \rho_i \geq b^{(t)}} z_j \right] = \Theta\left(\frac{1}{d}\right) \pm O\left(\frac{L_{t,j}}{d} + \frac{\sqrt{k}\sigma_\rho^{(t)} \log d}{d} + \frac{\sqrt{k}}{\beta d^{3/2}}\right)$$

*Proof of Lemma C.14.* First, without loss of generality, assuming that $\mathsf{sign}(\langle w_i^{(t)}, \mathbf{M}_j \rangle) = \mathsf{sign}(w_j^\star) = $

1. Let us define $z' = (z_1, \cdots, z_{j-1}, 0, z_{j+1}, \cdots, z_d)$ and $x' = \mathbf{M}z' + \xi$. Define
$$f_{t,i}(w^{(t)}; x, \rho) \stackrel{\text{def}}{=} \sum_{j \neq i} \left( \mathsf{ReLU}(\langle w_j^{(t)}, x \rangle + \rho_j + b_j^{(t)}) - \mathsf{ReLU}(-\langle w_j^{(t)}, x \rangle + \rho_j + b_j^{(t)}) \right)$$
$$+ \left( \mathsf{ReLU}(\langle w_i^{(t)}, x \rangle + b_i^{(t)}) - \mathsf{ReLU}(-\langle w_i^{(t)}, x \rangle + b_i^{(t)}) \right)$$
$$\ell'_{t,i}(w^{(t)}; x, y, \rho) \stackrel{\text{def}}{=} \frac{d}{ds}[\log(1+e^s)]\, |_{s=-yf_{t,i}(w^{(t)};x,\rho)}$$

Now, since $\frac{e^s}{1+e^s}$ is an $O(1)$-Lipschitz function in $s$, we know that w.p. at least $1 - e^{-\Omega(\log^2 d)}$
$$|\ell'_t(w^{(t)}; x, y, \rho) - \ell'_{t,i}(w^{(t)}; x', y, \rho)| = O(L_{t,j} \cdot |z_j| + \sigma_\rho^{(t)} \log d)$$
and this implies that
$$\left| \mathop{\mathbb{E}}_{x,y,\rho} \left[ y(\ell'_t(w^{(t)}; x, y, \rho) - \ell'_{t,i}(w^{(t)}; x', y, \rho)) \mathbf{1}_{\langle w_i^{(t)}, x \rangle + \rho_i \geq b^{(t)}} z_j \right] \right|$$
$$\leq O\left( L_{t,j} \mathbb{E}[z_j^2] + \sigma_\rho^{(t)} \log d \cdot \mathbb{E}[|z_j|] \right) + e^{-\Omega(\log^2 d)}$$
$$= O\left( \frac{L_{t,j}}{d} + \frac{\sigma_\rho^{(t)} \log d \cdot \sqrt{k}}{d} \right) + e^{-\Omega(\log^2 d)}$$
so we only need to bound $\mathbb{E}_{x,y,\rho} \left[ y\ell'_{t,i}(w^{(t)}; x', y, \rho) \mathbf{1}_{\langle w_i^{(t)}, x \rangle + \rho_i \geq b^{(t)}} z_j \right]$.

Let us first focus on the case that $|z_j| = 1$. As before, since $j \in \mathcal{S}_{j,\text{sure}}^{(t)}$, we have $\langle w_i^{(t)}, \mathbf{M}_j \rangle \geq b^{(t)} \sqrt{1 + \frac{c_2}{c_1}}$ so applying Lemma C.7,
$$\mathbf{Pr}[\langle w_i^{(t)}, x \rangle + \rho_i \geq b^{(t)} \mid z_j = 1] \geq 1 - 2\Gamma_t = 1 - o(1)$$
$$\mathbf{Pr}[\langle w_i^{(t)}, x \rangle + \rho_i \geq b^{(t)} \mid z_j = -1] \leq 2\Gamma_t = o(1)$$
This means
$$\mathop{\mathbb{E}}_{x,y,\rho} \left[ y\ell'_{t,i}(w^{(t)}; x', y, \rho) \mathbf{1}_{\langle w_i^{(t)}, x \rangle + \rho_i \geq b^{(t)}} z_j \mid |z_j| = 1 \right] \geq \frac{1}{2} \mathop{\mathbb{E}}_{x,y,\rho} \left[ y\ell'_{t,i}(w^{(t)}; x', y, \rho) \mid z_j = 1 \right] - o(1)$$
Now recall $y(z_j, z) = \mathsf{sign}(w_j^\star z_j + \langle w^\star, z \rangle)$.

- When $|\langle w^\star, z \rangle| > |w_j^\star|$, then we know that $y(z_j, z)$ and $y(z_j, -z)$ have different signs, but $\ell'_{t,i}(w^{(t)}; x', y, \rho) = \ell'_{t,i}(w^{(t)}; -x', -y, \rho)$ remains the same if we flip $z$ to $-z$. By symmetry, we have
$$\mathop{\mathbb{E}}_{x,y,\rho} \left[ y\ell'_{t,i}(w^{(t)}; x', y, \rho) \mid z_j = 1 \wedge |\langle w^\star, z \rangle| > |w_j^\star| \right] = 0$$

- Suppose otherwise $|\langle w^\star, z \rangle| \leq |w_j^\star|$. Since $|w_j^\star| = \Theta(1)$, by Lemma H.1, this event happens with at least constant probability. When it happens, we have $y(z_j, z) = y(z_j, -z) = +1$, but $\ell'_{t,i}(w^{(t)}; x', y, \rho) + \ell'_{t,i}(w^{(t)}; -x', -y, \rho) = 1$. Therefore,
$$\mathop{\mathbb{E}}_{x,y,\rho} \left[ y\ell'_{t,i}(w^{(t)}; x', y, \rho) \mid z_j = 1 \wedge |\langle w^\star, z \rangle| \leq |w_j^\star| \right] \geq \frac{1}{2}$$

Together, we have
$$\mathop{\mathbb{E}}_{x,y,\rho} \left[ y\ell'_{t,i}(w^{(t)}; x', y, \rho) \mathbf{1}_{\langle w_i^{(t)}, x \rangle + \rho_i \geq b^{(t)}} z_j \mid |z_j| = 1 \right] = \Omega(1) \tag{C.13}$$

Next, conditioning on $|z_j| = s$ for some $0 < s < 1$, we can apply Lemma C.5 with $Y = y\ell'_{t,i}(w^{(t)}; x', y, \rho)$ on $s = \mathsf{sign}(z_j)$, $\alpha = \langle w_i^{(t)}, \mathbf{M}_j \rangle z_j$, $S_1 = z$, $S_2 = \sum_{j' \neq j} \langle w_i^{(t)}, \mathbf{M}_{j'} \rangle z_{j'}$ and $\rho = \rho_i$.

Since $\ell'_t(w^{(t)}; x', y, \rho) \geq 0$, we can conclude that $Y$ is a monotone non-decreasing function of in $s$. One can verify that $L = O(s)$ using Lemma H.1. Moreover, since $i \in \mathcal{S}^{(t)}_{ept++}$, we have $\frac{\mathbb{E}[S_2^2]}{\left(\sigma_\rho^{(t)}\right)^2} \leq \frac{1}{d\beta^2}$.

Let us denote by
$$\Delta_s := \mathbb{E}_{x,y,\rho}\left[y\ell'_{t,i}(w^{(t)}; x', y, \rho)\mathbb{1}_{\langle w_i^{(t)},x\rangle + \rho_i \geq b^{(t)}}\mathsf{sign}(z_j) \mid |z_j| = s\right]$$
so according to Lemma C.5 we have
$$\Omega\left(\frac{1}{\beta\sqrt{d}}\right) \leq \Delta_s = O\left(\frac{1}{\beta\sqrt{d}} + s\right)$$

This implies, using $\mathbb{E}[|z_j|] \leq \frac{\sqrt{k}}{d}$, that

$$\mathbb{E}_{x,y,\rho}\left[y\ell'_{t,i}(w^{(t)}; x', y, \rho)\mathbb{1}_{\langle w_i^{(t)},x\rangle + \rho_i \geq b^{(t)}}z_j \cdot \mathbb{1}_{|z_j|<1}\right] \leq \mathbb{E}[\Delta_{z_j} \cdot |z_j| \cdot \mathbb{1}_{|z_j|<1}] \leq O\left(\frac{\mathbb{E}[|z_j|]}{\beta\sqrt{d}} + \frac{1}{d}\right) \leq O\left(\frac{\sqrt{k}}{\beta d^{1.5}} + \frac{1}{d}\right)$$

$$\mathbb{E}_{x,y,\rho}\left[y\ell'_{t,i}(w^{(t)}; x', y, \rho)\mathbb{1}_{\langle w_i^{(t)},x\rangle + \rho_i \geq b^{(t)}}z_j \cdot \mathbb{1}_{|z_j|<1}\right] \geq \mathbb{E}[\Delta_{z_j}|z_j| \cdot \mathbb{1}_{|z_j|<1}] \geq -\Omega\left(\frac{\mathbb{E}[|z_j|]}{\beta\sqrt{d}}\right) \geq -\Omega\left(\frac{\sqrt{k}}{\beta d^{1.5}}\right)$$

Combining this with (C.13), and using $\mathbf{Pr}[|z_j| = 1] \geq \Omega(1/d)$ finishes the proof. □

Similarly, we have the following Lemma

**Lemma C.15** (maximum growth II). *Suppose we $\mathcal{S}^{(t)}_{ept+} = \mathcal{S}^{(t)}_{ept++} = [m]$. Then, for every $j \in [d]$, every $i \in [m]$, the following holds:*
$$\left|\mathbb{E}_{x,y,\rho}\left[y\ell'_t(w^{(t)}; x, y, \rho)\mathbb{1}_{\langle w_i^{(t)},x\rangle + \rho_i \geq b^{(t)}}z_j\right]\right| = O\left(\frac{1 + L_{t,j}}{d} + \frac{\sqrt{k}\sigma_\rho^{(t)}\log d}{d} + \frac{\sqrt{k}}{\beta d^{3/2}}\right)$$

**Lemma C.16** (non-signal growth II). *Suppose we $\mathcal{S}^{(t)}_{ept+} = \mathcal{S}^{(t)}_{ept++} = [m]$. Then, for every $j \in [d]$, every $i \notin \mathcal{S}^{(t)}_{j,pot}$ and $i \in [m]$, the following holds:*
$$\left|\mathbb{E}_{x,y,\rho}\left[y\ell'_t(w^{(t)}; x, y, \rho)\mathbb{1}_{\langle w_i^{(t)},x\rangle + \rho_i \geq b^{(t)}}z_j\right]\right| \leq \Gamma_t \cdot O\left(\frac{\log d + L_{t,j}}{d} + \frac{\sqrt{k}\sigma_\rho^{(t)}\log d}{d}\right)$$

*Proof of Lemma C.16.* In the same notation as the proof of Lemma C.14, we have
$$\left|\mathbb{E}_{x,y,\rho}\left[y(\ell'_t(w^{(t)}; x, y, \rho) - \ell'_{t,i}(x', y))\mathbb{1}_{\langle w_i^{(t)},x\rangle + \rho_i \geq b^{(t)}}z_j\right]\right|$$
$$\leq O\left(L_{t,j} \mathbb{E}[z_j^2 \cdot \mathbb{1}_{\langle w_i^{(t)},x\rangle + \rho_i \geq b^{(t)}}] + \sigma_\rho^{(t)}\log d \cdot \mathbb{E}[|z_j| \cdot \mathbb{1}_{\langle w_i^{(t)},x\rangle + \rho_i \geq b^{(t)}}]\right) + e^{-\Omega(\log^2 d)}$$
$$\leq \Gamma_t O\left(\frac{L_{t,j}}{d} + \frac{\sigma_\rho^{(t)}\log d \cdot \sqrt{k}}{d}\right) + e^{-\Omega(\log^2 d)}$$

where the last inequality uses Lemma C.7 and the fact $i \notin \mathcal{S}^{(t)}_{j,pot}$ (which, as before, implies if we choose $\alpha = \langle w_i^{(t)}, \mathbf{M}_j\rangle \cdot z$ then $\alpha^2 \leq (c_1 - c_2)(\sigma_w^{(t)})^2 \log d \leq \left(\frac{b^{(t)}}{4}\right)^2$).

Thus, we only need to bound
$$\mathbb{E}_{x,y,\rho}\left[y\ell'_{t,i}(x', y)\mathbb{1}_{\langle w_i^{(t)},x\rangle + \rho_i \geq b^{(t)}}z_j\right] .$$

Conditioning on $|z_j| = s$ for some $0 < s \leq 1$, we can apply Lemma C.5 again with $Y = y\ell'_{t,i}(x', y)$ on $s = \mathsf{sign}(z_j)$, $\alpha = \langle w_i^{(t)}, \mathbf{M}_j\rangle z_j$, $S_1 = z$, $S_2 = \sum_{j' \neq j}\langle w_i^{(t)}, \mathbf{M}_{j'}\rangle z_{j'}$ and $\rho = \rho_i$. This time, we use

$\Gamma_y = \frac{1}{d^{10}}$ and $L_y \leq O(z \cdot \log^{1/4} d)$. Define

$$\Delta_s := \mathop{\mathbb{E}}_{x,y,\rho}\left[y\ell'_{t,i}(x',y)\mathbb{1}_{\langle w_i^{(t)},x\rangle+\rho_i\geq b^{(t)}}\operatorname{sign}(z_j) \mid |z_j|=s\right]$$

Since $\alpha < \frac{b^{(t)}}{4}$, Lemma C.5 tells us

$$|\Delta_s| \leq \left(e^{-\Omega(b^2/\sigma_\rho^2)} + \Gamma_t\right)\left(O(\frac{\alpha}{\sigma_\rho}) + L_y\right) + \Gamma_y \leq O(\Gamma_t \log d) \cdot z$$

and therefore

$$\left|\mathop{\mathbb{E}}_{x,y,\rho}\left[y\ell'_{t,i}(x',y)\mathbb{1}_{\langle w_i^{(t)},x\rangle+\rho_i\geq b^{(t)}} z_j\right]\right| = \left|\mathbb{E}[\Delta_{z_j}|z_j|]\right| \leq O(\Gamma_t \log d) \cdot \mathbb{E}[z_j^2] = O(\frac{\Gamma_t \log d}{d})$$

Combining this with (C.13), and using $\mathbf{Pr}[|z_j|=1]\geq \Omega(1/d)$ finishes the proof. $\square$

Finally, we derive a more fine-grind bound for the noise:

**Lemma C.17** (noise growth II). *Suppose we $\mathcal{S}_{ept+}^{(t)} = \mathcal{S}_{ept++}^{(t)} = [m]$. Then, for every $j \in [d]$,*

*(a) for every $j \in [d]$,*

$$\left|\mathop{\mathbb{E}}_{x,y,\rho}\left[y\ell'_t(w^{(t)};x,y,\rho)\mathbb{1}_{\langle w_i^{(t)},x\rangle+\rho_i\geq b^{(t)}}\langle \xi,\mathbf{M}_j\rangle\right]\right| = O\left(\frac{\Gamma_t}{d\sigma_\rho^{(t)}}|\langle w_i^{(t)},\mathbf{M}_j\rangle|\sigma_x^2 + \frac{\Gamma_t L_{t,j}\sigma_x^2}{d} + e^{-\Omega(\log^2 d)}\right)$$

*(b) suppose also $\mathcal{S}_{j,pot}^{(0)} \supseteq \mathcal{S}_{j,pot+}^{(t)}$ for every $j \in [d]$, then*

$$\sum_{j\in[d]}\left|\mathop{\mathbb{E}}_{x,y,\rho}\left[y\ell'_t(w^{(t)};x,y,\rho)\mathbb{1}_{\langle w_i^{(t)},x\rangle+\rho_i\geq b^{(t)}}\langle \xi,\mathbf{M}_j\rangle\right]\right| \leq O\left(\frac{k^3\Xi_2^4}{d^2}\sigma_x^2 + e^{-\Omega(\log^2 d)}\right)$$

*Proof of Lemma C.17.* We can first decompose the noise $\xi$ into

$$\xi = (\mathbf{I} - \mathbf{M}_j\mathbf{M}_j^\top)\xi + \langle \mathbf{M}_j,\xi\rangle\mathbf{M}_j =: \xi'_j + \langle \mathbf{M}_j,\xi\rangle\mathbf{M}_j \ .$$

Let us define $x'_j = \mathbf{M}z + \xi'_j$.

- On one hand we have with probability at least $1-\Gamma_t$, $|\langle w_i^{(t)}, x'_j\rangle| \leq \frac{b^{(t)}}{10}$ (using a variant of Lemma C.7). Using the randomness of $\langle \xi, \mathbf{M}_j\rangle$ and $\rho_i$ we also have with probability at least $1 - e^{-\Omega(\log^2 d)}$ it satisfies $|\langle w_i^{(t)}, \mathbf{M}_j\rangle \cdot \langle \mathbf{M}_j, \xi\rangle| + |\rho_i| \leq \frac{b^{(t)}}{10}$. Therefore, with probability at least $1 - \Gamma_t - e^{-\Omega(\log^2 d)}$, we have $\mathbb{1}_{\langle w_i^{(t)},x\rangle+\rho_i\geq b^{(t)}} = \mathbb{1}_{\langle w_i^{(t)},x'_j\rangle+\rho_i\geq b^{(t)}}$.

- Otherwise, in the event that $|\langle w_i^{(t)}, x'_j\rangle| \geq \frac{b^{(t)}}{10}$, using the randomness of $\rho_i$, we have that

$$\mathop{\mathbf{Pr}}_{\rho_i}\left[\mathbb{1}_{\langle w_i^{(t)},x\rangle+\rho_i\geq b^{(t)}} \neq \mathbb{1}_{\langle w_i^{(t)},x'_j\rangle+\rho_i\geq b^{(t)}}\right] \leq O\left(\frac{\mathbb{E}\left[|\langle \xi,\mathbf{M}_j\rangle\langle w_i^{(t)},\mathbf{M}_j\rangle|\right]}{\sigma_\rho^{(t)}}\right)$$

Together, we have

$$\left|\mathop{\mathbb{E}}_{x,y,\rho}\left[y\ell'_t(w^{(t)};x,y,\rho)\left(\mathbb{1}_{\langle w_i^{(t)},x\rangle+\rho_i\geq b^{(t)}} - \mathbb{1}_{\langle w_i^{(t)},x'_j\rangle+\rho_i\geq b^{(t)}}\right)\langle \xi,\mathbf{M}_j\rangle\right]\right|$$

$$\leq O\left(\Gamma_t\frac{\mathbb{E}\left[|\langle \xi,\mathbf{M}_j\rangle^2\langle w_i^{(t)},\mathbf{M}_j\rangle|\right]}{\sigma_\rho^{(t)}}\right) = O\left(\frac{\Gamma_t}{d\sigma_\rho^{(t)}}|\langle w_i^{(t)},\mathbf{M}_j\rangle|\sigma_x^2\right) \quad (C.14)$$

Using the coordinate Lipscthizness, we also have

$$\left| \mathbb{E}_{x,y,\rho} \left[ y(\ell'_t(w^{(t)}; x, y, \rho) - \ell'_t(x'_j, y)) \mathbb{1}_{\langle w_i^{(t)}, x'_j \rangle + \rho_i \geq b^{(t)}} \langle \xi, \mathbf{M}_j \rangle \right] \right|$$
$$\leq L_{t,j} \mathbb{E}[\langle \xi, \mathbf{M}_j \rangle^2] \mathbf{Pr}[\langle w_i^{(t)}, x'_j \rangle + \rho_i \geq b^{(t)}] = O\left( \frac{\Gamma_t L_{t,j} \sigma_x^2}{d} \right) \quad \text{(C.15)}$$

Finally, we have $\mathbb{E}_{x,y,\rho} \left[ y \ell'_t(x'_j, y) \mathbb{1}_{\langle w_i^{(t)}, x'_j \rangle + \rho_i \geq b^{(t)}} \langle \xi, \mathbf{M}_j \rangle \right] = 0$. We can thus combine (C.14) and (C.15) to complete the proof of Lemma C.17a.

Next, we want to prove Lemma C.17b. We have: denote

$$q_{i'} = (\mathbb{1}_{\langle w_{i'}^{(t)}, \mathbf{M}z \rangle + \rho_{i'} + b_{i'}^{(t)} \geq -|b^{(t)}|/10} + \mathbb{1}_{-\langle w_{i'}^{(t)}, \mathbf{M}z \rangle + \rho_{i'} + b_{i'}^{(t)} \geq -|b^{(t)}|/10}$$

Since w.p. at least $1 - e^{-\Omega(\log^2 d)}$, $|\langle w_{i'}^{(t)}, \xi \rangle| \leq \frac{|b^{(t)}|}{10}$, in this case, we know that

$$\sum_{j \in [d]} |\ell'_t(w^{(t)}; x, y, \rho) - \ell'_t(x'_j, y)| |\langle \xi, \mathbf{M}_j \rangle|$$
$$\leq \sum_{j \in [d]} \langle \mathbf{M}_j, \xi \rangle^2 \cdot \sum_{i' \in [m]} |\langle w_{i'}^{(t)}, \mathbf{M}_j \rangle| \cdot (\mathbb{1}_{\langle w_{i'}^{(t)}, x \rangle + \rho_{i'} + b_{i'}^{(t)} \geq 0} + \mathbb{1}_{-\langle w_{i'}^{(t)}, x \rangle + \rho_{i'} + b_{i'}^{(t)} \geq 0})$$
$$\leq \sum_{j \in [d]} \langle \mathbf{M}_j, \xi \rangle^2 \cdot \sum_{i' \in [m]} |\langle w_{i'}^{(t)}, \mathbf{M}_j \rangle| \cdot q_{i'}$$

Note also we have:

$$\sum_{j \in [d]} |\langle w_{i'}^{(t)}, \mathbf{M}_j \rangle| \leq O(1) \cdot \|w_{i'}^{(t)}\|_2 + \sum_{j:\, i' \notin \mathcal{S}_{j,pot+}} |\langle w_{i'}^{(t)}, \mathbf{M}_j \rangle|$$
$$\leq O(1) \cdot \frac{\sigma_w^{(t)}}{\beta} + d \cdot \frac{k}{d\beta} b^{(t)} \leq 2\frac{k}{\beta} \cdot b^{(t)}$$

By Lemma C.19, we have that

$$\sum_{j \in [d]} \left| \mathbb{E}_{x,y,\rho} \left[ y(\ell'_t(w^{(t)}; x, y, \rho) - \ell'_t(x'_j, y)) \mathbb{1}_{\langle w_i^{(t)}, x \rangle + \rho_i \geq b^{(t)}} \langle \xi, \mathbf{M}_j \rangle \right] \right|$$
$$\leq e^{-\Omega(\log^2 d)} + \sum_{j \in [d]} \mathbb{E}_{x,y,\rho} \left[ |(\ell'_t(w^{(t)}; x, y, \rho) - \ell'_t(x'_j, y))| |\langle \xi, \mathbf{M}_j \rangle| \mathbb{1}_{\langle w_i^{(t)}, x \rangle + \rho_i \geq b^{(t)}} \right]$$
$$\leq e^{-\Omega(\log^2 d)} + \sum_{j \in [d]} \mathbb{E}_{x,y,\rho} \left[ |(\ell'_t(w^{(t)}; x, y, \rho) - \ell'_t(x'_j, y))| |\langle \xi, \mathbf{M}_j \rangle| q_i \right]$$
$$\leq e^{-\Omega(\log^2 d)} + \sum_{j \in [d]} \mathbb{E}_{x,y,\rho} \left[ \langle \mathbf{M}_j, \xi \rangle^2 \cdot \sum_{i' \in [m]} |\langle w_{i'}^{(t)}, \mathbf{M}_j \rangle| \cdot q_{i'} q_i \right] \quad \cdots \text{ taking expectation w.r.t. } \xi \text{ first.}$$
$$\leq e^{-\Omega(\log^2 d)} + O\left( \frac{\sigma_x^2}{d} \right) \cdot \sum_{j \in [d]} \mathbb{E}_{x,y,\rho} \left[ \sum_{i' \in [m]} |\langle w_{i'}^{(t)}, \mathbf{M}_j \rangle| \cdot q_{i'} q_i \right]$$
$$\leq O\left( \frac{\sigma_x^2}{d} \right) \cdot O\left( \frac{k}{\beta} b^{(t)} \cdot k\Xi_2 \right) \cdot O\left( \frac{k}{d} \right) + e^{-\Omega(\log^2 d)} \leq O\left( \frac{k^3 \Xi_2^4}{d^2} \sigma_x^2 + e^{-\Omega(\log^2 d)} \right) \quad \text{(C.16)}$$

Next, similar to the (C.14), we also have

$$\sum_{j\in[d]} \left| \mathbb{E}_{x,y,\rho}\left[y\ell'_t(x'_j,y)\left(\mathbb{1}_{\langle w_i^{(t)},x\rangle+\rho_i\geq b^{(t)}} - \mathbb{1}_{\langle w_i^{(t)},x'_j\rangle+\rho_i\geq b^{(t)}}\right)\langle \xi, \mathbf{M}_j\rangle\right]\right|$$

$$\leq \sum_{j\in[d]} O\left(\frac{\Gamma_t}{d\sigma_\rho^{(t)}}|\langle w_i^{(t)},\mathbf{M}_j\rangle|\sigma_x^2\right) \leq O\left(\frac{k}{d^2\sigma_\rho^{(t)}}\sigma_x^2\right)\cdot\sum_{j\in[d]}|\langle w_i^{(t)},\mathbf{M}_j\rangle| \leq O\left(\frac{k^2\log d}{d^2\beta}\sigma_x^2\right) \quad (C.17)$$

Combining (C.16) and (C.17) we finish the proof of Lemma C.17b. □

### C.4.2 Growth Coupling

We also have the following lemma which says, essentially, that all those neurons $i \in [m]$ satisfying $|\langle w_i^{(t)}, \mathbf{M}_j\rangle| \geq 2\sqrt{k}b^{(t)}$ for the same $j$, grows roughly in the same direction that is *independent* of $i$.

**Lemma C.18** (growth coupling). *Suppose at iteration $t$, $\mathcal{S}_{ept+}^{(t)} = [m]$. Then, for every $j \in [d]$, every $i \in [m]$ such that $|\langle w_i^{(t)}, \mathbf{M}_j\rangle| \geq 2\sqrt{k}b^{(t)}$, we have:*

$$\mathbb{E}_{x,y,\rho}\left[y\ell'_t(w^{(t)};x,y,\rho)\left(\mathbb{1}_{\langle w_i^{(t)},x\rangle+\rho_i\geq b^{(t)}} + \mathbb{1}_{-\langle w_i^{(t)},x\rangle+\rho_i\geq b^{(t)}}\right)z_j\right] = \mathbb{E}_{x,y,\rho}\left[y\ell'_t(w^{(t)};x,y,\rho)z_j\right] \pm O\left(\frac{k^{3/2}}{d^2}\right)$$

*Proof of Lemma C.18.* We first focus on the case when $\langle w_i^{(t)}, \mathbf{M}_j\rangle| \geq 2\sqrt{k}b^{(t)}$ is positive, and the reverse case is analogous. Conditional on $|z_j| = s > 0$, we know that $s \geq \frac{1}{\sqrt{k}}$. Thus, when $|\langle w_i^{(t)}, \mathbf{M}_j\rangle| \geq 2\sqrt{k}b^{(t)}$, $|\langle w_i^{(t)}, \mathbf{M}_j\rangle s| \geq 2b^{(t)}$. Now, using $\mathcal{S}_{ept+}^{(t)} = [m]$ and Lemma C.7, we can conclude that

- when $z_j > 0$, $\mathbf{Pr}[\langle w_i^{(t)}, x\rangle + \rho_i \geq b^{(t)} \mid z_j = s] \geq 1 - O\left(\frac{k}{d}\right)$;
- when $z_j < 0$, $\mathbf{Pr}[\langle w_i^{(t)}, x\rangle + \rho_i \geq b^{(t)} \mid z_j = -s] \leq O\left(\frac{k}{d}\right)$.

Thus, we can obtain

$$\mathbb{E}_{x,y,\rho}\left[y\ell'_t(w^{(t)};x,y,\rho)\mathbb{1}_{\langle w_i^{(t)},x\rangle+\rho_i\geq b^{(t)}}z_j\right] = \mathbb{E}_{x,y,\rho}\left[y\ell'_t(w^{(t)};x,y,\rho)z_j\mathbb{1}_{z_j>0}\right] \pm O\left(\frac{k}{d}\right)\times\mathbb{E}|z_j|$$

$$= \mathbb{E}_{x,y,\rho}\left[y\ell'_t(w^{(t)};x,y,\rho)z_j\mathbb{1}_{z_j>0}\right] \pm O\left(\frac{k^{3/2}}{d^2}\right)$$

In the symmetric case, we also have

$$\mathbb{E}_{x,y,\rho}\left[y\ell'_t(w^{(t)};x,y,\rho)\mathbb{1}_{-\langle w_i^{(t)},x\rangle+\rho_i\geq b^{(t)}}z_j\right] = \mathbb{E}_{x,y,\rho}\left[y\ell'_t(w^{(t)};x,y,\rho)z_j\mathbb{1}_{z_j<0}\right] \pm O\left(\frac{k^{3/2}}{d^2}\right) \quad □$$

### C.4.3 Activation Probabilities

**Lemma C.19** (activation after ept+). *Suppose $\mathcal{S}_{ept+}^{(t)} = [m]$ and $\mathcal{S}_{j,pot}^{(0)} \supseteq \mathcal{S}_{j,pot+}^{(t)}$ for every $j \in [d]$. Then, with probability at least $1 - e^{-\Omega(\log^2 d)}$,*

- $\left|\left\{i \in [m] \text{ s.t. } |\langle w_i^{(t)}, x\rangle| \geq \frac{b^{(t)}}{10}\right\}\right| \leq O(k\Xi_2)$.

- $\left|\left\langle w_i^{(t)}, \sum_{j\in[d]:\, i\notin\mathcal{S}_{j,pot+}^{(t)}} \mathbf{M}_j z_j + \xi\right\rangle\right| \leq \frac{b^{(t)}}{10}$ *for every $i \in [m]$.*

*Proof.* For every $i \in [m]$ and $j \in [d]$ with $i \notin \mathcal{S}_{j,pot+}^{(t)}$, we have $|\langle w_i^{(t)}, \mathbf{M}_j\rangle| \leq \frac{k}{d\beta}b^{(t)}$. Therefore, by Bernstein's inequality (similar to Lemma C.4b), we know with probability at least $1 - e^{-\Omega(\log^2 d)}$, for every $i \in [m]$,

$$\left|\left\langle w_i^{(t)}, \sum_{j \in [d]:\, i \notin \mathcal{S}_{j,pot+}^{(t)}} \mathbf{M}_j z_j + \xi \right\rangle\right| \leq \frac{b^{(t)}}{10} \tag{C.18}$$

With probability at least $1 - e^{-\Omega(\log^2 d)}$ it satisfies $\sum_{j \in [d]} \mathbb{1}_{z_j \neq 0} \leq O(k)$ (since each $z_j \neq 0$ with probability at most $O(\frac{k}{d})$). Therefore, denoting by $\Lambda = \bigcup_{j \in [d]:\, z_j \neq 0} \mathcal{S}_{j,pot+}^{(t)}$, we have $|\Lambda| \leq O(k\Xi_2)$ (since every $|\mathcal{S}_{j,pot+}^{(t)}| \leq \Xi_2$). Now, for any $i \in [m] \setminus \Lambda$, inequality (C.18) immediately gives

$$\left|\langle w_i^{(t)}, x\rangle\right| \leq \frac{b^{(t)}}{10} \ .$$

Therefore, the number of $i \in [m]$ satisfying $\left|\langle w_i^{(t)}, x\rangle\right| \geq \frac{b^{(t)}}{10}$ cannot be more than $O(k\Xi_2)$. □

### C.4.4 Coordinate Lipscthizness Bound

**Lemma C.20** (coordinate Lipschitzness). *For every $j \in [d]$, let us define $\gamma_j^{(t)} = \sum_{i \in \mathcal{S}_{j,pot+}^{(t)}} |\langle w_i^{(t)}, \mathbf{M}_j\rangle|$. Then, suppose $\mathcal{S}_{ept+}^{(t)} = [m]$ and suppose $\mathcal{S}_{j,pot}^{(0)} \supseteq \mathcal{S}_{j,pot+}^{(t)}$, we have*

$$L_{t,j} \leq \gamma_j^{(t)} + O\left(\frac{k}{\sqrt{d}}\right) \leq \gamma_j^{(t)} + O\left(\frac{1}{\Xi_2^3}\right)$$

*Proof of Lemma C.20.* Clearly we have

$$L_{t,j} \leq \sum_{i \in \mathcal{S}_{j,pot+}^{(0)}} |\langle w_i^{(t)}, \mathbf{M}_j\rangle| + \sum_{i \notin \mathcal{S}_{j,pot+}^{(0)}} |\langle w_i^{(t)}, \mathbf{M}_j\rangle| \cdot \left(\mathbb{1}_{\langle w_i^{(t)}, x\rangle + \rho_i \geq 0.9b_i^{(t)}} + \mathbb{1}_{-\langle w_i^{(t)}, x\rangle + \rho_i \geq 0.9b_i^{(t)}}\right)$$

By Lemma C.19 and the randomness of $\rho_i$, we know with probability at least $1 - e^{-\Omega(\log^2 d)}$, the number of activate neurons $i \notin \mathcal{S}_{j,pot+}^{(t)}$—meaning $\langle w_i^{(t)}, x\rangle + \rho_i \geq 0.9b_i^{(t)}$ or $-\langle w_i^{(t)}, x\rangle + \rho_i \geq 0.9b_i^{(t)}$— is at most $O(k\Xi_2)$. On the other hand, when $i \notin \mathcal{S}_{j,pot+}^{(t)}$, we know that

$$|\langle w_i^{(t)}, \mathbf{M}_j\rangle| \leq \frac{k}{d\beta}b^{(t)} \leq \frac{k\Xi_2^2}{d}$$

Therefore, together, the total contribution from these active neurons with $i \notin \mathcal{S}_{j,pot+}^{(t)}$ is at most $\frac{k\Xi_2^2}{d} \cdot O(k\Xi_2) < O(\frac{k^2\Xi_2^3}{d}) < O\left(\frac{1}{\Xi_2^3}\right)$. This completes the proof. □

### C.4.5 Regularization

Following the same argument as (C.9) from phase I, we know at any iteration $t$, as long as $\mathcal{S}_{ept+}^{(t)} = [m]$,

$$\langle w_i^{(t+1)}, \mathbf{M}_j\rangle = \langle w_i^{(t)}, \mathbf{M}_j\rangle(1 - \eta\lambda - \eta\lambda\|w_i^{(t)}\|_2) \pm \frac{\eta}{\mathsf{poly}(d)}$$

$$+ \mathop{\mathbb{E}}_{x,y=y(x),\rho}\left[y\ell_t'(w^{(t)}; x, y, \rho) \sum_{i=1}^m \left(\mathbb{1}_{\langle w_i^{(t)}, x\rangle + \rho_i \geq b^{(t)}} + \mathbb{1}_{-\langle w_i^{(t)}, x\rangle + \rho_i \geq b^{(t)}}\right) \cdot \left(z_j + \langle \xi, \mathbf{M}_j\rangle\right)\right] \ . \tag{C.19}$$

In this and the next subsection, we shall repeatedly apply growth lemmas to (C.19). Before doing so, let us note $\sigma_\rho^{(t)} = o(b^{(t)} \log d) \leq o(\beta \Xi_2^2 \log d)$, so using our parameter choice of $\beta$ and using $k \leq d^{1-c_0}$,

$$\frac{\sqrt{k}\sigma_\rho^{(t)} \log d}{d} + \frac{\sqrt{k}}{\beta d^{3/2}} = o\left(\frac{\sqrt{k}\beta \Xi_2^2 \log^2 d}{d}\right) + \frac{\sqrt{k}}{\beta d^{3/2}} = o(\frac{1}{d}) \tag{C.20}$$

This means, when applying the aforementioned growth lemmas Lemma C.14, Lemma C.15, Lemma C.16, the additional terms $\frac{\sqrt{k}\sigma_\rho^{(t)} \log d}{d}$ and $\frac{\sqrt{k}}{\beta d^{3/2}}$ are negligible.

We also have the following regularity lemma:

**Lemma C.21** (regularity). *For every $T \leq d^{O(\log d)}/\eta$, suppose $\mathcal{S}_{ept+}^{(t)} = \mathcal{S}_{ept++}^{(t)} = [m]$ and $\mathcal{S}_{j,pot}^{(0)} \supseteq \mathcal{S}_{j,pot+}^{(t)}$ hold for every $t \leq T$ and $j \in [d]$. Then, we have for every $t \leq T$, with probability at least $1 - e^{-\Omega(\log^2 d)}$,*

$$\forall j \in [d], \forall i \in [m]: \quad L_{t,j} \leq O(\Xi_2^2), \quad \|w_i^{(t)}\|_2 \leq O(\Xi_2^2), \quad |f_t(x)| \leq O(\Xi_2^2 \log d) \ .$$

*Proof of Lemma C.21.* By substituting Lemma C.15, Lemma C.17a and (C.20) into (C.19), we have for every $j \in [d]$:

$$|\langle w_i^{(t+1)}, \mathbf{M}_j \rangle| \leq |\langle w_i^{(t)}, \mathbf{M}_j \rangle|(1 - \eta\lambda \|w_i^{(t)}\|_2) + \eta O\left(\frac{1 + L_{t,j}}{d}\right)$$

$$\leq |\langle w_i^{(t)}, \mathbf{M}_j \rangle| \left(1 - \eta\lambda |\langle w_i^{(t)}, \mathbf{M}_j \rangle|\right) + \eta O\left(\frac{1 + L_{t,j}}{d}\right)$$

Summing up over all $i \in \mathcal{S}_{j,pot}^{(0)}$, and using Cauchy-Schwarz inequality together with $|\mathcal{S}_{j,pot}^{(0)}| \leq \Xi_2$, we have

$$\sum_{i \in \mathcal{S}_{j,pot}^{(0)}} |\langle w_i^{(t+1)}, \mathbf{M}_j \rangle| \leq \sum_{i \in \mathcal{S}_{j,pot}^{(0)}} |\langle w_i^{(t)}, \mathbf{M}_j \rangle| - \frac{\eta\lambda}{\Xi_2} \left(\sum_{i \in \mathcal{S}_{j,pot}^{(0)}} |\langle w_i^{(t)}, \mathbf{M}_j \rangle|\right)^2 + \eta O\left(\frac{1 + L_{t,j}}{d}\right) \cdot \Xi_2$$

Combining this with $L_{t,j} \leq \sum_{i \in \mathcal{S}_{j,pot}^{(0)}} |\langle w_i^{(t)}, \mathbf{M}_j \rangle| + O(\frac{1}{\Xi_2^3})$ from Lemma C.20 and our choice $\lambda \geq \frac{1}{d}$, we have (for every $j \in [d]$ and $t \leq T$),

$$\sum_{i \in \mathcal{S}_{j,pot}^{(0)}} |\langle w_i^{(t)}, \mathbf{M}_j \rangle| \leq O(\Xi_2^2)$$

This also implies $L_{t,j} \leq O(\Xi_2^2)$ as well as

$$\|w_i^{(t)}\|^2 = \sum_{j:\, i \in \mathcal{S}_{j,pot+}^{(t)}} |\langle w_i^{(t)}, \mathbf{M}_j \rangle|^2 + \sum_{j:\, i \notin \mathcal{S}_{j,pot+}^{(t)}} |\langle w_i^{(t)}, \mathbf{M}_j \rangle|^2 \leq O(\Xi_2^4) + O(\frac{k^2}{d\beta^2}(b^{(t)})^2) \leq O(\Xi_2^4) \ .$$

Finally, for the objective value, we wish use $L_{t,j} \leq O(\Xi_2^2)$ and apply a high-probability Bernstein variant of the McDiarmid's inequality (see Lemma H.3).

Specifically, consider random $z, \xi, \rho$. For notation simplicity, let us write $\xi = \sum_{j \in [d]} \mathbf{M}_j \xi_j$ for i.i.d. random $\xi_j \sim \mathcal{N}(0, \frac{\sigma_x^2}{d})$.

Now, for every $j \in [m]$, suppose we change $z_j$ to $z_j'$ and $\xi_j$ to $\xi_j'$ with the same distribution. Then, with probability at least $1 - e^{-\Omega(\log^2 d)}$,

$$|f_t(z, \xi, \rho) - f_t(z_{-j}, z_j', \xi_{-j}, \xi_j', \rho)| \leq L_{t,j} \cdot (|z_j| + |z_j'| + |\xi_j| + |\xi_j'|)$$

This implies with probability at least $1 - e^{-\Omega(\log^2 d)}$,

$$\mathbb{E}_{z_j, z'_j} |f_t(z, \xi, \rho) - f_t(z_{-j}, z'_j, \xi, \rho)|^2 \leq O(L_{t,j}^2) \cdot \frac{1}{d}$$

Therefore, we can apply Lemma H.3 to derive that with probability at least $1 - e^{-\Omega(\log^2 d)}$,

$$|f_t(x, \rho) - \mathbb{E}_{z, \xi}[f_t(x, \rho)]| \leq O(\Xi_2^2 \log d)$$

Finally, noticing that for every $\rho$, by symmetry $\mathbb{E}_{z,\xi}[f_t(x, \rho)] = 0$. This finishes the bound on the objective value. $\square$

We also prove this Lemma, which gives a lower bound on the loss:

**Lemma C.22** (loss lower bound). *In every iteration $t$, define $L_{\max} := \max_{j \in [d]}\{L_{t,j}\}$ and suppose $L_{\max} \leq O(\Xi_2^2)$. Then we have:*

$$\mathbb{E}_{x,\rho}[\ell'_t(w^{(t)}; x, y, \rho)] = \Omega\left(\min\left\{1, \frac{1}{L_{\max}^2 \log^2 d}\right\}\right)$$

*Proof of Lemma C.22.* Let $\alpha \in \left[\frac{1}{(\Xi_2)^5}, 1\right]$ be a fixed value to be chosen later, and $\mathcal{S}_0 \subseteq [d]$ be an arbitrary subset of size $|\mathcal{S}_0| = \alpha d$. Consider a randomly sampled vector $z$ and let $x = \mathbf{M}z + \xi$ be the corresponding input. We construct another $z'$ that is generated from the following process

1. Let $\mathcal{S}_{re,z} \subseteq \mathcal{S}_0$ be the set consisting of all $i \in \mathcal{S}_0$ with $|z_i| = \Theta\left(\frac{1}{\sqrt{k}}\right)$.
2. For all $i \notin \mathcal{S}_{re,z}$, pick $z'_i = z_i$.
3. For all $i \in \mathcal{S}_{re,z}$, pick $z'_i = z_i$ or $z'_i = -z_i$ each with probability 0.5, independently at random.

Obviously, $z'$ has the same distribution as $z$. Now, let us define $x' = \mathbf{M}z' + \xi$, $y' = \text{sign}(\langle w^\star, z' \rangle)$.

Since $|\mathcal{S}_0| = \alpha d$, recalling the distribution property that $\mathbf{Pr}\left[|z_i| = \Theta\left(\frac{1}{\sqrt{k}}\right)\right] = \Omega\left(\frac{k}{d}\right)$, we know with probability at least $1 - e^{-\Omega(\log^2 d)}$ over the choice of $z$, $|\mathcal{S}_{re,z}| = \Theta(\alpha k)$. We call this event $\mathcal{E}_1(z)$.

Let us denote by $b_i = \frac{z'_i}{z_i} \in \{-1, 1\}$ for every $i \in \mathcal{S}_{re,z}$. We can therefore write $f_t(w^{(t)}; x', \rho) = f(z, b, \xi, \rho)$ to emphasize that the randomness comes from $z, b, \xi, \rho$. Using the definition of coordinate Lipscthizness, we know with probability at least $1 - e^{-\Omega(\log^2 d)}$ over $z, \xi, \rho$, it satisfies

$$\forall k \in \mathcal{S}_{re,z}, \forall b \in \{-1,1\}^{\mathcal{S}_{re,z}}, \forall b'_k \in \{-1, +1\}: \quad |f(z, b, \xi, \rho) - f(z, (b_{-k}, b_k), \xi, \rho)| \leq O\left(\frac{L_{\max}}{\sqrt{k}}\right)$$

Let $\mathcal{E}_2(z, \xi, \rho)$ denote the event where the above statement holds.

Now, conditioning on $\mathcal{E}_1(z)$ and $\mathcal{E}_2(z, \xi, \rho)$ both hold, we can apply standard MiDiarmid's inequality (see Lemma H.2) over the randomness of $b$, and derive that with probability at least $1 - e^{-\Omega(\log^2 d)}$ over $b$,

$$\left| f(z, b, \xi, \rho) - \mathbb{E}_b[f(z, b, \xi, \rho)] \right| \leq O\left(L_{\max} \sqrt{\alpha} \log d\right)$$

Let $\mathcal{E}_3(b||z, \xi, \rho)$ denote the (conditional) event where the above statement holds.

In sum, by combining $\mathcal{E}_1, \mathcal{E}_2, \mathcal{E}_3$, we know with probability at least $1 - e^{-\Omega(\log^2 d)}$ over $z, z', \xi, \rho$, it satisfies

$$\left| f_t(w^{(t)}; x', \rho) - \mathbb{E}_{z'}\left[f_t(w^{(t)}; x', \rho) \mid z, \xi, \rho\right] \right| \leq O\left(L_{\max} \sqrt{\alpha} \log d\right) \quad .$$

As a simple corollary, if we generate another copy $z''$ in the same way as $z'$, and denote by $x'' = \mathbf{M}z'' + \xi$, then with probability at least $1 - e^{-\Omega(\log^2 d)}$ over $z, z', z'', \xi, \rho$, it satisfies

$$\left| f_t(w^{(t)}; x', \rho) - f_t(w^{(t)}; x'', \rho) \right| \leq O\left(L_{\max}\sqrt{\alpha}\log d\right) \ . \tag{C.21}$$

Now, let us denote by $y' = \mathsf{sign}(\langle w^\star, z' \rangle)$ and $y'' = \mathsf{sign}(\langle w^\star, z'' \rangle)$ and compare them. Let us write

$$A = \sum_{i \in [d] \setminus \mathcal{S}_{re,z}} w_i^\star z_i \ , \quad B = \sum_{i \in \mathcal{S}_{re,z}} w_i^\star z_i' \ , \quad C = \sum_{i \in \mathcal{S}_{re,z}} w_i^\star z_i'' \ .$$

Thus, we have $y' = \mathsf{sign}(A + B)$ and $y'' = \mathsf{sign}(A + C)$.

First using a minor variant of Lemma H.1b, we have[20]

$$\mathbf{Pr}\left[A \in [0, \sqrt{\alpha}]\right] \geq \Omega(\sqrt{\alpha})$$

Denote this event by $\mathcal{E}_4(z)$.

Next, conditioning on any fixed $z$ which satisfies $\mathcal{E}_1(z)$ and $\mathcal{E}_4(z)$, we know that $B$ and $C$ become *independent*, each controlled by $|\mathcal{S}_{re,z}| = \Theta(\alpha k)$ random Bernoulli variables. Therefore, we can apply a Wasserstein distance version of the central limit theorem (that can be derived from [110], full statement see [6, Appendix A.2]) to derive that, for a Gaussian variable $g \sim (0, V^2)$ where $V^2 = \sum_{j \in \mathcal{S}_{re,z}} (z_j)^2 = \Theta(\alpha)$, the Wasserstein distance:

$$\mathcal{W}_2(B, g) \leq O\left(\frac{\log k}{\sqrt{k}}\right) \quad \text{and} \quad \mathcal{W}_2(C, g) \leq O\left(\frac{\log k}{\sqrt{k}}\right)$$

This means with probability at least $\Omega(1)$, it satisfies $B \in [0, \sqrt{\alpha}]$ and $C \leq -5\sqrt{\alpha}$.

To sum up, we know with probability at least $\Omega(\sqrt{\alpha})$, it satisfies $A, B \in [0, \sqrt{\alpha}]$ and $C \leq -5\sqrt{\alpha}$. This means $y' \neq y''$, or in symbols,

$$\mathbf{Pr}[y' \neq y''] \geq \Omega(\sqrt{\alpha}) \ . \tag{C.22}$$

Finally, conditioning on both (C.21) and (C.22) happen, we know that

- either $\mathsf{sign}(f_t(w^{(t)}; x', \rho)) = \mathsf{sign}(f_t(w^{(t)}; x'', \rho)$, in which case $\ell_t'(w^{(t)}; x', y', \rho) + \ell'(w^{(t)}; x'', y'', \rho) \geq \frac{1}{2}$,
- or $|f_t(w^{(t)}; x', \rho)| \leq O\left(L_{\max}\sqrt{\alpha}\log d\right)$ and $|f_t(w^{(t)}; x'', \rho)| \leq O\left(L_{\max}\sqrt{\alpha}\log d\right)$, in which case if we choose $\alpha = \min\{\frac{1}{2}, \frac{1}{L_{\max}^2 \log^2 d}\}$, then we have $|f_t(w^{(t)}; x', \rho)|, |f_t(w^{(t)}; x'', \rho)| \leq O(1)$ and therefore $\ell_t'(w^{(t)}; x', y', \rho) + \ell_t'(w^{(t)}; x'', y'', \rho) \geq \Omega(1)$.

To sum up, we have

$$\mathbb{E}_{x, y=y(x), \rho}[\ell_t'(w^{(t)}; x, y, \rho)] = \frac{1}{2} \mathbb{E}_{z, z', z'', \xi, \rho}[\ell_t'(w^{(t)}; x', y', \rho) + \ell_t'(w^{(t)}; x'', y'', \rho)]$$

$$\geq \Omega(\sqrt{\alpha}) = \Omega\left(\min\{1, \frac{1}{L_{\max}^2 \log^2 d}\}\right) \ . \qquad \square$$

### C.4.6 Proof of Theorem C.2

*Proof of Theorem C.2.* We first prove that for every $t \geq T_{\mathsf{b}}$,

$$\mathcal{S}_{j,pot}^{(t)} \subseteq \mathcal{S}_{j,pot+}^{(t)} \subseteq \mathcal{S}_{j,pot}^{(0)} \tag{C.23}$$

---

[20]To be precise, we can do so since we still have at least $(1-\alpha)d \geq \frac{d}{2}$ coordinates.

Note from the definitions the relationship $\mathcal{S}_{j,pot}^{(t)} \subseteq \mathcal{S}_{j,pot+}^{(t)}$ always holds, so we only need to prove the second inclusion.

Suppose (C.23) holds until iteration $t$. Then, for every $i \notin \mathcal{S}_{j,pot}^{(0)}$, let us apply Lemma C.16, Lemma C.17a together with (C.20) and $L_{t,j} \leq O(\Xi_2^2)$ (using Lemma C.21) to (C.19). We get

$$|\langle w_i^{(t+1)}, \mathbf{M}_j \rangle| \leq |\langle w_i^{(t)}, \mathbf{M}_j \rangle|(1 - \eta\lambda) + O\left(\frac{\eta k \Xi_2^2}{d^2}\right)$$

Therefore, for those $t$ that are sufficiently large so that $b^{(t+1)} = \beta \Xi_2^2$, we have (using $\lambda \geq \frac{1}{d}$)

$$|\langle w_i^{(t+1)}, \mathbf{M}_j \rangle| \leq O\left(\frac{\eta k \Xi_2^2}{d^2}\right) \cdot \frac{1}{\eta\lambda} = O\left(\frac{k \Xi_2^2}{d \cdot d\lambda}\right) \leq \frac{k}{d\beta} b^{(t+1)}$$

and for those $t$ that are still small so that $b^{(t+1)} = \Theta(\frac{\eta(t+1)}{d})$, we have

$$|\langle w_i^{(t+1)}, \mathbf{M}_j \rangle| \leq O\left(\frac{\eta k \Xi_2^2}{d^2} \cdot (t+1)\right) \ll \frac{k}{d\beta} b^{(t+1)}$$

Together, this means $i \notin \mathcal{S}_{j,pot+}^{(t+1)}$ so (C.23) holds for all $t \geq T_{\mathsf{b}}$ and $T \leq d^{O(\log d)}/\eta$.

**Phase II.1.** We will construct a threshold $T_{\mathsf{e}}$ and prove inductively for all $t \in [T_{\mathsf{b}}, T_{\mathsf{e}}]$. Initially at $t = T_{\mathsf{b}}$, by Lemma C.20 we have $L_{t,j} = o(1)$. As long as $L_{t,j} = o(1)$ holds for all $j \in [d]$, we have

- for every $i \in [m]$, substituting Lemma C.15, Lemma C.17a and (C.20) into (C.19),

$$|\langle w_i^{(t+1)}, \mathbf{M}_j \rangle| \leq |\langle w_i^{(t)}, \mathbf{M}_j \rangle| + O\left(\frac{\eta}{d}\right) \leq \cdots \leq O\left(\frac{\eta}{d} \cdot t\right)$$

- for every $i \notin \mathcal{S}_{j,pot}^{(t)}$, substituting Lemma C.16, Lemma C.17a and (C.20) into (C.19),

$$|\langle w_i^{(t+1)}, \mathbf{M}_j \rangle| \leq |\langle w_i^{(t)}, \mathbf{M}_j \rangle| + O\left(\frac{\eta k \log d}{d^2}\right) \leq \cdots \leq O\left(\frac{\eta k \log d}{d^2} \cdot t\right)$$

Since for each $i$, the number of $j$ satisfying $i \in \mathcal{S}_{j,pot}^{(t)}$ is at most $O(1)$ (using $\mathcal{S}_{j,pot}^{(t)} \subseteq \mathcal{S}_{j,pot}^{(0)}$ and $\mathcal{S}_{ept}^{(0)} = [m]$), we have

$$\|w_i^{(t+1)}\| \leq O(\frac{\eta}{d} \cdot t) \tag{C.24}$$

These bounds together mean several things:

- $L_{t,j} = o(1)$ for all $j \in [d]$ and $t \in [T_{\mathsf{b}}, T_{\mathsf{e}}]$ with $T_{\mathsf{e}} = \Theta\left(\frac{d}{\eta \Xi_2 \log d}\right)$.

  Indeed, (C.24) gives $|\langle w_i^{(t)}, \mathbf{M}_j \rangle| \leq O(\frac{1}{\Xi_2 \log d})$, but the number of $j$ satisfying $i \in \mathcal{S}_{j,pot}^{(t)}$ is at most $O(1)$. So we can apply Lemma C.20 to get $L_{t,j} = o(1)$.

- $\mathcal{S}_{ept++}^{(t)} = [m]$ for all $t \in [T_{\mathsf{b}}, T_{\mathsf{e}}]$.

  Indeed,

  - for those $t$ that are small so that $\sigma_w^{(t)} = \Theta(\frac{\eta}{d\sqrt{\log d}} t)$, we have (C.24) implies $\|w_i^{(t)}\| \leq O(\sqrt{\log d} \cdot \sigma_w^{(t)}) \ll \frac{\sigma_w^{(t)}}{\beta}$; and
  - for those $t$ that are large so that $\sigma_w^{(t)} = \Theta(\frac{\beta \Xi_2^2}{\sqrt{\log d}})$, we have (C.24) implies $\|w_i^{(t)}\| \leq O(\frac{1}{\Xi_2 \log d}) \ll \frac{\sigma_w^{(t)}}{\beta}$.

  Together we have $i \in \mathcal{S}_{ept++}^{(t)}$.

- $\mathcal{S}_{ept+}^{(t)} = [m]$ for all $t \in [T_{\mathsf{b}}, T_{\mathsf{e}}]$.

  This is a direct corollary of $\mathcal{S}_{ept++}^{(t)} = [m]$ together with the property that the number of $j$ satisfying $i \in \mathcal{S}_{j,pot+}^{(t)}$ is at most $O(1)$.

Next, let us consider any $j \in [d]$ with $i \in \mathcal{S}_{j,sure}^{(0)}$. At any iteration $t \in [T_{\mathsf{b}}, T_{\mathsf{e}}]$, substituting Lemma C.14, Lemma C.17a, (C.24), and (C.20) into (C.19),

$$|\langle w_i^{(t+1)}, \mathbf{M}_j\rangle| \geq |\langle w_i^{(t)}, \mathbf{M}_j\rangle|(1 - \eta\lambda - \eta\lambda\|w_i^{(t)}\|_2) + \Omega\left(\frac{\eta}{d}\right)$$
$$\geq |\langle w_i^{(t)}, \mathbf{M}_j\rangle|(1 - 2\eta\lambda) + \Omega\left(\frac{\eta}{d}\right)$$

This means two things:

- The value $|\langle w_i^{(t)}, \mathbf{M}_j\rangle|$ keeps increasing as $t$ increases, until it reaches $\Theta(\frac{\eta}{d} \cdot \frac{1}{\eta\lambda}) = \Theta(\frac{1}{d\lambda})$ and at that point it may decrease but will not fall below $\Theta(\frac{1}{d\lambda})$. This ensures $i \in \mathcal{S}_{j,sure}^{(t)}$.

- At $t = T_{\mathsf{e}}$, we must have $i \in \mathcal{S}_{j,sure+}^{(t)}$ because

$$|\langle w_i^{(t)}, \mathbf{M}_j\rangle| \geq \Omega(\frac{\eta}{d}T_{\mathsf{e}}) \geq \Omega\left(\frac{1}{\Xi_2 \log d}\right) \geq \Omega\left(\frac{1}{\Xi_2 \log d}\right) \cdot \frac{(b^{(t)})^2}{\beta^2 \Xi_2^4}$$
$$\geq \Omega\left(\frac{1}{k\beta^2\Xi_2^5 \log d}\right) \cdot 4k(b^{(t)})^2 \geq 4k\left(b^{(t)}\right)^2$$

To sum up, at iteration $t = T_{\mathsf{e}}$, we have

- for $i \in \mathcal{S}_{j,sure}^{(0)}$, $|\langle w_i^{(t)}, \mathbf{M}_j\rangle| \geq \Omega\left(\frac{1}{\Xi_2 \log d}\right)$;

- for $i \in \mathcal{S}_{j,pot}^{(0)}$, $|\langle w_i^{(t)}, \mathbf{M}_j\rangle| \leq O(\frac{1}{\Xi_2 \log d})$

- for $i \notin \mathcal{S}_{j,pot}^{(0)}$, $|\langle w_i^{(t)}, \mathbf{M}_j\rangle| \leq O(\frac{k}{d\Xi_2})$

**Phase II.2.** We first make a quick observation that

- $\mathcal{S}_{ept+}^{(t)} = \mathcal{S}_{ept++}^{(t)} = [m]$ for all $t \geq T_{\mathsf{e}}$.

  Indeed, from iteration $t = T_{\mathsf{e}}$ on, we have $b^{(t)} = \beta\Xi_2^2$. Using Lemma C.21 we have for every $i \in [m]$, $\|w_i^{(t)}\| \leq O(\Xi_2^2) \leq \frac{\sigma_w^{(t)}}{\beta}$. Thus, $\mathcal{S}_{ept++}^{(t)} = [m]$ holds for all $t \geq T_{\mathsf{e}}$. As for $\mathcal{S}_{ept+}^{(t)} = [m]$, it is a simple corollary of $\mathcal{S}_{ept++}^{(t)} = [m]$ together with the property that the number of $j$ satisfying $i \in \mathcal{S}_{j,pot+}^{(t)}$ is at most $O(1)$.

Next, we claim for every $i \in \mathcal{S}_{j,sure+}^{(T_{\mathsf{e}})}$ and every $t \geq T_{\mathsf{e}}$, it must hold that

$$|\langle w_i^{(t)}, \mathbf{M}_j\rangle| \geq \frac{1}{C'}\left(\max_{i' \in [m]} |\langle w_{i'}^{(t)}, \mathbf{M}_j\rangle|\right) \quad \text{and} \quad i \in \mathcal{S}_{j,sure+}^{(t)} \tag{C.25}$$

for some sufficiently large constant $C' > 1$. We prove by induction. Suppose (C.25) holds for $t$ and we consider $t + 1$. By the definition of $i \in \mathcal{S}_{j,sure+}^{(t)}$, we know $|\langle w_i^{(t)}, \mathbf{M}_j\rangle| \geq 2\sqrt{k}b^{(t)}$. Now, consider every other $i' \in [m] \setminus \{i\}$

- if $|\langle w_{i'}^{(t)}, \mathbf{M}_j\rangle| < 2C'|\langle w_i^{(t)}, \mathbf{M}_j\rangle|$, then after one iteration we still have $|\langle w_{i'}^{(t+1)}, \mathbf{M}_j\rangle| < C'|\langle w_i^{(t+1)}, \mathbf{M}_j\rangle|$.

- if $|\langle w_{i'}^{(t)}, \mathbf{M}_j \rangle| > 2C'|\langle w_i^{(t)}, \mathbf{M}_j \rangle|$, then we have

$$\|w_i^{(t)}\|^2 = |\langle w_i^{(t)}, \mathbf{M}_j \rangle|^2 + \sum_{j' \neq j} |\langle w_i^{(t)}, \mathbf{M}_{j'} \rangle|^2 \leq |\langle w_i^{(t)}, \mathbf{M}_j \rangle|^2 + (d-1) \cdot \frac{k^2}{d^2 \beta^2}(b^{(t)})^2$$

$$\leq 10|\langle w_i^{(t)}, \mathbf{M}_j \rangle|^2 \leq \frac{10}{2C'}|\langle w_{i'}^{(t)}, \mathbf{M}_j \rangle|^2 \leq \|w_{i'}^{(t)}\|^2 \tag{C.26}$$

Therefore, applying Lemma C.18 and Lemma C.17a (for $i$ and $i'$), and using $\beta \leq \frac{1}{\sqrt{k}}$, we have

$$|\langle w_i^{(t+1)}, \mathbf{M}_j \rangle| = |\langle w_i^{(t)}, \mathbf{M}_j \rangle|(1 - \eta\lambda - \eta\lambda\|w_i^{(t)}\|_2) + \eta \mathop{\mathbb{E}}_{x,y,\rho}\left[y\ell_t'(w^{(t)}; x, y, \rho)z_j\right] \pm O\left(\frac{\eta k^{1.5}}{d^2}\right)$$

$$|\langle w_{i'}^{(t+1)}, \mathbf{M}_j \rangle| = |\langle w_{i'}^{(t)}, \mathbf{M}_j \rangle|(1 - \eta\lambda - \eta\lambda\|w_{i'}^{(t)}\|_2) + \eta \mathop{\mathbb{E}}_{x,y,\rho}\left[y\ell_t'(w^{(t)}; x, y, \rho)z_j\right] \pm O\left(\frac{\eta k^{1.5}}{d^2}\right)$$

Taking the difference and using (C.26), we have

$$|\langle w_{i'}^{(t+1)}, \mathbf{M}_j \rangle| - |\langle w_i^{(t+1)}, \mathbf{M}_j \rangle| \leq \left(|\langle w_{i'}^{(t)}, \mathbf{M}_j \rangle| - |\langle w_i^{(t)}, \mathbf{M}_j \rangle|\right)(1 - \eta\lambda - \eta\lambda\|w_i^{(t)}\|_2) + O\left(\frac{\eta k^{1.5}}{d^2}\right)$$

$$\leq \left(|\langle w_{i'}^{(t)}, \mathbf{M}_j \rangle| - |\langle w_i^{(t)}, \mathbf{M}_j \rangle|\right) - \Omega(\eta\lambda(\sqrt{k}b^{(t)})^3) + O\left(\frac{\eta k^{1.5}}{d^2}\right)$$

$$\leq \left(|\langle w_{i'}^{(t)}, \mathbf{M}_j \rangle| - |\langle w_i^{(t)}, \mathbf{M}_j \rangle|\right)$$

thus we continue to have $|\langle w_{i'}^{(t+1)}, \mathbf{M}_j \rangle| \leq C'|\langle w_i^{(t+1)}, \mathbf{M}_j \rangle|$.

Putting these together we show that the first half of (C.25) holds at $t+1$.

As for why $i \in \mathcal{S}_{j,sure+}^{(t+1)}$, we consider two cases.

- If $|\langle w_i^{(t)}, \mathbf{M}_j \rangle| \geq 4\sqrt{k}b^{(t)}$, then in one iteration we should still have $|\langle w_i^{(t+1)}, \mathbf{M}_j \rangle| \geq 2\sqrt{k}b^{(t)} = 2\sqrt{k}b^{(t+1)}$.
- If $|\langle w_i^{(t)}, \mathbf{M}_j \rangle| \leq 4\sqrt{k}b^{(t)}$, then by the first half of (C.25) together with Lemma C.20, we know the Lipscthizness $L_{t,j} \leq O(\sqrt{k}b^{(t)} \cdot \Xi_2) + O\left(\frac{1}{\Xi_2^3}\right) \leq o(1)$. In this case, we also have (see (C.26)) $\|w_i^{(t)}\|^2 \leq O(k(b^{(t)})^2) = o(1)$. Applying Lemma C.14 and Lemma C.17a again we have

$$|\langle w_i^{(t+1)}, \mathbf{M}_j \rangle| \geq |\langle w_i^{(t)}, \mathbf{M}_j \rangle|(1 - \eta\lambda - \eta\lambda\|w_i^{(t)}\|_2) + \Omega\left(\frac{\eta}{d}\right)$$

$$\geq |\langle w_i^{(t)}, \mathbf{M}_j \rangle|(1 - 2\eta\lambda) + \Omega\left(\frac{\eta}{d}\right) \geq |\langle w_i^{(t)}, \mathbf{M}_j \rangle|$$

Putting both cases together we have $i \in \mathcal{S}_{j,sure+}^{(t+1)}$ so the second half of (C.25) holds at $t+1$.

□

## D  Clean Accuracy Convergence Analysis

In this section we show the upper bound on how the clean training of a two-layer neural network can learn the labeling function from $N$ training samples $\{x_i, y_i\}_{i=1}^N$ up to small generalization error.

**Theorem D.1.** *Suppose the high-probability initialization event in Lemma B.2 holds, and suppose $\eta, \sigma_0 \in (0, \frac{1}{\mathsf{poly}(d)})$ and $N \geq \mathsf{poly}(d)$. With probability at least $1 - e^{-\Omega(\log^2 d)}$, for any $T \geq T_{\mathsf{c}} \stackrel{\text{def}}{=} \Omega(\frac{d\Xi_2^6}{\eta})$ and $T \leq d^{O(\log d)}/\eta$, if we run the Algorithm 1 for $T_{\mathsf{f}} = T_{\mathsf{e}} + T$ iterations, we have*

$$\frac{1}{T} \sum_{t=T_{\mathsf{e}}}^{T_{\mathsf{e}}+T-1} \mathop{\mathbb{E}}_{x,y,\rho} \mathbf{Obj}_t(w^{(t)}; x, y, \rho) \leq o(1)$$

*In other words, at least 99% of the iterations $t = T_{\mathsf{e}}, T_{\mathsf{e}}+1, \ldots, T_{\mathsf{e}}+T-1$ will have population risk $o(1)$ and clean population accuracy $\geq 1 - o(1)$.*

*Remark* D.2. With additional efforts, one can also prove that Theorem D.1 holds with high probability $1 - e^{-\Omega(\log^2 d)}$ for all $T$ in the range $T \in [T_{\mathsf{c}}, T_{\mathsf{f}}]$. We do not prove it here since it completes the notations and is not beyond the scope of this paper.

### D.1 Proof of Theorem D.1: Convergence Theorem

Our convergence analysis will rely on the following (what we call) *coupling* function which is the first-order approximation of the neural network.

**Definition D.3** (coupling). *At every iteration $t$, we define a linear function in $\mu$*

$$g_t(\mu; x, \rho) \stackrel{\text{def}}{=} \sum_{i=1}^{m} \left( \mathbb{1}_{\langle w_i^{(t)}, x \rangle + \rho_i \geq b^{(t)}} \cdot (\langle \mu_i, x \rangle + \rho_i - b^{(t)}) - \mathbb{1}_{-\langle w_i^{(t)}, x \rangle + \rho_i \geq b^{(t)}} \cdot (-\langle \mu_i, x \rangle + \rho_i - b^{(t)}) \right)$$

*and it equals the output of the real network at point $\mu = w^{(t)}$ both on zero and first order:*

$$g_t(w^{(t)}; x, \rho) = f_t(w^{(t)}; x, \rho) \quad \text{and} \quad \nabla_\mu g_t(\mu; x, \rho)\big|_{\mu=w^{(t)}} = \nabla_w f_t(w; x, \rho)\big|_{w=w^{(t)}}$$

In the analysis, we shall also identify a special choice $\mu^\star$ defined as follows.

**Definition D.4.** *Recall $\mathcal{S}_{1,\mathsf{sure}}^{(0)}, \ldots, \mathcal{S}_{d,\mathsf{sure}}^{(0)} \subseteq [m]$ are disjoint, so we construct $\mu_1^\star, \ldots, \mu_m^\star$ by*

$$\mu_i^\star \stackrel{\text{def}}{=} \begin{cases} \alpha \left( \frac{w_j^\star}{|\mathcal{S}_{j,\mathsf{sure}}^{(0)}|} \right) \mathbf{M}_j, & i \in \mathcal{S}_{j,\mathsf{sure}}^{(0)} \text{ for some } j \in [d]; \\ \vec{0}, & \text{otherwise.} \end{cases}$$

Above, $\alpha = o(1)$ is a parameter to be chosen later. One can easily check (using Lemma B.2) that

**Claim D.5.** $\sum_{i \in [m]} \|\mu_i^\star\|^2 \leq O(\frac{\alpha^2}{\Xi_1} d)$ *and* $\sum_{i \in [m]} \|\mu_i^\star\|^3 \leq O(\frac{\alpha^3}{\Xi_1^2} d)$

More interestingly, our so-constructed $\mu^\star$ satisfies (to be proved in Section D.2)

**Lemma D.6.** *Suppose $\mathcal{S}_{\mathsf{ept}+}^{(t)} = \mathcal{S}_{\mathsf{ept}++}^{(t)} = [m]$, $\mathcal{S}_{j,\mathsf{pot}}^{(0)} \supseteq \mathcal{S}_{j,\mathsf{pot}+}^{(t)}$ and $\mathcal{S}_{j,\mathsf{sure}}^{(0)} \subseteq \mathcal{S}_{j,\mathsf{sure}+}^{(t)}$ for every $j \in [d]$. Then,*

*(a) with probability at least $1 - e^{-\Omega(\log^2 d)}$ over $x, \rho$, $g_t(\mu^\star; x, \rho) = \alpha \langle w^\star, z \rangle \pm O(\frac{1}{\Xi_2^2})$*

*(b) $\mathbb{E}_{x, y=y(x), \rho} \left[ \log \left( 1 + e^{-y \cdot g_t(\mu^\star; x, \rho)} \right) \right] \leq O\left( \frac{1}{\alpha^2} + \frac{1}{\Xi_2^2} \right)$*

We are now ready to prove Theorem D.1. Since $w_i^{(t+1)} = w_i^{(t)} - \eta \nabla_{w_i} \widetilde{\mathbf{Obj}}_t(w^{(t)})$, we have the identity

$$\eta \langle \nabla \widetilde{\mathbf{Obj}}_t(w^{(t)}), w^{(t)} - \mu^\star \rangle = \frac{\eta^2}{2} \|\nabla \widetilde{\mathbf{Obj}}_t(w^{(t)})\|_F^2 + \frac{1}{2} \|w^{(t)} - \mu^\star\|_F^2 - \frac{1}{2} \|w^{(t+1)} - \mu^\star\|_F^2$$

Applying Lemma A.2, we know that by letting $\mathbf{Obj}_t(w^{(t)}) = \mathbb{E}_{x,y,\rho} \mathbf{Obj}_t(w^{(t)}; x, y, \rho)$, it satisfies

$$\eta \langle \nabla \mathbf{Obj}_t(w^{(t)}), w^{(t)} - \mu^\star \rangle \leq \eta^2 \cdot \mathsf{poly}(d) + \frac{1}{2} \|w^{(t)} - \mu^\star\|_F^2 - \frac{1}{2} \|w^{(t+1)} - \mu^\star\|_F^2 + \frac{\eta}{\mathsf{poly}(d)}$$

Let us define a pseudo objective

$$\mathbf{Obj}'_t(\mu) \stackrel{\text{def}}{=} \mathop{\mathbb{E}}_{x, y=y(x), \rho} \left[ \log(1 + e^{-y \cdot g_t(\mu; x, \rho)}) \right] + \lambda \sum_{i \in [m]} \mathbf{Reg}(\mu_i) \enspace ,$$

which is a convex function in $\mu$ because $g_t(\mu; x, \rho)$ is linear in $\mu$. We have, for every $t \geq T_\mathsf{e}$,

$$\langle \nabla \mathbf{Obj}_t(w^{(t)}), w^{(t)} - \mu^\star \rangle \stackrel{①}{=} \langle \nabla \mathbf{Obj}'_t(w^{(t)}), w^{(t)} - \mu^\star \rangle$$
$$\geq \mathbf{Obj}'_t(w^{(t)}) - \mathbf{Obj}'_t(\mu^\star) = \mathbf{Obj}_t(w^{(t)}) - \mathbf{Obj}'_t(\mu^\star)$$
$$\stackrel{②}{\geq} \mathbf{Obj}_t(w^{(t)}) - \lambda \sum_{i \in [m]} \left( \frac{\|\mu_i^\star\|^3}{3} + \frac{\|\mu_i^\star\|^2}{2} \right) - O\left( \frac{1}{\alpha^2} + \frac{1}{\Xi_2^2} \right)$$
$$\stackrel{③}{\geq} \mathbf{Obj}_t(w^{(t)}) - O\left( \frac{\alpha^3 \log d}{\Xi_1^2} + \frac{\alpha^2 \log d}{\Xi_1} + \frac{1}{\alpha^2} + \frac{1}{\Xi_2^2} \right)$$
$$\geq \mathbf{Obj}_t(w^{(t)}) - O\left( \frac{\sqrt{\log d}}{\sqrt{\Xi_1}} \right)$$

Above, ① uses the definition of $g_t$, ② uses Lemma D.6b (and Theorem C.2 for the prerequisite for Lemma D.6b), and ③ uses Claim D.5 for the bound on $\|\mu_i^\star\|^2$ and $\|\mu_i^\star\|^3$.

Putting these together, we have

$$\eta \left( \mathbf{Obj}_t(w^{(t)}) - O\left( \frac{\sqrt{\log d}}{\sqrt{\Xi_1}} \right) \right) \leq \eta^2 \cdot \mathsf{poly}(d) + \frac{1}{2} \|w^{(t)} - \mu^\star\|_F^2 - \frac{1}{2} \|w^{(t+1)} - \mu^\star\|_F^2 + \frac{\eta}{\mathsf{poly}(d)}$$

Therefore, after telescoping for $t = T_\mathsf{e}, T_\mathsf{e} + 1, \ldots, T_\mathsf{e} + T - 1$, and using $\eta \leq \frac{1}{\mathsf{poly}(d)}$, we have

$$\frac{1}{T} \sum_{t=T_\mathsf{e}}^{T_\mathsf{e}+T-1} \left( \mathbf{Obj}_t(w^{(t)}) - O\left( \frac{\sqrt{\log d}}{\sqrt{\Xi_1}} \right) \right) \leq \frac{O(\|w^{(T_\mathsf{e})} - \mu^\star\|_F^2)}{\eta T} \leq \frac{O(\Xi_2^4 m)}{\eta T}$$

Finally, we calculate

$$\|w^{(T_\mathsf{e})}\|_F^2 \leq \sum_{i \in \bigcup_j \mathcal{S}_{j,pot}^{(0)}} \|w_i^{(T_\mathsf{e})}\|_2^2 + \sum_{i \notin \bigcup_j \mathcal{S}_{j,pot}^{(0)}} \|w_i^{(T_\mathsf{e})}\|_2^2$$
$$\leq d\Xi_2 \cdot O(\Xi_2^4) + m \cdot O(\frac{k^2}{d^2} \Xi_2^4) \leq O(d\Xi_2^5)$$

and this finishes the proof. ∎

### D.2 Proof of Claim D.6: Main Coupling

The proof of Lemma D.6a comes from Claim D.7 and Claim D.8 below. In the two claims, we split $g_t(\mu^\star; x) = g_{t,1} + g_{t,4}$ into two terms, and bound them separately. Define

$$g_{t,1}(\mu^\star; x, \rho) = \sum_{j \in [d]} \frac{\alpha \cdot w_j^\star}{|\mathcal{S}_{j,sure}^{(0)}|} \sum_{i \in \mathcal{S}_{j,sure}^{(0)}} \left( \mathbb{1}_{\langle w_i^{(t)}, x \rangle + \rho_i \geq b^{(t)}} + \mathbb{1}_{-\langle w_i^{(t)}, x \rangle + \rho_i \geq b^{(t)}} \right) \cdot z_j$$

$$g_{t,4}(x, \rho) = \sum_{i \in [m]} \left( (\rho_i - b^{(t)}) \mathbb{1}_{\langle w_i^{(t)}, x \rangle + \rho_i \geq b^{(t)}} + (b^{(t)} - \rho_i) \mathbb{1}_{-\langle w_i^{(t)}, x \rangle + \rho_i \geq b^{(t)}} \right)$$

**Claim D.7.** $\mathbf{Pr}_{x,\rho}[g_{t,1}(\mu^\star; x, \rho) = \alpha \langle w^\star, z \rangle] \geq 1 - e^{-\Omega(\log^2 d)}$

*Proof of Claim D.7.* Recall for each $i \in \mathcal{S}^{(0)}_{j,sure}$,

- it satisfies $i \in \mathcal{S}^{(t)}_{j,sure+}$ so $|\langle w_i^{(t)}, \mathbf{M}_j \rangle| \geq 2\sqrt{k}b^{(t)}$;
- it also implies $i \notin \mathcal{S}^{(t)}_{j',pot+}$ for any $j' \neq j$, so $|\langle w_i^{(t)}, \mathbf{M}_{j'} \rangle| \leq \frac{k}{d\bar{\beta}} b^{(t)}$;
- recall $\rho_i \sim \mathcal{N}(0, (\sigma_\rho^{(t)})^2)$ for $\sigma_\rho^{(t)} = \Theta(b^{(t)} \cdot \frac{(\log\log\log d)^3}{\log d})$.
- recall $\langle w_i^{(t)}, \xi \rangle$ is a variable with variance at most $O(\frac{\|w_i^{(t)}\|^2 \sigma_x^2}{d})$ for $\sigma_x = O(1)$.

Applying Lemma C.19, we know with probability at least $1 - e^{-\Omega(\log^2 d)}$ it satisfies

$$\left| \langle w_i^{(t)}, \sum_{j' \neq j} \mathbf{M}_{j'} z_{j'} + \xi \rangle \right| + |\rho_i| \leq \frac{b^{(t)}}{2} = \frac{\beta \Xi_2^2}{2} \tag{D.1}$$

and when this happens it satisfies, whenever $z_j \neq 0$,

$$\mathbb{1}_{\langle w_i^{(t)}, x \rangle + \rho_i \geq b^{(t)}} + \mathbb{1}_{-\langle w_i^{(t)}, x \rangle + \rho_i \geq b^{(t)}} = 1$$

Summing up over all $i \in \mathcal{S}^{(0)}_{j,sure}$ and $j \in [d]$, we have with probability at least $1 - e^{-\Omega(\log^2 d)}$ over $x, \rho$: $g_{t,1}(\mu^\star; x) = \alpha \langle w^\star, z \rangle$. □

**Claim D.8.** $\mathbf{Pr}_{x,\rho}[|g_{t,4}(x, \rho)| \leq O(\frac{1}{\Xi_2^2})] \geq 1 - e^{-\Omega(\log^2 d)}$

*Proof of Claim D.8.* Let us write $\xi = \sum_{j \in [d]} \mathbf{M}_j \xi_j$ where each $\xi_j$ is i.i.d. Let us write

$$g_{t,4}(x, \rho) = \sum_{i \in [m]} g_{t,4,i}(x, \rho_i)$$

$$\text{for} \quad g_{t,4,i}(x, \rho_i) \stackrel{\text{def}}{=} \left( (\rho_i - b^{(t)}) \mathbb{1}_{\langle w_i^{(t)}, x \rangle + \rho_i \geq b^{(t)}} + (b^{(t)} - \rho_i) \mathbb{1}_{-\langle w_i^{(t)}, x \rangle + \rho_i \geq b^{(t)}} \right)$$

We note that $g_{t,4}(x, \rho)$ is a random variable that depends on independent variables

$$\xi_1, \ldots, \xi_d, z_1, \ldots, z_d, \rho_1, \ldots, \rho_m \ ,$$

so we also want to write it as $g_{t,4}(z, \xi, \rho)$ and $g_{t,4,i}(z, \xi, \rho_i)$.

We can without loss of generality assume as if $|\rho_i| \leq \frac{b^{(t)}}{10}$ and $|\xi_j| \leq \frac{b^{(t)}}{\Xi_2^{10}} := B$ always hold, both of which happen with probability at least $1 - e^{-\Omega(\log^2 d)}$. In the rest of the proof we condition on this happens. By symmetry we have

$$\forall \rho: \mathop{\mathbb{E}}_{\xi, z}[g_{t,4}(z, \xi, \rho)] = 0 \ .$$

We wish to apply a high-probability version of the McDiarmid's inequality (see Lemma H.3) to bound $g_{t,4}$. In order to do so, we need to check the sensitivity of $g_{t,4}(x, \rho)$ regarding every random variable.

- For every $z_j$, suppose we perturb it to an arbitrary $z'_j \in [-1, 1]$. We also write $z' = (z_{-j}, z'_j)$ and $x' = \mathbf{M}z' + \xi$.

  - Now, for every $i \in \mathcal{S}^{(t)}_{j,pot+}$, we have the naive bound
  
    $$|g_{t,4,i}(z, \xi, \rho_i) - g_{t,4,i}(z', \xi, \rho_i)| \leq 2b^{(t)} \cdot \mathbb{1}_{z_j \neq z'_j}$$
    
    and there are at most $|\mathcal{S}^{(t)}_{j,pot+}| \leq \Xi_2$ such neurons $i$.

- For every $i \notin \mathcal{S}_{j,pot+}^{(t)}$, we have $|\langle w_i^{(t)}, \mathbf{M}_j \rangle| \leq \frac{k}{d\beta} b^{(t)}$. Define event

$$\mathcal{E}_i = \left\{ |\langle w_i^{(t)}, \sum_{j' \neq j} \mathbf{M}_{j'} z_{j'} + \xi \rangle| \geq \frac{b^{(t)}}{2} \right\}$$

  * When event $\mathcal{E}_i$ does not happen, we have $\mathbb{1}_{\langle w_i^{(t)}, x \rangle + \rho_i \geq b^{(t)}} = \mathbb{1}_{\langle w_i^{(t)}, x' \rangle + \rho_i \geq b^{(t)}}$, and thus

  $$g_{t,4,i}(z, \xi, \rho_i) = g_{t,4,i}(z', \xi, \rho_i) \ .$$

  * When $\mathcal{E}_i$ happens, using the randomness of $\rho_i$, we have

  $$\Pr_{\rho_i} \left[ \mathbb{1}_{\langle w_i^{(t)}, x \rangle + \rho_i \geq b^{(t)}} \neq \mathbb{1}_{\langle w_i^{(t)}, x' \rangle + \rho_i \geq b^{(t)}} \right] \leq O\left( \frac{|\langle w_i^{(t)}, \mathbf{M}_j \rangle| \cdot |z_j - z_j'|}{\sigma_\rho^{(t)}} \right)$$

  $$\leq O\left( \frac{k \log d}{d\beta} \right) \cdot |z_j - z_j'|$$

  and thus

  $$|g_{t,4,i}(z, \xi, \rho_i) - g_{t,4,i}(z', \xi, \rho_i)| = \begin{cases} 0, & \text{w.p. } \geq 1 - O\left( \frac{k \log d}{d\beta} |z_j - z_j'| \right) \text{ over } \rho_i; \\ O(b^{(t)}), & \text{otherwise.} \end{cases}$$

  Note with probability at least $1 - e^{-\Omega(\log^2 d)}$, the number of $i \in [m]$ with $\mathcal{E}_i$ holds is at most $O(k\Xi_2)$ (using Lemma C.19). Therefore, by applying Chernoff bound, we know

  $$|g_{t,4}(z, \xi, \rho) - g_{t,4}(z', \xi, \rho)| \leq O(b^{(t)}) \cdot \left( \frac{k \log d}{d\beta} |z_j - z_j'| \cdot k\Xi_2 + \Xi_2 \right)$$

  This means two things that both hold with probability at least $1 - e^{-\Omega(\log^2 d)}$ over $z_{-j}, \xi, \rho$:

  - For all $z_j, z_j'$, $|g_{t,4}(x, \rho) - g_{t,4}(x', \rho)| \leq O(b^{(t)}) \cdot \left( \frac{k \log d}{d\beta} \cdot k\Xi_2 + \Xi_2 \right) \leq O(\sqrt{k} b^{(t)}) < o(\frac{1}{\Xi_2^2})$
  - $\mathbb{E}_{z_j, z_j'} |g_{t,4}(x, \rho) - g_{t,4}(x', \rho)|^2 \leq O((b^{(t)})^2) \cdot \mathbb{E}_{z_j, z_j'} \left( \left( \frac{k \log d}{d\beta} \cdot k\Xi_2 \right)^2 |z_j - z_j'|^2 + \Xi_2^2 \mathbb{1}_{z_j \neq z_j'} \right) \leq$
    $O\left( \left( \frac{k^4 \Xi_2^2 \log^2 d}{d^2 \beta^2} \frac{1}{d} + \Xi_2^2 \frac{k}{d} \right) (b^{(t)})^2 \right) < o(\frac{1}{d\Xi_2^4})$

- For every $\xi_j$, suppose we perturb it to $\xi_j' \in [-B, B]$. We write $\xi' = \xi + \mathbf{M}_j(\xi_j' - \xi_j)$ and $x' = \mathbf{M}z + \xi'$.

  - Now, for every $i \in \mathcal{S}_{j,pot+}^{(t)}$, with probability at least $1 - e^{-\Omega(\log^2 d)}$ we have $|\langle w_i^{(t)}, \sum_{j' \neq j} \mathbf{M}_{j'} \xi_{j'} \rangle| \leq \frac{b^{(t)}}{10}$. Therefore, if it also happens that $|\langle w_i^{(t)}, \mathbf{M}z \rangle| \leq \frac{b^{(t)}}{10}$, then $g_{t,4,i}(z, \xi, \rho_i) = g_{t,4,i}(z, \xi', \rho_i)$. In other words, we have

  $$|g_{t,4,i}(z, \xi, \rho_i) - g_{t,4,i}(z', \xi, \rho_i)| \leq 2b^{(t)} \cdot \mathbb{1}_{|\langle w_i^{(t)}, \mathbf{M}z \rangle| \geq \frac{b^{(t)}}{10}} \ .$$

  Summing up over $i \in \mathcal{S}_{j,pot+}^{(t)}$, and taking expectation in $z$, we have

  $$\left| \mathbb{E}_z \sum_{i \in \mathcal{S}_{j,pot+}^{(t)}} g_{t,4,i}(z, \xi, \rho_i) - g_{t,4,i}(z', \xi, \rho_i) \right| \leq 2b^{(t)} \cdot \mathbb{E}_z \left[ \sum_{i \in \mathcal{S}_{j,pot+}^{(t)}} \mathbb{1}_{|\langle w_i^{(t)}, \mathbf{M}z \rangle| \geq \frac{b^{(t)}}{10}} \right] \leq O\left( b^{(t)} \frac{k \Xi_2}{d} \right)$$

  where the last inequality uses a variant of Lemma C.7 and $|\mathcal{S}_{j,pot+}^{(t)}| \leq \Xi_2$.

- For every $i \notin \mathcal{S}_{j,pot+}^{(t)}$, we have $|\langle w_i^{(t)}, \mathbf{M}_j \rangle| \leq \frac{k}{d\beta} b^{(t)}$. Define event

$$\mathcal{E}_i = \left\{ |\langle w_i^{(t)}, \mathbf{M}z + \sum_{j' \neq j} \mathbf{M}_{j'} \xi_{j'} \rangle| \geq \frac{b^{(t)}}{2} \right\}$$

  * When event $\mathcal{E}_i$ does not happen, we have $\mathbb{1}_{\langle w_i^{(t)}, x \rangle + \rho_i \geq b^{(t)}} = \mathbb{1}_{\langle w_i^{(t)}, x' \rangle + \rho_i \geq b^{(t)}}$, and thus
  $$g_{t,4,i}(z, \xi, \rho_i) = g_{t,4,i}(z', \xi, \rho_i) \ .$$

  * When $\mathcal{E}_i$ happens, using the randomness of $\rho_i$, we have
  $$\Pr_{\rho_i}\left[ \mathbb{1}_{\langle w_i^{(t)}, x \rangle + \rho_i \geq b^{(t)}} \neq \mathbb{1}_{\langle w_i^{(t)}, x' \rangle + \rho_i \geq b^{(t)}} \right] \leq O\left( \frac{|\langle w_i^{(t)}, \mathbf{M}_j \rangle| \cdot |\xi_j - \xi'_j|}{\sigma_\rho^{(t)}} \right)$$
  $$\leq O\left( \frac{k \log d}{d\beta} \right) \cdot |\xi_j - \xi'_j|$$

  and thus
  $$|g_{t,4,i}(z, \xi, \rho_i) - g_{t,4,i}(z, \xi', \rho_i)| = \begin{cases} 0, & \text{w.p. } \geq 1 - O\left(\frac{k \log d}{d\beta} |\xi_j - \xi'_j|\right) \text{ over } \rho_i; \\ O(b^{(t)}), & \text{otherwise.} \end{cases}$$

  Note with probability at least $1 - e^{-\Omega(\log^2 d)}$, the number of $i \in [m]$ with $\mathcal{E}_i$ holds is at most $O(k\Xi_2)$ (using a minor variant of Lemma C.19). Therefore, by applying Chernoff bound, we know with probability at least $1 - e^{-\Omega(\log^2 d)}$ over $z, \xi_{-j}, \rho$

  $$|\sum_{i \notin \mathcal{S}_{j,pot+}^{(t)}} g_{t,4,i}(z, \xi, \rho_i) - g_{t,4,i}(z, \xi', \rho_i)| \leq O(b^{(t)}) \cdot \left( \frac{k \log d}{d\beta} |\xi_j - \xi'_j| \cdot k\Xi_2 \right)$$

  Taking expectation over $z$, we have with probability at least $1 - e^{-\Omega(\log^2 d)}$ over $\xi_{-j}, \rho$:

  $$\left| \mathbb{E}_z \sum_{i \notin \mathcal{S}_{j,pot+}^{(t)}} g_{t,4,i}(z, \xi, \rho_i) - g_{t,4,i}(z, \xi', \rho_i) \right| \leq O(b^{(t)}) \cdot \left( \frac{\sqrt{k} \log d}{\sqrt{d}} |\xi_j - \xi'_j| \cdot k\Xi_2 \right)$$

Putting the two cases together, we have with probability at least $1 - e^{-\Omega(\log^2 d)}$ over $\xi_{-j}, \rho$:

$$\left| \mathbb{E}_z g_{t,4}(z, \xi, \rho) - \mathbb{E}_z g_{t,4}(z, \xi', \rho) \right| \leq O(b^{(t)}) \cdot O\left( \frac{k \log d}{d\beta} |\xi_j - \xi'_j| \cdot k\Xi_2 + \frac{k\Xi_2}{d} \right)$$

This means two things that both hold with probability at least $1 - e^{-\Omega(\log^2 d)}$ over $\xi_{-j}, \rho$:

- For all $\xi_j, \xi'_j$, $|\mathbb{E}_z g_{t,4}(z, \xi, \rho) - \mathbb{E}_z g_{t,4}(z, \xi', \rho)| \leq O(b^{(t)}) \cdot \left( \frac{k \log d}{d\beta} B \cdot k\Xi_2 + \frac{k\Xi_2}{d} \right) \ll o(\frac{1}{\Xi_2^2})$
- $\mathbb{E}_{\xi_j, \xi'_j} |\mathbb{E}_z g_{t,4}(z, \xi, \rho) - \mathbb{E}_z g_{t,4}(z, \xi', \rho)|^2 \leq O((b^{(t)})^2) \cdot \mathbb{E}_{z_j, z'_j} \left( \left( \frac{k \log d}{d\beta} \cdot k\Xi_2 \right)^2 |\xi_j - \xi'_j|^2 + \frac{k^2}{d^2} \Xi_2^2 \right) \leq O\left( \left( \frac{k^4 \Xi_2^2 \log^2 d}{d^2 \beta^2} \frac{1}{d} + \frac{k^2}{d^2} \Xi_2^2 \right) (b^{(t)})^2 \right) \ll o(\frac{1}{d\Xi_2^4})$

We are now ready to apply the high-probability version of the McDiarmid's inequality (see Lemma H.3). We apply it twice. In the first time, we use the perturbation on $z$ to derive that, with probability at least $1 - e^{-\Omega(\log^2 d)}$ over $z, \xi, \rho$:

$$|g_{t,4}(z, \xi, \rho) - \mathbb{E}_z g_{t,4}(z, \xi, \rho)| \leq O(\frac{1}{\Xi_2^2})$$

In the second time, we use the perturbation on $\xi$ to derive that, with probability at least $1 - e^{-\Omega(\log^2 d)}$ over $\xi, \rho$,

$$|\mathbb{E}_z g_{t,4}(z,\xi,\rho) - \mathbb{E}_{z,\xi} g_{t,4}(z,\xi,\rho)| \leq O(\frac{1}{\Xi_2^2})$$

Finally, noticing that $\mathbb{E}_{z,\xi} g_{t,4}(z,\xi,\rho) = 0$ for every $\rho$, we finish the proof. □

This finishes the proof of Lemma D.6a. We are only left to prove Lemma D.6b.

By Lipscthiz continuity of the $\log(1 + e^{-x})$ function, we know with probability at least $1 - e^{-\Omega(\log^2 d)}$,

$$\log\left(1 + e^{-y(x) \cdot g_t(\mu^\star; x)}\right) = \log\left(1 + e^{-y(x) \cdot \alpha \langle w^\star, z \rangle}\right) \pm O(\frac{1}{\Xi_2^2}) = \log\left(1 + e^{-\alpha|\langle w^\star, z \rangle|}\right) \pm O(\frac{1}{\Xi_2^2})$$

Taking expectation (and using the exponential tail) we have

$$\mathbb{E}\left[\log\left(1 + e^{-y(x) \cdot g_t(\mu^\star; x)}\right)\right] = \mathbb{E}\left[\log\left(1 + e^{-\alpha|\langle w^\star, z \rangle|}\right)\right] \pm O(\frac{1}{\Xi_2^2})$$

Note if we take expectation over $z$, we have

$$\mathbb{E}_z[\log\left(1 + e^{-\alpha|\langle w^\star, z \rangle|}\right)] \leq \int_{t \geq 0} \log\left(1 + e^{-\alpha t}\right) \cdot \mathbf{Pr}[|\langle w^\star, z \rangle| \leq t] dt$$

$$\overset{\textcircled{1}}{\leq} O(1) \cdot \int_{t \geq 0} e^{-\alpha t} \cdot \left(t + \frac{1}{\sqrt{k}}\right) \leq O(\frac{1}{\alpha^2} + \frac{1}{\sqrt{k}})$$

where ① uses Lemma H.1a. This finishes the proof of Lemma D.6b. ■

## E  Why Clean Training is Non-Robust

In this section we shall show that clean training will not achieve robustness against $\ell_2$ perturbation of size $\tau = \Omega\left(\frac{d^{0.4999}}{k^2}\right)$ as long as $k = \widetilde{\Omega}(d^{0.3334})$. Recall $\|\mathbf{M}\|_1 = \sum_{j \in [d]} \|\mathbf{M}_j\|_\infty$.

**Theorem E.1** (clean training is non-robust)**.** *Suppose the high-probability initialization event in Lemma B.2 holds. Suppose $k > d^{(1-c_0)/3}$ and consider any iteration $t \geq \Omega(\frac{1}{\eta \lambda \Xi_2^2})$ and $t \leq \min\{d^{\log d}/\eta, T_f\}$. With probability at least $1 - e^{-\Omega(\log^2 d)}$ the following holds. If we perturb every input $x$ by $\delta = -y\Xi_2^{10}(\mathbf{M}w^\star)/k^2$, then the accuracy drops below $e^{-\Omega(\log^2 d)}$:*

$$\Pr_{x,y=y(x),\rho}\left[\mathsf{sign}(f_t(x-\delta)) = y\right] \leq e^{-\Omega(\log^2 d)} \ ,$$

$$\Pr_{x,y=y(x)}\left[\mathsf{sign}(\mathbb{E}_\rho[f_t(x-\delta)]) = y\right] \leq e^{-\Omega(\log^2 d)} \ .$$

*Note that*

$$\|\delta\|_2 \leq \frac{\Xi_2^{10} \sqrt{d}}{k^2} \quad and \quad \|\delta\|_\infty \leq \frac{\Xi_2^{10} \|\mathbf{M}\|_1}{k^2}.$$

The proof of Theorem E.1 relies on the following main lemma (to be proved in Section E.1). It says that towards the end of clean training, neurons $w_i^{(t)}$ have a (small) common direction in $\mathbf{M}w^\star$.

**Lemma E.2** (non-robust)**.** *For any iteration $t \geq \Omega(\frac{1}{\eta \lambda \Xi_2^2})$, let subset $\mathcal{S} \overset{\text{def}}{=} \cup_{j \in [d]} \mathcal{S}_{j,sure}^{(0)}$, then*

$$\sum_{i \in \mathcal{S}} \langle w_i^{(t)}, \mathbf{M}w^\star \rangle = \Omega\left(\frac{kd}{\Xi_2^7}\right) \quad and \quad \forall i \in [m] \colon \langle w_i^{(t)}, \mathbf{M}w^\star \rangle \geq -O\left(\frac{1}{\lambda} \frac{k^3 \Xi_2^4}{d^2} \sigma_x^2\right)$$

With the help of Lemma E.2, one can calculate that by perturbing input in this direction $-y \cdot \mathbf{M} w^\star$, the output label of the network can change dramatically. This is the proof of Theorem E.1 and details can be found in Section E.2.

### E.1 Proof of Lemma E.2: Common Direction Among Neurons

Before proving Lemma E.2, let us first present Claim E.3.

**Claim E.3.** *We have*
$$\langle w_i^{(t+1)}, \mathbf{M} w^\star \rangle \geq \langle w_i^{(t)}, \mathbf{M} w^\star \rangle \cdot (1 - \eta\lambda - \eta\lambda\|w_i^{(t)}\|) - O\left(\eta \frac{k^2}{d^{1.5}} \sigma_x^2 + \frac{\eta}{\mathsf{poly}(d)}\right)$$
$$+ \eta \mathop{\mathbb{E}}_{x,\rho} \left[ \ell_t'(w^{(t)}; x, y, \rho) \mathbb{1}_{\langle w_i^{(t)}, x \rangle + \rho_i \geq b^{(t)}} |\langle w^\star, z \rangle| \right]$$

*Proof of Claim E.3.* Let us recall from (C.19) that
$$\langle w_i^{(t+1)}, \mathbf{M}_j \rangle = \langle w_i^{(t)}, \mathbf{M}_j \rangle \cdot (1 - \eta\lambda - \eta\lambda\|w_i^{(t)}\|) \pm \frac{\eta}{\mathsf{poly}(d)}$$
$$+ \eta \mathop{\mathbb{E}}_{x, y=y(x), \rho} \left[ y \ell_t'(w^{(t)}; x, y, \rho) \big( \mathbb{1}_{\langle w_i^{(t)}, x \rangle + \rho_i \geq b^{(t)}} + \mathbb{1}_{-\langle w_i^{(t)}, x \rangle + \rho_i \geq b^{(t)}} \big) \big( z_j + \langle \xi, \mathbf{M}_j \rangle \big) \right]$$

and therefore
$$\langle w_i^{(t+1)}, \sum_{j \in [d]} w_j^\star \mathbf{M}_j \rangle \geq \langle w_i^{(t)}, \sum_{j \in [d]} w_j^\star \mathbf{M}_j \rangle \cdot (1 - \eta\lambda - \eta\lambda\|w_i^{(t)}\|) - \frac{\eta}{\mathsf{poly}(d)}$$
$$+ \eta \mathop{\mathbb{E}}_{x,\rho} \left[ y \ell_t'(w^{(t)}; x, y, \rho) \big( \mathbb{1}_{\langle w_i^{(t)}, x \rangle + \rho_i \geq b^{(t)}} + \mathbb{1}_{-\langle w_i^{(t)}, x \rangle + \rho_i \geq b^{(t)}} \big) \langle w^\star, z \rangle \right]$$
$$- O(\eta) \cdot \sum_{j \in [d]} \left| \mathop{\mathbb{E}}_{x,\rho} \left[ y \ell_t'(w^{(t)}; x, y, \rho) \big( \mathbb{1}_{\langle w_i^{(t)}, x \rangle + \rho_i \geq b^{(t)}} + \mathbb{1}_{-\langle w_i^{(t)}, x \rangle + \rho_i \geq b^{(t)}} \big) \langle \xi, \mathbf{M}_j \rangle \right] \right|$$

Applying Lemma C.17b and using $y \langle w^*, z \rangle = |\langle w^*, z \rangle|$ and $\|w_i^{(t)}\| \leq O(\Xi_2^2)$ (see Lemma C.21), we have
$$\langle w_i^{(t+1)}, \sum_{j \in [d]} w_j^\star \mathbf{M}_j \rangle \geq \langle w_i^{(t)}, \sum_{j \in [d]} w_j^\star \mathbf{M}_j \rangle \cdot (1 - \eta\lambda - \eta\lambda\|w_i^{(t)}\|) - O\left(\eta \frac{k^3 \Xi_2^4}{d^2} \sigma_x^2 + \frac{\eta}{\mathsf{poly}(d)}\right)$$
$$+ \eta \mathop{\mathbb{E}}_{x,\rho} \left[ \ell_t'(w^{(t)}; x, y, \rho) \big( \mathbb{1}_{\langle w_i^{(t)}, x \rangle + \rho_i \geq b^{(t)}} + \mathbb{1}_{-\langle w_i^{(t)}, x \rangle + \rho_i \geq b^{(t)}} \big) |\langle w^\star, z \rangle| \right]$$
$$\geq \langle w_i^{(t)}, \sum_{j \in [d]} w_j^\star \mathbf{M}_j \rangle \cdot (1 - \eta\lambda - \eta\lambda\|w_i^{(t)}\|) - O\left(\eta \frac{k^3 \Xi_2^4}{d^2} \sigma_x^2 + \frac{\eta}{\mathsf{poly}(d)}\right)$$
$$+ \eta \mathop{\mathbb{E}}_{x,\rho} \left[ \ell_t'(w^{(t)}; x, y, \rho) \mathbb{1}_{\langle w_i^{(t)}, x \rangle + \rho_i \geq b^{(t)}} |\langle w^\star, z \rangle| \right] \quad .$$
$\square$

*Proof of Lemma E.2.* Recall Claim E.3 says that
$$\sum_{i \in \mathcal{S}} \langle w_i^{(t+1)}, \mathbf{M} w^\star \rangle \geq \sum_{i \in \mathcal{S}} \langle w_i^{(t)}, \mathbf{M} w^\star \rangle \cdot (1 - \eta\lambda - \eta\lambda\|w_i^{(t)}\|) - O\left(\eta |\mathcal{S}| \frac{k^3 \Xi_2^4}{d^2} \sigma_x^2 \right)$$
$$+ \eta \mathop{\mathbb{E}}_{x,\rho} \left[ \ell_t'(w^{(t)}; x, y, \rho) |\langle w^\star, z \rangle| \sum_{i \in \mathcal{S}} \mathbb{1}_{\langle w_i^{(t)}, x \rangle + \rho_i \geq b^{(t)}} \right]$$

Using $\mathcal{S}_{j,sure}^{(0)} \subseteq \mathcal{S}_{j,sure+}^{(t)}$, and a similar analysis to Lemma C.19, we know with probability at least $1 - e^{-\Omega(\log^2 d)}$ it satisfies $\sum_{i \in \mathcal{S}} \mathbb{1}_{\langle w_i^{(s)}, x \rangle + \rho_i \geq b^{(s)}} \geq \Omega(k)$. Therefore, the above inequality gives

$$\sum_{i \in \mathcal{S}} \langle w_i^{(t+1)}, \mathbf{M} w^\star \rangle \geq \sum_{i \in \mathcal{S}} \langle w_i^{(t)}, \mathbf{M} w^\star \rangle \cdot (1 - \eta\lambda - \eta\lambda \|w_i^{(t)}\|) - O\left(\eta |\mathcal{S}| \frac{k^3 \Xi_2^4}{d^2} \sigma_x^2\right)$$
$$+ \Omega(\eta k) \cdot \mathop{\mathbb{E}}_{x,\rho} \left[\ell_t'(w^{(t)}; x, y, \rho) | \langle w^\star, z \rangle |\right]$$

Now, using small ball probability Lemma H.1a we have

$$\mathbf{Pr}\left[|\langle w^*, z \rangle| \leq 0.01 \, \mathbb{E}[\ell_s'(w^{(t)}; x, y, \rho)]\right] \leq \frac{1}{2} \, \mathbb{E}[\ell_s'(w^{(t)}; x, y, \rho)] + O(\frac{1}{\sqrt{k}}) \enspace .$$

Therefore, let us abbreviate by writing $\ell_s' = \ell_s'(w^{(t)}; x, y, \rho)$, then

$$\mathbb{E}\left[\ell_s'(w^{(t)}; x, y, \rho) \cdot |\langle w^*, z \rangle|\right]$$
$$\geq \mathbb{E}\left[\ell_s' \cdot |\langle w^*, z \rangle| \, \Big| \, |\langle w^*, z \rangle| \geq 0.01 \, \mathbb{E}[\ell_s']\right] \cdot \mathbf{Pr}\left[|\langle w^*, z \rangle| \geq 0.01 \, \mathbb{E}[\ell_s']\right]$$
$$\geq 0.01 \, \mathbb{E}[\ell_s'] \cdot \mathbb{E}\left[\ell_s' \, \Big| \, |\langle w^*, z \rangle| \geq 0.01 \, \mathbb{E}[\ell_s']\right] \cdot \mathbf{Pr}\left[|\langle w^*, z \rangle| \geq 0.01 \, \mathbb{E}[\ell_s']\right]$$
$$= 0.01 \, \mathbb{E}[\ell_s'] \cdot \left(\mathbb{E}\left[\ell_s'\right] - \mathbb{E}\left[\ell_s' \, \Big| \, |\langle w^*, z \rangle| < 0.01 \, \mathbb{E}[\ell_s']\right] \cdot \mathbf{Pr}\left[|\langle w^*, z \rangle| < 0.01 \, \mathbb{E}[\ell_s']\right]\right)$$
$$\geq 0.01 \, \mathbb{E}[\ell_s'] \cdot \left(\mathbb{E}\left[\ell_s'\right] - \mathbf{Pr}\left[|\langle w^*, z \rangle| < 0.01 \, \mathbb{E}[\ell_s']\right]\right)$$
$$\geq 0.01 \, \mathbb{E}[\ell_s'] \cdot \left(\frac{1}{2} \, \mathbb{E}\left[\ell_s'\right] - O(\frac{1}{\sqrt{k}})\right) \overset{\text{①}}{\geq} \frac{1}{\Xi_2^5}$$

where the last inequality ① uses Lemma C.21 and Lemma C.22.

Therefore, using $|\mathcal{S}| \leq d\Xi_2$, we have

$$\sum_{i \in \mathcal{S}} \langle w_i^{(t+1)}, \mathbf{M} w^\star \rangle \geq \sum_{i \in \mathcal{S}} \langle w_i^{(t)}, \mathbf{M} w^\star \rangle \cdot (1 - \eta\lambda - \eta\lambda \|w_i^{(t)}\|) - O\left(\eta |\mathcal{S}| \frac{k^3 \Xi_2^4}{d^2} \sigma_x^2\right) + \Omega(\frac{\eta k}{\Xi_2^5})$$
$$\geq \sum_{i \in \mathcal{S}} \langle w_i^{(t)}, \mathbf{M} w^\star \rangle \cdot (1 - O(\eta\lambda \Xi_2^2)) + \Omega(\frac{\eta k}{\Xi_2^5})$$

so we conclude for every $t \geq \Omega(\frac{1}{\eta\lambda\Xi_2^2})$ it satisfies

$$\sum_{i \in \mathcal{S}} \langle w_i^{(t)}, \mathbf{M} w^\star \rangle \geq \Omega\left(\frac{kd}{\Xi_2^7}\right) \enspace .$$

As for any arbitrary $i \in [m]$, we have

$$\langle w_i^{(t+1)}, \mathbf{M} w^\star \rangle \geq \langle w_i^{(t)}, \mathbf{M} w^\star \rangle \cdot (1 - \eta\lambda - \eta\lambda \|w_i^{(t)}\|) - O\left(\eta \frac{k^3 \Xi_2^4}{d^2} \sigma_x^2\right)$$
$$\geq \cdots \geq -O\left(\frac{1}{\lambda} \frac{k^3 \Xi_2^4}{d^2} \sigma_x^2\right) \enspace . \qquad \square$$

### E.2 Proof of Theorem E.1

*Proof of Theorem E.1.* Recall $\mathcal{S} \overset{\text{def}}{=} \cup_{j \in [d]} \mathcal{S}_{j,sure}^{(0)}$ from Lemma E.2 and $\mathcal{S}_{j,sure}^{(0)} \subseteq \mathcal{S}_{j,sure+}^{(t)}$ from Theorem C.2. For every $i \in \mathcal{S}_{j,sure}^{(0)} \subseteq \mathcal{S}_{j,sure+}^{(t)}$, we know with probability at least $1 - e^{-\Omega(\log^2 d)}$, the same proof of (D.1) gives

$$\mathbb{1}_{\langle w_i^{(t)},x\rangle+\rho_i\geq 10b^{(t)}} = \mathbb{1}_{w_j^\star z_j>0} \quad \text{and} \quad \mathbb{1}_{-\langle w_i^{(t)},x\rangle+\rho_i\geq 10b^{(t)}} = \mathbb{1}_{w_j^\star z_j<0}$$

Therefore, setting $\delta = \delta_0 \mathbf{M} w^\star$ for some $\delta_0 \in (0, \frac{\beta}{\sqrt{d}})$, and using $\|w_i^{(t)}\|_2 \leq \Xi_2^2$ (since $\mathcal{S}_{ept++}^{(t)} = [m]$), we have $|\langle w_i^{(t)}, \delta\rangle| \leq \delta_0 \Xi_2^2 \sqrt{d} \leq b^{(t)}$. Using this, we can sum up over all $i \in \mathcal{S}$:

$$\sum_{i \in \mathcal{S}} \mathsf{ReLU}(\langle w_i^{(t)}, x-\delta\rangle + \rho_i - b^{(t)}) - \mathsf{ReLU}(-\langle w_i^{(t)}, x-\delta\rangle + \rho_i - b^{(t)})$$
$$= \sum_{i \in \mathcal{S}} \mathsf{ReLU}(\langle w_i^{(t)}, x\rangle + \rho_i - b^{(t)}) - \mathsf{ReLU}(-\langle w_i^{(t)}, x\rangle + \rho_i - b^{(t)}) - \underbrace{\sum_{i \in \mathcal{S}} \left(\mathbb{1}_{w_{j_i}^\star z_{j_i}>0} + \mathbb{1}_{w_{j_i}^\star z_{j_i}<0}\right)\langle w_i^{(t)}, \delta\rangle}_{\clubsuit}$$

where $j_i \in [d]$ is the unique index such that $i \in \mathcal{S}_{j_i,sure}^{(0)}$. We can rewrite the decrement

$$\clubsuit = \delta_0 \sum_{j \in [d]} \left(\mathbb{1}_{w_j^\star z_j>0} + \mathbb{1}_{w_j^\star z_j<0}\right) \sum_{i \in \mathcal{S}_{j,sure}^{(0)}} \langle w_i^{(t)}, \mathbf{M} w^\star\rangle$$

Using $|\sum_{i \in \mathcal{S}_{j,sure}^{(0)}} \langle w_i^{(t)}, \mathbf{M} w^\star\rangle| \leq \Xi_2^2 \sqrt{d}|\mathcal{S}_{j,sure}^{(0)}| \leq \Xi_2^3 \sqrt{d}$ and $\left(\mathbb{1}_{w_j^\star z_j>0} + \mathbb{1}_{w_j^\star z_j<0}\right) = 1$ with probability $\Theta(\frac{k}{d})$, we can apply Bernstein's inequality and derive

$$\mathbf{Pr}\left[|\clubsuit - \mathbb{E}_z[\clubsuit]| > \delta_0 \cdot \Xi_2^3 \sqrt{kd}\log d\right] \leq e^{-\Omega(\log^2 d)}$$

Also using $\left(\mathbb{1}_{w_j^\star z_j>0} + \mathbb{1}_{w_j^\star z_j<0}\right) = 1$ with probability $\Theta(\frac{k}{d})$, we can derive using Lemma E.2 that

$$\mathbb{E}_z[\clubsuit] \geq \delta_0 \cdot \left(\Omega(\frac{k}{d})\sum_{i \in \mathcal{S}}\langle w_i^{(t)}, \mathbf{M} w^\star\rangle - O(k\Xi_2) \cdot \frac{k^2}{\lambda d^{1.5}}\sigma_x^2\right)$$
$$\geq \delta_0 \cdot \left(\Omega(\frac{k^2}{\Xi_2^7}) - O(k\Xi_2) \cdot \frac{1}{\lambda}\frac{k^3 \Xi_2^4}{d^2}\sigma_x^2\right) \geq \delta_0 \cdot \Omega(\frac{k^2}{\Xi_2^7})$$

Combining the above equations and using $k > d^{(1-c_0)/3}$, we have with probability at least $1 - e^{-\Omega(\log^2 d)}$,

$$\clubsuit \geq \delta_0 \cdot \Omega\left(\frac{k^2}{\Xi_2^7}\right)$$

For the remainder terms, we using $|\langle w_i^{(t)}, \delta\rangle| \leq \delta_0 \Xi_2^2 \sqrt{d} \leq b^{(t)}/2$, we have

$$\sum_{i \in [m]\setminus \mathcal{S}} \mathsf{ReLU}(\langle w_i^{(t)}, x-\delta\rangle + \rho_i - b^{(t)}) - \mathsf{ReLU}(-\langle w_i^{(t)}, x-\delta\rangle + \rho_i - b^{(t)})$$
$$\leq \sum_{i \in [m]\setminus \mathcal{S}} \mathsf{ReLU}(\langle w_i^{(t)}, x\rangle + \rho_i - b^{(t)}) - \mathsf{ReLU}(-\langle w_i^{(t)}, x\rangle + \rho_i - b^{(t)}) - \underbrace{\sum_{i \in [m]\setminus \mathcal{S}} \mathbb{1}_{|\langle w_i^{(t)},x\rangle|+|\rho_i|>\frac{b^{(t)}}{2}} \min\left\{0, \langle w_i^{(t)}, \delta\rangle\right\}}_{\spadesuit}$$

Using Lemma E.2 and Lemma C.19 we have with probability at least $1 - e^{\Omega(\log^2 d)}$,

$$\spadesuit \geq -\delta_0 \cdot O\left(\frac{1}{\lambda}\frac{k^3 \Xi_2^4}{d^2}\sigma_x^2\right) \cdot \sum_{i \in [m]\setminus \mathcal{S}} \mathbb{1}_{|\langle w_i^{(t)},x\rangle|+|\rho_i|>\frac{b^{(t)}}{2}} \geq -\delta_0 \cdot O\left(\frac{1}{\lambda}\frac{k^3 \Xi_2^4}{d^2}\sigma_x^2\right) \cdot O(k\Xi_2) \geq -\frac{\clubsuit}{2} \ .$$

Putting together the bounds for ♣ and ♠ we have

$$f_t(x - \delta) = \sum_{i \in [m]} \mathsf{ReLU}(\langle w_i^{(t)}, x - \delta \rangle + \rho_i - b^{(t)}) - \mathsf{ReLU}(-\langle w_i^{(t)}, x - \delta \rangle + \rho_i - b^{(t)})$$

$$\leq \sum_{i \in [m]} \mathsf{ReLU}(\langle w_i^{(t)}, x \rangle + \rho_i - b^{(t)}) - \mathsf{ReLU}(-\langle w_i^{(t)}, x \rangle + \rho_i - b^{(t)}) - \frac{\clubsuit}{2} \leq f_t(x) - \delta_0 \cdot \Omega\left(\frac{k^2}{\Xi_2^7}\right)$$

In other words, choosing $\delta_0 = \frac{\Xi_2^{10}}{k^2}$, then combining with $|f_t(x)| \leq O(\Xi_2^2 \log d)$ from Lemma C.21, we immediately have $f_t(x - \delta) < 0$.

Using an analogous proof, one can also show that $f_t(x + \delta) > 0$. Therefore, if we choose a perturb direction $-\delta_0 y \mathbf{M} w^* = -y\delta$, we have

$$\Pr_{x, y=y(x), \rho}\left[\mathsf{sign}(f_t(x - \delta_0 y \mathbf{M} w^*)) = y\right] \leq e^{-\Omega(\log^2 d)} \ .$$

This means the robust accuracy is below $e^{-\Omega(\log^2 d)}$. Finally, using $\|\mathbf{M} w^*\|_2 \leq O(\sqrt{d})$ and $\|\mathbf{M} w^*\|_\infty \leq O(\sum_{j \in [d]} \|\mathbf{M}_j\|_\infty) = O(\|\mathbf{M}\|_1)$ finishes the proof.

Note that a similar proof as above also shows

$$\Pr_{x, y=y(x)}\left[\mathsf{sign}\big(\mathbb{E}_\rho[f_t(x - \delta_0 y \mathbf{M} w^*)]\big) = y\right] \leq e^{-\Omega(\log^2 d)} \ . \qquad \square$$

# F Robust Training Through Local Feature Purification

Suppose we run clean training for $T_{\mathsf{f}} \geq \Omega(\frac{d\Xi_2^6}{\eta})$ iterations following Theorem D.1. From this iteration on, let us perform $T$ more steps of robust training.

During the robust training phase, let us consider an arbitrary (norm-bounded) adversarial perturbation algorithm $A$. Recall from Definition 4.2 that, given the current network $f$ (which includes hidden weights $\{w_i\}$, output weights $\{a_i\}$, bias $\{b_i\}$ and smoothing parameter $\sigma_\rho$), an input $x$, a label $y$, and some internal random string $r$, the perturbation algorithm $A$ outputs a vector satisfying

$$\|A(f, x, y, r)\|_p \leq \tau \ .$$

for some $\ell_p$ norm. Our two main theorems below apply to all such perturbation algorithms $A$ (including the Fast Gradient Method, FGM).

> **Theorem F.1** ($\ell_2$-adversarial training). *In the same setting as Theorem D.1, suppose we first run $T_{\mathsf{f}}$ iterations of clean training with $\Omega(\frac{d\Xi_2^6}{\eta}) \leq T_{\mathsf{f}} \leq d^{\log d}/\eta$ and obtain*
> 
> $$\mathbf{Obj}_{clean} = \mathbb{E}_{x, y=y(x), \rho}\left[\mathbf{Obj}_{T_{\mathsf{f}}}(w^{(T_{\mathsf{f}})}; x, y, \rho)\right] \leq o(1) \ .$$
> 
> *Next, suppose $\sigma_x \leq \min\{O(1), \frac{d^{2(1-2c_0)}}{k^{5.5}}\}$ and $k^{2.5} < d^{1-2c_0}/\log d$. Starting from iteration $T_{\mathsf{f}}$, suppose we perform robust training for additional $T = T_{\mathsf{g}} = \Theta(\frac{k^2 \Xi_2^4 m \log d}{\eta d}) \leq O(\frac{k^2}{d^{1-2c_0}})$ iterations, against some $\ell_2$ perturbation algorithm $A$ with radius $\tau \stackrel{\mathsf{def}}{=} \frac{1}{\sqrt{k} \cdot d^{c_0}}$. With probability $\geq 1 - e^{-\Omega(\log^2 d)}$,*
> 
> $$\frac{1}{T} \sum_{t=T_{\mathsf{f}}}^{T_{\mathsf{f}}+T-1} \mathbb{E}_{x, y=y(x), \delta=A(f_t, x, y, r), \rho}\left[\mathbf{Obj}_t(w^{(t)}; x + \delta, y, \rho)\right] \leq \mathbf{Obj}_{clean} + o(1)$$

**Corollary F.2.** *In Theorem F.1, if $A$ is Fast Gradient Method (FGM) with $\ell_2$ radius $\tau$, and*

$$\mathop{\mathbb{E}}_{x,y=y(x),\delta=A(f_t,x,y),\rho}\left[\mathbf{Obj}_t(w^{(t)};x+\delta,y,\rho)\right] \leq o(1)$$

*for some $t \in [T_{\mathsf{f}}, T_{\mathsf{f}} + T_{\mathsf{g}}]$. Then,*

$$\mathop{\mathbf{Pr}}_{x,y=y(x)}\left[\exists \delta \in \mathbb{R}^d, \|\delta\|_2 \leq \tau \colon \mathsf{sign}(\mathop{\mathbb{E}}_\rho f_t(w^{(t)};x+\delta,\rho)) \neq y\right] \leq o(1) \ .$$

**Corollary F.3.** *Consider for instance $\sigma_x = 0$, $c_0 = 0.00001$, and sufficiently large $d > 1$.*

- *For $k \in [d^{0.0001}, d^{0.3999}]$, robust training gives $99.9\%$ accuracy against $\ell_2$ perturbation $\geq \frac{1}{k^{0.5} \cdot d^{0.0001}}$.*
- *For $k \geq d^{0.3334}$, clean training gives $0.01\%$ accuracy against $\ell_2$ perturbation radius $\leq \frac{d^{0.5001}}{k^2}$.*
- *For $k \in [d^{0.3334}, d^{0.3999}]$, robust training provably beats clean training in $\ell_2$ robust accuracy.*

---

**Theorem F.4** ($\ell_\infty$-adversarial training). *In the same setting as Theorem D.1, suppose we first run $T_{\mathsf{f}}$ iterations of clean training with $\Omega(\frac{d\Xi_2^6}{\eta}) \leq T_{\mathsf{f}} \leq d^{\log d}/\eta$ and obtain*

$$\mathbf{Obj}_{clean} = \mathop{\mathbb{E}}_{x,y=y(x),\rho}\left[\mathbf{Obj}_{T_{\mathsf{f}}}(w^{(T_{\mathsf{f}})};x,y,\rho)\right] \leq o(1) \ .$$

*Next, suppose $\sigma_x \leq O(1)$ and $k^{2.5} < d^{1-2c_0}/\log d$. Starting from iteration $T_{\mathsf{f}}$, suppose we perform robust training for additional $T = T_{\mathsf{g}} = \Theta(\frac{k^2 \Xi_2^4 m \log d}{\eta d}) \leq O(\frac{k^2}{d^{1-2c_0}})$ iterations, against some $\ell_\infty$ perturbation algorithm $A$ of radius $\tau \stackrel{\text{def}}{=} \frac{1}{k^{1.75} \cdot \|\mathbf{M}\|_\infty \cdot d^{c_0}}$. Then, with probability $\geq 1 - e^{-\Omega(\log^2 d)}$,*

$$\frac{1}{T} \sum_{t=T_{\mathsf{f}}}^{T_{\mathsf{f}}+T-1} \mathop{\mathbb{E}}_{x,y=y(x),\delta=A(f_t,x,y,r),\rho}\left[\mathbf{Obj}_t(w^{(t)};x+\delta,y,\rho)\right] \leq \mathbf{Obj}_{clean} + o(1)$$

---

**Corollary F.5.** *In Theorem F.4, if $A$ is Fast Gradient Method (FGM) with $\ell_\infty$ radius $\tau$, and*

$$\mathop{\mathbb{E}}_{x,y=y(x),\delta=A(f_t,x,y),\rho}\left[\mathbf{Obj}_t(w^{(t)};x+\delta,y,\rho)\right] \leq o(1)$$

*for some $t \in [T_{\mathsf{f}}, T_{\mathsf{f}} + T_{\mathsf{g}}]$. Then,*

$$\mathop{\mathbf{Pr}}_{x,y=y(x)}\left[\exists \delta \in \mathbb{R}^d, \|\delta\|_\infty \leq \tau \colon \mathsf{sign}(\mathop{\mathbb{E}}_\rho f_t(w^{(t)};x+\delta,\rho)) \neq y\right] \leq o(1) \ .$$

**Corollary F.6.**

- *For $k \in [d^{0.0001}, d^{0.3999}]$, robust training gives $99.9\%$ accuracy against $\ell_\infty$ perturbation $\geq \frac{1}{k^{1.75} \cdot d^{0.0001} \cdot \|\mathbf{M}\|_\infty}$.*
- *For $k \geq d^{0.3334}$, clean training gives $0.01\%$ accuracy against $\ell_\infty$ perturbation $\leq \frac{d^{0.0001} \|\mathbf{M}\|_1}{k^2}$.*
- *For $k \in [d^{0.3334}, d^{0.3999}]$ and when $\|\mathbf{M}\|_\infty, \|\mathbf{M}\|_1 \leq d^{0.1248}$, robust training provably beats clean training in $\ell_\infty$ robust accuracy.*

*Remark* F.7. With additional efforts, one can also prove that Theorem F.1 and Theorem F.4 holds with high probability for all $T$ in the range $T = \Theta(T_{\mathsf{g}})$. We do not prove it here since it is not beyond the scope of this paper.

## F.1 Some Notations

We first note some simple structural properties that are corollaries of Theorem C.2.

**Proposition F.8.** *At iteration $t = T_f$, for every neuron $i \in [m]$, we can write*

$$w_i^{(t)} \stackrel{\text{def}}{=} g_i + u_i \stackrel{\text{def}}{=} \sum_{j \in \mathcal{S}_i} \alpha_{i,j} \mathbf{M}_r + u_i$$

*where $\mathcal{S}_i \subseteq \{j \in [d] \mid i \in \mathcal{S}_{j,pot}^{(0)}\}$ with $|\mathcal{S}_i| = O(1)$, $|\alpha_{i,j}| \leq O(\Xi_2^2)$ and $\max_{j \in [d]}\{|\langle u_i, \mathbf{M}_j \rangle|\} = \frac{k\Xi_2^2}{d}$.*

*Proof.* We can let $\alpha_{i,j} = \langle w_i^{(t)}, \mathbf{M}_j \rangle$ and let $u_i$ be the remaining part. We have $|\mathcal{S}_i| \leq O(1)$ because $\mathcal{S}_{ept}^{(0)} = [m]$. We have $|\alpha_{i,j}| = |\langle w_i^{(t)}, \mathbf{M}_j \rangle| \leq \|w_i^{(t)}\| \leq O(\Xi_2^2)$. We also have

$$\max_{j \in [d]}\{|\langle u_i, \mathbf{M}_j \rangle|\} = \max_{j \in [d]:\, i \notin \mathcal{S}_{j,pot+}^{(0)}}\{|\langle w_i^{(t)}, \mathbf{M}_j \rangle|\} \leq \frac{k}{d\beta} b^{(t)} \leq \frac{k\Xi_2^2}{d} \ . \qquad \square$$

We next introduce an important notation that shall be used throughout the proofs of this section.

**Definition F.9.** *For every $t \geq T_f$, we write $w_i^{(t)} = g_i + v_i^{(t)}$ by defining $v_i^{(t)} \stackrel{\text{def}}{=} w_i^{(t)} - g_i$.*

## F.2 Robust Coupling

**Definition F.10** (robust coupling). *At every iteration $t$, recalling $w_i^{(t)} = g_i + v_i^{(t)}$, we define a linear function in $\mu$*

$$g_t(\mu; x, x_0, \rho) = \sum_{i=1}^m \Big( \mathbb{1}_{\langle g_i + v_i^{(t)}, x_0 \rangle + \rho_i \geq b^{(t)}} \cdot (\langle g_i + \mu_i, x \rangle + \rho_i - b^{(t)})$$
$$- \mathbb{1}_{-\langle g_i + v_i^{(t)}, x_0 \rangle + \rho_i \geq b^{(t)}} \cdot (-\langle g_i + \mu_i, x \rangle + \rho_i - b^{(t)}) \Big)$$

*and it equals the output of the real network at point $\mu = v^{(t)}$ both on its zero and first order:*

$$g_t(\mu; x + \delta, x + \delta, \rho)\big|_{\mu = v^{(t)}} = f_t(w; x + \delta, \rho)\big|_{w = w^{(t)}}$$
$$\nabla_\mu g_t(\mu; x + \delta, x + \delta, \rho)\big|_{\mu = v^{(t)}} = \nabla_w f_t(w; x + \delta, \rho)\big|_{w = w^{(t)}}$$

We shall show in this section that, recalling $w^{(t)} = g + v^{(t)}$, then

$$g_t(v^{(t)}; x + \delta, x, \rho) \approx f_t(w^{(t)}; x + \delta, \rho)$$
$$g_t(0; x + \delta, x, \rho) \approx f_t(w^{(T_f)}, x, \rho)$$

However, as regarding how close they are, it depends on whether we have an $\ell_2$ bound or $\ell_\infty$ bound on $\delta$, so we shall prove the two cases separately in Section F.2.1 and F.2.2.

It is perhaps worth nothing that the "closeness" of the above terms depend on two things,

- One is regarding how small $\sum_{i \in [m]} \|v_i^{(t)}\|_2^2$ is, and this shall later be *automatically* guaranteed via implicit regularization of first-order methods.

- The other is regarding how small $\|v_i^{(t)}\|_2$ or $\|v_i^{(t)}\|_1$ is for *every* individual neuron $i \in [m]$. This is a bit non-trivial to prove, and we shall spend the entire Section F.3 to deal with this.

### F.2.1 Robust Coupling for $\ell_2$ Perturbation

**Lemma F.11.** *Suppose at iteration $t$, $\sum_{i\in[m]} \|v_i^{(t)}\|_2^2 \leq r^2 m$ for some $r \leq 1$, and suppose $\max_{i\in[m]} \|v_i^{(t)}\|_2 \leq r'$. Then for any vector $\delta \in \mathbb{R}^d$ that can depend on $x$ (but not on $\rho$) with $\|\delta\|_2 \leq \tau$ for some $\tau \leq o(\frac{b}{\Xi_2^2 + r'})$, we have*

$$\mathbb{E}_{x,\rho}\left[\left|g_t(v^{(t)}; x+\delta, x, \rho) - f_t(w^{(t)}; x+\delta, \rho)\right|\right] \leq O(\tau^2) \cdot \left(\frac{\Xi_2^5}{\sigma_\rho} + \frac{(\Xi_2^2 + r')^2 r^2 m}{db^2 \sigma_\rho}\right)$$

*As a corollary, in the event of $r \leq O\left(\frac{k\Xi_2^2}{\sqrt{d}}\right)$ and $r' \leq 1$ and using $m = d^{1+c_0}$, we have*

$$\mathbb{E}_{x,\rho}\left[\left|g_t(v^{(t)}; x+\delta, x, \rho) - f_t(w^{(t)}; x+\delta, \rho)\right|\right] \leq O(\tau^2) \cdot \left(\frac{\Xi_2^5}{\sigma_\rho} + \frac{k^{3.5}}{d^{1-2c_0}}\right)$$

*Proof of Lemma F.11.* Let us abbreviate the notations by setting $v_i = v_i^{(t)}$ and $b = b^{(t)}$.

To upper bound $|g_t(v^{(t)}; x+\delta, x, \rho) - f_t(w^{(t)}; x+\delta, \rho)|$ it suffices to upper bound $|V_1 - V_2|$ for

$$V_1 := \sum_{i\in[m]} (\langle g_i + v_i, x+\delta \rangle + \rho_i - b) \mathbb{1}_{\langle g_i+v_i, x+\delta\rangle + \rho_i \geq b}$$

$$V_2 := \sum_{i\in[m]} (\langle g_i + v_i, x+\delta \rangle + \rho_i - b) \mathbb{1}_{\langle g_i+v_i, x\rangle + \rho_i \geq b}$$

(and one also needs to take into account the reverse part, whose proof is analogous).

We first make some calculations. Using the definition of $g_i$, we have $\mathbf{Pr}_x[\langle g_i, x\rangle \geq |b|/10] \leq O\left(\frac{k}{d}\right)$ for every $i \in [m]$. Thus, we can easily calculate that[21]

$$\mathbb{E}_x\left[\sum_{i\in[m]} \langle v_i, \delta\rangle^2 \mathbb{1}_{\langle g_i, x\rangle \geq |b|/10}\right] \leq \tau^2 \sum_{i\in[m]} \|v_i\|^2 \mathbb{E}_x\left[\mathbb{1}_{\langle g_i, x\rangle \geq |b|/10}\right] = O\left(\tau^2 \cdot r^2 m \cdot \frac{k}{d}\right) \quad \text{(F.1)}$$

$$\mathbb{E}_x\left[\sum_{i\in[m]} (\langle v_i, \delta\rangle^2 + \langle g_i, \delta\rangle^2) \mathbb{1}_{\langle v_i, x\rangle \geq |b|/10}\right] \leq \tau^2 \cdot O(\Xi_2^4 + (r')^2) \sum_{i\in[m]} \mathbb{E}_x\left[\mathbb{1}_{\langle v_i, x\rangle \geq |b|/10}\right]$$

$$\leq \tau^2 \cdot O((\Xi_2^2 + r')^2) \cdot \sum_{i\in[m]} O\left(\mathbb{E}_x \frac{\langle v_i, x\rangle^2}{b^2}\right)$$

$$= O\left(\tau^2 (\Xi_2^2 + r')^2 \cdot \frac{r^2 m}{db^2}\right) \quad \text{(F.2)}$$

$$\sum_{i\in[m]} \langle g_i, \delta\rangle^2 \mathbb{1}_{\langle g_i, x\rangle \geq |b|/10} \leq \sum_{i\in[m]} \langle g_i, \delta\rangle^2 \leq \tau^2 \left\|\sum_{i\in[m]} g_i g_i^\top\right\|_{spectral-norm} \leq O(\tau^2 \Xi_2^5) \quad \text{(F.3)}$$

Now, for every $i \in [m]$,

- Case 1, $|\langle v_i, x\rangle| \leq \frac{b}{10}$ and $|\langle g_i, x\rangle| \leq \frac{b}{10}$ both happen. In this case, it must satisfy $|\langle g_i + v_i, \delta\rangle| \leq (\|g_i\| + \|v_i\|) \cdot \tau \leq O(\Xi_2^2 + r') \cdot \tau \leq \frac{b}{10}$. . Also, with probability at least $1 - e^{-\Omega(\log^2 d)}$, it satisfies

---

[21] Here, the spectral norm bound of $\sum_{i\in[m]} g_i g_i^\top$ holds for the following reason. Each $g_i$ is a sparse vector supported only on $|\mathcal{S}_i| = O(1)$ coordinates, and thus $g_i g_i^\top \preceq \mathbf{D}_i$ holds for a diagonal matrix $\mathbf{D}_i$ that where $[\mathbf{D}_i]_{j,j} = \|g_i\|^2 \leq O(\Xi_2^4)$ for $j \in \mathcal{S}_i$ and $[\mathbf{D}_i]_{j,j} = 0$ otherwise. Now, using the fact that $|\mathcal{S}_{j,pot}^{(0)}| \leq \Xi_2$, we immediately have that $\mathbf{D}_1 + \cdots + \mathbf{D}_m \preceq O(\Xi_2^5) \cdot \mathbf{I}_{d\times d}$.

$|\rho_i| \leq \frac{b}{10}$. To sum up, with high probability we have
$$\mathbb{1}_{\langle g_i+v_i, x+\delta\rangle+\rho_i \geq b} = \mathbb{1}_{\langle g_i+v_i, x\rangle+\rho_i \geq b} = 0$$

- Case 2, either $|\langle v_i, x\rangle| > \frac{b}{10}$ or $|\langle g_i, x\rangle| > \frac{b}{10}$. In this case, to satisfy $\mathbb{1}_{\langle g_i+v_i, x+\delta\rangle+\rho_i \geq b} \neq \mathbb{1}_{\langle g_i+v_i, x\rangle+\rho_i \geq b}$, one must have $|\langle g_i + v_i, x+\delta\rangle - b + \rho_i| \leq |\langle v_i, \delta\rangle| + |\langle g_i, \delta\rangle|$. Also, using the randomness of $\rho_i$, we have
$$\Pr_{\rho_i}\left[\mathbb{1}_{\langle g_i+v_i, x+\delta\rangle+\rho_i \geq b} \neq \mathbb{1}_{\langle g_i+v_i, x\rangle+\rho_i \geq b}\right] \leq O\left(\frac{|\langle v_i, \delta\rangle| + |\langle g_i, \delta\rangle|}{\sigma_\rho}\right)$$

Together, we have

$$\begin{aligned}
\mathbb{E}_{x,\rho,\delta}[|V_1 - V_2|] &\leq \mathbb{E}_{x,\delta,\rho}\left[\sum_{i\in[m]} \left(|\langle v_i, \delta\rangle| + |\langle g_i, \delta\rangle|\right)\left|\mathbb{1}_{\langle g_i+v_i, x+\delta\rangle+\rho_i \geq b} - \mathbb{1}_{\langle g_i+v_i, x\rangle+\rho_i \geq b}\right|\right] \\
&\leq O(1) \cdot \mathbb{E}_{x,\delta,\rho}\left[\sum_{i\in[m]} \frac{\langle v_i, \delta\rangle^2 + \langle g_i, \delta\rangle^2}{\sigma_\rho}\left(\mathbb{1}_{\langle v_i, x\rangle \geq |b|/10} + \mathbb{1}_{\langle g_i, x\rangle \geq |b|/10}\right)\right] + \frac{1}{\mathsf{poly}(d)} \quad (\text{F.4}) \\
&\leq O(\tau^2) \cdot \left(\frac{\Xi_2^5}{\sigma_\rho} + \frac{kr^2m}{d\sigma_\rho} + \frac{(\Xi_2^2 + r')^2 r^2 m}{db^2 \sigma_\rho}\right) \\
&\leq O(\tau^2) \cdot \left(\frac{\Xi_2^5}{\sigma_\rho} + \frac{(\Xi_2^2 + r')^2 r^2 m}{db^2 \sigma_\rho}\right) \quad \square
\end{aligned}$$

**Lemma F.12.** *Suppose at iteration $t$, $\max_{i\in[m], j\in[d]}\{|\langle u_i, \mathbf{M}_j\rangle|\} \leq \frac{r}{\sqrt{d}}$ and $\sum_{i\in[m]} \|v_i^{(t)}\|_2^2 \leq r^2 m$ with $r \leq 1$. Then for any vector $\delta \in \mathbb{R}^d$ that can depend on $x$ (but not on $\rho$) with $\|\delta\|_2 \leq \tau$, we have*

$$\mathbb{E}_{x,\rho}[|g_t(0; x+\delta, x, \rho) - f_t(w^{(T_\mathfrak{f})}, x, \rho)|] \leq O\left(\frac{mr^2}{d\sigma_\rho} + k\Xi_2 \frac{r\log d}{\sqrt{d}} + \tau\Xi_2^3 \sqrt{k\Xi_2 + \frac{r^2 m}{db^2}}\right)$$

*As a corollary, in the event of $r \leq O\left(\frac{k\Xi_2^2}{\sqrt{d}}\right)$ and using $m = d^{1+c_0}$, we have*

$$\mathbb{E}_{x,\rho}[|g_t(0; x+\delta, x, \rho) - f_t(w^{(T_\mathfrak{f})}, x, \rho)|] \leq O\left(\frac{k^{2.5}}{d^{1-2c_0}} + \tau \cdot \sqrt{k\Xi_2^7}\right)$$

*Proof of Lemma F.12.* To upper bound $|g_t(0; x+\delta, x, \rho) - f_t(v^{(T_\mathfrak{f})}, x)|$ it suffices to upper bound $|V_3 - V_4|$ for

$$V_3 := \sum_{i\in[m]} (\langle g_i, x+\delta\rangle - b + \rho_i)\mathbb{1}_{\langle g_i + v_i, x\rangle + \rho_i \geq b}$$
$$V_4 := \sum_{i\in[m]} (\langle g_i + u_i, x\rangle - b + \rho_i)\mathbb{1}_{\langle g_i + u_i, x\rangle + \rho_i \geq b}$$

(and one also needs to take into account the reverse part, whose proof is analogous).

Let us first define
$$V_5 := \sum_{i\in[m]} (\langle g_i, x\rangle - b + \rho_i)\mathbb{1}_{\langle g_i + v_i, x\rangle + \rho_i \geq b}$$

Let us define $s = \sum_{i\in[m]} \mathbb{1}_{\langle g_i+v_i, x\rangle+\rho_i \geq b}$. By the properties that (1) $g_i$ is only supported on $\mathcal{S}_i$ with $|\mathcal{S}_i| \leq O(1)$, (2) for each $j \in [d]$ at most $\Xi_2$ of the $g_i$ are supported on $i$, and (3) $\|g_i\|_2 \leq O(\Xi_2^2)$,

we can obtain

$$\mathbb{E}_{x,\rho}\left\|\sum_{i\in[m]} g_i \mathbb{1}_{\langle g_i+v_i,x\rangle+\rho_i\geq b}\right\|_2 \leq O(\Xi_2^3)\mathbb{E}_{x,\rho}[\sqrt{s}] \leq O(\Xi_2^3)\sqrt{\mathbb{E}_{x,\rho}[s]}$$

At the same time, we know

$$\mathbb{E}_{x,\rho}[s] \leq \mathbb{E}_{x,\rho}\left[\sum_{i\in[m]} \mathbb{1}_{\langle g_i,x\rangle\geq \frac{b}{4}} + \mathbb{1}_{\langle v_i,x\rangle\geq \frac{b}{4}}\right] + \frac{1}{\mathsf{poly}(d)} \leq O(k\Xi_2) + O\left(\sum_{i\in[m]} \frac{\mathbb{E}[\langle v_i,x\rangle^2]}{b^2}\right)$$

$$\leq O\left(k\Xi_2 + r^2 m \cdot \frac{1}{db^2}\right)$$

Putting them together, we have

$$\mathbb{E}_{x,\rho}|V_3 - V_5| \leq \tau \cdot \mathbb{E}_{x,\rho}\left\|\sum_{i\in[m]} g_i \mathbb{1}_{\langle g_i+v_i,x\rangle+\rho_i\geq b}\right\|_2 \leq O\left(\tau\Xi_2^3\sqrt{k\Xi_2 + \frac{r^2 m}{db^2}}\right) \quad \text{(F.5)}$$

Next, let us define

$$V_6 := \sum_{i\in[m]} (\langle g_i, x\rangle - b + \rho_i) \mathbb{1}_{\langle g_i,x\rangle+\rho_i\geq b}$$

Using a similar analysis to (F.4), we have

$$\mathbb{E}_{x,\rho}|V_5 - V_6| \leq \mathbb{E}_{x,\rho}\sum_{i\in[m]} |\langle g_i, x\rangle - b + \rho_i| \cdot \left|\mathbb{1}_{\langle g_i,x\rangle+\rho_i\geq b} - \mathbb{1}_{\langle g_i+v_i,x\rangle+\rho_i\geq b}\right|$$

$$\leq \mathbb{E}_{x,\rho}\sum_{i\in[m]} |\langle v_i, x\rangle|\mathbb{E}[|\mathbb{1}_{\langle g_i,x\rangle+\rho_i\geq b} - \mathbb{1}_{\langle g_i+v_i,x\rangle+\rho_i\geq b}|]$$

$$\leq O\left(\mathbb{E}_{x,\rho}\frac{\langle v_i,x\rangle^2}{\sigma_\rho}\right) \leq O\left(\frac{r^2 m}{d\sigma_\rho}\right) \quad \text{(F.6)}$$

Finally, we also have

$$\mathbb{E}_{x,\rho}|V_6 - V_4| = \mathbb{E}_{x,\rho}\left|\sum_{i\in[m]} (\langle g_i+u_i, x\rangle - b + \rho_i)\mathbb{1}_{\langle g_i+u_i,x\rangle+\rho_i\geq b} - (\langle g_i, x\rangle - b + \rho_i)\mathbb{1}_{\langle g_i,x\rangle+\rho_i\geq b}\right|$$

$$\leq \mathbb{E}_{x,\rho}\left|\sum_{i\in[m]} \langle u_i, x\rangle \mathbb{1}_{\langle g_i+u_i,x\rangle+\rho_i\geq b}\right| + \mathbb{E}_{x,\rho}\left|\sum_{i\in[m]} (\langle g_i,x\rangle - b + \rho_i)\left(\mathbb{1}_{\langle g_i+u_i,x\rangle+\rho_i\geq b} - \mathbb{1}_{\langle g_i,x\rangle+\rho_i\geq b}\right)\right|$$

$$\overset{①}{\leq} \mathbb{E}_{x,\rho}\sum_{i\in[m]} |\langle u_i, x\rangle|\mathbb{1}_{\langle g_i+u_i,x\rangle+\rho_i\geq b} + O\left(\frac{mr^2}{d\sigma_\rho}\right)$$

$$\overset{②}{\leq} \mathbb{E}_{x,\rho}\sum_{i\in[m]} |\langle u_i, x\rangle|\left(\mathbb{1}_{\langle g_i,x\rangle\geq b/4} + \mathbb{1}_{\langle u_i,x\rangle\geq b/4}\right) + O\left(\frac{mr^2}{d\sigma_\rho}\right) + \frac{1}{\mathsf{poly}(d)}$$

Above, inequality ① is due to a similar analysis as (F.6), and inequality ② is because $|\rho_i| \leq b/4$ with probability at least $1 - e^{-\Omega(\log^2 d)}$. Next, let us recall $\langle u_i, \mathbf{M}_j\rangle \leq \frac{r}{\sqrt{d}}$ and thus, by Bernstein's inequality, with probability at least $1 - e^{-\Omega(\log^2 d)}$,

$$|\langle u_i, x\rangle| \leq O\left(\frac{r\log^2 d}{\sqrt{d}}\right) \ll \frac{b}{4} \ .$$

Putting this back we have

$$\mathbb{E}_{x,\rho} |V_6 - V_4| \leq O\left(\frac{mr^2}{d\sigma_\rho} + k\Xi_2 \frac{r\log d}{\sqrt{d}}\right)$$

Combining the bounds on $|V_6 - V_4|$, $|V_3 - V_5|$, and $|V_5 - V_6|$ finishes the proof. □

### F.2.2 Robust Coupling for $\ell_\infty$ Perturbation

**Lemma F.13.** *Suppose at iteration $t$, $\sum_{i\in[m]} \|v_i^{(t)}\|_2^2 \leq r^2 m$ for some $r \leq 1$, and suppose $\max_{i\in[m]} \|v_i^{(t)}\|_1 \leq r'$. Then for any vector $\delta \in \mathbb{R}^d$ that can depend on $x$ (but not on $\rho$) with $\|\delta\|_\infty \leq \tau$ for some $\tau \leq o(\frac{b}{\Xi_2^2 + r'})$, we have*

$$\mathbb{E}_{x,\rho}\left[|g_t(v^{(t)}; x+\delta, x, \rho) - f_t(w^{(t)}; x+\delta, \rho)|\right] \leq O(\tau^2) \cdot \left(\frac{k\Xi_2^5}{\sigma_\rho} + \frac{(\Xi_2^2 + r')^2 r^2 m}{db^2 \sigma_\rho} + \frac{(r')^2 k\Xi_2}{\sigma_\rho}\right)$$

*As a corollary, in the event of $r \leq O\left(\frac{k\Xi_2^2}{\sqrt{d}}\right)$ and $r' \leq O(k\Xi_2^2 \cdot \|\mathbf{M}\|_\infty)$ and using $m = d^{1+c_0}$, we have*

$$\mathbb{E}_{x,\rho}\left[|g_t(v^{(t)}; x+\delta, x, \rho) - f_t(w^{(t)}; x+\delta, \rho)|\right] \leq O(\tau^2) \cdot k^{3.5} d^{c_0} \cdot \|\mathbf{M}\|_\infty^2$$

*Proof of Lemma F.13.* The proof is analogous to Lemma F.11 so we only highly the differences. In fact, we only need to change (F.1), (F.2) and (F.3) with the following calculations.

Using the definition of $g_i$, we have $\mathbf{Pr}_x[\langle g_i, x\rangle \geq |b|/10] \leq O\left(\frac{k}{d}\right)$ for every $i \in [m]$ as well as $\sum_{i\in[m]} \mathbb{E}_x[\mathbb{1}_{\langle g_i,x\rangle \geq |b|/10}] \leq O(k\Xi_2)$. Thus, we can easily calculate that[22]

$$\mathbb{E}_x\left[\sum_{i\in[m]} \langle v_i,\delta\rangle^2 \mathbb{1}_{\langle g_i,x\rangle \geq |b|/10}\right] \leq \tau^2 \sum_{i\in[m]} (r')^2 \mathbb{E}_x\left[\mathbb{1}_{\langle g_i,x\rangle \geq |b|/10}\right] = O\left(\tau^2 \cdot (r')^2 \cdot k\Xi_2\right)$$

$$\mathbb{E}_x\left[\sum_{i\in[m]} (\langle v_i,\delta\rangle^2 + \langle g_i,\delta\rangle^2)\mathbb{1}_{\langle v_i,x\rangle \geq |b|/10}\right] \leq \tau^2 \cdot O(\Xi_2^4 + (r')^2) \sum_{i\in[m]} \mathbb{E}_x\left[\mathbb{1}_{\langle v_i,x\rangle \geq |b|/10}\right]$$

$$\leq \tau^2 \cdot O((\Xi_2^2 + r')^2) \cdot \sum_{i\in[m]} O\left(\mathbb{E}_x \frac{\langle v_i,x\rangle^2}{b^2}\right)$$

$$= O\left(\tau^2(\Xi_2^2 + r')^2 \cdot \frac{r^2 m}{db^2}\right)$$

$$\mathbb{E}_x\left[\sum_{i\in[m]} \langle g_i,\delta\rangle^2 \mathbb{1}_{\langle g_i,x\rangle \geq |b|/10}\right] \leq O(\tau^2 \Xi_2^4) \sum_{i\in[m]} \mathbb{E}_x\left[\mathbb{1}_{\langle g_i,x\rangle \geq |b|/10}\right] \leq O(\tau^2 k\Xi_2^5)$$

Putting those into the rest of the proof (to replace (F.1), (F.2) and (F.3)) finishes the proof. □

**Lemma F.14.** *Suppose at iteration $t$, $\max_{i\in[m],j\in[d]}\{|\langle u_i, \mathbf{M}_j\rangle|\} \leq \frac{r}{\sqrt{d}}$ and $\sum_{i\in[m]} \|v_i^{(t)}\|_2^2 \leq r^2 m$ with $r \leq 1$. Then for any vector $\delta \in \mathbb{R}^d$ that can depend on $x$ (but not on $\rho$) with $\|\delta\|_\infty \leq \tau$, we*

---

[22]Here, the spectral norm bound of $\sum_{i\in[m]} g_i g_i^\top$ holds for the following reason. Each $g_i$ is a sparse vector supported only on $|\mathcal{S}_i| = O(1)$ coordinates, and thus $g_i g_i^\top \preceq \mathbf{D}_i$ holds for a diagonal matrix $\mathbf{D}_i$ that where $[\mathbf{D}_i]_{j,j} = \|g_i\|^2 \leq O(\Xi_2^4)$ for $j \in \mathcal{S}_i$ and $[\mathbf{D}_i]_{j,j} = 0$ otherwise. Now, using the fact that $|\mathcal{S}_{j,pot}^{(0)}| \leq \Xi_2$, we immediately have that $\mathbf{D}_1 + \cdots + \mathbf{D}_m \preceq O(\Xi_2^5) \cdot \mathbf{I}_{d\times d}$.

*have*

$$\mathbb{E}_{x,\rho}[|g_t(0; x+\delta, x, \rho) - f_t(w^{(T_f)}; x, \rho)|] \leq O\left(\frac{mr^2}{d\sigma_\rho} + k\Xi_2 \frac{r\log d}{\sqrt{d}} + \tau \Xi_2^3\left(k\Xi_2 + \frac{r^2 m}{db^2}\right)\right)$$

*As a corollary, in the event of $r \leq O\left(\frac{k\Xi_2^2}{\sqrt{d}}\right)$ and using $m = d^{1+c_0}$, we have*

$$\mathbb{E}_{x,\rho}[|g_t(0; x+\delta, x, \rho) - f_t(w^{(T_f)}; x, \rho)|] \leq O\left(\frac{k^{2.5}}{d^{1-2c_0}} + \tau \cdot k\Xi_2^4\right)$$

*Proof of Lemma F.14.* The proof is analogous to Lemma F.12 so we only highly the differences. Recall we have defined

$$V_3 := \sum_{i \in [m]} (\langle g_i, x+\delta\rangle - b + \rho_i) \mathbb{1}_{\langle g_i + v_i, x\rangle + \rho_i \geq b}$$

$$V_4 := \sum_{i \in [m]} (\langle g_i + u_i, x\rangle - b + \rho_i) \mathbb{1}_{\langle g_i + u_i, x\rangle + \rho_i \geq b}$$

$$V_5 := \sum_{i \in [m]} (\langle g_i, x\rangle - b + \rho_i) \mathbb{1}_{\langle g_i + v_i, x\rangle + \rho_i \geq b}$$

$$V_6 := \sum_{i \in [m]} (\langle g_i, x\rangle - b + \rho_i) \mathbb{1}_{\langle g_i, x\rangle + \rho_i \geq b}$$

The bounds on $\mathbb{E}|V_5 - V_6|$ and $\mathbb{E}|V_6 - V_4|$ state exactly the same comparing to Lemma F.12 (because they do not have $\delta$ involved). Let us now recalculate the difference $|V_3 - V_5|$.

Let us define $s = \sum_{i \in [m]} \mathbb{1}_{\langle g_i + v_i, x\rangle + \rho_i \geq b}$. By the properties that (1) $g_i$ is only supported on $\mathcal{S}_i$ with $|\mathcal{S}_i| \leq O(1)$, (2) for each $j \in [d]$ at most $\Xi_2$ of the $g_i$ are supported on $i$, and (3) $\|g_i\|_2 \leq O(\Xi_2^2)$, we can obtain

$$\mathbb{E}_{x,\rho}\left\|\sum_{i \in [m]} g_i \mathbb{1}_{\langle g_i + v_i, x\rangle + \rho_i \geq b}\right\|_1 \leq O(\Xi_2^3) \mathbb{E}_{x,\rho}[s]$$

At the same time, we know

$$\mathbb{E}_{x,\rho}[s] \leq \mathbb{E}_{x,\rho}\left[\sum_{i \in [m]} \mathbb{1}_{\langle g_i, x\rangle \geq \frac{b}{4}} + \mathbb{1}_{\langle v_i, x\rangle \geq \frac{b}{4}}\right] + \frac{1}{\mathsf{poly}(d)} \leq O(k\Xi_2) + O\left(\sum_{i \in [m]} \frac{\mathbb{E}[\langle v_i, x\rangle^2]}{b^2}\right)$$

$$\leq O\left(k\Xi_2 + r^2 m \cdot \frac{1}{db^2}\right)$$

Putting them together, we have

$$\mathbb{E}_{x,\rho,\delta}|V_3 - V_5| \leq \tau \cdot \mathbb{E}_{x,\rho}\left\|\sum_{i \in [m]} g_i \mathbb{1}_{\langle g_i + v_i, x\rangle + \rho_i \geq b}\right\|_1 \leq O\left(\tau \Xi_2^3\left(k\Xi_2 + \frac{r^2 m}{db^2}\right)\right)$$

Using this new bound on $\mathbb{E}[V_3 - V_5]$ to replace the old one (F.5), the rest of the proof follows. $\square$

### F.3 Individual Neuron Growth Lemma

As mentioned earlier, the purpose of this section is to upper bound $\max_{i \in [m]} \|v_i^{(T_f+T)}\|_2$ (if it is $\ell_2$ perturbation) or $\max_{i \in [m]} \|v_i^{(T_f+T)}\|_1$ (if it is $\ell_\infty$ perturbation) during the course of robust training. We have two subsections to deal with the two cases.

### F.3.1 Growth Lemma for $\ell_2$ Perturbation

We first bound $\|v_i^{(t)}\|_2$ during $\ell_2$ robust training.

**Lemma F.15** (movement bound). *Suppose at iteration $t$, $\max_{i \in [m]} \|v_i\|_2 \leq r'$. Let $\ell \in [-1, 1]$ be any random variable that can depend on $x, \rho$, and $\delta \in \mathbb{R}^d$ be any random vector that can depend on $x$ with $\|\delta\|_2 \leq \tau$. Then, for every $i \in [m]$,*

$$\left\| \mathop{\mathbb{E}}_{x,y=y(x),\rho} \left[ \ell \mathbb{1}_{\langle g_i + v_i, x+\delta \rangle + \rho_i \geq b}(x+\delta) \right] \right\|_2 \leq O\left( \left( \frac{k}{d} + \frac{(r')^2 \log d}{db^2} \right) \tau + \frac{\sqrt{k}}{d} + \frac{(r')^2}{db^2} \left( \frac{\sqrt{k}}{\sqrt{d}} + \sigma_x \log d \right) + \frac{r'}{db} \right)$$

*As a corollary, suppose we run robust training from iteration $T_{\mathsf{f}}$ to $T_{\mathsf{f}} + T$ with $T\eta \leq o(db)$, $\tau \leq \frac{1}{\sqrt{k} \log d}$ and $\sigma_x \leq o(\frac{d^2 b^2}{(T\eta)^2 \sqrt{k} \log d})$, then*

$$\max_{i \in [m]} \|v_i^{(T_{\mathsf{f}}+T)}\|_2 \leq O\left( \frac{k \Xi_2^2}{\sqrt{d}} + T\eta \cdot \frac{\sqrt{k}}{d} \right) \leq o(1)$$

*Proof of Lemma F.15.* First of all we can reuse the analysis of (F.4) and derive that

$$\mathbf{Pr}[\langle g_i + v_i, x + \delta \rangle + \rho_i \geq b] \leq \frac{1}{\mathsf{poly}(d)} + \mathbf{Pr}[\langle g_i, x \rangle \geq |b|/10] + \mathbf{Pr}[\langle v_i, x \rangle \geq |b|/10]$$

$$\leq O\left( \frac{k}{d} + \frac{(r')^2}{db^2} \right) =: \kappa$$

This immediately gives

$$\| \mathbb{E}[\ell \mathbb{1}_{\langle g_i + v_i, x+\delta \rangle + \rho_i \geq b} \delta] \|_2 \leq \tau \cdot \mathbb{E}\left[ \mathbb{1}_{\langle g_i + v_i, x+\delta \rangle + \rho_i \geq b} \right] \leq \kappa \tau \ .$$

Next, in order to bound the norm of $\phi \stackrel{\text{def}}{=} \mathbb{E}\left[ \ell \mathbb{1}_{\langle g_i + v_i, x+\delta \rangle + \rho_i \geq b} x \right]$, we first inner product it with $\mathbf{M}_j$ for each $j \in [d]$. This gives

$$|\langle \phi, \mathbf{M}_j \rangle| = \left| \mathbb{E}\left[ \ell \mathbb{1}_{\langle g_i + v_i, x+\delta \rangle + \rho_i \geq b} \langle x, \mathbf{M}_j \rangle \right] \right|$$

$$\leq \mathbb{E}[(\mathbb{1}_{\langle g_i, x \rangle \geq b/10} + \mathbb{1}_{\langle v_i, x \rangle \geq b/10}) \cdot |\langle x, \mathbf{M}_j \rangle|] + \frac{1}{\mathsf{poly}(d)}$$

$$\leq \mathbb{E}[(\mathbb{1}_{\langle g_i, x \rangle \geq b/10} + \mathbb{1}_{\langle v_i, x - \mathbf{M}_j z_j \rangle \geq b/20} + \mathbb{1}_{\langle v_i, \mathbf{M}_j \rangle z_j \geq b/20}) \cdot |\langle x, \mathbf{M}_j \rangle|] + \frac{1}{\mathsf{poly}(d)} \quad (F.7)$$

We bound the three terms separately.

- For the first term,

$$\mathbb{E}[\mathbb{1}_{\langle g_i, x \rangle \geq b/10} \cdot |\langle x, \mathbf{M}_j \rangle|] \leq \mathbb{E}[\mathbb{1}_{\langle g_i, x \rangle \geq b/10} \cdot (|z_j| + O(\frac{\log d}{\sqrt{d}}))] \leq \mathbb{E}[\mathbb{1}_{\langle g_i, x \rangle \geq b/10} \cdot |z_j|] + O(\frac{k \log d}{d^{1.5}}))$$

Using the property of $g_i$ we have $\mathbb{1}_{\langle g_i, x \rangle \geq b/10} \leq \sum_{j' \in \mathcal{S}_i} \mathbb{1}_{z_{j'} \neq 0}$ for $|\mathcal{S}_i| \leq O(1)$ and therefore

$$\mathbb{E}[\mathbb{1}_{\langle g_i, x \rangle \geq b/10} \cdot |z_j|] \leq \mathbb{E}\left[ \sum_{j' \in \mathcal{S}_i} \mathbb{1}_{z_{j'} \neq 0} \cdot |z_j| \right] = \begin{cases} \sqrt{k}/d, & \text{if } j \in \mathcal{S}_i; \\ k^{1.5}/d^2, & \text{if } j \notin \mathcal{S}_i. \end{cases}$$

Therefore, we have

$$\sum_{j \in [d]} \left( \mathbb{E}[\mathbb{1}_{\langle g_i, x \rangle \geq b/10} \cdot |\langle x, \mathbf{M}_j \rangle|] \right)^2 \leq O\left( \frac{k}{d^2} \right) \quad (F.8)$$

- For the second term,

$$\mathbb{E}\left[\mathbb{1}_{\langle v_i, x-\mathbf{M}_j z_j\rangle \geq b/20} \cdot |\langle x, \mathbf{M}_j\rangle|\right] = \mathbb{E}\left[\mathbb{1}_{\langle v_i, x-\mathbf{M}_j z_j\rangle \geq b/20} \cdot |z_j + \langle \mathbf{M}_j, \xi\rangle|\right]$$

$$\leq \mathbb{E}\left[\mathbb{1}_{\langle v_i, x-\mathbf{M}_j z_j\rangle \geq b/20}\right] \cdot O\left(\mathbb{E}[|z_j|] + \frac{\log d}{\sqrt{d}}\sigma_x\right)$$

$$\leq O\left(\frac{\mathbb{E}\left[\langle v_i, x-\mathbf{M}_j z_j\rangle^2\right]}{b^2}\right) \cdot O\left(\frac{\sqrt{k}}{d} + \frac{\log d}{\sqrt{d}}\sigma_x\right)$$

$$\leq O\left(\frac{(r')^2}{db^2}\right) \cdot O\left(\frac{\sqrt{k}}{d} + \frac{\log d}{\sqrt{d}}\sigma_x\right)$$

and therefore

$$\sum_{j\in[d]}\left(\mathbb{E}\left[\mathbb{1}_{\langle v_i, x-\mathbf{M}_j z_j\rangle \geq b/20} \cdot |\langle x, \mathbf{M}_j\rangle|\right]\right)^2 \leq O\left(\frac{(r')^4}{d^2 b^4}\right) \cdot \left(\frac{k}{d} + \sigma_x^2 \log^2 d\right) \qquad \text{(F.9)}$$

- For the third term,

$$\mathbb{E}[\mathbb{1}_{\langle v_i, \mathbf{M}_j\rangle z_j \geq b/20} \cdot |\langle x, \mathbf{M}_j\rangle|] \leq \mathbb{E}\left[\mathbb{1}_{\langle v_i, \mathbf{M}_j\rangle z_j \geq b/20} \cdot \left(|z_i| + O(\frac{\log d}{\sqrt{d}})\right)\right]$$

$$\leq \mathbb{E}\left[\left(\frac{|\langle v_i, \mathbf{M}_j\rangle z_j|}{b} \cdot |z_i| + \frac{|\langle v_i, \mathbf{M}_j\rangle z_j|}{b} \cdot O(\frac{\log d}{\sqrt{d}})\right)\right]$$

$$\leq |\langle v_i, \mathbf{M}_j\rangle| \cdot O\left(\frac{1}{db} + \frac{\sqrt{k}\log d}{bd^{1.5}}\right) \leq |\langle v_i, \mathbf{M}_j\rangle| \cdot O\left(\frac{1}{db}\right)$$

and therefore

$$\sum_{j\in[d]}\left(\mathbb{E}[\mathbb{1}_{\langle v_i, \mathbf{M}_j\rangle z_j \geq b/20} \cdot |\langle x, \mathbf{M}_j\rangle|]\right)^2 \leq O\left(\frac{(r')^2}{d^2 b^2}\right) \qquad \text{(F.10)}$$

Putting (F.7), (F.8), (F.9), (F.10) these together, we have

$$\|\phi\|^2 \leq \sum_{j\in[d]} |\langle \phi, \mathbf{M}_j\rangle|^2 \leq O\left(\frac{k}{d^2} + \frac{(r')^4}{d^2 b^4}\left(\frac{k}{d} + \sigma_x^2 \log^2 d\right) + \frac{(r')^2}{d^2 b^2}\right)$$

Summing everything up, we have

$$\left\|\mathbb{E}_{x,y=y(x),\rho}\left[\ell \mathbb{1}_{\langle g_i+v_i, x+\delta\rangle + \rho_i \geq b}(x+\delta)\right]\right\|_2 \leq O\left(\left(\frac{k}{d} + \frac{(r')^2}{db^2}\right)\tau + \frac{\sqrt{k}}{d} + \frac{(r')^2}{db^2}\left(\frac{\sqrt{k}}{\sqrt{d}} + \sigma_x \log d\right) + \frac{r'}{db}\right)$$

Now, suppose we run robust training for $t = T_f, T_f+1, \ldots, T_f+T-1$ and suppose for all of them we have $\|v_i^{(T)}\|_2 \leq r'$ satisfied. Then, using the gradient update formula (see e.g. (C.19))

$$\|v_i^{(T_f+T)}\|_2 \leq \|v_i^{(T_f)}\|_2 + T\eta \cdot O\left(\left(\frac{k}{d} + \frac{(r')^2}{db^2}\right)\tau + \frac{\sqrt{k}}{d} + \frac{(r')^2}{db^2}\left(\frac{\sqrt{k}}{\sqrt{d}} + \sigma_x \log d\right) + \frac{r'}{db}\right)$$

This means, in order to show $\|v_i^{(T_f+T)}\|_2 \leq r'$ we can choose any $r' > 0$ satisfying

$$O(\frac{k\Xi_2^2}{\sqrt{d}}) + T\eta \cdot O\left(\left(\frac{k}{d} + \frac{(r')^2}{db^2}\right)\tau + \frac{\sqrt{k}}{d} + \frac{(r')^2}{db^2}\left(\frac{\sqrt{k}}{\sqrt{d}} + \sigma_x \log d\right) + \frac{r'}{db}\right) \leq r' \ .$$

Using the assumption of $T\eta \leq o(db)$ (which also implies $(T\eta)^2 \leq o(\frac{d^{2.5}b^2}{k})$), $\tau \leq \frac{1}{\sqrt{k}\log d}$ (which also implies $\tau \leq o(\frac{d^2 b^2}{(T\eta)^2 \sqrt{k} \log d}))$, and $\sigma_x \leq o(\frac{d^2 b^2}{(T\eta)^2 \sqrt{k} \log d})$, we can choose
$$r' \leq O(\frac{k\Xi_2^2}{\sqrt{d}}) + T\eta \cdot O\left(\frac{\sqrt{k}}{d}\right) \quad.$$

$\square$

### F.3.2 Growth Lemma for $\ell_\infty$ Perturbation

We now bound $\|v_i^{(t)}\|_1$ during $\ell_\infty$ robust training. Recall $\|\mathbf{M}\|_\infty \stackrel{\text{def}}{=} \max_{j \in [d]} \|\mathbf{M}_j\|_1$.

**Lemma F.16** (movement bound). *Suppose at iteration $t$, $\max_{i \in [m]} \|v_i\|_1 \leq r'$. Let $\ell \in [-1, 1]$ be any random variable that can depend on $x, \rho$, and $\delta \in \mathbb{R}^d$ be any random vector that can depend on $x$ with $\|\delta\|_\infty \leq \tau$. Then, for every $i \in [m]$,*
$$\| \mathbb{E}\left[\ell \mathbb{1}_{\langle g_i + v_i, x+\delta \rangle + \rho_i \geq b}(x + \delta)\right] \|_1 \leq O\left(\frac{k}{d} + \frac{(r')^2}{db^2}\right) \cdot (\tau d + \|\mathbf{M}\|_\infty \log d)$$

*As a corollary, suppose we run robust training from iteration $T_{\mathsf{f}}$ to $T_{\mathsf{f}} + T$ with $T\eta \leq \frac{db^2}{\|\mathbf{M}\|_\infty^2 k \Xi_2^3}$ and $\tau \leq o(\frac{b^2}{T\eta \cdot k \Xi_2^2 \|\mathbf{M}\|_\infty})$, then*
$$\max_{i \in [m]} \|v_i^{(T_{\mathsf{f}}+T)}\|_1 \leq O(k\Xi_2^2 \cdot \|\mathbf{M}\|_\infty)$$

*Proof of Lemma F.16.* Similar to the proof of Lemma F.15, and using $\|v_i\|_2 \leq \|v_i\|_1 \leq r'$, we have
$$\mathbf{Pr}[\langle g_i + v_i, x + \delta\rangle + \rho_i \geq b] \leq \frac{1}{\mathsf{poly}(d)} + \mathbf{Pr}[\langle g_i, x \rangle \geq |b|/10] + \mathbf{Pr}[\langle v_i, x \rangle \geq |b|/10]$$
$$\leq O\left(\frac{k}{d} + \frac{(r')^2}{db^2}\right) =: \kappa$$

This implies
$$\| \mathbb{E}\left[\ell \mathbb{1}_{\langle g_i + v_i, x+\delta \rangle + \rho_i \geq b} \delta\right] \|_1 \leq \kappa \cdot \tau d$$

On the other hand, let us look at $h := \mathbb{E}\left[\ell \mathbb{1}_{\langle g_i + v_i, x+\delta \rangle + \rho_i \geq b} x\right]$ and set $u = \mathsf{sign}(h) \in \{-1, 1\}^d$. We have
$$\|h\|_1 = \mathbb{E}\left[\ell \mathbb{1}_{\langle g_i + v_i, x+\delta \rangle + \rho_i \geq b} \langle x, u \rangle\right]$$

Since with probability at least $1 - e^{-\Omega(\log d)}$ it satisfies $|\langle x, u \rangle| = O(\max_{j \in [d]} \|\mathbf{M}_j\|_1 \log d)$, we can conclude that $\|h\|_1 = O(\kappa \|\mathbf{M}\|_\infty \log d)$. Together we have
$$\left\| \mathbb{E}\left[\ell \mathbb{1}_{\langle g_i + v_i, x+\delta \rangle + \rho_i \geq b}(x + \delta)\right] \right\|_1 \leq O\left(\frac{k}{d} + \frac{(r')^2}{db^2}\right) \cdot (\tau d + \|\mathbf{M}\|_\infty \log d)$$

Now, suppose we run robust training for $t = T_{\mathsf{f}}, T_{\mathsf{f}} + 1, \ldots, T_{\mathsf{f}} + T - 1$ and suppose for all of them we have $\|v_i^{(T)}\|_1 \leq r'$ satisfied. Then, using the gradient update formula (see e.g. (C.19))
$$\|v_i^{(T_{\mathsf{f}}+T)}\|_1 \leq \|v_i^{(T_{\mathsf{f}})}\|_1 + T\eta \cdot O\left(\frac{k}{d} + \frac{(r')^2}{db^2}\right) \cdot (\tau d + \|\mathbf{M}\|_\infty \log d)$$

Recalling $|\langle v_i^{(T_{\mathsf{f}})}, \mathbf{M}_j \rangle| \leq \frac{k\Xi_2^2}{d}$ from (F.8), we have
$$\left\|v_i^{(T_{\mathsf{f}})}\right\|_1 = \left\|\sum_{j \in [d]} \langle v_i^{(T_{\mathsf{f}})}, \mathbf{M}_j \rangle \cdot \mathbf{M}_j \right\|_1 \leq k\Xi_2^2 \cdot \|\mathbf{M}\|_\infty$$

This means, to prove that $\left\|v_i^{(T_f)}\right\|_1 \leq r'$, we can choose any $r'$ satisfying

$$k\Xi_2^2 \cdot \|\mathbf{M}\|_\infty + T\eta \cdot O\left(\frac{k}{d} + \frac{(r')^2}{db^2}\right) \cdot (\tau d + \|\mathbf{M}\|_\infty \log d) \leq r'$$

and using the assumption of $T\eta \leq \frac{db^2}{\|\mathbf{M}\|_\infty^2 k\Xi_2^3}$ (which implies $T\eta \leq O(d)$), and $\tau \leq o\left(\frac{b^2}{T\eta \cdot k\Xi_2^2\|\mathbf{M}\|_\infty}\right)$ (which implies $\tau \leq \frac{1}{\eta T}$), we can choose

$$r' \leq 2k\Xi_2^2 \cdot \|\mathbf{M}\|_\infty + T\eta \cdot O(\tau k) \leq O(k\Xi_2^2 \cdot \|\mathbf{M}\|_\infty) \ . \qquad \square$$

### F.4 Robust Convergence

We are now ready to prove the main convergence theorem (that is, Theorem F.1 and F.4) for robust learning. Let us first calculate a simple bound:

**Claim F.17.** $\left|\sum_{i\in[m]} \mathbf{Reg}(g_i) - \mathbf{Reg}(w_i^{(T_f)})\right| \leq O(k\sqrt{d}\Xi_2^4)$

*Proof.* Recalling $\|g_i\|_2 \leq O(\Xi_2^2)$ and $\|u_i\| \leq O(\frac{k\Xi_2^2}{\sqrt{d}})$ from Proposition F.8, we have

$$\left|\|g_i\|^3 - \|g_i + u_i\|^3\right| \leq O\left(\|u_i\| \cdot (\|u_i\|^2 + \|g_i\|^2)\right) \leq O(\frac{k\Xi_2^4}{\sqrt{d}}) \ . \qquad \square$$

#### F.4.1 Robust Convergence for $\ell_2$ Perturbation

*Proof of Theorem F.1.* Since $w_i^{(t+1)} = w_i^{(t)} - \eta \nabla_{w_i} \widetilde{\mathbf{RobObj}}_t(w^{(t)})$, we have the identity

$$\eta \langle \nabla \widetilde{\mathbf{RobObj}}_t(w^{(t)}), w^{(t)} - g \rangle = \frac{\eta^2}{2} \|\nabla \widetilde{\mathbf{RobObj}}_t(w^{(t)})\|_F^2 + \frac{1}{2}\|w^{(t)} - g\|_F^2 - \frac{1}{2}\|w^{(t+1)} - g\|_F^2$$

Applying (a variant of) Lemma A.2 (which requires us to use the Lipscthiz continuity assumption on $A$, see Definition 4.2), we know that by letting

$$\mathbf{RobObj}_t(w) = \mathop{\mathbb{E}}_{x,y=y(x),\delta,\rho} \left[\mathbf{Obj}_t(w; x+\delta, y, \rho)\right] \ ,$$

it satisfies

$$\eta \langle \nabla \mathbf{RobObj}_t(w^{(t)}), w^{(t)} - g \rangle \leq \eta^2 \cdot \mathsf{poly}(d) + \frac{1}{2}\|w^{(t)} - g\|_F^2 - \frac{1}{2}\|w^{(t+1)} - g\|_F^2 + \frac{\eta}{\mathsf{poly}(d)} \quad (\text{F.11})$$

Let us also define the clean objective and the pseudo objective as follows:

$$\mathbf{Obj}_t(w) = \mathop{\mathbb{E}}_{x,y=y(x),\rho} \left[\mathbf{Obj}_t(w; x, y, \rho)\right]$$

$$\mathbf{RobObj}'_t(\mu) = \mathop{\mathbb{E}}_{x,y=y(x),\delta,\rho} \left[\log(1 + e^{-y \cdot g_t(\mu; x+\delta, x, \rho)})\right] + \lambda \sum_{i\in[m]} \mathbf{Reg}(g_i + \mu_i) \ ,$$

which is a convex function in $\mu$ because $g_t(\mu; x+\delta, x, \rho)$ is linear in $\mu$.

Now, we inductively prove that at every iteration $t \in [T_f, T_f + T]$, it satisfies

$$\sum_{i\in[m]} \|v_i^{(t)}\|_2^2 \leq r^2 m \quad \text{for} \quad r = \Theta\left(\frac{k\Xi_2^2}{\sqrt{d}}\right) \quad (\text{F.12})$$

$$\max_{i\in[m]} \|v_i^{(t)}\|_2 \leq r' \quad \text{for} \quad r' = 1 \quad (\text{F.13})$$

In the base case $t = T_f$ this is obvious due to Proposition F.8. Next, suppose (F.12) and (F.13) hold at iteration $t$. Using the notation $w_i^{(t)} = g_i + v^{(t)}$ and the Lipscthiz continuity of $\log(1+e^t)$, we have[23]

$$\mathbb{E}[|\mathbf{RobObj}'_t(v^{(t)}) - \mathbf{RobObj}_t(w^{(t)})|] \leq \mathbb{E}\left[|g_t(v^{(t)}; x+\delta, x, \rho) - f_t(w^{(t)}, x+\delta)|\right]$$
$$\leq O(\tau^2) \cdot \left(\frac{\Xi_2^5}{\sigma_\rho} + \frac{k^{3.5}}{d^{1-2c_0}}\right) \quad \text{(using Lemma F.11)}$$
$$\leq O\left(\frac{1}{\log d}\right) \quad \text{(using } \tau \leq \tfrac{1}{\sqrt{k} \cdot d^{c_0}}\text{)}$$

$$\mathbb{E}[|\mathbf{RobObj}'_t(0) - \mathbf{Obj}_t(w^{(T_f)})|] \leq \mathbb{E}[|g_t(0; x+\delta, x) - f_t(w^{(T_f)}, x)|] + \lambda \left|\sum_{i \in [m]} \mathbf{Reg}(w_i^{(T_f)}) - \mathbf{Reg}(g_i)\right|$$
$$\leq O\left(\frac{k^{2.5}}{d^{1-2c_0}} + \tau\sqrt{k\Xi_2^7}\right) + O\left(\frac{k\Xi_2^4 \log d}{\sqrt{d}}\right)$$
$$\text{(using Lemma F.12 and Claim F.17)}$$
$$\leq O\left(\frac{1}{\log d} + \tau\sqrt{k\Xi_2^7}\right) \quad \text{(using } k^{2.5} < d^{1-2c_0}/\log d \text{)}$$
$$\leq O\left(\frac{1}{\log d}\right) \quad \text{(using } \tau \leq \tfrac{1}{\sqrt{k} \cdot d^{c_0}}\text{)}$$

Therefore, we can bound the left hand side of (F.11) as follows:
$$\langle \nabla \mathbf{RobObj}_t(w^{(t)}), w^{(t)} - g \rangle = \langle \nabla \mathbf{RobObj}'_t(v^{(t)}), v^{(t)} \rangle$$
$$\geq \mathbf{RobObj}'_t(v^{(t)}) - \mathbf{RobObj}'_t(0) \geq \mathbf{RobObj}_t(w^{(t)}) - \mathbf{Obj}_t(w^{(T_f)}) - O\left(\frac{1}{\log d}\right) .$$

Putting this back to (F.11) and telescoping for $t = T_f, T_f+1, \ldots, T_f+T_0-1$ for any $T_0 \leq T$, we have

$$\frac{1}{T_0} \sum_{t=T_f}^{T_f+T_0-1} \left(\mathbf{RobObj}_t(w^{(t)}) - \mathbf{Obj}_t(w^{(T_f)}) - O\left(\frac{1}{\log d}\right)\right)$$
$$\leq \frac{1}{2\eta T_0}\|w^{(T_f)} - g\|_F^2 - \frac{1}{2\eta T_0}\|w^{(T_f+T_0)} - g\|_F^2 \leq \frac{1}{\eta T_0} \cdot O\left(\frac{k^2 \Xi_2^4}{d}m\right) - \frac{1}{2\eta T_0}\|w^{(T_f+T_0)} - g\|_F^2 \tag{F.14}$$

Inequality (F.14) now implies that
$$\sum_{i \in [m]} \|v_i^{(T_f+T_0)}\|^2 = \|w^{(T_f+T_0)} - g\|_F^2 \leq O\left(\frac{k^2 \Xi_2^4}{d}m\right)$$

so (F.12) holds at iteration $t = T_f + T_0$. We can then also apply Lemma F.15 which ensures (F.13) holds at iteration $t = T_f + T_0$.

Finally, let us go back to (F.14) and choose $T_0 = T = \Theta(\frac{k^2 \Xi_2^4 m \log d}{\eta d})$. It implies

$$\frac{1}{T} \sum_{t=T_f}^{T_f+T-1} \left(\mathbf{RobObj}_t(w^{(t)}) - \mathbf{Obj}_t(w^{(T_f)}) - O\left(\frac{1}{\log d}\right)\right) \leq O(\frac{1}{\log d})$$

---
[23] Note to apply Lemma F.11 we also need to check $\tau \leq o(\frac{b}{\Xi_2^2+r'})$ but this is automatically satisfied under our parameter choice $\tau \leq \frac{1}{\sqrt{k} \cdot d^{c_0}}$.

Note that our final choice of $T$ also ensures that the pre-requisite $T\eta \leq o(db)$ and $\tau, \sigma_x \leq o(\frac{d^2b^2}{(T\eta)^2\sqrt{k}\log d})$ of Lemma F.15 hold. $\square$

### F.4.2 Robust Convergence for $\ell_\infty$ Perturbation

*Proof of Theorem F.4.* The proof is nearly identical to that of Theorem F.1. In particular, we want to inductively prove that at every iteration $t \in [T_{\mathsf{f}}, T_{\mathsf{f}} + T]$, it satisfies

$$\sum_{i \in [m]} \|v_i^{(t)}\|_2^2 \leq r^2 m \quad \text{for} \quad r = \Theta\left(\frac{k\Xi_2^2}{\sqrt{d}}\right) \tag{F.15}$$

$$\max_{i \in [m]} \|v_i^{(t)}\|_1 \leq r' \quad \text{for} \quad r' = \Theta(k\Xi_2^2 \cdot \|\mathbf{M}\|_\infty) \tag{F.16}$$

We also need to redo the following calculations:[24]

$$\mathbb{E}[|\mathbf{RobObj}'_t(v^{(t)}) - \mathbf{RobObj}_t(w^{(t)})|] \leq \mathbb{E}\left[|g_t(v^{(t)}; x+\delta, x, \rho) - f_t(w^{(t)}, x+\delta)|\right]$$

$$\leq O(\tau^2) \cdot k^{3.5} d^{c_0} \cdot \|\mathbf{M}\|_\infty^2 \quad \text{(using Lemma F.13)}$$

$$\leq O\left(\frac{1}{\log d}\right) \quad \text{(using } \tau \leq \frac{1}{k^{1.75} \cdot \|\mathbf{M}\|_\infty \cdot d^{c_0}}\text{)}$$

$$\mathbb{E}[|\mathbf{RobObj}'_t(0) - \mathbf{Obj}_t(w^{(T_{\mathsf{f}})})|] \leq \mathbb{E}[|g_t(0; x+\delta, x) - f_t(w^{(T_{\mathsf{f}})}, x)|] + \lambda \left|\sum_{i \in [m]} \mathbf{Reg}(w_i^{(T_{\mathsf{f}})}) - \mathbf{Reg}(g_i)\right|$$

$$\leq O\left(\frac{k^{2.5}}{d^{1-2c_0}} + \tau \cdot k\Xi_2^4\right) + O\left(\frac{k\Xi_2^4 \log d}{\sqrt{d}}\right)$$
(using Lemma F.14 and Claim F.17)

$$\leq O\left(\frac{1}{\log d} + \tau\sqrt{k\Xi_2^7}\right) \quad \text{(using } k^{2.5} < d^{1-2c_0}/\log d\text{)}$$

$$\leq O\left(\frac{1}{\log d}\right) \quad \text{(using } \tau \leq \frac{1}{k^{1.75} \cdot \|\mathbf{M}\|_\infty \cdot d^{c_0}}\text{)}$$

$\square$

## F.5 Fast Gradient Method (FGM) Robust Training

Let us prove Corollary F.2 only for the $\ell_2$ case, and the other $\ell_\infty$ case Corollary F.5 is completely analogous.

*Proof of Corollary F.2.* At any iteration $t \in [T_{\mathsf{f}}, T_{\mathsf{f}} + T_{\mathsf{g}}]$, consider any perturbation vector $\delta \in \mathbb{R}^d$ which may depend on $x$ but not on $\rho$, with $\|\delta\|_2 \leq \tau$.

Recall from Lemma F.11 that

$$\mathbb{E}_x\left[\left|\mathbb{E}_\rho g_t(v^{(t)}; x+\delta, x, \rho) - \mathbb{E}_\rho f_t(w^{(t)}; x+\delta, \rho)\right|\right] \leq O(\tau^2) \cdot \left(\frac{\Xi_2^5}{\sigma_\rho} + \frac{k^{3.5}}{d^{1-2c_0}}\right) \leq O\left(\frac{1}{\log^2 d}\right)$$

This means for at least $1 - O(\frac{1}{\log d})$ probability mass of inputs $x$, we have

$$\left|\mathbb{E}_\rho g_t(v^{(t)}; x+\delta, x, \rho) - \mathbb{E}_\rho f_t(w^{(t)}; x+\delta, \rho)\right| \leq O\left(\frac{1}{\log d}\right)$$

---
[24]Note to apply Lemma F.13 we also need to check $\tau \leq o(\frac{b}{\Xi_2^2 + r'})$ but this is automatically satisfied under our parameter choice for $\tau$.

For those choices of $x$, using the fact that $g_t$ is linear in $\delta$, we also have

$$\mathbb{E}_\rho g_t(v^{(t)}; x + \delta, x, \rho) - \mathbb{E}_\rho f_t(w^{(t)}; x, \rho)$$
$$= \mathbb{E}_\rho g_t(v^{(t)}; x + \delta, x, \rho) - \mathbb{E}_\rho g_t(v^{(t)}; x, x, \rho) = \langle \nabla_x \mathbb{E}_\rho g_t(v^{(t)}; x, x, \rho), \delta \rangle$$
$$= \langle \nabla_x \mathbb{E}_\rho f_t(w^{(t)}; x, \rho), \delta \rangle$$

Putting them together we have

$$\left[ -y(x) \mathbb{E}_\rho f_t(w^{(t)}; x + \delta, \rho) \right] = \left[ -y(x) \mathbb{E}_\rho f_t(w^{(t)}; x, \rho) - \langle y(x) \nabla_x \mathbb{E}_\rho f_t(w^{(t)}; x, \rho), \delta \rangle \right] \pm O\left( \frac{1}{\log d} \right)$$
(F.17)

$$\leq \left[ -y(x) \mathbb{E}_\rho f_t(w^{(t)}; x, \rho) - \langle y(x) \nabla_x \mathbb{E}_\rho f_t(w^{(t)}; x, \rho), \delta^\star \rangle \right] + O\left( \frac{1}{\log d} \right)$$
(F.18)

where $\delta^\star = A(f_t, x, y)$ is the perturbation obtained by the fast gradient method with $\ell_2$ radius $\tau$. This means two things.

On the other hand, by applying Markov's inequality and Jensen's inequality to $\mathbb{E}_{x,y=y(x),\rho} \left[ \mathbf{Obj}_t(w^{(t)}; x + \delta^\star, y, \rho) \right] \leq o(1)$, we know for at least $1 - o(1)$ probability mass of the choices of $x$, it satisfies

$$\log(1 + e^{-y(x) \mathbb{E}_\rho f_t(w^{(t)}; x + \delta^\star, \rho)}) \leq o(1)$$
$$\implies -y(x) \mathbb{E}_\rho f_t(w^{(t)}; x + \delta^\star, \rho) \leq -10$$

Therefore, for all of those $x$ (with total mass $\geq 1 - o(1)$) satisfying both, we can first apply (F.17) (with $\delta = \delta^\star$) to derive

$$-y(x) \mathbb{E}_\rho f_t(w^{(t)}; x, \rho) - \langle y(x) \nabla_x \mathbb{E}_\rho f_t(w^{(t)}; x, \rho), \delta^\star \rangle \leq -9$$

Applying (F.18) then we obtain (for any $\delta$)

$$-y(x) \mathbb{E}_\rho f_t(w^{(t)}; x + \delta, \rho) \leq -8$$

This means, the output of the network $f_t$ is robust at point $x$ against *any* perturbation $\delta$ with radius $\tau$. We finish the proof of Corollary F.2. $\square$

## G  NTK Lower Bound For $\ell_\infty$ Perturbation

Recall from Definition 5.5 that the feature mapping of the neural tangent kernel for our two-layer network $f$ is

$$\Phi(x) = \left( x \mathbb{E}_{\rho_i} \left( \mathbb{1}_{\langle w_i, x \rangle + \rho_i \geq b_i} - \mathbb{1}_{-\langle w_i, x \rangle + \rho_i \geq b_i} \right) \right)_{i=1}^m$$

Therefore, given weights $\{v_i\}_{i \in [m]}$, the NTK function $p(x)$ is given as

$$p(x) = \sum_{i \in [m]} \langle x, v_i \rangle \mathbb{E}_{\rho_i \sim \mathcal{N}(0, \sigma_{\rho_i}^2)} \left( \mathbb{1}_{\langle w_i, x \rangle + \rho_i \geq b_i} - \mathbb{1}_{-\langle w_i, x \rangle + \rho_i \geq b_i} \right)$$

To make our lower bound stronger, in this section, we consider the simplest input distribution with $\mathbf{M} = \mathbf{I}$ and $\sigma_x = 0$ (so $\xi \equiv 0$). Our main theorem is the following.

**Theorem G.1.** *Suppose $w_1, \ldots, w_m \in \mathbb{R}^d$ are i.i.d. sampled from $\mathcal{N}(0, \mathbf{I})$ with $m \leq d^C$ for some constant $C > 1$; and suppose $\rho_i \sim \mathcal{N}(0, \sigma_{\rho_i}^2)$ with $|\sigma_{\rho_i}| \leq d^{o(1)}$ and $|b_i| \leq d^{o(1)}$. Then, there exists constant $c_6 > 0$ so that, with probability at least $1 - e^{-\Omega(\log^2 d)}$, choosing $\tau = \frac{1}{d^{c_6}}$, then for any $k \in \left[d^{\frac{c_6}{100}}, d^{0.5 - \frac{c_6}{100}}\right]$ and sufficiently large $d$.*

$$\Pr_{x, y = y(x)}\left[\exists \delta \in \mathbb{R}^d, \|\delta\|_\infty \leq \tau : \mathsf{sign}(p(x + \delta)) \neq y\right] \geq \frac{1 - o(1)}{2} \ .$$

## G.1 Proof of Theorem G.1

We first note the following:

**Claim G.2.** *Suppose at point $z \in \mathbb{R}^d$, some function $p(z)$ gives the correct label $y(z) = \mathsf{sign}(\langle w^\star, z\rangle)$ against any $\ell_\infty$ perturbation of radius $\tau$. Then, letting $\zeta \sim \mathcal{N}(0, \frac{\tau^2}{\log^8 d} \mathbf{I}_{d \times d})$ be a random vector, and $\delta \in \mathbb{R}^d$ be any vector with $\|\delta\|_\infty \leq \tau/2$, it satisfies*

$$\mathbb{E}_\zeta \left[p(z + \delta + \zeta)\right] \cdot \mathsf{sign}(\langle w^\star, z\rangle) \geq -e^{-\Omega(\log^2 d)} \max_{i \in [m]}\{\|v_i\|_2\}_{i \in [m]}$$

*Proof of Claim G.2.* With probability at least $1 - e^{-\Omega(\log^2 d)}$ it satisfies $\|\zeta + \delta\|_\infty \leq \tau$. When this happens, we must have $p(z + \delta + \zeta) \cdot \mathsf{sign}(\langle w^\star, z\rangle) \geq 0$. □

Therefore, for the analysis purpose (by sacrificing $\ell_\infty$ norm radius from $\tau$ to $\tau/2$), we can imagine as if the input is *randomly perturbed* by $\zeta$. This serves for the purpose of smoothing the NTK function $p(\cdot)$, which originally has indicator functions in it so may be trickier to analyze.

Next, we define

$$M_W(x) \stackrel{\text{def}}{=} \max_{i \in [m]} |\langle w_i, x\rangle| \ .$$

One can carefully apply the Taylor expansion of the smoothed indicator function (using the randomness of $\zeta$), to derive the following claim. (Detailed proof in Section G.4.)

**Claim G.3.** *Consider any NTK function $p(x)$ with parameters $\|w_i\|_2 \geq \frac{\sqrt{d}}{2}$, $\|w_i\|_\infty \leq \log^2 d$, $\rho_i \sim \mathcal{N}(0, \sigma_{\rho_i}^2)$ with $|\sigma_{\rho_i}| \leq d^{o(1)}$ and $|b_i| \leq d^{o(1)}$. Suppose $\tau \in [\frac{1}{d^{1/5}}, 1]$, then there exists coefficients $\{c_{i,r}, c'_{i,r}, c''_i\}_{i \in [m], r \geq 0}$ with*
- *each $|c_{i,r}|, |c'_{i,r}| \leq O(1)$,*
- *each $|c'_{i,r}| \leq |c_{i,r}| \cdot O(d^{-0.1} r)$,*
- *each $|c_{i,r}| \geq \Omega\left(\frac{1}{d^2}\right)$ for every odd constant $r \geq 1$*
- *each $|c''_i| \leq O(d^{-1/4})$.*

*so that, for every $z$ with $\|z\|_1 \leq d^{1/4}$ and every $\delta$ with $\|\delta\|_\infty \leq \tau/2$ and $M_W(\delta) \leq \tau d^{1/4}$, we have:*

$$\mathbb{E}_{\zeta \sim \mathcal{N}(0, \frac{\tau^2}{\log^8 d}\mathbf{I}_{d \times d})} [p(z + \delta + \zeta)]$$

$$= \sum_{r \geq 0} \|v_i\|_2 \left(c''_i + \sum_{i \in [m]} \left(c_{i,r} \langle \frac{v_i}{\|v_i\|_2}, z + \delta\rangle + c'_{i,r} \langle \frac{w_i}{\|w_i\|_2}, z + \delta\rangle\right) \left(\frac{\langle w_i, z + \delta\rangle}{\tau \|w_i\|_2}\right)^r\right)$$

Using the above formula, we can write

$$\mathbb{E}_\zeta[p(x+\zeta)] = CONST + \sum_{r\geq 0} T_{r+1}(x^{\otimes r+1})$$

$$\text{for} \quad T_{r+1}(x^{\otimes r+1}) \stackrel{\text{def}}{=} \sum_{i\in[m]} \|v_i\|_2 \left\langle c_{i,r}\frac{v_i}{\|v_i\|_2} + c'_{i,r}\frac{w_i}{\|w_i\|_2}, x \right\rangle \left(\frac{\langle w_i, x\rangle}{\tau\|w_i\|_2}\right)^r$$

Using $|c_{i,r}| \geq \Omega\left(\frac{1}{d^2}\right)$ for odd constant $r \geq 1$, and $|c'_{i,r}| \leq O(d^{-0.1}) \cdot |c_{i,r}|$, by applying Lemma G.4,[25] we know that when $r = 3C+3$ (say wlog. $3C+3$ is odd),

$$\|T_{3C+4}\|_F = \Omega\left(\frac{1}{d^3}\max_{i\in[m]}\{\|v_i\|_2\}\right) \tag{G.1}$$

Also, for a parameter $q = \sqrt{d}$, let us apply Lemma G.5 to derive

$$\lambda_r \stackrel{\text{def}}{=} \max_{\delta\in\mathbb{R}^d: \|\delta\|_\infty\leq\tau, M_W(\delta)\leq\tau\sqrt{q}} |T_r(\delta^{\otimes r})| \geq \Omega\left(\frac{1}{(\tau)^r}\|T_r\|_F\right) \tag{G.2}$$

Let $R \geq 3C+3$ be a constant to be chosen later, $\lambda_{\max} = \max_{r<R}\{\lambda_r\}$, and let $\delta_{\max}$ be the choice of $\delta$ which maximizes the value of $\lambda_{\max}$.

Consider the high probability event that $M_W(z) = \widetilde{O}(1)$, then using $M_W(\delta_{\max}) \leq \tau\sqrt{q}$, we have

$$\left|\sum_{r\geq R} T_{r+1}((z+\delta_{\max})^{\otimes r+1})\right| = \left|\sum_{r\geq R}\sum_{i\in[m]} c_{i,r}\langle v_i, z+\delta_{\max}\rangle\left(\frac{\langle w_i, z+\delta_{\max}\rangle}{\tau\|w_i\|_2}\right)^r\right|$$

$$\leq d^4 m \max_{i\in[m]}\{\|v_i\|_2\}_{i\in[m]} \sum_{r\geq R}\left(\frac{\widetilde{O}(1)+\tau\sqrt{q}}{\tau\sqrt{d}}\right)^r$$

$$\leq d^5 m \max_{i\in[m]}\{\|v_i\|_2\}_{i\in[m]} \sum_{r\geq R}\left(\frac{O(1)}{d^{1/4}}\right)^r$$

When $R \geq 10000(C+1)$, we have

$$\left|\sum_{r\geq R} T_{r+1}((z+\delta_{\max})^{\otimes r+1})\right| \leq O\left(\frac{\max_{i\in[m]}\{\|v_i\|_2\}_{i\in[m]}}{d^{100C}}\right) \tag{G.3}$$

Next, for every $s \in [1/2, 1]$, let us define

$$q_{<R}(z, s) := CONST + \sum_{r<R} T_{r+1}((z+s\delta_{\max})^{\otimes r+1})$$

- On one hand, by applying Lemma G.5 twice for each $r$, we know for every set of vectors $z_1, \cdots, z_q$ with $\|z_i\|_\infty \leq 1$, $M_W(z_i) = \widetilde{O}(1)$ and $\text{supp}(z_i) \cap \text{supp}(z_j) = \varnothing$ for $i \neq j$, it satisfies

$$\sum_{j\in[q]} (|q_{<R}(z_j, s) - q_{<R}(0, s)| + |q_{<R}(-z_j, s) - q_{<R}(0, s)|) \leq \sum_{r<R}\widetilde{O}\left(\frac{\lambda_{\max}}{\tau^r}\right) \leq \widetilde{O}\left(\frac{\lambda_{\max}}{\tau^R}\right)$$

This means by Markov's inequality, for at least $(1 - \frac{1}{\log d})$ fraction of the indices $j \in [q]$, denoting them by $\Lambda \subseteq [q]$, it satisfies

$$|q_{<R}(z_j, s) - q_{<R}(0, s)| + |q_{<R}(-z_j, s) - q_{<R}(0, s)| \leq \frac{1}{q}\widetilde{O}\left(\frac{\lambda_{\max}}{\tau^R}\right)$$

---

[25]Specifically, one should substitute $\|v_i\|_2\left(c_{i,r}\frac{v_i}{\|v_i\|_2} + c'_{i,r}\frac{w_i}{\|w_i\|_2}\right)$ as the new $v_i$ when applying Lemma G.4.

- On the other hand, by Claim G.7, we know that there is an $s \in [1/2, 1]$ such that
$$|q_{<R}(0, s)| \geq \Omega(\lambda_{\max})$$
Without loss of generality, suppose $q_{<R}(0, s)$ is positive and $q_{<R}(0, s) \geq \Omega(\lambda_{\max})$.

Combining the two, when $\tau^{100000(C+1)} \geq \frac{1}{d}$, we derive that for those $j \in \Lambda$,
$$q_{<R}(z_j, s) \geq \Omega(\lambda_{\max}) \quad \text{and} \quad q_{<R}(-z_j, s) \geq \Omega(\lambda_{\max})$$

Thus, combining with (G.1), (G.2) and (G.3), we have for those $j \in \Lambda$,

$$\mathbb{E}_\zeta[p(z_j + s\delta_{\max} + \zeta)] = CONST + \sum_{r \geq 0} T_{r+1}((z_j + s\delta_{\max})^{\otimes r+1}) \quad \geq \Omega\left(\frac{1}{d^4} \max_{i \in [m]}\{\|v_i\|_2\}\right)$$

$$\mathbb{E}_\zeta[p(-z_j + s\delta_{\max} + \zeta)] = CONST + \sum_{r \geq 0} T_{r+1}((-z_j + s\delta_{\max})^{\otimes r+1}) \quad \geq \Omega\left(\frac{1}{d^4} \max_{i \in [m]}\{\|v_i\|_2\}\right)$$

but according to Claim G.2 (see end of this section), this means the NTK function $p(\cdot)$ outputs the wrong label either $z_j$ or for $-z_j$. Therefore, among those $2q$ data points $z_1, \ldots, z_q, -z_1, \ldots, -z_q$, at least $50\% \cdot (1 - o(1))$ of them must be wrong under $\ell_2$ perturbation with radius $\tau$.

Finally, recall when $z$ is generated from the data distribution, with high probability $z \in \mathbb{R}^d$ is $O(k)$-sparse. Therefore, we can divide (nearly) all possible choices of $z$ into $q = \sqrt{d} \ll O(d/k)$ groups, in a way that when we generate $z_1, \ldots, z_q$ from those groups, they have disjoint support and they together match the overall distribution. Using this argument one can prove that, for at least $50\% \cdot (1 - o(1))$ of the probability mass of $z$ from the data distribution, the prediction must be wrong under $\ell_2$ perturbation with radius $\tau$.

This finishes the proof of Theorem G.1 ∎

### G.2 Tensor Lower Bound

Next, for each degree-$r$ homogenous part of the polynomial expansion of Claim G.3, we can write it as a tensor and lower bound its Frobenius norm as follows.

**Lemma G.4.** *Suppose $w_1, \ldots, w_m \in \mathbb{R}^d$ are i.i.d. sampled from $\mathcal{N}(0, \mathbf{I})$ with $m \leq d^C$ for some constant $C > 0$. Let $v_1, \ldots, v_m \in \mathbb{R}^d$ be arbitrary vectors that can depend on the randomness of $w_1, \ldots, w_m$. Let us denote by $T_{r+1}$ the symmetric tensor $\mathbb{R}^{d \times (r+1)} \to \mathbb{R}$ such that*

$$T_{r+1}(x^{\otimes r+1}) = \sum_{i \in [m]} \langle v_i, x\rangle \left(\frac{\langle w_i, x\rangle}{\|w_i\|_2}\right)^r$$

*We have as long as $r \geq 3C$, then w.p. $\geq 1 - e^{-\Omega(\log^2 d)}$ over the randomness of $\{w_i\}_{i \in [m]}$, for every $\{v_i\}_{i \in [m]}$ we have*

$$\|T_{r+1}\|_F \geq \Omega\left(\frac{1}{\sqrt{d}} \max_{i \in [m]}\{\|v_i\|_2\}\right)$$

*Proof of Lemma G.4.* Consider any fixed $j \in [m]$, and some $\gamma \in [-1, 1]$ to be chosen later.

Let us define $x = \frac{w_j}{2\|w_j\|_2} + \gamma \frac{v_j}{2\|v_j\|_2}$ which satisfies $\|x\|_2 \leq 1$. We have

$$T_{r+1}(x^{\otimes r+1}) = \sum_{i \in [m]} \langle v_i, x\rangle \left(\frac{\langle w_i, x\rangle}{\|w_i\|_2}\right)^r$$

$$= \sum_{i \in [m] \setminus \{j\}} \langle v_i, x\rangle \left(\frac{\langle w_i, x\rangle}{\|w_i\|_2}\right)^r + \left(\frac{\gamma}{2}\|v_j\|_2 + \frac{\langle v_j, w_j\rangle}{2\|w_j\|_2}\right)\left(\frac{1}{2} + \gamma\frac{\langle w_j, v_j\rangle}{2\|v_j\|_2\|w_j\|_2}\right)^r$$

Note that for every $j \neq i$, with probability at least $1 - e^{-\Omega(\log^2 d)}$,

$$\left|\frac{\langle w_i, x\rangle}{\|w_i\|_2}\right| = \left|\frac{\langle w_i, w_j\rangle}{2\|w_i\|_2\|w_j\|_2} + \gamma\frac{\langle w_i, v_j\rangle}{2\|v_j\|_2\|w_i\|_2}\right| \leq O\left(\frac{\log d}{\sqrt{d}} + \gamma\right) .$$

This implies that as long as $|\gamma| \leq \frac{1}{\sqrt{d}}$,

$$|T_{r+1}(x^{\otimes r+1})| \geq \frac{1}{3r}\left|\frac{\gamma}{2}\|v_j\|_2 + \frac{\langle v_j, w_j\rangle}{2\|w_j\|_2}\right| - \left(\frac{O(\log d)}{\sqrt{d}}\right)^r m \cdot \max_{i\in[m]}\{\|v_i\|_2\}$$

Since the above lower bound holds for every $|\gamma| \leq \frac{1}{\sqrt{d}}$ and every $j \in [d]$, we immediately know

$$\max_{\|x\|_2 \leq 1}\{T_{r+1}(x^{\otimes r+1})\} \geq \Omega\left(\frac{1}{\sqrt{d}}\max_{i\in[m]}\{\|v_i\|_2\}\right)$$

This implies our bound on the Frobenius norm as well. □

### G.3 Tensor Perturbation

We present the following critical lemma, which serves as the major step to prove the non-robustness of Neural Tangent Kernel:

**Lemma G.5** (Tensor difference). *For every $m = \mathsf{poly}(d)$, every set of vectors $W = \{w_i\}_{i\in[m]}$ with each $\|w_i\|_2 = O(\sqrt{d})$, for every constant $r > 0$, every $q \in [0, d]$, every symmetric tensor $T$ of degree $r$: $\mathbb{R}^{d\times r} \to \mathbb{R}$, for every $\tau > 0$,*

1. *The following is true*

$$\lambda := \max_{\delta \in \mathbb{R}^d: \|\delta\|_\infty \leq \tau, M_W(\delta) \leq \tau\sqrt{q}} |T(\delta^{\otimes r})| \geq \Omega\left(\frac{1}{(\tau)^r}\|T\|_F\right)$$

2. *For every vectors $z_1, \cdots, z_q \in \mathbb{R}^d$ with $\|z_i\|_\infty \leq 1$, $M_W(z_i) = \widetilde{O}(1)$ and $\mathrm{supp}(z_i) \cap \mathrm{supp}(z_j) = \varnothing$ for $i \neq j$, for every $y$ such that $\|y\|_\infty \leq \tau$ and $M_W(y) \leq \tau\sqrt{q}$, the following holds:*

$$\sum_{i\in[q]}\left|T(y^{\otimes r}) - T((y + z_i)^{\otimes r})\right| \leq \widetilde{O}\left(\frac{\lambda}{\tau^r}\right)$$

*Proof of Lemma G.5.* For the first item, we can simply let $\delta \sim \mathcal{N}\left(0, \frac{\tau^2}{\log^2 d}\right)$. This choice of $\delta$ satisfies $\|\delta\|_\infty \leq \tau$ and $M_W(\delta) \leq \tau\sqrt{q}$ with high probability. Furthermore, by applying anti-concentration of Gaussian polynomials (see for instance [5, Lemma I.1]), we know with at least constant probability $|T(\delta^{\otimes r})| \geq \frac{\Omega(\|T\|_F)}{\tau^r}$. This proves the first item.

To see the second item, we first note by tensor $r$-linearity and symmetry,

$$\sum_{i\in[q]}|T(y^{\otimes r}) - T((y+z_i)^{\otimes r})| \leq \sum_{r'=1}^{r}\binom{r}{r'}\sum_{j\in[q]}|T(z_j^{\otimes r'}, y^{\otimes(r-r')})|$$

and therefore we only need to bound the terms on the right hand side for any fixed $r' \in [r]$.

Define random variable $\{\xi_{i,j}\}_{i\in[r'-1], j\in[q]}$ where each $\xi_{i,j}$ is i.i.d. uniformly at random chosen from $\{-\tau, \tau\}$. Consider arbitrary fixed values $\gamma_j \in \{-1, 1\}$ for $j \in [q]$. Let us define random variables $Z_1, Z_2, \cdots, Z_{r'} \in \mathbb{R}^d$ as:

$$\forall i \in [r'-1]: Z_i := \sum_{j\in[q]}\xi_{i,j}z_j, \qquad Z_{r'} := \sum_{j\in[q]}\gamma_j\left(\prod_{i\in[r'-1]}\xi_{i,j}\right)z_j$$

From these notions one can directly calculate that
$$\mathbb{E}_{\xi}[T(Z_1, Z_2, \cdots, Z_{r'}, y^{\otimes(r-r')})] = \tau^{r'} \sum_{j \in [q]} \gamma_j T(z_j^{\otimes r'}, y^{\otimes(r-r')})$$

On the other hand, we have $\|Z_i\|_\infty \leq \tau$ and moreover, using the randomness of $\xi_{i,j}$, we know w.h.p. $|M_W(Z_i)| = \widetilde{O}(\tau\sqrt{q})$ for every $i \in [q]$. Hence, by Claim G.8, we know that
$$|\mathbb{E}[T(Z_1, Z_2, \cdots, Z_{r'}, y^{\otimes(r-r')})]| = \widetilde{O}(\lambda)$$

Putting them together, we have $\left|\sum_{i \in [q]} \gamma_i T(z_i^{\otimes r'}, y^{\otimes(r-r')})\right| = \widetilde{O}\left(\frac{\lambda}{\tau^{r'}}\right)$, and since this holds for every $\gamma_i \in \{-1, 1\}$, we conclude that:
$$\sum_{i \in [q]} |T(z_i^{\otimes r'}, y^{\otimes(r-r')})| = \widetilde{O}\left(\frac{\lambda}{\tau^{r'}}\right)$$

Putting this back to the binomial expansion finishes the proof. □

### G.4 Smoothed ReLU Taylor Series: Proof of Claim G.3

We first note the following Taylor expansion formula for smoothed ReLU.

**Claim G.6** (smoothed ReLU). *Let $a \geq 0$ be any real and $\rho \sim \mathcal{N}(0, \sigma^2)$ for $\sigma \geq a$. Then, for every $x \in [-a, a]$,*
$$\mathbb{E}_{\rho}[\rho \mathbb{1}_{\rho+x \geq 0}] = \sigma \sum_{i=0}^{\infty} c_{2i} \left(\frac{x}{\sigma}\right)^{2i} \quad \text{and} \quad \mathbb{E}_{\rho}[\mathbb{1}_{\rho+x \geq 0}] = \frac{1}{2} + \sum_{i=0}^{\infty} c'_{2i+1} \left(\frac{x}{\sigma}\right)^{2i+1}$$
*where $|c_{2i}| = \Theta\left(\frac{1}{i!}\right), |c'_{2i+1}| = \Theta\left(\frac{1}{(i+1)!}\right)$*

*Proof of Claim G.6.* We can directly calculate that
$$\mathbb{E}[\rho \mathbb{1}_{\rho+x \geq 0}] = \frac{1}{\sqrt{2\pi}\sigma} \int_{\rho \geq -x} \rho e^{-\frac{\rho^2}{2\sigma^2}} d\rho = \frac{1}{\sqrt{2\pi}} e^{-\frac{x^2}{2\sigma^2}} \sigma$$

so using Taylor expansion of $e^{-\frac{x^2}{2\sigma^2}}$ we prove the first equation. As for the second equation, we have
$$\mathbb{E}[\mathbb{1}_{\rho+x \geq 0}] = \frac{1}{\sqrt{2\pi}\sigma} \int_{\rho \geq -x} e^{-\frac{\rho^2}{2\sigma^2}} d\rho$$

This implies that
$$\frac{d}{dx} \mathbb{E}[\mathbb{1}_{\rho+x \geq 0}] = -\frac{1}{\sqrt{2\pi}\sigma} e^{-\frac{x^2}{2\sigma^2}}$$

Using Taylor expansion and integrating once, we prove the second equation. □

We are now ready to prove Claim G.3.

*Proof of Claim G.3.* Specifically, for each $i \in [m]$, denoting by $x = z + \delta$, we wish to apply Claim G.6 to
$$\mathbb{E}_{\rho_i, \zeta} \langle x + \zeta, v_i \rangle \left(\mathbb{1}_{\langle w_i, x+\zeta \rangle + \rho_i \geq b_i} - \mathbb{1}_{-\langle w_i, x+\zeta \rangle + \rho_i \geq b_i}\right)$$
$$= \underbrace{\mathbb{E}_{\rho_i, \zeta} \langle x, v_i \rangle \left(\mathbb{1}_{\langle w_i, x+\zeta \rangle + \rho_i \geq b_i} - \mathbb{1}_{-\langle w_i, x+\zeta \rangle + \rho_i \geq b_i}\right)}_{\heartsuit} + \underbrace{\mathbb{E}_{\rho_i, \zeta} \langle \zeta, v_i \rangle \left(\mathbb{1}_{\langle w_i, x+\zeta \rangle + \rho_i \geq b_i} - \mathbb{1}_{-\langle w_i, x+\zeta \rangle + \rho_i \geq b_i}\right)}_{\diamond}$$

Note that $g \stackrel{\text{def}}{=} \langle w_i, \zeta \rangle + \rho_i \sim \mathcal{N}(0, \sigma^2)$ for $\sigma^2 = \frac{\tau^2}{\log^{16} d} \|w_i\|_2^2 + \sigma_{\rho_i}^2 \in [\frac{\tau^2}{\log^{16} d} \|w_i\|_2^2, \frac{2\tau^2}{\log^{16} d} \|w_i\|_2^2]$.

- We first deal with the $\heartsuit$ part. Using Claim G.6, we have

$$\mathbb{E}_{\rho_i, \zeta} \langle x, v_i \rangle \mathbb{1}_{\langle w_i, x+\zeta \rangle + \rho_i \geq b_i} = \langle x, v_i \rangle \mathbb{E}_g \mathbb{1}_{\langle w_i, x \rangle - b_i + g \geq 0} = \langle x, v_i \rangle \left( \frac{1}{2} + \sum_{r=0}^{\infty} c'_{2r+1} \left( \frac{\langle w_i, x \rangle - b_i}{\sigma} \right)^{2r+1} \right)$$

for $|c'_{2r+1}| = \Theta\left(\frac{1}{(r+1)!}\right)$. Similarly, we also have

$$-\mathbb{E}_{\rho_i, \zeta} \langle x, v_i \rangle \mathbb{1}_{-\langle w_i, x+\zeta \rangle + \rho_i \geq b_i} = \langle x, v_i \rangle \left( -\frac{1}{2} - \sum_{r=0}^{\infty} c'_{2r+1} \left( \frac{-\langle w_i, x \rangle - b_i}{\sigma} \right)^{2r+1} \right)$$

Putting them together, and using the fact that $b_i \ll d^{-0.2} \ll \sigma$, we can write

$$\heartsuit = \mathbb{E}_{\rho_i, \zeta} \langle x, v_i \rangle \left[ \mathbb{1}_{\langle w_i, x+\zeta \rangle + \rho_i \geq b_i} - \mathbb{1}_{-\langle w_i, x+\zeta \rangle + \rho_i \geq b_i} \right]$$

$$= \langle x, v_i \rangle \left( \sum_{r=0}^{\infty} c'_{2r+1} \left( \frac{\langle w_i, x \rangle - b_i}{\sigma} \right)^{2r+1} - c'_{2r+1} \left( \frac{-\langle w_i, x \rangle - b_i}{\sigma} \right)^{2r+1} \right)$$

$$= \langle x, v_i \rangle \sum_{r \geq 0} c''_r \left( \frac{\langle w_i, x \rangle}{\sigma} \right)^r$$

for $|c''_{2r}| \leq O\left(\frac{1}{(r)!}\right)$ for every $r \geq 0$ and $|c''_{2r+1}| \geq \Omega\left(\frac{1}{(r+1)!}\right)$.

- Let us now focus on the $\diamondsuit$ part. Let $v_i^{\parallel}$ be the part of $v_i$ that is parallel to $w_i$. Then obviously we have

$$\mathbb{E}_{\zeta, \rho_i} \langle \zeta, v_i \rangle \mathbb{1}_{\langle w_i, x+\zeta \rangle + \rho_i \geq b_i} = \mathbb{E}_{\zeta, \rho_i} \langle \zeta, v_i^{\parallel} \rangle \mathbb{1}_{\langle w_i, x+\zeta \rangle + \rho_i \geq b_i} = \frac{\|v_i^{\parallel}\|_2}{\|w_i\|_2} \mathbb{E}_{\zeta, \rho_i} \langle \zeta, w_i \rangle \mathbb{1}_{\langle w_i, x+\zeta \rangle + \rho_i \geq b_i}$$

$$\stackrel{①}{=} \frac{\|v_i^{\parallel}\|_2}{\|w_i\|_2} \mathbb{E}_{\zeta, \rho_i} (\langle \zeta, w_i \rangle + \rho_i) \mathbb{1}_{\langle w_i, x+\zeta \rangle + \rho_i \geq b_i} \pm \|v_i^{\parallel}\|_2 \cdot O(d^{-1/4})$$

Above, the last ① is due to $\sigma_{\rho_i} \leq d^{o(1)}$ and $\|w_i\|_2 \geq \Omega(\sqrt{d})$.

Next, we again treat $g \stackrel{\text{def}}{=} \langle \zeta, w_i \rangle + \rho_i \sim \mathcal{N}(0, \sigma^2)$ and apply Claim G.6. We have

$$\mathbb{E}_{\rho_i, \zeta} (\langle \zeta, w_i \rangle + \rho_i) \mathbb{1}_{\langle w_i, x+\zeta \rangle + \rho_i \geq b_i} = \mathbb{E}_g g \mathbb{1}_{\langle w_i, x \rangle - b_i + g \geq 0} = \sigma \sum_{r=0}^{\infty} c_{2r} \left( \frac{\langle w_i, x \rangle - b_i}{\sigma} \right)^{2r}$$

for $|c_{2r}| = \Theta\left(\frac{1}{r!}\right)$. Putting them together, and doing the same thing for the symmetric part, we have

$$\diamondsuit = \mathbb{E}_{\rho_i, \zeta} \langle \zeta, v_i \rangle \left( \mathbb{1}_{\langle w_i, x+\zeta \rangle + \rho_i \geq b_i} - \mathbb{1}_{-\langle w_i, x+\zeta \rangle + \rho_i \geq b_i} \right)$$

$$= \frac{\|v_i^{\parallel}\|_2 \sigma}{\|w_i\|_2} \sum_{r=0}^{\infty} \left( c_{2r} \left( \frac{\langle w_i, x \rangle - b_i}{\sigma} \right)^{2r} - c_{2r} \left( \frac{-\langle w_i, x \rangle - b_i}{\sigma} \right)^{2r} \right) \pm \|v_i^{\parallel}\|_2 \cdot O(d^{-1/4})$$

$$\stackrel{①}{=} \frac{\|v_i^{\parallel}\|_2 \sigma}{\|w_i\|_2} \sum_{r=1}^{\infty} c'''_r \left( \frac{\langle w_i, x \rangle}{\sigma} \right)^r \pm \|v_i^{\parallel}\|_2 \cdot O(d^{-1/4})$$

$$= \|v_i^{\parallel}\|_2 \langle \frac{w_i}{\|w_i\|_2}, x \rangle \sum_{r=0}^{\infty} c'''_{r+1} \left( \frac{\langle w_i, x \rangle}{\sigma} \right)^r \pm \|v_i^{\parallel}\|_2 \cdot O(d^{-1/4})$$

Above, using the property of $b_i \ll d^{-0.2} \ll \sigma$, equation ① holds for some $|c'''_{2r}| \leq O(\frac{d^{-0.1}}{r!})$ and $|c'''_{2r+1}| \leq O(\frac{d^{-0.1}}{(r+1)!})$.

Finally, putting the bounds for ♡ and ◇ together, and using $\frac{\tau\|w_i\|_2}{\log^8 d} \leq \sigma \leq \tau\|w_i\|_2$, we derive that

$$\heartsuit + \diamondsuit = \|v_i\|_2 \cdot \sum_{r \geq 0} \left( c''''_{i,r} \langle x, \frac{v_i}{\|v_i\|_2} \rangle + c'''''_{i,r} \langle x, \frac{w_i}{\|w_i\|_2} \rangle \right) \left( \frac{\langle w_i, x \rangle}{\tau\|w_i\|_2} \right)^r \pm \|v_i\|_2 \cdot O(d^{-1/4})$$

for $|c''''_{i,r}| \leq O(1)$ for every $r \geq 0$, $c'''''_r \leq O(d^{-0.1}r) \cdot c''''_r$ for every $r \geq 0$, and $|c''''_{i,r}| \geq \Omega\left(\frac{1}{d^2}\right)$ for every odd constant $r \geq 1$. This finishes the proof of Claim G.3. □

### G.5 Simple Lemmas

We have the following claim relating polynomial value with its coefficients:

**Claim G.7** (low degree polynomial). *Let $p: \mathbb{R} \to \mathbb{R}$ be a constant-degree polynomial $p(x) = \sum_{r=0}^{R} c_r x^r$, then there exists $x \in [1/2, 1]$ such that*

$$|p(x)| \geq \Omega\left( \max_{r=0,1,2,\cdots,R} |c_r| \right) .$$

*Proof of Claim G.7.* Let us define $q(x) \stackrel{\text{def}}{=} p\left(x + \frac{1}{2}\right)$ and write accordingly $q(x) = \sum_{r=0}^{R} c'_r x^r$. Using the identity formula $\sum_{r=0}^{R} c'_r x^r = \sum_{r=0}^{R} c_r (x + 0.5)^r$ we can derive (recalling $R$ is constant)

$$\max_{r=0,1,2,\cdots,R} |c'_r| \leq O\left( \max_{r=0,1,2,\cdots,R} |c_r| \right)$$

Conversely, by writing $\sum_{r=0}^{R} c'_r (x - 0.5)^r = \sum_{r=0}^{R} c_r x^r$, we also have the other direction and therefore

$$\max_{r=0,1,2,\cdots,R} |c'_r| = \Theta\left( \max_{r=0,1,2,\cdots,R} |c_r| \right)$$

Now, notice that $\left|\frac{d^r}{dx^r} q(x) |_{x=0}\right| = \Theta(|c'_r|)$, so we can apply Markov brother's inequality to derive that

$$\max_{x \in [0, 1/2]} |q(x)| \geq \Omega\left( \max_{r=0,1,2,\cdots,R} |c'_r| \right) .$$

This finishes the proof. □

Using this Claim, we also have the following claim about symmetric tensor:

**Claim G.8** (symmetric tensor norms). *For every constant $r > 0$, every fixed $a_1, a_2 > 0$, every symmetric tensor $T$ of degree $r$ of the form $T: \mathbb{R}^{d \times r} \to \mathbb{R}$, let*

$$\lambda_1 := \max_{x: \|x\|_\infty \leq a_1, M_W(x) \leq a_2} \{|T(x^{\otimes r})|\}, \quad \lambda_2 := \max_{\{x_i\}_{i \in [r]}: \|x_i\|_\infty \leq a_1, M_W(x_i) \leq a_2} \{|T(x_1, x_2, \cdots, x_r)|\}$$

*then we have:*

$$\lambda_1 \leq \lambda_2 \leq O(\lambda_1)$$

*Proof of Claim G.8.* $\lambda_1 \leq \lambda_2$ is obvious so let us prove the other direction. Define polynomial

$$p(s) = T\left( \left( x_1 + s^{r+1} x_2 + s^{(r+1)^2} x_3 + \cdots + s^{(r+1)^{r-1}} x_r \right)^{\otimes r} \right)$$

The coefficient of $p(s)$ at degree $\sum_{r' \in [r]} (r+1)^{r'-1}$ is $\Theta(T(x_1, x_2, \cdots, x_r))$. Thus, applying Claim G.7 and appropriately scaling the operator, we complete the proof. □

# H  Appendix for Probability Theory

## H.1  Small ball probability: The basic property

We also have the following property:

**Lemma H.1** (small ball probability, 1-d case).

*(a) For every subset $\Lambda \subseteq [d]$, every $r$, and every $t > 0$,*

$$\mathbf{Pr}\left[|\sum_{j \in \Lambda} w_j^\star \cdot z_j - r| \leq t\right] \leq O(\frac{t}{\sqrt{|\Lambda|/d}} + \frac{1}{\sqrt{|\Lambda|k/d}})$$

*(b) For every subset $\Lambda \subseteq [d]$ with $|\Lambda| \geq \Omega(d)$, and every $t > 0$,*

$$\mathbf{Pr}\left[|\sum_{j \in \Lambda} w_j^\star \cdot z_j| \leq t\right] \geq \Omega(t) - O\left(\frac{\log k}{\sqrt{k}}\right)$$

*Proof of Lemma H.1.* Recall we have $\mathbf{Pr}_{z_j}[z_j \neq 0] \geq \Omega(\frac{k}{d})$ for each $j \in \Lambda$. Let $\Lambda' \subseteq \Lambda$ be the subset of such indices $j$ with non-zero $z_j$, so by our assumption we have $|z_j| \geq \frac{1}{\sqrt{k}}$ for each $j \in \Lambda'$. By Chernoff bound, with probability at least $1 - e^{-\Omega(|\Lambda|k/d)}$, we know $|\Lambda'| \geq \Omega(\frac{k}{d}) \cdot |\Lambda|$.

Conditioning on such $\Lambda'$, by the Littlewood-Offord problem (a.k.a. small ball probability theorem, or anti-concentration for sum of Bernoulli variables, see [30]), we know

$$\mathbf{Pr}\left[|\sum_{j \in \Lambda} w_j^\star \cdot z_j - r| \leq t \,\big|\, \mathcal{E}\right] \leq O(\frac{\sqrt{k}t + 1}{\sqrt{|\Lambda'|}}) = O(\frac{\sqrt{k}t + 1}{\sqrt{|\Lambda|k/d}})$$

As for the lower bound, let us denote by $\Lambda'' \subseteq \Lambda$ be the subset of such indices $j$ with non-zero $z_j$ and $|z_j| \leq O(\frac{1}{\sqrt{k}})$. We know with high probability $|\Lambda''| \geq \Omega(k)$. Using $\mathbb{E}\left[|\sum_{j \in \Lambda \setminus \Lambda''} w_j^\star \cdot z_j|\right] \leq O(1)$, we can apply Markov's inequality and get

$$\mathbf{Pr}\left[|\sum_{j \in \Lambda \setminus \Lambda''} w_j^\star \cdot z_j| \leq B\right] \geq 0.6 \quad \text{for some constant } B = O(1)$$

Now, for the sum over $\Lambda''$, we can apply a Wasserstein distance version of the central limit theorem (that can be derived from [110], full statement see [6, Appendix A.2]) to derive that, for a Gaussian variable $g \sim (0, V^2)$ where $V = \sum_{j \in \Lambda''} (w_j^\star)^2 \mathbb{E}[(z_j)^2] \geq \Omega(1)$, the Wasserstein distance:

$$\mathcal{W}_2\left(\sum_{j \in \Lambda''} w_j^\star \cdot z_j, g\right) \leq O\left(\frac{\log k}{\sqrt{k}}\right)$$

Using the property of Gaussian variables and $B = O(1)$, we have

$$\mathbf{Pr}\left[g \in \left[\sum_{j \in \Lambda \setminus \Lambda''} w_j^\star \cdot z_j - \frac{t}{2}, \sum_{j \in \Lambda \setminus \Lambda''} w_j^\star \cdot z_j + \frac{t}{2}\right]\right] \geq \Omega(t)$$

and using the above Wasserstein distance bound, we have

$$\mathbf{Pr}\left[\sum_{j \in \Lambda''} w_j^\star \cdot z_j \in \left[\sum_{j \in \Lambda \setminus \Lambda''} w_j^\star \cdot z_j - t, \sum_{j \in \Lambda \setminus \Lambda''} w_j^\star \cdot z_j + t\right]\right] \geq \Omega(t) - O\left(\frac{\log k}{\sqrt{k}}\right)$$

$\square$

## H.2 McDiarmid's Inequality and An Extension

We state the standard McDiarmid's inequality,

**Lemma H.2** (McDiarmid's inequality). *Consider independent random variables $x_1, \cdots, x_n \in \mathcal{X}$ and a mapping $f : \mathcal{X}^n \to \mathsf{R}$. If for all $i \in [n]$ and for all $y_1, \cdots, y_n, y_i' \in \mathcal{X}$, the function $f$ satisfies*

$$|f(y_1, \cdots, y_{i-1}, y_i, y_{i+1}, \cdots, y_n) - f(y_1, \cdots, y_{i-1}, y_i', y_{i+1}, \cdots, y_n)| \leq c_i.$$

*Then*

$$\mathbf{Pr}[f(x_1, \cdots, x_n) - \mathbb{E}\, f \geq t] \geq \exp(\frac{-2t^2}{\sum_{i=1}^n c_i^2}),$$

$$\mathbf{Pr}[f(x_1, \cdots, x_n) - \mathbb{E}\, f \leq -t] \geq \exp(\frac{2t^2}{\sum_{i=1}^n c_i^2}).$$

We prove a more general version of McDiarmid's inequality,

**Lemma H.3** (McDiarmid extension). *Let $w_1, \ldots, w_N$ be independent random variables and $f \colon (w_1, \ldots, w_N) \mapsto [0, B]$. Suppose it satisfies for every $k \in \{2, 3, \ldots, N\}$,*

- *with probability at least $1 - p$ over $w_1, \ldots, w_N$, it satisfies*

$$\forall w_k'' \colon \left| f(w_{-k}, w_k) - f(w_{-k}, w_k'') \right| \leq c$$

- *with probability at least $1 - p$ over $w_1, \ldots, w_{k-1}, w_{k+1}, \ldots, w_N$, it satisfies*

$$\mathop{\mathbb{E}}_{w_k', w_k''}\left[ (f(w_{-k}, w_k') - f(w_{-k}, w_k''))^2 \right] \leq V_k^2$$

*Then,*

$$\mathop{\mathbf{Pr}}_{w_1, \ldots, w_N}\left[ \left| f(w_1, \ldots, w_N) - \mathop{\mathbb{E}}_{w_2, \ldots, w_N}[f(w_1, \ldots, w_N) \mid w_1] \right| \geq t \right]$$

$$\leq O(N\sqrt{p}) + \exp\left( \frac{-\Omega(t^2)}{t(c + \sqrt{p}B) + \sum_{t=2}^N (V_t^2 + \sqrt{p}B^2)^2} \right).$$

*Proof of Lemma H.3.* For each $t = 1, \ldots, N-1$, we have with probability at least $1 - \sqrt{p}$ over $w_1, \ldots, w_t$, it satisfies

$$\mathop{\mathbf{Pr}}_{w_{t+1}, \ldots, w_N}\left[ \forall w_{t+1}'' \colon \left| f(w_{\leq t}, w_{t+1}, w_{>t+1}) - f(w_{\leq t}, w_{t+1}'', w_{>t+1}) \right| \leq c \right] \geq 1 - \sqrt{p}\ .$$

We also have with probability at least $1 - \sqrt{p}$ over $w_1, \ldots, w_t$, it satisfies

$$\mathop{\mathbf{Pr}}_{w_{t+2}, \ldots, w_N}\left[ \mathop{\mathbb{E}}_{w_{t+1}', w_{t+1}''} \left( f(w_{\leq t}, w_{t+1}', w_{>t+1}) - f(w_{\leq t}, w_{t+1}'', w_{>t+1}) \right)^2 \leq V_{t+1}^2 \right] \geq 1 - \sqrt{p}\ .$$

We denote by $K_{\leq t}$ the event (over $w_{\leq t} = (w_1, \ldots, w_t)$) that the above two statements hold. We know that $\mathbf{Pr}[w_{\leq t} \in K_{\leq t}] \geq 1 - 2\sqrt{p}$. For notational simplicity, we denote by $K_{\leq N}$ the full set over all possible $(w_1, \ldots, w_N)$.

Define random variable $X_t$ (which depends only on $w_1, \ldots, w_t$) as

$$X_t := \mathop{\mathbb{E}}_{w_{>t}}[f(\vec{w}) \mid w_{\leq t}]\, \mathbb{1}_{(w_{\leq 1}, \ldots, w_{\leq t}) \in K_{\leq 1} \times \cdots \times K_{\leq t}} \in [0, B]$$

For every $t$ and fixed $w_1, \ldots, w_{t-1}$.

- If $(w_{\leq 1}, \ldots, w_{<t}) \notin K_{\leq 1} \times \cdots \times K_{<t}$, then $X_t = X_{t-1} = 0$.

- If $(w_{\leq 1}, \ldots, w_{<t}) \in K_{\leq 1} \times \cdots \times K_{<t}$,

  - If $w_{\leq t} \notin K_{\leq t}$, then $X_t - X_{t-1} = 0 - X_{t-1} \leq 0$.
  - If $w_{\leq t} \in K_{\leq t}$, then
  $$X_t - X_{t-1} = \mathbb{E}_{w_{>t}}[f(w_{<t}, w_t, w_{>t}) \mid w_{\leq t}] - \mathbb{E}_{w_{\geq t}}[f(w_{<t}, w_t, w_{>t}) \mid w_{<t}]$$
  Recall the property $w_{<t} \in K_{<t}$, we know with probability at least $1 - \sqrt{p}$ over $w_t$ and $w_{>t}$, it satisfies
  $$\forall w_t'': \left| f(w_{<t}, w_t'', w_{>t}) - f(w_{<t}, w_t, w_{>t}) \right| \leq c$$
  Taking expectation over $w_t$ and $w_{>t}$, we have
  $$\forall w_t'': \mathbb{E}_{w_{>t}}[f(w_{<t}, w_t'', w_{>t})] - \mathbb{E}_{w_{\geq t}}[f(w_{<t}, w_t, w_{>t})] \leq (c + \sqrt{p}B)$$
  This precisely means $X_t - X_{t-1} \leq c + \sqrt{p}B$.

  - Using the property $w_{<t} \in K_{<t}$, we know with probability at least $1 - \sqrt{p}$ over $w_{>t}$, it satisfies
  $$\mathbb{E}_{w_t, w_t''} \left( f(w_{<t}, w_t, w_{>t}) - f(w_{<t}, w_t'', w_{>t}) \right)^2 \leq V_t^2$$
  Taking expectation also over $w_{>t}$, we have
  $$\mathbb{E}_{w_t, w_t'', w_{>t}} \left( f(w_{<t}, w_t, w_{>t}) - f(w_{<t}, w_t'', w_{>t}) \right)^2 \leq V_t^2 + \sqrt{p}B^2$$
  $$\implies \mathbb{E}_{w_t} \left( \mathbb{E}_{w_{>t}}[f(w_{<t}, w_t, w_{>t})] - \mathbb{E}_{w_t'', w_{>t}}[f(w_{<t}, w_t'', w_{>t})] \right)^2 \leq V_t^2 + \sqrt{p}B^2$$
  Now observe that, since $(w_{\leq 1}, \ldots, w_{<t}) \in K_{\leq 1} \times \cdots \times K_{<t}$, we have $X_{t-1} = \mathbb{E}_{w_t'', w_{>t}}[f(w_{<t}, w_t'', w_{>t})]$. We also have that as long as $w_{\leq t} \in K_{\leq t}$, then $X_t = \mathbb{E}_{w_{>t}}[f(w_{<t}, w_t, w_{>t})]$. Putting these together, and using the fact $\mathbf{Pr}[w_{\leq t} \in K_{\leq t}] \geq 1 - 2\sqrt{p}$, we have
  $$\mathbb{E}_{w_k}\left[(X_{t+1} - X_t)^2 \mid w_{<t}\right] \leq V_t^2 + 3\sqrt{p}B^2$$

In sum, we have just shown that for all choices of $w_1, \ldots, w_{t-1}$,
$$X_t - X_{t-1} \leq (c + \sqrt{p}B) \quad \text{and} \quad \mathbb{E}_{w_k}\left[(X_{t+1} - X_t)^2 \mid w_{<t}\right] \leq V_t^2 + 3\sqrt{p}B^2$$
always holds. Note in addition we also have $\mathbb{E}_{w_t}[X_t \mid w_{<t}] \leq X_{t-1}$. Therefore, by applying martingale concentration (with its one-sided and Bernstein form, see Lemma H.4),
$$\mathbf{Pr}[X_N - X_1 > t] \leq \exp\left( \frac{-\Omega(t^2)}{t(c + \sqrt{p}B) + \sum_{t=2}^{N}(V_t^2 + \sqrt{p}B^2)^2} \right)$$
Recalling
$$X_N := f(\vec{w}) \mathbb{1}_{(w_{\leq 1}, \ldots, w_{\leq t}) \in K_{\leq 1} \times \cdots \times K_{\leq t}}$$
and we have $X_N = f(w_1, \ldots, w_N)$ with probability at least $1 - 2N\sqrt{p}$ (and $X_N = 0$ with the remaining probability). Also recalling
$$X_1 := \mathbb{E}_{w_2, \ldots, w_N}[f(\vec{w}) \mid w_1] \mathbb{1}_{w_{\leq 1}}$$
and we have $X_1 = \mathbb{E}_{w_2, \ldots, w_N}[f(\vec{w}) \mid w_1]$ with probability at least $1 - 2\sqrt{p}$ (and $X_1 = 0$ with the remaining probability).

Together, we have the desired theorem.

$\square$

Let us state, for completeness' sake, a simple one-sided Bernstein form of martingale concentration (that we do not know a good reference to it).

**Lemma H.4.** *Suppose we have a submartingale sequence $X_0, X_1, \ldots, X_N$, satisfying:*

- $X_0 = 0$ and $\mathbb{E}[X_t \mid X_{t-1}] \leq X_t$,
- $X_t - X_{t-1} \leq c$ *always holds, and*
- $\mathbb{E}_{X_t}[(X_t - X_{t-1})^2 \mid X_{t-1}] \leq V_t^2$ *always holds.*

*Then,*
$$\mathbf{Pr}[X_N > t] \leq e^{-\Omega(\frac{t^2}{tc + \sum_t V_t^2})}$$

*Proof.* Define potential function $\Psi_t = e^{\frac{\eta}{2c} X_t}$ for some $\eta \in (0, 1)$ to be chosen later. We have
$$\Psi_t = \Psi_{t-1} \cdot e^{\frac{\eta}{2c}(X_t - X_{t-1})} \leq \Psi_{t-1} \cdot \left(1 + \left(\frac{\eta(X_t - X_{t-1})}{2c}\right) + \left(\frac{\eta(X_t - X_{t-1})}{2c}\right)^2\right)$$

where the inequality is due to $e^y \leq 1 + y + y^2$ which holds for all $-\infty < y \leq 0.5$. Taking conditional expectation, we have
$$\mathbb{E}[\Psi_t \mid X_{t-1}] \leq \Psi_{t-1} \cdot \left(1 + \eta \mathbb{E}\left[\frac{X_t - X_{t-1}}{2c} \mid X_{t-1}\right] + \eta^2 \mathbb{E}\left[\left(\frac{X_t - X_{t-1}}{2c}\right)^2 \mid X_{t-1}\right]\right)$$
$$\leq \Psi_{t-1} \cdot \left(1 + \eta^2 \frac{V_t^2}{4c^2}\right) \leq \Psi_{t-1} \cdot e^{\eta^2 \frac{V_t^2}{4c^2}}.$$

After telescoping, we have $\mathbb{E}[\Psi_N] \leq e^{\eta^2 \frac{\sum_t V_t^2}{4c^2}}$, and therefore
$$\mathbf{Pr}[X_N > t] \leq \frac{\mathbb{E}[e^{\eta X_N/(2c)}]}{e^{\eta t/(2c)}} \leq e^{\eta^2 \frac{\sum_t V_t^2}{4c^2} - \eta \frac{t}{2c}}$$

Choosing the optimal $\eta \in (0, 1)$ gives us bound
$$\mathbf{Pr}[X_N > t] \leq e^{-\Omega(\frac{t^2}{tc + \sum_t V_t^2})}$$
□

[7] Zeyuan Allen-Zhu, Yuanzhi Li, and Zhao Song. On the convergence rate of training recurrent neural networks. In *NeurIPS*, 2019. Full version available at http://arxiv.org/abs/1810.12065.

[8] Zeyuan Allen-Zhu, Yuanzhi Li, and Zhao Song. A convergence theory for deep learning via overparameterization. In *ICML*, 2019. Full version available at http://arxiv.org/abs/1811.03962.

[9] Sanjeev Arora, Rong Ge, and Ankur Moitra. New algorithms for learning incoherent and overcomplete dictionaries. In *Conference on Learning Theory*, pages 779–806, 2014.

[10] Sanjeev Arora, Rong Ge, Tengyu Ma, and Ankur Moitra. Simple, efficient, and neural algorithms for sparse coding. *Journal of Machine Learning Research*, 40(2015), 2015.

[11] Sanjeev Arora, Yuanzhi Li, Yingyu Liang, Tengyu Ma, and Andrej Risteski. A latent variable model approach to pmi-based word embeddings. *Transactions of the Association for Computational Linguistics*, 4:385–399, 2016.

[12] Sanjeev Arora, Yuanzhi Li, Yingyu Liang, Tengyu Ma, and Andrej Risteski. Linear algebraic structure of word senses, with applications to polysemy. *Transactions of the Association for Computational Linguistics*, 6:483–495, 2018.

[13] Sanjeev Arora, Simon S Du, Wei Hu, Zhiyuan Li, Ruslan Salakhutdinov, and Ruosong Wang. On exact computation with an infinitely wide neural net. *arXiv preprint arXiv:1904.11955*, 2019.

[14] Sanjeev Arora, Simon S. Du, Wei Hu, Zhiyuan Li, and Ruosong Wang. Fine-grained analysis of optimization and generalization for overparameterized two-layer neural networks. *CoRR*, abs/1901.08584, 2019. URL http://arxiv.org/abs/1901.08584.

[15] Anish Athalye, Nicholas Carlini, and David Wagner. Obfuscated gradients give a false sense of security: Circumventing defenses to adversarial examples. *arXiv preprint arXiv:1802.00420*, 2018.

[16] Ainesh Bakshi, Rajesh Jayaram, and David P Woodruff. Learning two layer rectified neural networks in polynomial time. *arXiv preprint arXiv:1811.01885*, 2018.

[17] Boaz Barak, Jonathan A Kelner, and David Steurer. Dictionary learning and tensor decomposition via the sum-of-squares method. In *Proceedings of the forty-seventh annual ACM symposium on Theory of computing*, pages 143–151, 2015.

[18] Richard Beigel. The polynomial method in circuit complexity. In *[1993] Proceedings of the Eigth Annual Structure in Complexity Theory Conference*, pages 82–95. IEEE, 1993.

[19] Battista Biggio, Igino Corona, Davide Maiorca, Blaine Nelson, Nedim Šrndić, Pavel Laskov, Giorgio Giacinto, and Fabio Roli. Evasion attacks against machine learning at test time. In *Joint European conference on machine learning and knowledge discovery in databases*, pages 387–402. Springer, 2013.

[20] Digvijay Boob and Guanghui Lan. Theoretical properties of the global optimizer of two layer neural network. *arXiv preprint arXiv:1710.11241*, 2017.

[21] Jehoshua Bruck and Roman Smolensky. Polynomial threshold functions, ac^0 functions, and spectral norms. *SIAM Journal on Computing*, 21(1):33–42, 1992.

[22] Alon Brutzkus and Amir Globerson. Globally optimal gradient descent for a convnet with gaussian inputs. *arXiv preprint arXiv:1702.07966*, 2017.

[23] Alon Brutzkus, Amir Globerson, Eran Malach, and Shai Shalev-Shwartz. Sgd learns overparameterized networks that provably generalize on linearly separable data. *arXiv preprint arXiv:1710.10174*, 2017. URL https://arxiv.org/abs/1710.10174.

[24] Sébastien Bubeck, Eric Price, and Ilya Razenshteyn. Adversarial examples from computational constraints. *arXiv preprint arXiv:1805.10204*, 2018.

[25] Mark Bun and Justin Thaler. Improved bounds on the sign-rank of ac^ 0. In *43rd International Colloquium on Automata, Languages, and Programming (ICALP 2016)*. Schloss Dagstuhl-Leibniz-Zentrum fuer Informatik, 2016.

[26] Yuan Cao and Quanquan Gu. Generalization bounds of stochastic gradient descent for wide and deep neural networks. In *Advances in Neural Information Processing Systems*, pages 10835–10845, 2019.

[27] Amit Daniely, Roy Frostig, and Yoram Singer. Toward deeper understanding of neural networks: The power of initialization and a dual view on expressivity. In *Advances in Neural Information Processing*
88

coding. In *Proceedings of the 26th annual international conference on machine learning*, pages 689–696, 2009.

[69] Julien Mairal, Francis Bach, Jean Ponce, and Guillermo Sapiro. Online learning for matrix factorization and sparse coding. *Journal of Machine Learning Research*, 11(Jan):19–60, 2010.

[70] Alexander Mordvintsev. Deepdreaming with tensorflow. https://github.com/tensorflow/tensorflow/blob/master/tensorflow/examples/tutorials/deepdream/deepdream.ipynb, 2016.

[71] Alexander Mordvintsev, Christopher Olah, and Mike Tyka. Inceptionism: Going deeper into neural networks. https://research.googleblog.com/2015/06/inceptionism-going-deeper-into-neural.html, 2015.

[72] Anh Nguyen, Jason Yosinski, and Jeff Clune. Deep neural networks are easily fooled: High confidence predictions for unrecognizable images. In *Proceedings of the IEEE Conference on Computer Vision and Pattern Recognition*, pages 427–436, 2015.

[73] Anh Nguyen, Alexey Dosovitskiy, Jason Yosinski, Thomas Brox, and Jeff Clune. Synthesizing the preferred inputs for neurons in neural networks via deep generator networks. In *Advances in neural information processing systems*, pages 3387–3395, 2016.

[74] Anh Nguyen, Jeff Clune, Yoshua Bengio, Alexey Dosovitskiy, and Jason Yosinski. Plug & play generative networks: Conditional iterative generation of images in latent space. In *Proceedings of the IEEE Conference on Computer Vision and Pattern Recognition*, pages 4467–4477, 2017.

[75] Ryan O'Donnell and Rocco A Servedio. New degree bounds for polynomial threshold functions. *Combinatorica*, 30(3):327–358, 2010.

[76] Chris Olah, Alexander Mordvintsev, and Ludwig Schubert. Feature visualization. *Distill*, 2017. doi: 10.23915/distill.00007. https://distill.pub/2017/feature-visualization.

[77] Bruno A Olshausen and David J Field. Sparse coding with an overcomplete basis set: A strategy employed by v1? *Vision research*, 37(23):3311–3325, 1997.

[78] Bruno A Olshausen and David J Field. Sparse coding of sensory inputs. *Current opinion in neurobiology*, 14(4):481–487, 2004.

[79] Audun Øygard. Visualizing googlenet classes. https://www.auduno.com/2015/07/29/visualizing-googlenet-classes, 2015.

[80] Aditi Raghunathan, Sang Michael Xie, Fanny Yang, John C Duchi, and Percy Liang. Adversarial training can hurt generalization. *arXiv preprint arXiv:1906.06032*, 2019.

[81] Alexander A Razborov and Alexander A Sherstov. The sign-rank of ac ˆ0. *SIAM Journal on Computing*, 39(5):1833–1855, 2010.

[82] Hadi Salman, Jerry Li, Ilya Razenshteyn, Pengchuan Zhang, Huan Zhang, Sebastien Bubeck, and Greg Yang. Provably robust deep learning via adversarially trained smoothed classifiers. In *Advances in Neural Information Processing Systems*, pages 11289–11300, 2019.

[83] Pouya Samangouei, Maya Kabkab, and Rama Chellappa. Defense-gan: Protecting classifiers against adversarial attacks using generative models. *arXiv preprint arXiv:1805.06605*, 2018.

[84] Ludwig Schmidt, Shibani Santurkar, Dimitris Tsipras, Kunal Talwar, and Aleksander Madry. Adversarially robust generalization requires more data. In *Advances in Neural Information Processing Systems*, pages 5014–5026, 2018.

[85] Karin Schnass. On the identifiability of overcomplete dictionaries via the minimisation principle underlying k-svd. *Applied and Computational Harmonic Analysis*, 37(3):464–491, 2014.

[86] Ali Shafahi, W Ronny Huang, Christoph Studer, Soheil Feizi, and Tom Goldstein. Are adversarial examples inevitable? *arXiv preprint arXiv:1809.02104*, 2018.

[87] David Silver, Aja Huang, Chris J Maddison, Arthur Guez, Laurent Sifre, George Van Den Driessche, Julian Schrittwieser, Ioannis Antonoglou, Veda Panneershelvam, Marc Lanctot, et al. Mastering the game of go with deep neural networks and tree search. *nature*, 529(7587):484, 2016.

[88] Mahdi Soltanolkotabi, Adel Javanmard, and Jason D Lee. Theoretical insights into the optimization landscape of over-parameterized shallow neural networks. *arXiv preprint arXiv:1707.04926*, 2017.
91